\documentclass{article}

\PassOptionsToPackage{square, comma, numbers, sort&compress}{natbib}
\usepackage[preprint]{neurips_data_2022}

\usepackage[T1]{fontenc}
\usepackage[font=small,labelfont=bf,tableposition=top]{caption}
\DeclareCaptionLabelFormat{andtable}{#1~#2  \&  \tablename~\thetable}

\usepackage[utf8]{inputenc} 
\usepackage[T1]{fontenc}    
\usepackage{hyperref}       
\usepackage{url}            
\usepackage{booktabs}       
\usepackage{amsfonts}       
\usepackage{nicefrac}       
\usepackage{microtype}      
\usepackage{xcolor}         
\usepackage{multirow}
\usepackage{graphicx}
\usepackage{xspace}
\usepackage{amsmath}
\usepackage{amsthm}
\usepackage{booktabs}
\usepackage{multicol}
\usepackage{float}
\usepackage{enumitem}

\usepackage{caption}
\usepackage{subcaption}
\usepackage[group-separator={,},group-minimum-digits={3}]{siunitx}

\newcommand{\ours}{\textsc{FedHPO-B}\xspace}
\newcommand{\TBD}[1]{\textcolor{orange}{#1}\xspace}
\newcommand{\subj}{FedHPO\xspace}
\newtheorem{proposition}{Proposition}

\title{\ours: A Benchmark Suite for Federated Hyperparameter Optimization}

\begin{document}

\author{
Zhen Wang$^{1,*}$, 
Weirui Kuang$^{1,*}$,
Ce Zhang$^{2}$, 
Bolin Ding$^{1}$, 
Yaliang Li$^{1,\dagger}$\\
\textsuperscript{\rm 1}Alibaba Group, \textsuperscript{\rm 2}ETH Z{\"u}rich\\
\{jones.wz, weirui.kwr, bolin.ding, yaliang.li\}@alibaba-inc.com, ce.zhang@inf.ethz.ch
}
\renewcommand*{\thefootnote}{\fnsymbol{footnote}}
\footnotetext[1]{Co-first authors.}
\footnotetext[2]{Corresponding author}
\renewcommand*{\thefootnote}{\arabic{footnote}}

\maketitle

\begin{abstract}
Hyperparameter optimization (HPO) is crucial for machine learning algorithms to achieve satisfactory performance, whose progress has been boosted by related benchmarks. Nonetheless, existing efforts in benchmarking all focus on HPO for traditional centralized learning while ignoring federated learning (FL), a promising paradigm for collaboratively learning models from dispersed data. In this paper, we first identify some uniqueness of HPO for FL algorithms from various aspects. Due to this uniqueness, existing HPO benchmarks no longer satisfy the need to compare HPO methods in the FL setting. To facilitate the research of HPO in the FL setting, we propose and implement a benchmark suite \ours that incorporates comprehensive FL tasks, enables efficient function evaluations, and eases continuing extensions. We also conduct extensive experiments based on \ours to benchmark a few HPO methods. We open-source \ours at \href{https://github.com/alibaba/FederatedScope/tree/master/benchmark/FedHPOB}{https://github.com/alibaba/FederatedScope/tree/master/benchmark/FedHPOB}.
\end{abstract}

\vspace{-0.05in}
\section{Introduction}
\label{sec:intro}
\vspace{-0.05in}
Most machine learning algorithms expose many design choices, which can drastically impact the ultimate performance.
Hyperparameter optimization (HPO)~\cite{hpobook} aims at making the right choices without human intervention.
Formally, HPO can be described as the problem $\min_{\lambda\in\Lambda_1 \times \cdots \times \Lambda_K}f(\lambda)$, where each $\Lambda_k$ corresponds to the candidate choices of a specific hyperparameter, e.g., taking the learning rate from $\Lambda_1 = [0.01,1.0]$ and the batch size from $\Lambda_2 = \{16, 32, 64\}$.
For each specified $\lambda$, $f(\lambda)$ is the output result (e.g., validation loss) of executing the considered algorithm configured by $\lambda$.

Research in this line has been facilitated by HPO benchmarks~\cite{automlbenchmark,hpobench,hpob}, which encourage reproducible and fair comparisons between different HPO methods. To this end, their primary efforts are two-fold: One is to keep the results of the same function evaluation consistent across different runtime environments, e.g., by containerizing its execution; The other is to simplify the evaluations, e.g., by evaluating a function via querying a readily available lookup table or a fitted surrogate model.

However, existing HPO benchmarks all focus on traditional learning paradigms, where the functions to be optimized correspond to centralized learning tasks.
Federated learning (FL)~\cite{deffl,flsurvey}, as a privacy-preserving paradigm for collaboratively learning a model from distributed data, has not been considered.
Actually, along with the increasing privacy concerns from the whole society, FL has been gaining more attentions from academia and industry.
Meanwhile, HPO for FL algorithms (denoted by \subj from now on) is identified as a critical and promising open problem in FL~\cite{flsurvey1}.

As an emerging topic, the community lacks a thorough understanding of how traditional HPO methods perform in the FL setting. Meanwhile, the recently proposed \subj methods have not been well benchmarked. Before attempting to fill this gap, it is helpful to gain some insights into the difference between \subj and traditional HPO. We elaborate on such differences from various aspects in Section~\ref{sec:background}, which essentially come from the distributed nature of FL and the heterogeneity among FL's participants.
In summary, the function to be optimized in \subj has an augmented domain that introduces new hyperparameter and fidelity dimensions, with the intricate correlations among them; The FL setting poses both opportunities and challenges in concurrently exploring the search space with a stricter budget constraint.

Due to \subj's uniqueness, existing HPO benchmarks cannot standardize the comparisons between HPO methods regarding FL tasks.
Firstly, their integrated functions correspond to non-FL tasks, which may lead to performances of compared methods inconsistent with their actual performances in optimizing FL algorithms.
Moreover, those recently proposed FedHPO methods need to incorporate into the procedure of function evaluation and thus can not be evaluated against existing benchmarks.
Motivated by \subj's uniqueness and the successes of previous HPO benchmarks, we summarize the desiderata of \subj benchmarks as follows.

\noindent\textbf{Comprehensiveness}. FL tasks are diverse in terms of data, model architecture, the level of heterogeneity among participants, etc. As their corresponding functions to be optimized by HPO methods are thus likely to be diverse, including a comprehensive collection of FL tasks is necessary for drawing an unbiased conclusion from comparisons.

\noindent\textbf{Efficiency}. As exact function evaluations are costly in the FL setting, an ideal benchmark is expected to provide tabular and surrogate modes for approximate but efficient function evaluations. When accurate results are required, the benchmark should enable simulated execution while reasonably estimating the corresponding deployment cost.

\noindent\textbf{Extensibility}. As a developing field, new FL tasks and novel \subj methods constantly emerge, and FL's best practice continuously evolves. Thus, what the community desired is more of a benchmarking tool that can effortlessly incorporate novel ingredients.

Towards these desiderata, we propose and implement \ours, a dedicated benchmark suite, to facilitate the research and application of \subj. \ours incorporates rich FL tasks from various domains with respective model architectures, providing realistic and, more importantly, comprehensive \subj problems for studying the related methods. In addition to the tabular and surrogate modes, \ours provides a configurable system model so that function evaluations can be efficiently executed via simulation while keeping the tracked time consumption meaningful. Last but not least, we build \ours upon a recently open-sourced FL platform FederatedScope (FS), which provides solid infrastructure and many off-the-shelf FL-related functionalities.
Thus, it is easy for the community to extend \ours with more and more tasks and \subj methods.
\vspace{-0.1in}
\section{Background and Motivations}
\label{sec:background}
\vspace{-0.1in}
We first give a brief introduction to the settings of HPO and its related benchmarks.
Then we present and explain the uniqueness of \subj to show the demand for dedicated \subj benchmarks.

\vspace{-0.1in}
\subsection{Problem Settings}
\label{subsec:problem}
\vspace{-0.1in}
As mentioned in Section~\ref{sec:intro}, HPO aims at solving $\min_{\lambda\in\Lambda_1 \times\cdots\times\Lambda_K}f(\lambda)$, where each $\Lambda_k$ corresponds to candidate choices of a specific hyperparameter, and their Cartesian product (denoted by $\times$) constitute the search space.
In practice, such $\Lambda_k$ is often bounded and can be continuous (e.g., an interval of real numbers) or discrete (e.g., a set of categories/integers).
Each function evaluation with a specified hyperparameter configuration $\lambda$ means to execute the corresponding algorithm accordingly, which results in $f(\lambda)$.
HPO methods, e.g., Gaussian process, generally solves this problem with a series of function evaluations.
To save the time and energy consumed by a full-fidelity function evaluation, multi-fidelity methods exploit low-fidelity function evaluation, e.g., training for fewer epochs~\cite{lowerrounds1,lowerrounds2} or on a subset of data~\cite{subset1,subset2,subset3}, to approximate the exact result.
Thus, it would be convenient to treat $f$ as $f(\lambda,b),\lambda\in\Lambda_1 \times \cdots \times \Lambda_K, b\in \mathcal{B}_1 \times \cdots \times \mathcal{B}_L$, where each $\mathcal{B}_l$ corresponds to the possible choices of a specific fidelity dimension, e.g., taking \textit{\#epoch} from $\{10,\ldots,50\}$.

For the purpose of benchmarking different HPO methods, it is necessary to integrate diverse HPO problems wherein the function to be optimized exhibits the same or at least similar characteristics as that in realistic applications.
To evaluate these functions, HPO benchmarks, e.g., HPOBench~\cite{hpobench}, often provide three modes:
(1) ``Raw'' means truly execute the corresponding algorithm;
(2) ``Tabular'' means querying a lookup table, where each entry corresponds to a specific $f(\lambda,b)$;
(3) ``Surrogate'' means querying a surrogate model that might be trained on the tabular data.

\subsection{Uniqueness of Federated Hyperparameter Optimization}
\label{subsec:fedhpo}
\noindent\textbf{Function evaluation in FL}.
Despite the various scenarios in FL literature, we restrict our discussion about \subj to one of the most general FL settings that has also been adopted in existing \subj works~\cite{fedex,fedhposys}.
Conceptually, there are $N$ clients, each of which has its specific data, and a server coordinates them to learn a model $\theta$ collaboratively.
Most FL algorithms are designed under this setting, including FedAvg~\cite{deffl} and FedOPT~\cite{fedopt}.
Such FL algorithms are iterative.
In the $t$-th round, the server broadcasts the global model $\theta^{(t)}$; then the clients make local updates and send the updates back; finally, the server aggregates the updates to produce $\theta^{(t+1)}$.
Obviously, this procedure consists of two subroutines---local updates and aggregation.
Thus, $\lambda$ can be divided into client-side and server-side hyperparameters according to which subroutine each hyperparameter influences.
After executing an algorithm configured by $\lambda$ for $T$ such rounds, what $\theta^{(T)}$ achieves on the validation set (e.g., its validation loss) is regarded as $f(\lambda)$.

The execution of an FL algorithm is essentially a distributed machine learning procedure while distinguishing from general non-FL cases by the heterogeneity among clients~\cite{fs}.
These characteristics make \subj unique against HPO for traditional machine learning algorithms.
We summarize the uniqueness from the following perspectives:

\noindent\textbf{Hyperparameter dimensions}.
Despite the server-side hyperparameters newly introduced by FL algorithms (e.g., FedOPT), some client-side hyperparameters, e.g., the \textit{\#local\_update\_step}, do not exist in the non-FL setting.
Moreover, these new hyperparameter dimensions bring in correlations that do not exist in HPO for traditional machine learning algorithms.
For example, \textit{\#local\_update\_step}, client-side \textit{learning\_rate}, and server-side \textit{learning\_rate} together determine the step size of each round's update.
Besides, their relationships are not only determined by the landscape of aggregated objective function but also the statistical heterogeneity of clients, which is a unique factor for FL.

\noindent\textbf{Fidelity dimensions}.
\subj introduces a new fidelity dimension---\textit{sample\_rate}, which determines the fraction of clients sampled for training in each round.
The larger \textit{sample\_client} is, the smaller the variance of each aggregation is, and the more resource each round consumes.
As existing fidelity dimensions, \textit{sample\_rate} allows to trade accuracy for efficiency.
Moreover, it correlates with other fidelity dimensions, such as \textit{\#round} $T$, where, in general, aggregation with smaller variance is believed to need fewer rounds for convergence.
This correlation encourages people to balance these quantities w.r.t. their system conditions, e.g., choosing large $T$ but small \textit{sample\_rate} when straggler issue is severe, to achieve more economical accuracy-efficiency trade-offs.

\noindent\textbf{Concurrent exploration}.
Unlike centralized learning, where each execution can only try a specific $\lambda$, some \subj works, such as FedEx~\cite{fedex}, concurrently explores different client-side configurations in each round and updates a policy w.r.t. the feedback from all these clients.
FedEx regards this strategy as a \subj counterpart to the weight-sharing strategy in neural architecture search.
However, the heterogeneity among clients is likely to make them have different optimal configurations~\cite{adaptivelr}, where making decisions by the same policy would become unsatisfactory.
In the same spirit as personalized FL~\cite{fedbn,ditto}, a promising direction is to decide on personalized hyperparameters in \subj.

\noindent\textbf{One-shot optimization}.
As each round in an FL course corresponds to two times of communication among participants (i.e., download and upload the model), the consumed resource, in terms of both time and carbon emission, is larger than that in centralized learning by orders of magnitude.
As a result, most traditional black-box optimizers that require more than one full-fidelity trials are impractical in the FL setting~\cite{robustadaptive}.
Thus, multi-fidelity methods, particularly those capable of one-shot optimization~\cite{fedex,flora}, are more in demand in \subj.

Due to the uniqueness mentioned above, existing HPO benchmarks are inappropriate for studying \subj.
\subj calls for dedicated benchmarks that incorporate functions corresponding to FL algorithms and respect realistic FL settings.
\section{Our Proposed Benchmark Suite: \ours}
\label{sec:ourmethod}
\begin{figure}
\centering
\includegraphics[width=0.75\textwidth]{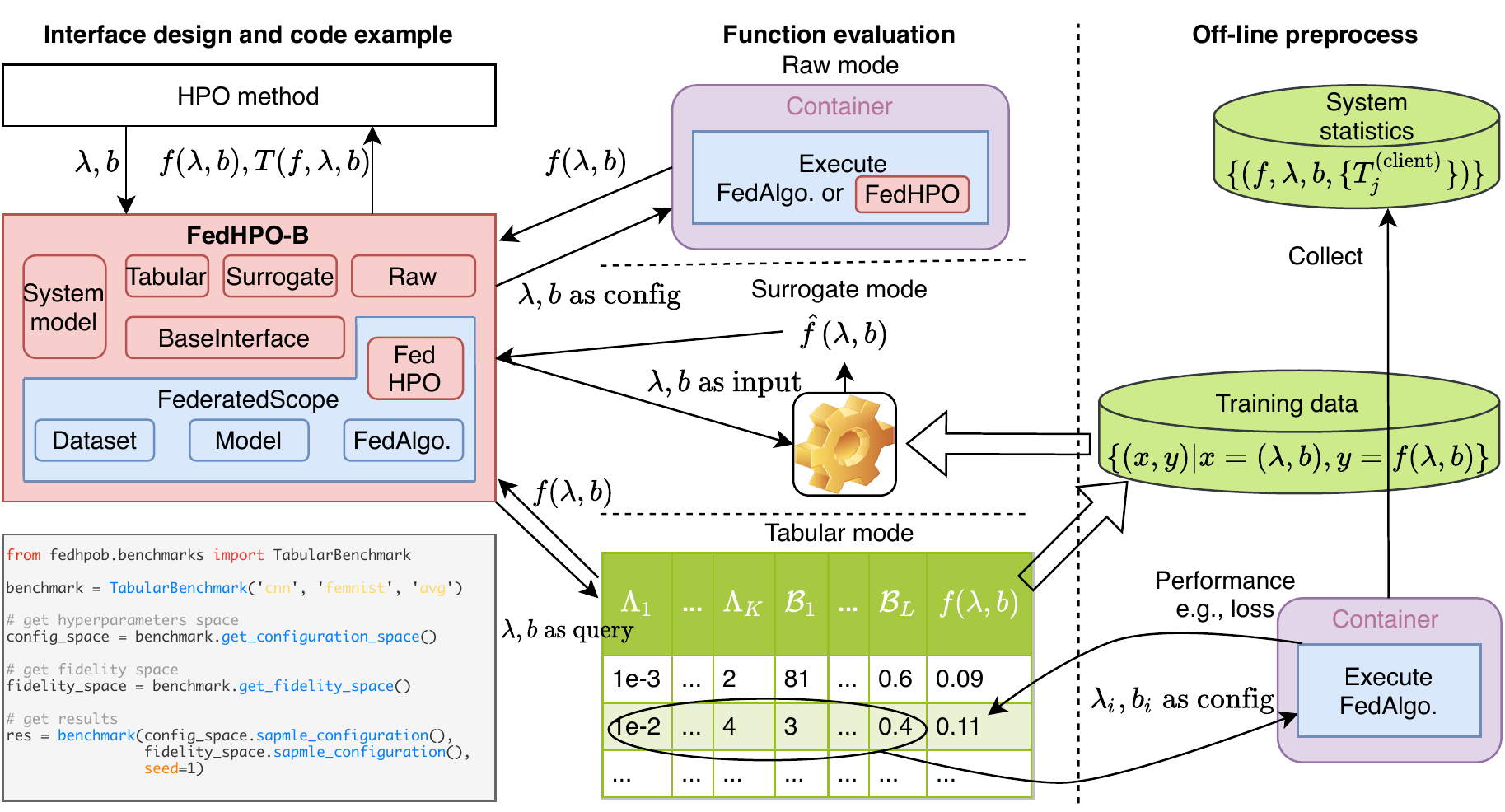}
\caption{Overview of \ours.}
\label{fig:overview}
\end{figure}

We present an overview of \ours in Figure~\ref{fig:overview}.
Conceptually, \ours encapsulates functions to be optimized and provides a unified interface for HPO methods to access.
As the incorporated functions correspond to FL tasks, we build \ours upon an FL platform---FederatedScope (FS)~\cite{fs}.
It offers many off-the-shelf and pluggable FL-related ingredients, which enable us to prepare a comprehensive collection of FL tasks (see Section~\ref{subsec:comprehensive}).
Besides, FS's event-driven framework and well-designed APIs allow us to easily incorporate more FL tasks and \subj methods into \ours, which is valuable for this nascent research direction (see Section~\ref{subsec:extensive}).

In \ours, function evaluations can be conducted in either of the three modes---``tabular'', ``surrogate'', and ``raw'', following the convention mentioned in Section~\ref{subsec:problem}.
To create the lookup table for tabular mode, we truly execute the corresponding FL algorithms with the grids of search space as their configurations.
These lookup tables are adopted as training data for the surrogate models, which are expected to approximate the functions of interest.
Meanwhile, we collect clients' execution time from these executions to form system statistics for our system model (see Section~\ref{subsec:efficient}).
As all our FL tasks and algorithms are implemented in FS, and FS has provided its docker images, we can containerize \ours effortlessly, i.e., the function evaluation in the ``raw'' mode is executed in an FS docker container.

\subsection{Comprehensiveness}
\label{subsec:comprehensive}
There is no universally best HPO method~\cite{automlbenchmark}.
Therefore, it is necessary to compare related methods on multiple HPO problems that correspond to diverse functions and thus can comprehensively evaluate their performances.

To satisfy this need, we leverage FS to prepare various FL tasks, where their considered datasets and model architectures are quite different. Specifically, the data can be images, sentences, graphs, or tabular data. Some datasets are provided by existing FL benchmarks, including FEMNIST (from LEAF~\cite{leaf}) and split Cora (from FS-GNN~\cite{fsg}), which are readily distributed and thus conform to the FL setting. Some are centralized initially (e.g., those from OpenML~\cite{openml,automlbenchmark} and Hugging Face~\cite{nlpdataset}), which we partition by FS's splitters to construct their FL version with Non-IIDness among clients. All these datasets are publicly available and can be downloaded and preprocessed by our prepared scripts. The corresponding suitable neural network model is applied to handle each dataset. Thus, these FL tasks involve fully-connected networks, convolutional networks, and the latest attention-based model. For each such FL task, we employ two FL algorithms---FedAvg and FedOPT to handle it, respectively, where it is worth mentioning that FedOPT has server-side hyperparameters.
\begin{table}[bthp]
    \centering
    \caption{Summary of benchmarks in current \ours: \#Cont. and \#Disc. denote the number of hyperparameter dimensions corresponding to continuous and discrete candidate choices, respectively.}
    \label{tab:probstat}
    \begin{tabular}{c|c|c|c|c|c|c|c}
    \toprule
    \multicolumn{1}{c}{Model} & \multicolumn{1}{c}{\#Dataset} & \multicolumn{1}{c}{Domain} & \multicolumn{1}{c}{\#Client} & \multicolumn{1}{c}{\#FL Algo.} & \multicolumn{1}{c}{\#Cont.} & \multicolumn{1}{c}{\#Disc.} & \multicolumn{1}{c}{Opt. budget} \\
    \hline
    \multicolumn{1}{c}{CNN} & \multicolumn{1}{c}{2} & \multicolumn{1}{c}{CV} & \multicolumn{1}{c}{200} & \multicolumn{1}{c}{2} & \multicolumn{1}{c}{4} & \multicolumn{1}{c}{2} & \multicolumn{1}{c}{20 days} \\
    \multicolumn{1}{c}{BERT~\cite{BERT}} & \multicolumn{1}{c}{2} & \multicolumn{1}{c}{NLP} & \multicolumn{1}{c}{5} & \multicolumn{1}{c}{2} & \multicolumn{1}{c}{4} & \multicolumn{1}{c}{2} & \multicolumn{1}{c}{20 days} \\
    \multicolumn{1}{c}{GNN} & \multicolumn{1}{c}{3} & \multicolumn{1}{c}{Graph} & \multicolumn{1}{c}{5} & \multicolumn{1}{c}{2} & \multicolumn{1}{c}{4} & \multicolumn{1}{c}{1} & \multicolumn{1}{c}{1 days} \\
    \multicolumn{1}{c}{LR} & \multicolumn{1}{c}{7} & \multicolumn{1}{c}{Tabular} & \multicolumn{1}{c}{5} & \multicolumn{1}{c}{2} & \multicolumn{1}{c}{3} & \multicolumn{1}{c}{1} & \multicolumn{1}{c}{21,600 seconds} \\
    \multicolumn{1}{c}{MLP} & \multicolumn{1}{c}{7} & \multicolumn{1}{c}{Tabular} & \multicolumn{1}{c}{5} & \multicolumn{1}{c}{2} & \multicolumn{1}{c}{4} & \multicolumn{1}{c}{3} & \multicolumn{1}{c}{43,200 seconds} \\
    \bottomrule
    \end{tabular}
\end{table}

Then the \subj problem is defined as optimizing the design choices of the FL algorithm on each specific FL task.
We are more interested in FL tasks' unique hyperparameter dimensions that are not involved in traditional centralized learning. Thus, client-side \textit{learning\_rate}, \textit{\#local\_ update\_step}, and server-side \textit{learning\_rate} are optimized in all provided \subj problems. Besides, in addition to \textit{\#round}, the unique fidelity dimension, \textit{sample\_rate}, is adopted. We summarize our currently provided \subj problems in Table~\ref{tab:probstat}. More details can be found in Appendix~\ref{sec:datasets} and Appendix~\ref{sec:benchmarks}.

\begin{figure}[htbp]
\centering
\begin{subfigure}[b]{0.193\textwidth}
\centering
\includegraphics[width=\textwidth]{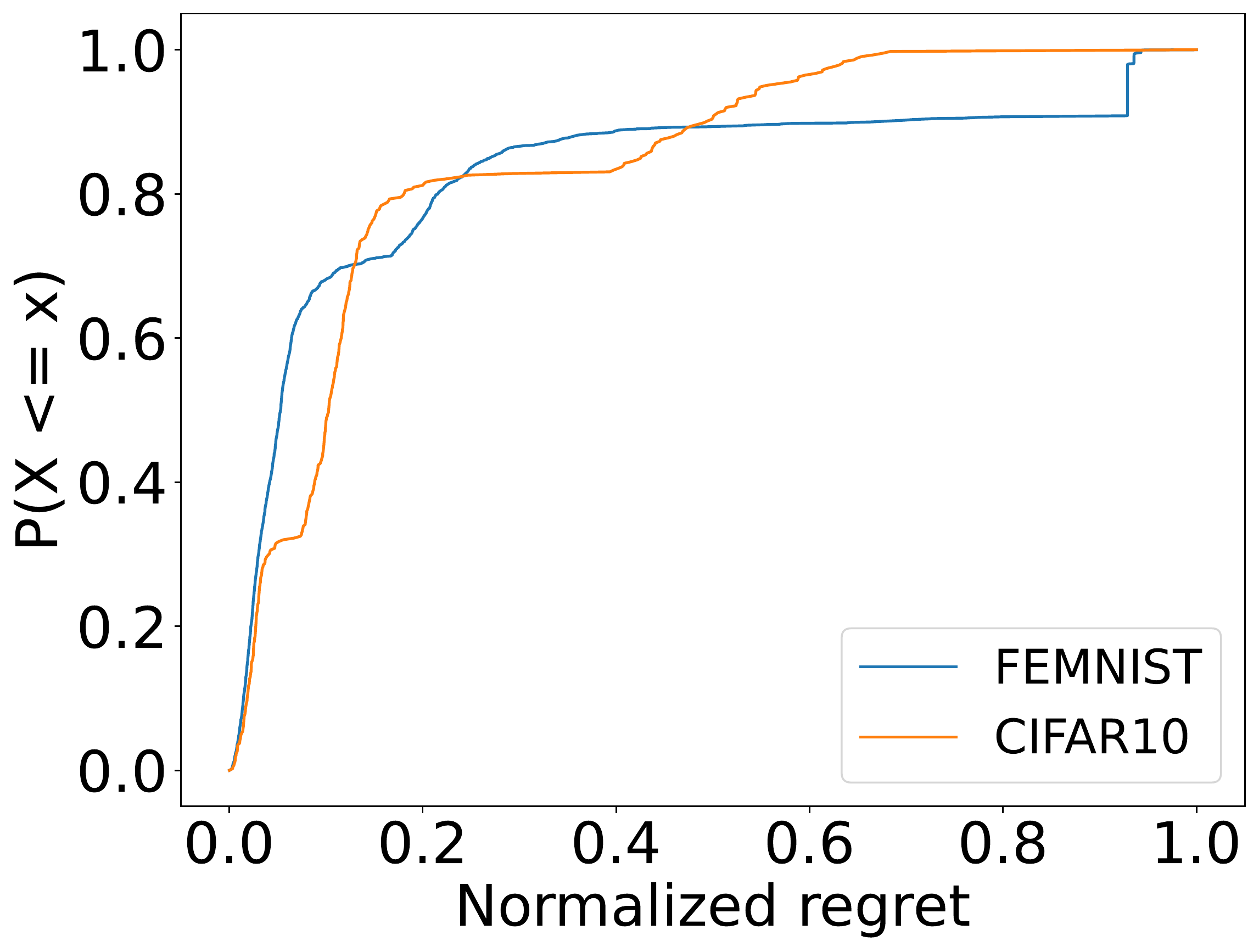}
\caption{CNN}
\end{subfigure}
\hfill
\begin{subfigure}[b]{0.193\textwidth}
\centering
\includegraphics[width=\textwidth]{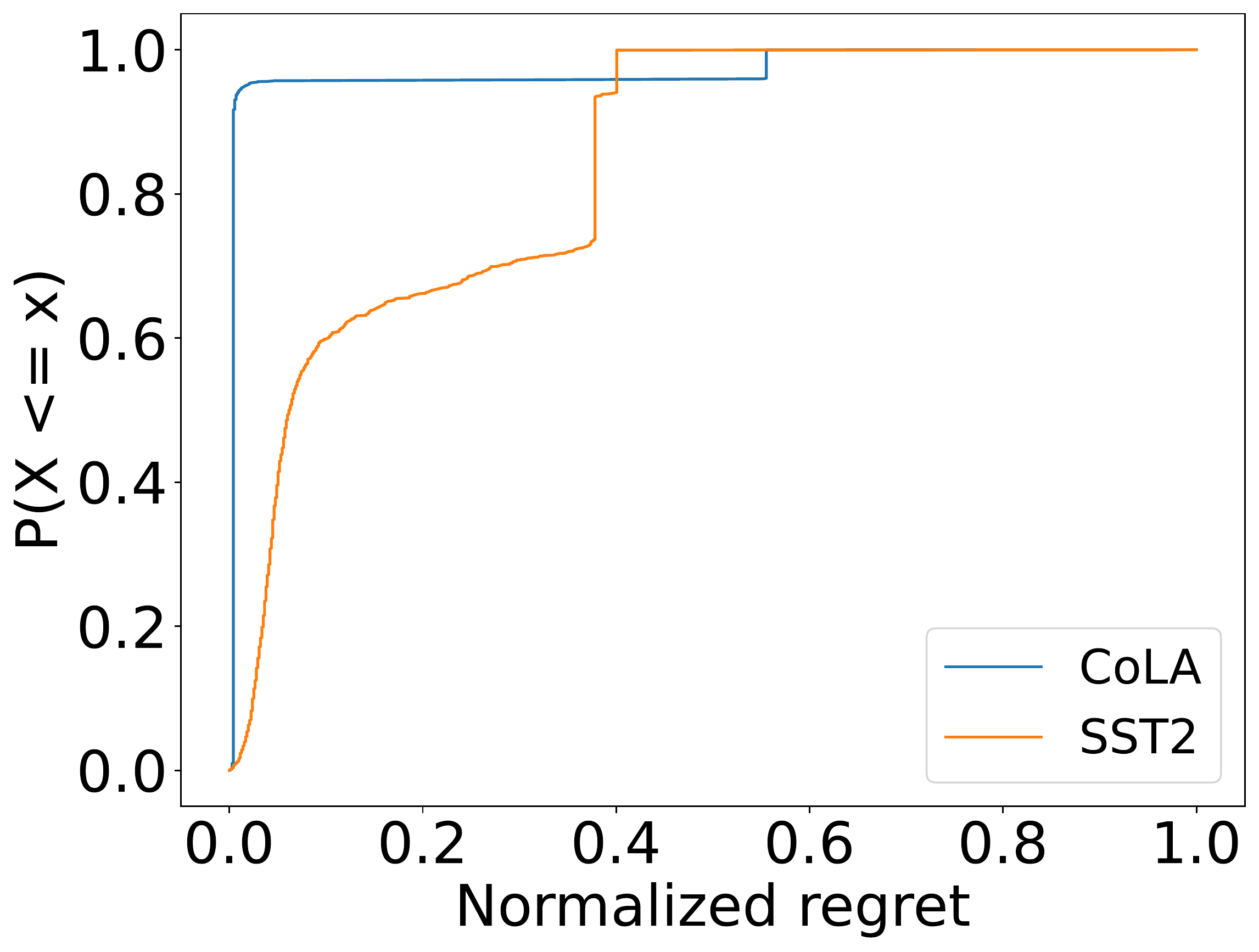}
\caption{BERT}
\end{subfigure}
\hfill
\begin{subfigure}[b]{0.193\textwidth}
\centering
\includegraphics[width=\textwidth]{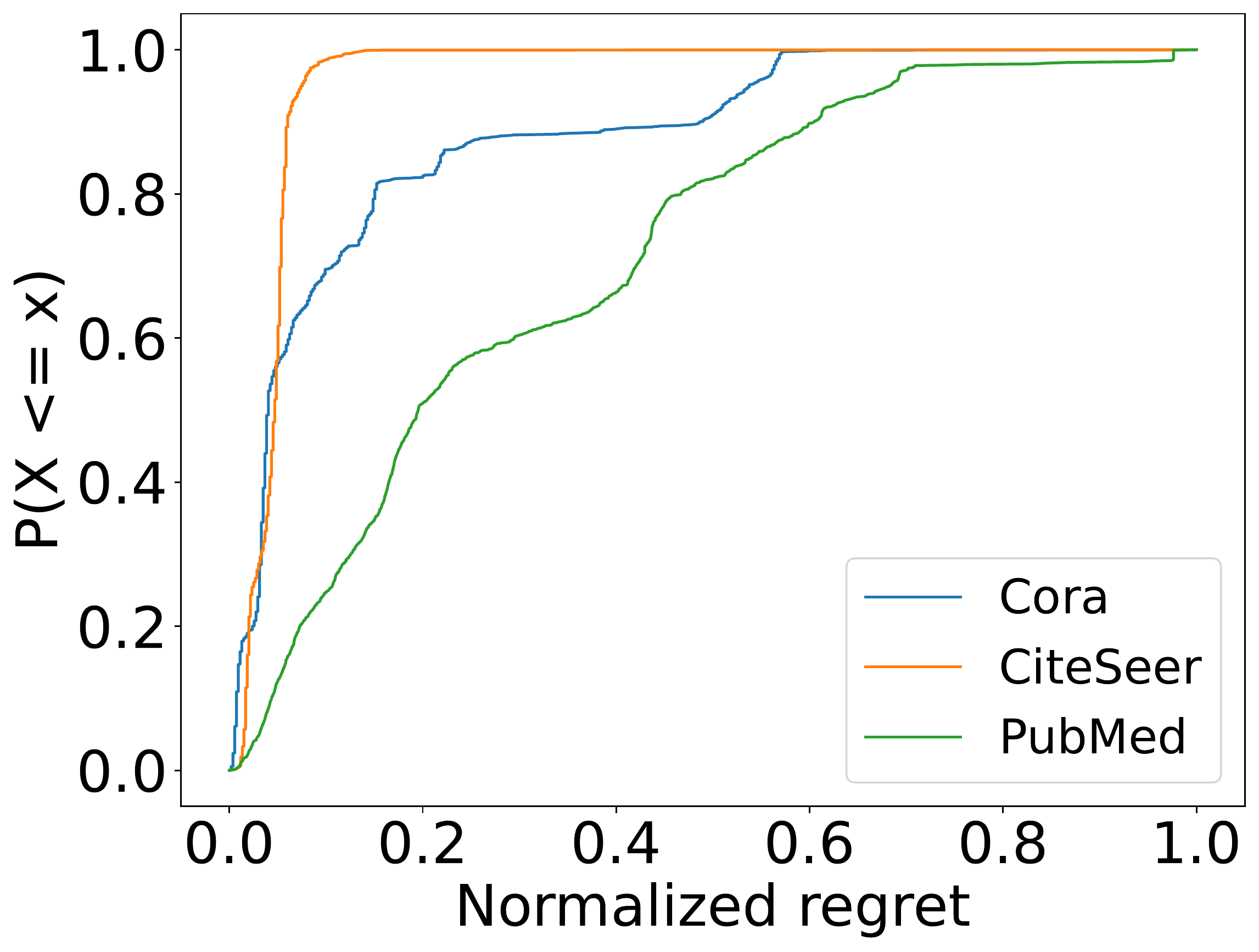}
\caption{GNN}
\end{subfigure}
\hfill
\begin{subfigure}[b]{0.193\textwidth}
\centering
\includegraphics[width=\textwidth]{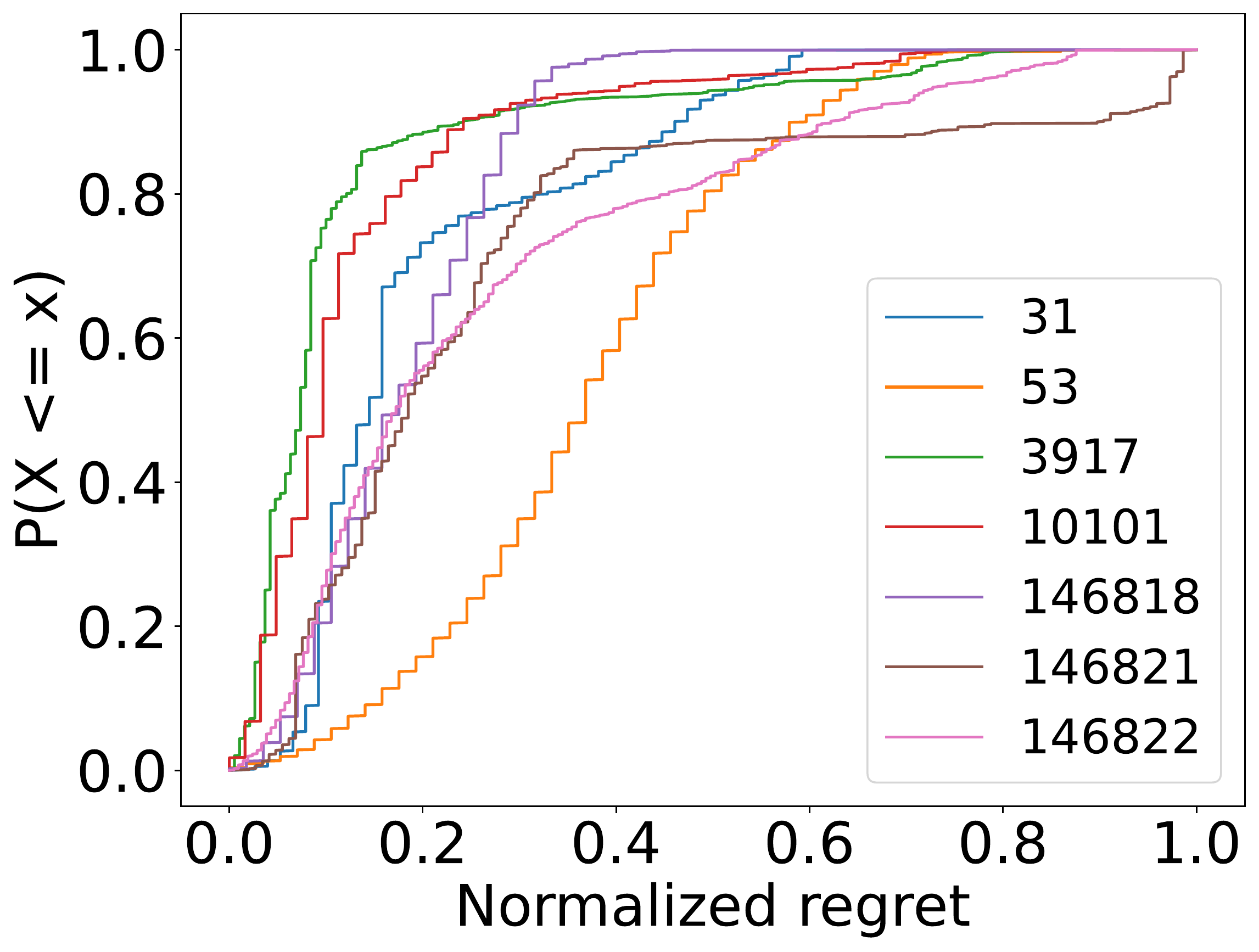}
\caption{LR}
\end{subfigure}
\begin{subfigure}[b]{0.193\textwidth}
\centering
\includegraphics[width=\textwidth]{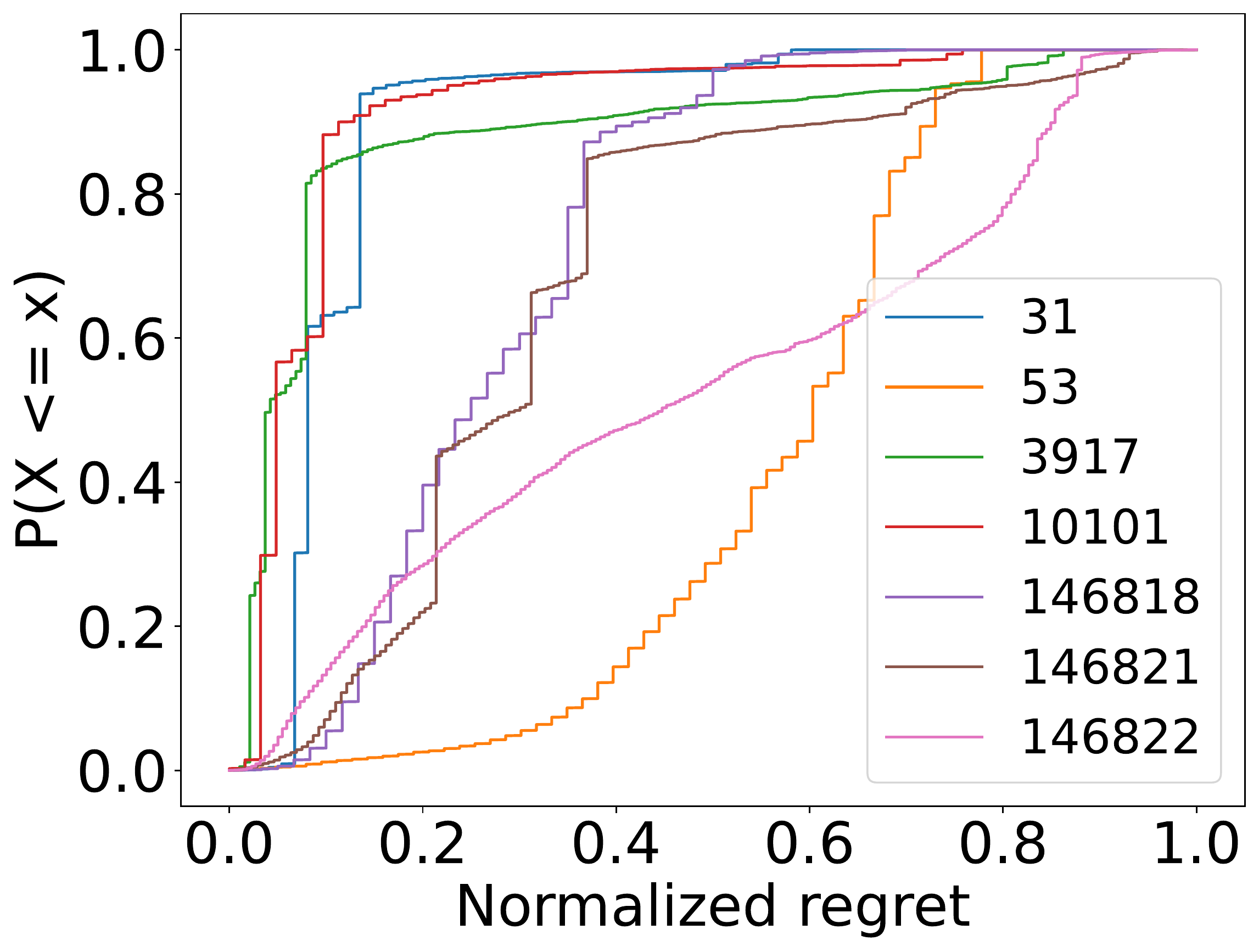}
\caption{MLP}
\end{subfigure}
\hfill
\caption{Empirical Cumulative Distribution Functions: The normalized regret is calculated for all evaluated configurations of the respective model on the respective FL task with FedAvg.}
\label{fig:cdf_avg}
\end{figure}

We study the empirical cumulative distribution function (ECDF) for each model type in \ours.
Specifically, in creating the lookup table for tabular mode, we have conducted function evaluations for the grid search space, resulting in a finite set $\{(\lambda, f(\lambda))\}$ for each benchmark.
Then we normalize the performances (i.e., $f(\lambda)$) and show their ECDF in Figure~\ref{fig:cdf_avg}, where these curves exhibit different shapes.
For example, the amounts of top-tier configurations for GNN on PubMed are remarkably less than on other graph datasets, which might imply a less smoothed landscape and difficulty in seeking the optimal configuration.
As the varying shapes of ECDF curves have been regarded as an indicator of the diversity of benchmarks~\cite{hpobench}, we can conclude from Figure~\ref{fig:cdf_avg} that \ours enables evaluating HPO methods comprehensively. We defer more studies about function landscape from the perspective of ECDF to Appendix~\ref{sec:more_results}.

We are continuously integrating more and more benchmarks into \ours to improve its comprehensiveness. Notably, we will incorporate the emerging learning paradigms, including federated reinforcement learning~\cite{fedrl}, federated unsupervised representation learning~\cite{fedunsupervised}, and federated hetero-task~\cite{fedhetero}, whose HPO problems have not been studied by the community.

\subsection{Efficiency}
\label{subsec:efficient}
For efficient function evaluation, we implement the tabular mode of \ours by running the FL algorithms configured by the grid search space in advance. Each specific configuration $\lambda$ is repeated five times with different random seeds, and the resulted performances, including loss, accuracy and f1-score under train/validation/test splits, are averaged and adopted as the results of $f(\lambda)$. Besides, we provide not only the results of $f(\lambda)$ (i.e., that with full-fidelity) but also results of $f(\lambda,b)$, where $b$ is enumerated across different \textit{\#round} and different \textit{sample\_rate}. Since executing function evaluation is much more costly in FL than traditional centralized learning, such lookup tables are precious. In creating them, we spent about two months of computation time on six machines, each with four Nvidia V100 GPUs. Now we make them publicly accessible via the tabular mode of \ours.

As tabular mode has discretized the original search space and thus cannot respond to queries other than the grids, we train random forest models on these lookup tables, i.e., $\{(\lambda, b), f(\lambda,b))\}$. These models serve as a surrogate of the functions to be optimized and can answer any query $\lambda$ by simply making an inference. More details about implementing the tabular and surrogate modes of \ours are deferred to Appendix~\ref{sec:benchmarks}.

When an HPO method interacts with \ours in raw mode, each function evaluation is to run the corresponding FL course, which can be conducted by indeed executing it on a cluster of FL participants or simulating this execution in a standalone machine. Simulation is preferred, as it can provide consistent results as running on a cluster while saving time and energy. However, the time consumed by simulation cannot reasonably reflect that by actual execution, which makes the HPO method fail to track the depleted budget. Hence, a system model that can estimate the time consumed by evaluating $f(\lambda, b)$ in realistic scenarios is indispensable. Meanwhile, such a system model should be configurable so that users with different system conditions can calibrate the model to their cases.
Therefore, we propose and implement a novel system model based on a basic one~\cite{fedoptimizationsurvey}.
Formally, the execution time for each FL round in our model is estimated as follow:

\vspace{-0.1in}
\begin{equation}
\label{eq:sysmodel}
\begin{split}
T(f,\lambda,b) &= T_{\text{comm}}(f,\lambda,b) + T_{\text{comp}}(f,\lambda,b),\\
T_{\text{comm}}(f,\lambda,b) &= \max(\frac{N\times S_{\text{down}}(f,\lambda)}{B_{\text{up}}^{(\text{server})}}, \frac{S_{\text{down}}(f,\lambda)}{B_{\text{down}}^{(\text{client})}}) + \frac{S_{\text{up}}(f,\lambda)}{B_{\text{up}}^{(\text{client})}},\\
T_{\text{comp}}(f,\lambda,b) &= \mathbb{E}_{T_{i}^{(\text{client})}\sim\text{Exp}(\cdot |\frac{1}{c(f,\lambda,b)}),i=1,\ldots,N}[\max(\{T_{i}^{(\text{client})}\})] + T^{(\text{server})}(f,\lambda,b),
\end{split}
\end{equation}
\vspace{-0.05in}

where $N$ denotes the number of clients sampled in this round, $S(f,\lambda)$ denotes the download/upload size, $B$ denotes the download/upload bandwidth of server/client, $T^{(\text{server})}$ is the time consumed by server-side computation, and $T_{i}^{(\text{client})}$ denotes the computation time consumed by $i$-th client, which is sampled from an exponential distribution with $c(f,\lambda,b)$ as its mean.
Compared with the existing basic model, one ingredient we add is to reflect the bottleneck issue of the server.
Specifically, the server broadcasts model parameters for $N$ clients in each round, which might become the bottleneck of the communication.
And $N$ is determined by the total number of clients in the considered FL task and \textit{sample\_rate} ($b$ specified).
Another ingredient is to consider the heterogeneity among clients' computational capacity, where the assumed exponential distribution has been widely adopted in system designs~\cite{fedoptimizationsurvey} and is consistent with real-world applications~\cite{papaya}.
As the local updates are not sent back simultaneously, there is no need to consider the bottleneck issue for the server twice.

To implement our system model, we use the following proposition to calculate Eq.~\eqref{eq:sysmodel} analytically.
\begin{proposition}
When the computation time of clients are identically independently distributed, following an exponential distribution $\text{Exp}(\cdot|\frac{1}{c})$, then the expected time for the straggler of $N$ uniformly sampled clients is $\sum_{i=1}^{N}\frac{c}{i}$.
\label{prop:1}
\end{proposition}
Proof can be found in Appendix~\ref{sec:proof}.
We provide default parameters of our system model, including $c$, $B$, and $T^{(\text{server})}$, based on observations collected from the executions in Section~\ref{subsec:efficient}.
Users are allowed to specify these parameters according to their scenarios or other system statistic providers, e.g., estimating the computation time of stragglers by sampling from FedScale~\cite{fedscale}.

\subsection{Extensibility}
\label{subsec:extensive}
\begin{figure}[b]
    \centering
    \includegraphics[width=0.95\textwidth]{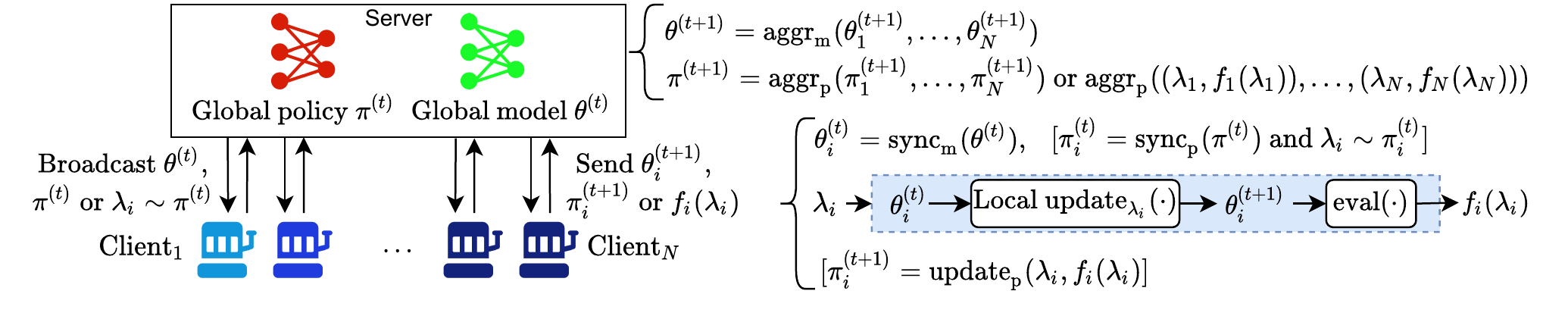}
    \caption{A general algorithmic view for \subj methods: They are allowed to concurrently explore different client-side configurations in the same round of FL, but the clients are heterogeneous, i.e., corresponding to different functions $f_i(\cdot)$. Operators in brackets are optional.}
    \label{fig:fedhpoalgoarch}
\end{figure}
Traditional HPO methods are decoupled from the procedure of function evaluation, with a well-defined interface for interaction (see Figure~\ref{fig:overview}). Thus, any novel method is readily applicable to optimizing the prepared functions and could be integrated into \ours without further development. However, \subj methods, including FTS~\cite{fedbo} and FedEx~\cite{fedex}, are often coupled with the FL procedure, which needs to be implemented in FS if we want to incorporate them into \ours, as the red color ``FedHPO'' module in Figure~\ref{fig:overview} shows. As \subj is springing up, we must ease the development of novel \subj methods so that \ours is extensible.

We present a general algorithmic view in Figure~\ref{fig:fedhpoalgoarch}, which unifies several related methods and thus would benefit \ours's extensibility. In this view, \subj follows what an FL round is framed: (1) server broadcasts information; (2) clients make local updates and send feedback; (3) server aggregates feedback. At the server-side, we maintain the global policy for determining hyperparameter configurations. In addition to the model parameters, either the policy or configurations sampled from it are also broadcasted. If the $i$-th client receives the global policy, it will update its local policy w.r.t. the global one and then sample a configuration from its local policy. Either received or locally sampled, the configuration $\lambda_i$ is specified for the local update procedure, which results in updated local model parameters $\theta^{(t+1)}_i$. Then $\theta^{(t+1)}_i$ is evaluated, and its performance is regarded as the result of (client-specific) function evaluation on $\lambda_i$, i.e., $f_i(\lambda_i)$. Finally, both $\theta^{(t+1)}_i$ and $(\lambda_i, f_i(\lambda_i))$ are sent back to the server, which will be aggregated for updating the global model and policy, respectively.

We have implemented FedEx in FS with such a view, where $\lambda_i$ is independently sampled from the global policy, and the ``$\text{aggr}_{\text{p}}$'' operator is exponential gradient descent. Other \subj methods, e.g., FTS, can also be implemented with our view. In FTS, the broadcasted policy $\pi^{(t)}$ is the samples drawn from all clients' posterior beliefs. The ``$\text{sync}_{\text{p}}$'' operator can be regarded as mixing Gaussian process (GP) models.
``$\text{update}_{\text{p}}$'' operator corresponds to updating local GP model.
Then a sample drawn from local GP posterior belief is regarded as $\pi^{(t+1)}_i$ and sent back. The ``$\text{aggr}_{p}$'' operator corresponds to packing received samples together.

We choose to build \ours on FS as it allows developers to flexibly customize the message exchanged among FL participants. Meanwhile, the native procedures to handle a received message could be modularized. These features make it easy to express novel \subj methods with the above view. Last but not least, FS's rich off-the-shelf datasets, splitters, models, and trainers have almost eliminated the effort of introducing more FL tasks into \ours.
\section{Experiments}
\label{sec:exp}
We conduct extensive empirical studies with our proposed \ours. Basically, we exemplify the use of \ours in comparing HPO methods, which, in the meantime, can somewhat validate the correctness of \ours. Moreover, we aim to gain more insights into \subj, answering three research questions: \textbf{(RQ1)} How do traditional HPO methods perform in the FL setting? \textbf{(RQ2)} Do recently proposed methods that exploit ``concurrent exploration'' (see Section~\ref{sec:background}) significantly improve traditional methods? \textbf{(RQ3)} How can we leverage the new fidelity dimension of \subj?
All scripts concerning the studies here will be committed to \ours so that the community can quickly reproduce our established benchmarks.

\subsection{Studies about Applying Traditional HPO Methods in the FL Setting}
\label{subsec:exp1}
To answer RQ1, we largely follow the experiment conducted in HPOBench~\cite{hpobench} but focus on the \subj problems \ours provided.

\noindent\textbf{Protocol.}
We employ up to ten optimizers (i.e., HPO methods) from widely adopted libraries (see Table~\ref{tab:hpo_optimizers} for more details).
For black-box optimizers (\textit{BBO}), we consider random search (\textit{RS}), the evolutionary search approach of differential evolution (\textit{DE}~\cite{de1,de2}), and bayesian optimization with: a GP model ($\textit{BO}_{\textit{GP}}$), a random forest model ($\textit{BO}_{\textit{RF}}$~\cite{borf}), and a kernel density estimator ($\textit{BO}_{\textit{KDE}}$~\cite{kde}), respectively.
For multi-fidelity optimizers (\textit{MF}), we consider Hyperband (\textit{HB}~\cite{hyperband}), its model-based extensions with KDE-based model (\textit{BOHB}~\cite{bohb}), and differential evolution (\textit{DEHB}~\cite{dehb}), and Optuna's implementations of TPE with median stopping ($\textit{TPE}_{\textit{MD}}$) and TPE with Hyperband ($\textit{TPE}_{\textit{HB}}$)~\cite{optuna}.
We apply these optimizers to optimize the design choices of FedAvg and FedOPT on 20 FL tasks drawn from what \ours currently provides (see Table~\ref{tab:probstat}). These FL tasks involve five model types and four data domains. To compare the optimizers uniformly and fairly, we repeat each setting five times in the same runtime environment but with different random seeds. The best-seen validation loss is monitored for each optimizer (for multi-fidelity optimizers, higher fidelity results are preferred over lower ones). We sort the optimizers by their best-seen results and compare their mean ranks on these 20 FL tasks. Following HPOBench~\cite{hpobench}, we use sign tests to judge whether advanced methods outperform their baselines and whether multi-fidelity methods outperform their single-fidelity counterparts. We refer our readers to Appendix~\ref{sec:optimizers} for more details.

\begin{figure}[htbp]
	\centering
	\begin{subfigure}{0.32\linewidth}
		\centering
		\includegraphics[width=0.95\linewidth]{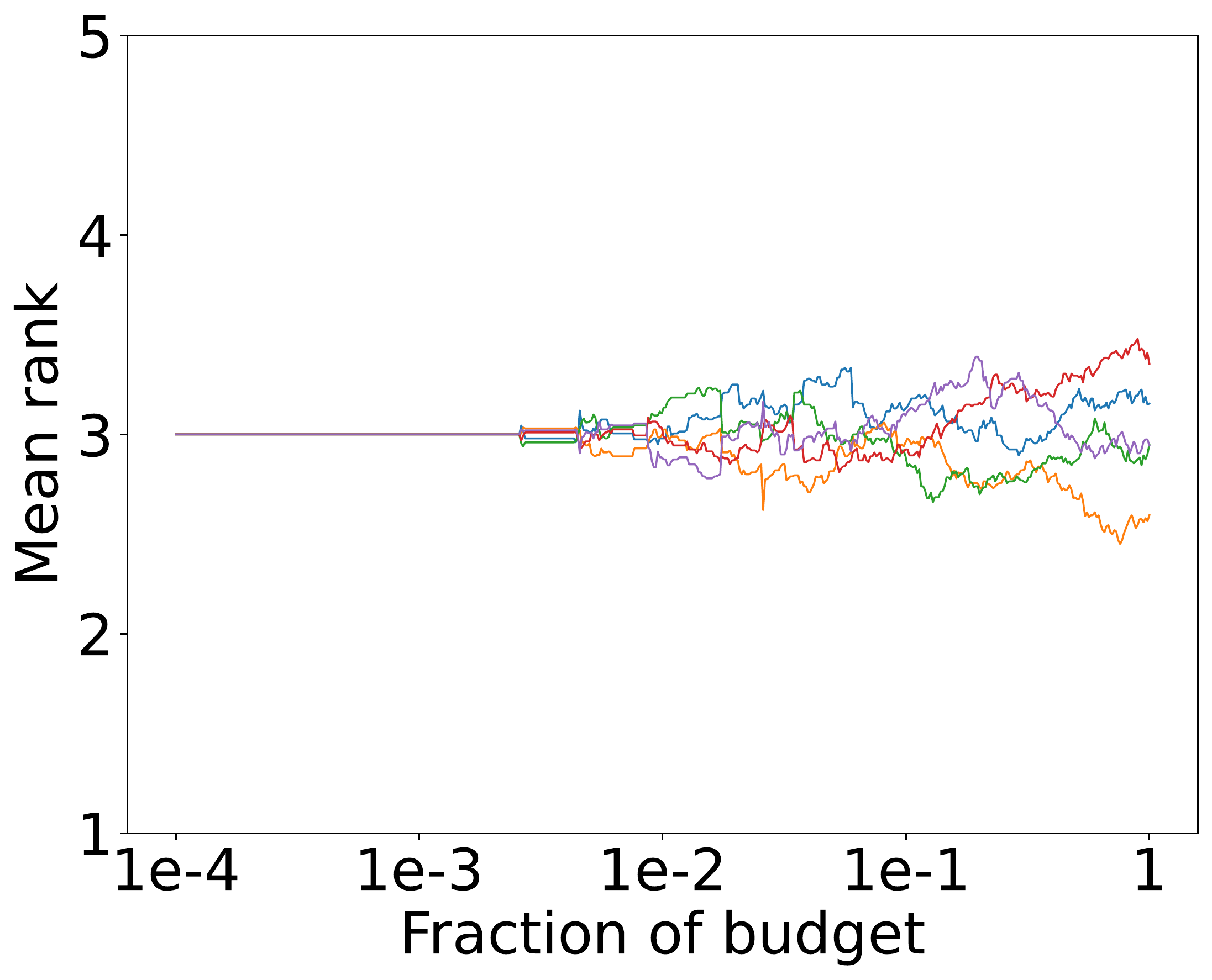}
		\caption{\textit{BBO}}
		\label{fig:BBO_All_avg}
	\end{subfigure}
	\begin{subfigure}{0.32\linewidth}
		\centering
		\includegraphics[width=0.95\linewidth]{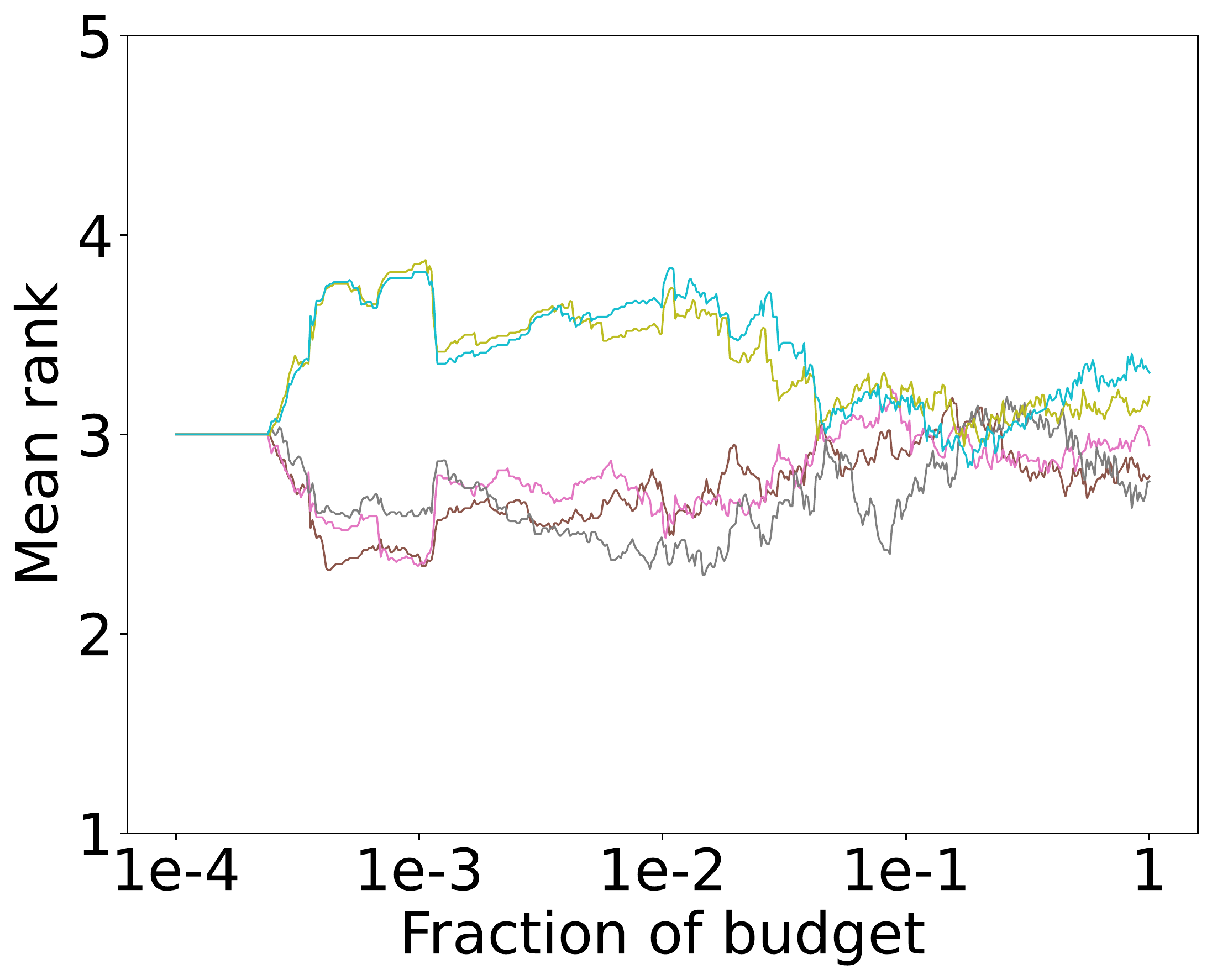}
		\caption{\textit{MF}}
		\label{fig:MF_All_avg}
	\end{subfigure}
	\begin{subfigure}{0.32\linewidth}
		\centering
		\includegraphics[width=0.95\linewidth]{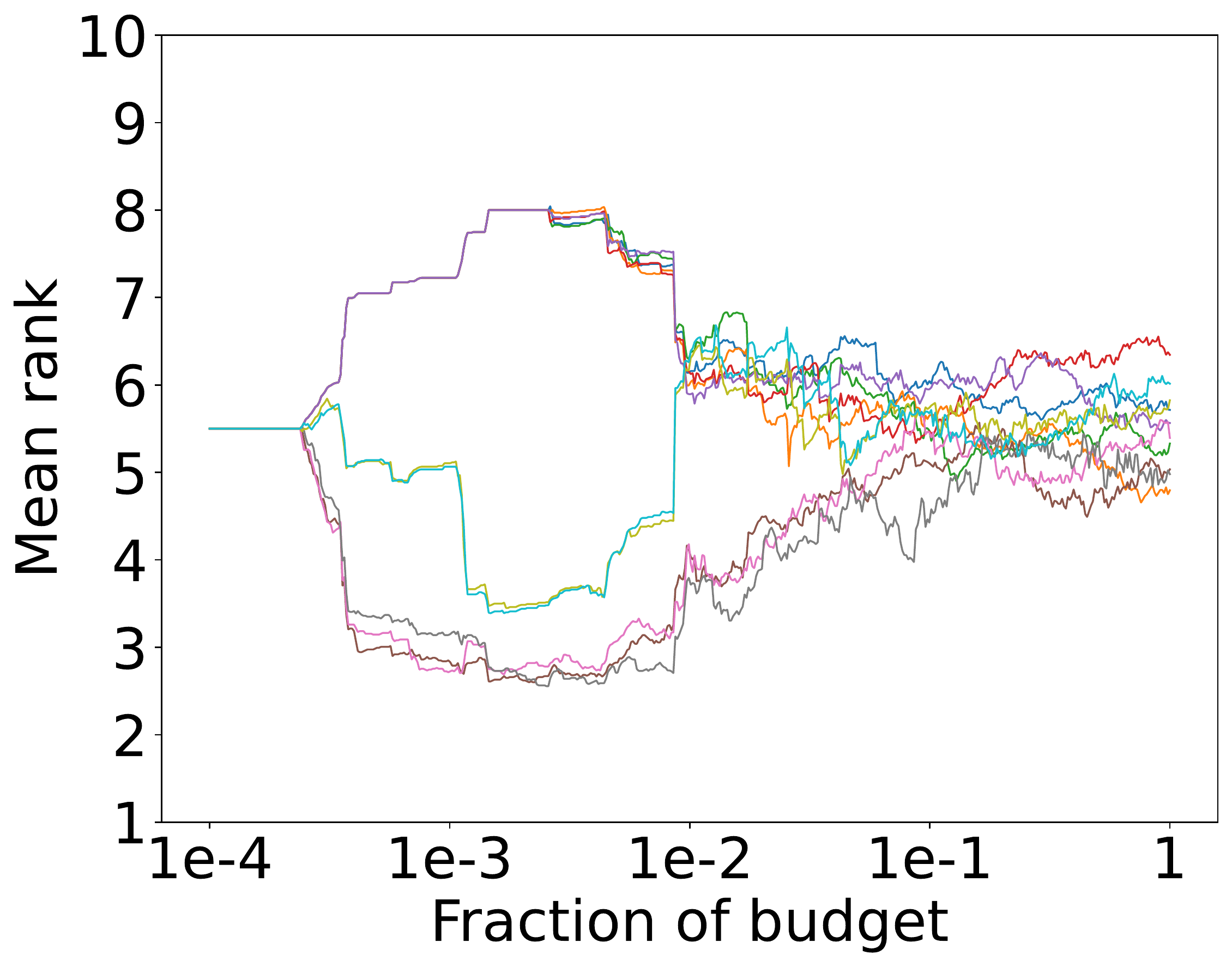}
		\caption{All}
		\label{fig:All_All_avg}
	\end{subfigure}
	\centering
	\hspace*{1.2cm}\begin{subfigure}{1.0\linewidth}
		\centering
		\includegraphics[width=0.95\linewidth]{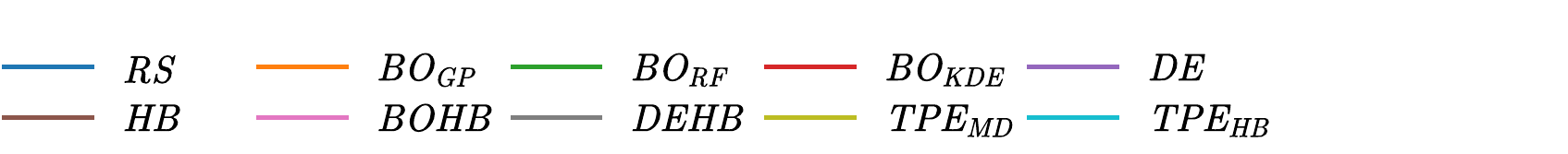}
	\end{subfigure}
	\vspace{-0.1in}
	\caption{Mean rank over time on all \subj problems (with FedAvg).}
	\label{fig:entire_all_tabular_avg_rank}
	\vspace{-0.2in}
\end{figure}

\noindent\textbf{Results and Analysis.}
We show the results in Figure~\ref{fig:entire_all_tabular_avg_rank}. Overall, their eventual mean ranks do not deviate remarkably. For \textit{BBO}, the performances of optimizers are close at the beginning but become more distinguishable along with their exploration. Ultimately, $\textit{BO}_{\textit{GP}}$ has successfully sought better configurations than other optimizers. In contrast to \textit{BBO}, \textit{MF} optimizers perform pretty differently in the early stage, which might be rooted in the vast variance of low-fidelity function evaluations. Eventually, \textit{HB} and $\textit{BOHB}$ become superior to others while achieving a very close mean rank.
We consider optimizers' final performances on these 20 tasks, where, for each pair of optimizers, one may win, tie, or lose against the other. Then we can conduct sign tests to compare pairs of optimizers, where results are presented in Table~\ref{tab:p-value} and Table~\ref{tab:p-value2}. Comparing these advanced optimizers with their baselines, only $\textit{BO}_{\textit{GP}}$, $\textit{BO}_{\textit{RF}}$, and \textit{DE} win on more than half of the FL tasks but have no significant improvement. Meanwhile, no \textit{MF} optimizers show any advantage in exploiting experience. These observations differ from non-FL cases, where we presume the reason lies in the distribution of configurations' performances (see Figure~\ref{fig:cdf_avg}). From Table~\ref{tab:p-value2}, we see that \textit{MF} optimizers always outperform their corresponding single-fidelity version, which is consistent with non-FL settings.

\begin{table}[htbp]
\centering
\vspace{-0.1in}
\caption{P-value of a sign test for the hypothesis---these advanced methods surpass the baselines (\textit{RS} for \textit{BBO} and \textit{HB} for \textit{MF}).}
\label{tab:p-value}
\begin{tabular}{ccccc}
\toprule
                     & $\textit{BO}_{\textit{GP}}$ & $\textit{BO}_{\textit{RF}}$ & $\textit{BO}_{\textit{KDE}}$   & $\textit{DE}$         \\
p-value agains \textit{RS}  & 0.0637         & 0.2161         & 0.1649            & 0.7561            \\
win-tie-loss         & 13 / 0 / 7         & 12 / 0/ 8         & 7 / 0 / 13            & 11 / 0 / 9            \\ \hline
                     & $\textit{BOHB}$    & $\textit{DEHB}$    & $\textit{TPE}_{\textit{MD}}$ & $\textit{TPE}_{\textit{HB}}$ \\
p-value against \textit{HB} & 0.4523         & 0.9854         & 0.2942            & 0.2454            \\
win-tie-loss         & 7 / 0 / 13         & 9 / 0 / 11         & 9 / 0 / 11            & 9 / 0 / 11            \\ \bottomrule
\end{tabular}%
\end{table}

\begin{table}[hbp]
\centering
\vspace{-0.2in}
\caption{P-value of a sign test for the hypothesis---\textit{MF} methods surpass corresponding \textit{BBO} methods.}
\label{tab:p-value2}
\begin{tabular}{cccc}
\toprule
                     & $\textit{HB}$ vs. $\textit{RS}$ & $\textit{DEHB}$ vs. $\textit{DE}$ & $\textit{BOHB}$ vs. $\textit{BO}_{\textit{KDE}}$          \\
p-value  & 0.1139         & 0.2942         & \textbf{0.0106}                  \\
win-tie-loss         & 13 / 0 / 7         & 13 / 0 / 7         & 16 / 0 / 4            \\  \bottomrule
\end{tabular}%
\vspace{-0.1in}
\end{table}

\subsection{Studies about Concurrent Exploration}
\label{subsec:exp2}
As mentioned in Section~\ref{sec:background}, the cost of communication between FL participants has made acquiring multiple full-fidelity function evaluations unaffordable, posing a stricter budget constraint to HPO methods. Yet FL, at the same time, allows HPO methods to take advantage of concurrent exploration, which somewhat compensates for the number of function evaluations. We are interested in methods designed regarding these characteristics of \subj and design this experiment to see how much concurrent exploration contributes.

\noindent\textbf{Protocol.} We consider the FL tasks where FedAvg and FedOPT are applied to learn a 2-layer CNN on FEMNIST. As a full-fidelity function evaluation consumes 500 rounds on this dataset, we carefully specify \textit{RS} and successive halving algorithm (\textit{SHA}) to limit their total budget as a one-shot optimization in terms of \textit{\#round}. Precisely, \textit{RS} consists of ten trials, each running for 50 rounds. \textit{SHA}, initialized with 27 candidate configurations, consists of three stages with budgets to be 12, 13, and 19 rounds. Then we adopt \textit{RS}, \textit{SHA}, FedEx wrapped by \textit{RS} (\textit{RS+FedEx}), and FedEx wrapped by \textit{SHA} (\textit{SHA+FedEx}) to optimize the design choices of FedAvg and FedOPT, respectively. The wrapper is responsible for (1) determining the server-side \textit{learning\_rate} for FedOPT and (2) determining the arms for FedEx. We consider validation loss the metric of interest, and function evaluations are conducted in the raw mode. We repeat each method three times and report the averaged best-seen value at the end of each trial. Meanwhile, for each considered method, we entirely run the FL course with the optimal configuration it seeks. Their averaged test accuracies are compared.

\noindent\textbf{Results and Analysis.} We present the results in Figure~\ref{fig:exp2} and Table~\ref{tab:exp2}. For FedAvg, the best-seen mean validation losses of all wrapped FedEx decrease slower than their corresponding wrapper. However, their searched configurations' generalization performances are significantly better than their wrappers, which strongly confirms the effectiveness of concurrent exploration. As for FedOPT, all wrapped FedEx show better regrets than their corresponding wrapper. However, as the one-shot setting has drastically limited the number of initial configurations, all searched configurations cannot lead to satisfactory performances. Notably, the most crucial hyperparameter, server-side \textit{learning\_rate}, cannot be well specified.

\begin{table}[ht]
\begin{minipage}[b]{0.6\linewidth}
\centering
\captionsetup{justification=centering}
	\centering
\begin{subfigure}{0.45\linewidth}
	\centering
	\includegraphics[width=0.98\linewidth]{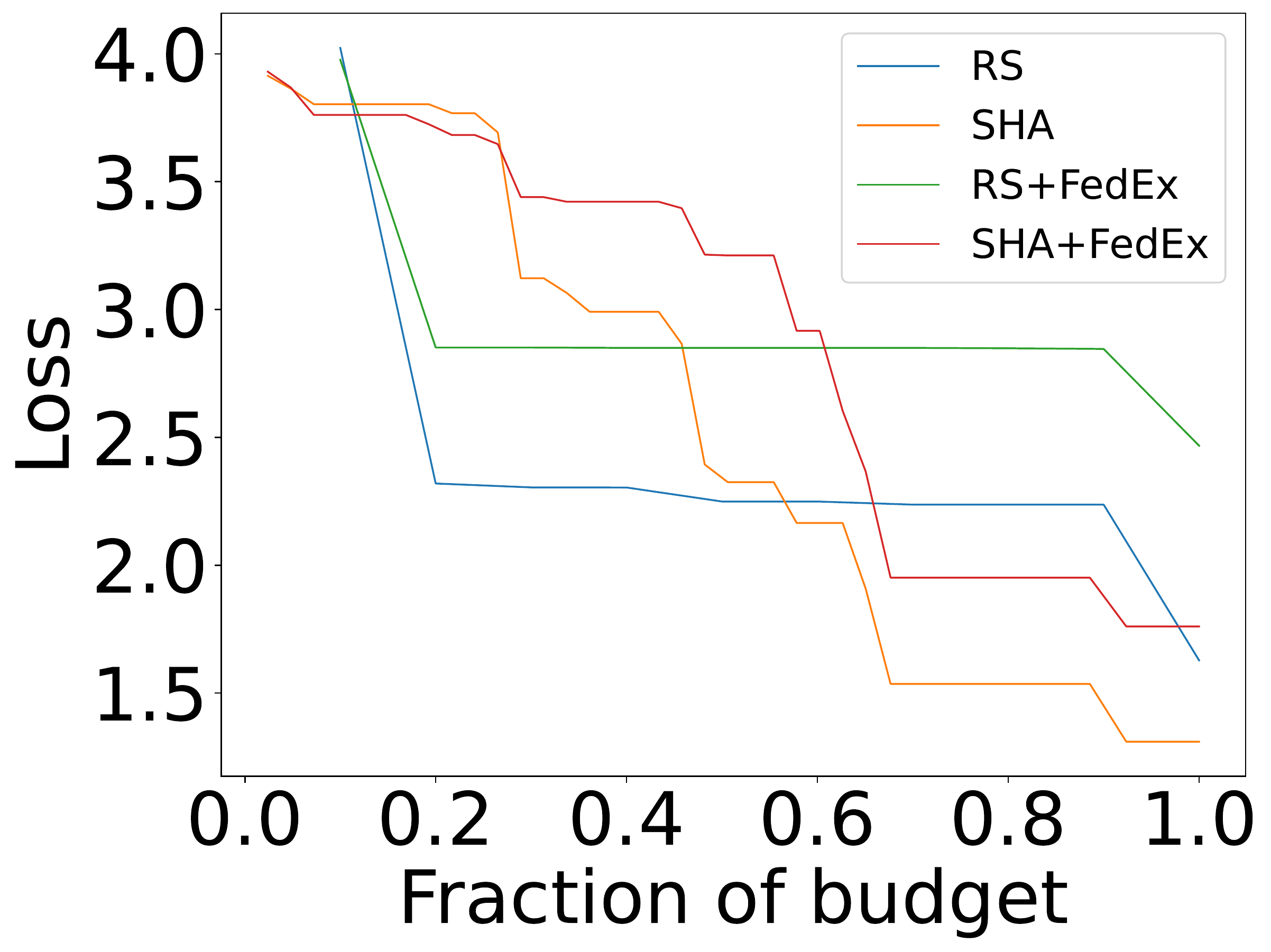}
\end{subfigure}
\begin{subfigure}{0.45\linewidth}
	\centering
	\includegraphics[width=0.98\linewidth]{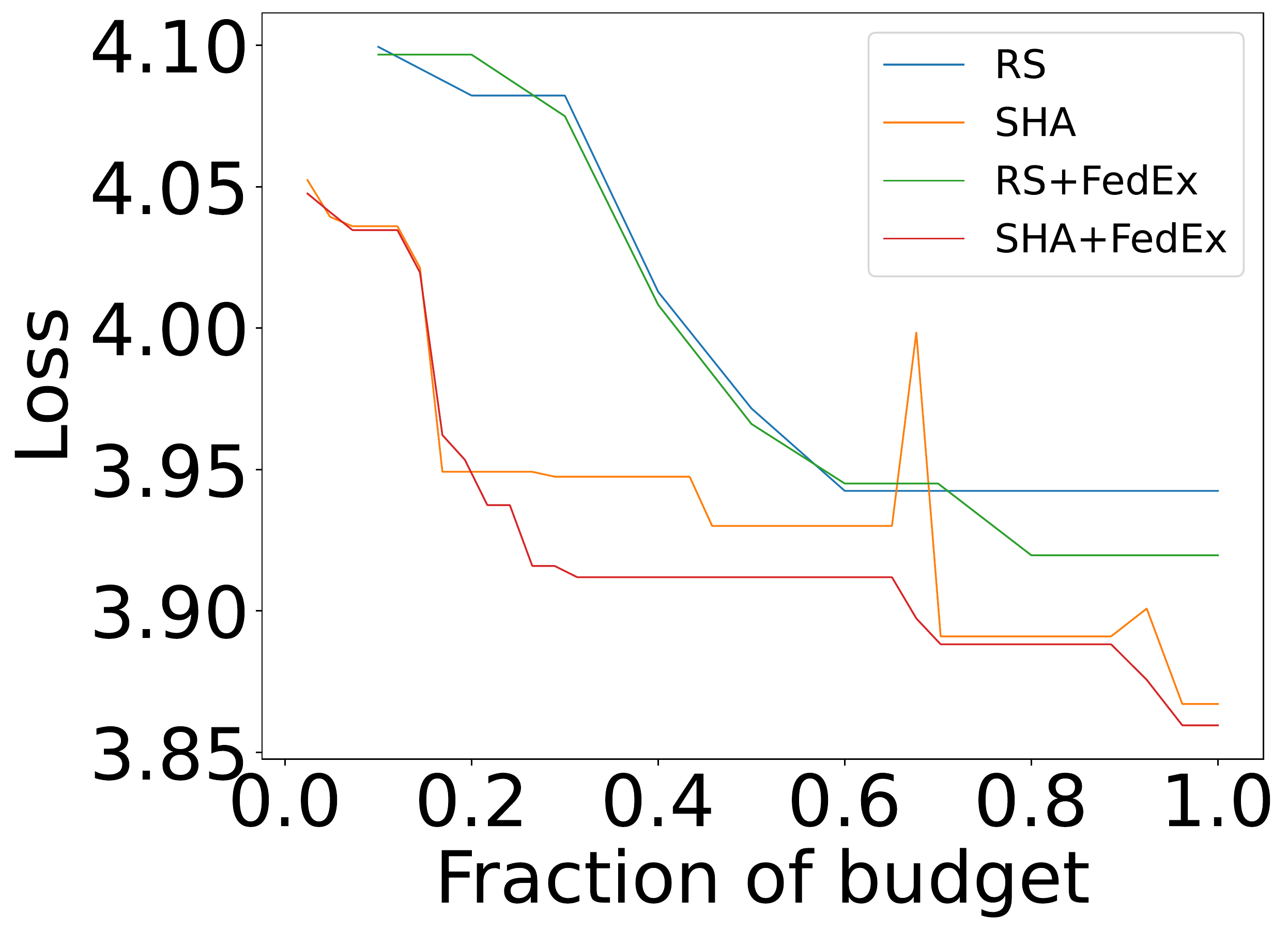}
\end{subfigure}
\captionof{figure}{Mean validation loss over time. \\  \textbf{Left}: FedAvg. \textbf{Right}: FedOPT.}
\label{fig:exp2}
\end{minipage}
\begin{minipage}[b]{0.4\linewidth}
\centering
\begin{tabular}{ll}
    \toprule
                                    Methods    & Test Accuracy \\ \midrule
    \textit{RS}        & 67.14 ± 8.46           \\
    \textit{SHA}       & 75.15 ± 3.44           \\
    \textit{RS+FedEx}  & 71.25 ± 8.79           \\
    \textit{SHA+FedEx} & \textbf{77.46 ± 1.78}           \\ \bottomrule
    \end{tabular}%
    \caption{Evaluation about the searched configurations: Mean test accuracy (\%) ± standard deviation.}
    \label{tab:exp2}
\end{minipage}\hfill
\end{table}

\subsection{Studies about the New Fidelity}
\label{subsec:exp3}
\begin{figure}[htbp]
	\centering
	\begin{subfigure}{0.36\linewidth}
		\centering
		\includegraphics[width=0.9\linewidth]{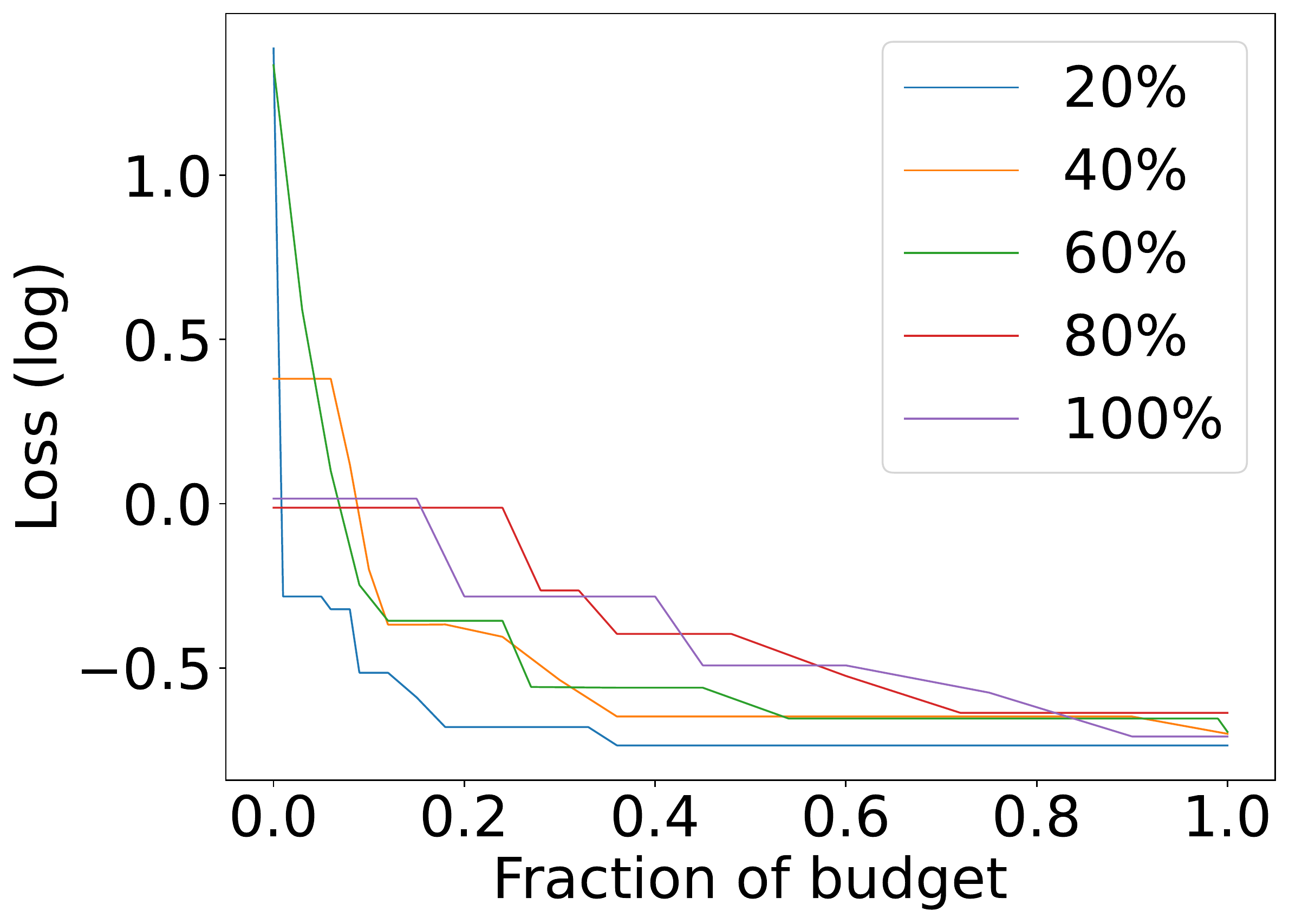}
		\caption{With bad network status}
		\label{fig:exp3_slow}
	\end{subfigure}
	\begin{subfigure}{0.36\linewidth}
		\centering
		\includegraphics[width=0.9\linewidth]{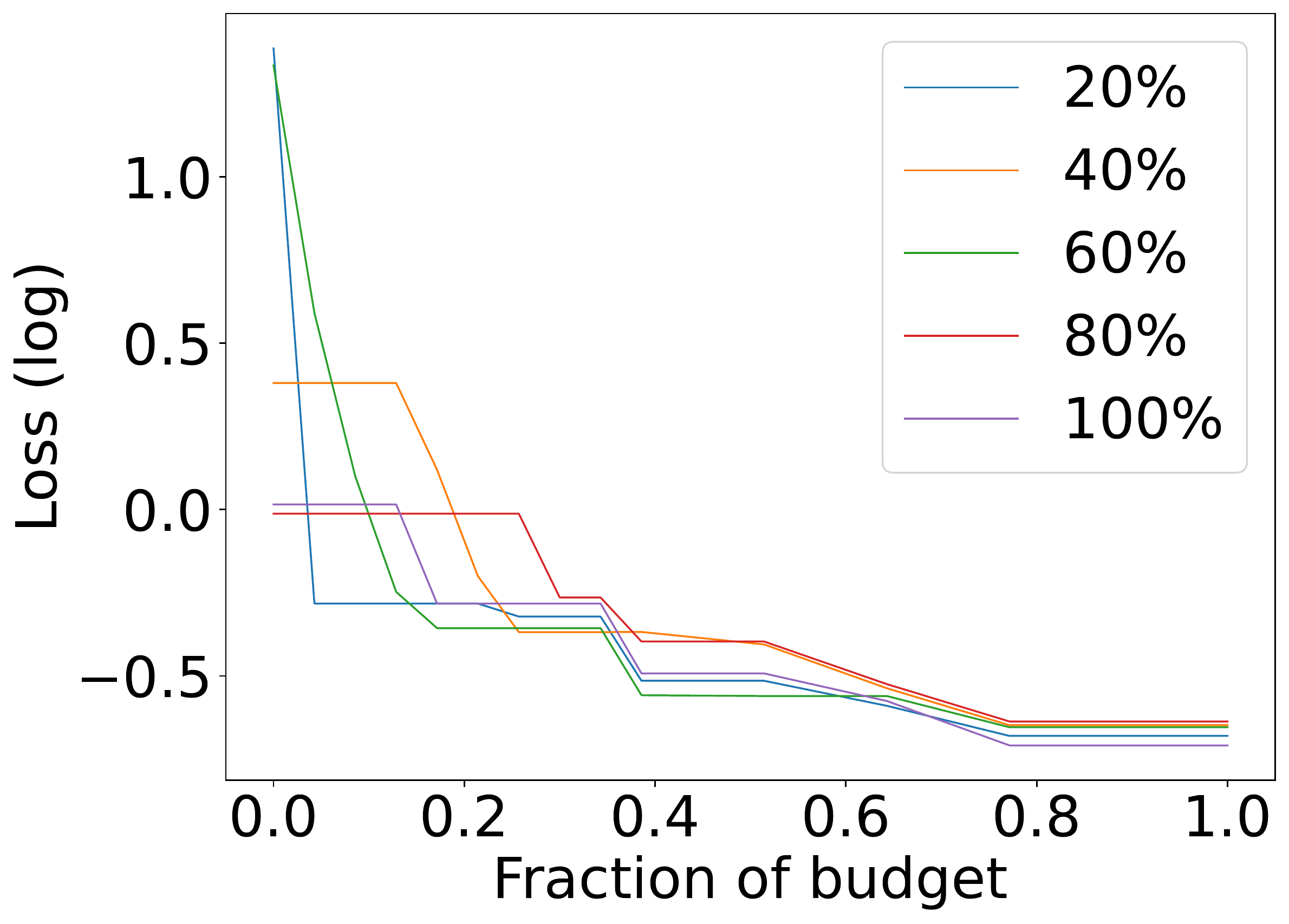}
		\caption{With good network status}
		\label{fig:exp3_quick}
	\end{subfigure}
	\vspace{-0.1in}
	\caption{Performances of different \textit{sample\_rate} under different system conditions.}
	\vspace{-0.3in}
	\label{fig:exp3}
\end{figure}

We simulate distinct system conditions by specifying different parameters for our system model. Then we show the performances of \textit{HB} with varying \textit{sample\_rates} in Figure~\ref{fig:exp3}, where which \textit{sample\_rate} is in favor depends on the system condition. Such a phenomenon supports the idea of pursuing a more economic accuracy-efficiency trade-off by balancing \textit{sample\_rate} with \textit{\#rounds}, w.r.t. the system condition. More details about this experiment are deferred to Appendix~\ref{sec:new_fidelity}.
\section{Conclusion and Future Work}
\vspace{-0.05in}
\label{sec:conclusion}
In this paper, we first identify the uniqueness of \subj, which we ascribe to the distributed nature of FL and its heterogeneous clients. This uniqueness prevents \subj research from leveraging existing HPO benchmarks, which has led to inconsistent comparisons between some recently proposed methods. Hence, we suggest and implement a comprehensive, efficient, and extensible benchmark suite, \ours. We further conduct extensive HPO experiments on \ours, validating its correctness and applicability to comparing traditional and federated HPO methods. We have open-sourced \ours with an Apache-2.0 license and will actively maintain it in the future. We believe \ours can serve as the stepping stone to developing reproducible \subj works, which is indispensable for such a nascent direction.

As mentioned in Section~\ref{subsec:comprehensive}, tasks other than federated supervised learning will be incorporated. At the same time, we aim to extend \ours to include different FL settings, e.g., HPO for vertical FL~\cite{flora}. Another issue the current version has not touched on is the risk of privacy leakage caused by HPO methods~\cite{adaptivelr}, which we should provide related metrics and testbeds in the future.

\bibliography{neurips_data_2022.bib}
\bibliographystyle{neurips}

\clearpage
\appendix
\section{Maintenance of \ours}
\label{sec:app}
In this section, we present our plan for maintaining \ours following~\cite{hpobench}.
\begin{itemize}
    \item \textbf{Who is maintaining the benchmarking library?} \ours is developed and maintained by Data Analytics and Intelligence Lab (DAIL) of DAMO Academy.
    \item \textbf{How can the maintainer of the dataset be contacted (e.g., email address)?} Users can reach out the maintainer via creating issues on Github repository at \url{https://github.com/alibaba/FederatedScope} with \ours label.
    \item \textbf{Is there an erratum?} No.
    \item \textbf{Will the benchmarking library be updated?} Yes, as we discussed in Section~\ref{sec:conclusion}, we will add more FedHPO problems and introduce more FL tasks to existing benchmark. We will track updates and Github release on the \href{https://federatedscope.io/}{website} and README. In addition, we will fix potential issues regularly.
    \item \textbf{Will older versions of the benchmarking library continue to be supported/hosted/maintained?} All older versions are available and maintained by the Github release, but limited support will be provided for older versions. Containers will be versioned and available via AliyunOSS.
    \item \textbf{If others want to extend/augment/build on/contribute to the dataset, is there a mechanism for them to do so?} Any contribution is welcome, and all commits to \ours must follow the guidance and regulations at \url{https://federatedscope.io/docs/contributor/}.
\end{itemize}

\section{HPO Methods}
\label{sec:optimizers}
As shown in Table~\ref{tab:hpo_optimizers}, we provide an overview of the optimizers (i.e., HPO methods) we use in this paper.

\begin{table}[htbp]
\centering
\caption{Overview of the optimizers from widely adopted libraries.}
\label{tab:hpo_optimizers}
\begin{tabular}{lllll}
\toprule
Name                & Model & Packages   & version \\ \midrule
\textit{RS}~\cite{rs}     & -     & \href{https://github.com/automl/HpBandSter}{\textit{HPBandster}} & 0.7.4          \\
$\textit{BO}_{\textit{GP}}$~\cite{bogp1,bogp2}            & \textit{GP}    & \href{https://github.com/automl/SMAC3}{\textit{SMAC3}}      & 1.3.3      \\
$\textit{BO}_{\textit{RF}}$~\cite{borf}           & \textit{RF}    & \href{https://github.com/automl/SMAC3}{\textit{SMAC3}}      & 1.3.3         \\
$\textit{BO}_{\textit{KDE}}$~\cite{kde}          & \textit{KDE}   & \href{https://github.com/automl/HpBandSter}{\textit{HPBandster}} & 0.7.4         \\
\textit{DE}~\cite{de1,de2}                & -     & \href{https://github.com/automl/DEHB}{\textit{DEHB}}       & git commit            \\
\hline
\textit{HB}~\cite{hyperband}         & -     & \href{https://github.com/automl/HpBandSter}{\textit{HPBandster}} & 0.7.4           \\
\textit{BOHB}~\cite{bohb}              & \textit{KDE}   & \href{https://github.com/automl/HpBandSter}{\textit{HPBandster}} & 0.7.4           \\
\textit{DEHB}~\cite{dehb}              & -     & \href{https://github.com/automl/DEHB}{\textit{DEHB}}       & git commit          \\
$\textit{TPE}_{\textit{MD}}$~\cite{optuna} & \textit{TPE}   & \href{https://github.com/optuna/optuna}{\textit{Optuna}}     & 2.10.0            \\
$\textit{TPE}_{\textit{HB}}$~\cite{optuna} & \textit{TPE}   & \href{https://github.com/optuna/optuna}{\textit{Optuna}}     & 2.10.0            \\ \bottomrule
\end{tabular}%
\end{table}

\subsection{Black-box Optimizers}
\label{subsec:bbo}
\noindent\textbf{\textit{RS}} (\textit{Random search}) is a priori-free HPO method, i.e., each step of the search does not exploit the already explored configuration. The random search outperforms the grid search within a small fraction of the computation time.

\noindent\textbf{$\textit{BO}_{\textit{GP}}$} is a Bayesian optimization with a Gaussian process model. $\textit{BO}_{\textit{GP}}$ uses a Matérn kernel for continuous hyperparameters, and a hamming kernel for categorical hyperparameters. In addition, the acquisition function is expected improvement (EI).

\noindent\textbf{$\textit{BO}_{\textit{RF}}$} is a Bayesian optimization with a random forest model. We set the hyperparameters of the random forest as follows: the number of trees is 10, the max depth of each tree is 20, and we use the default setting of the minimal samples split, which is 3.

\noindent\textbf{$\textit{BO}_{\textit{KDE}}$} is a Bayesian optimization with kernel density estimators (KDE), which is used in \textit{BOHB}~\cite{bohb}. It models objective function as $\Pr(x\mid y_{\text{good}})$ and $\Pr(x\mid y_{\text{bad}})$. We set the hyperparameters for $\textit{BO}_{\textit{KDE}}$ as follows: the number of samples to optimize EI is 64, and $1/3$ of purely random configurations are sampled from the prior without the model; the bandwidth factor is 3 to encourage diversity, and the minimum bandwidth is 1e-3 to keep diversity.

\noindent\textbf{\textit{DE}} uses the evolutionary search approach of Differential Evolution. We set the mutation strategy to \textit{rand1} and the binomial crossover strategy to \textit{bin}~\footnote{Please refer to \url{https://github.com/automl/DEHB/blob/master/README.md} for details.}. In addition, we use the default settings for the other hyperparameters of \textit{DE}, where the mutation factor is 0.5, crossover probability is 0.5, and the population size is 20.

\subsection{Multi-fidelity Optimizers}
\label{subsec:mf}
\noindent\textbf{\textit{HB}} (\textit{Hyperband}) is an extension on top of successive halving algorithms for the pure-exploration nonstochastic infinite-armed bandit problem. Hyperband makes a trade-off between the number of hyperparameter configurations and the budget allocated to each hyperparameter configuration. We set $\eta$ to 3, which means only a fraction of $1/\eta$ of hyperparameter configurations goes to the next round.

\noindent\textbf{\textit{BOHB}} combines \textit{HB} with the guidance and guarantees of convergence of Bayesian optimization with kernel density estimators. We set the hyperparameter of the \textit{BO} components and the \textit{HB} components of \textit{BOHB} to be the same as $\textit{BO}_{\textit{KDE}}$ and \textit{HB} described above, respectively.

\noindent\textbf{\textit{DEHB}} combines the advantages of the bandit-based method \textit{HB} and the evolutionary search approach of \textit{DE}. The hyperparameter of \textit{DE} components and \textit{BO} components are set to be exactly the same as \textit{DE} and \textit{HB} described above, respectively.

\noindent\textbf{$\textit{TPE}_{\textit{MD}}$} is implemented in \textit{Optuna} and uses Tree-structured Parzen Estimator (\textit{TPE}) as a sampling algorithm, where on each trial, TPE fits two Gaussian Mixture models for each hyperparameter. One is to the set of hyperparameters with the best performance, and the other is to the remaining hyperparameters. In addition, it uses the median stopping rule as a pruner, which means that it will prune if the trial’s best intermediate result is worse than the median (\textit{MD}) of intermediate results of previous trials at the same step. We use the default settings for both \textit{TPE} and \textit{MD}.

\noindent\textbf{$\textit{TPE}_{\textit{HB}}$} is similar to $\textit{TPE}_{\textit{MD}}$ described above, which uses \textit{TPE} as a sampling algorithm and \textit{HB} as pruner. We set the reduction factor to 3 for \textit{HB} pruner, and all other settings use the default ones.

\section{Datasets}
\label{sec:datasets}
As shown in Table~\ref{tab:hpo_datasets}, we provide a detailed description of the datasets we use in current \ours. Following FS~\cite{fs}, FS-G~\cite{fsg}, and HPOBench~\cite{hpobench}, we use 14 FL datasets from 4 domains, including CV, NLP, graph, and tabular.
Some of them are inherently real-world FL datasets, while others are simulated FL datasets split by the splitter modules of FS.
Notably, the name of datasets from OpenML is the ID of the corresponding task.

\begin{table}[htbp]
\centering
\caption{Statistics of the datasets used in current \ours.}
\label{tab:hpo_datasets}
\begin{tabular}{lcccccc}
\toprule
Name                                        & \#Client & Subsample & \#Instance & \#Class & Split by  \\ \midrule
FMNIST                                      & \num{3550}   & 5\%     & \num{805263}        & 62         & Writer         \\
CIFAR-10                                    & 5       & 100\%       & \num{60000}        & 10         & LDA         \\ \hline
CoLA                                        & 5       & 100\%       & \num{10657}       & 2         & LDA         \\
SST-2                                       & 5       & 100\%       & \num{70042}       & 2         & LDA         \\ \hline
Cora                                        & 5       & 100\%       & \num{2708}       & 7         & Community        \\
CiteSeer                                    & 5       & 100\%       & \num{4230}       & 6        & Community        \\
PubMed                                      & 5       & 100\%       & \num{19717}       & 5        & Community        \\ \hline
$31_{OpenML}$                                          & 5       & 100\%       &  \num{1000}         & 2       & LDA        \\
$53_{OpenML}$                                          & 5       & 100\%       & 846         & 4       & LDA         \\
$3917_{OpenML}$                                        & 5       & 100\%       & \num{2109}         & 2        & LDA         \\
$10101_{OpenML}$                                       & 5       & 100\%       & 748         & 2       & LDA         \\
$146818_{OpenML}$                                      & 5       & 100\%       & 690         & 2       & LDA         \\
$146821_{OpenML}$                                      & 5       & 100\%       & \num{1728}         & 4        & LDA         \\
$146822_{OpenML}$                                      & 5       & 100\%       & \num{2310}         & 7       & LDA        \\ 
\bottomrule
\end{tabular}%
\end{table}

\noindent\textbf{FEMNIST} is an FL image dataset from LEAF~\cite{leaf}, whose task is image classification. Following \cite{leaf}, we use a subsample of FEMNIST with 200 clients, which is round 5\%. And we use the default train/valid/test splits for each client, where the ratio is $60\%:20\%:20\%$.

\noindent\textbf{CIFAR-10}~\cite{cifar10} is from Tiny Images dataset and consists of \num{60000} $32\times32$ color images, whose task is image classification. We split images into 5 clients by latent dirichlet allocation (LDA) to produce statistical heterogeneity among these clients. We split the raw training set to training and validation sets with a ratio $4:1$, so that ratio of final train/valid/test splits is 66.7\%:16.67\%:16.67\%.

\noindent\textbf{SST-2} is a dataset from GLUE~\cite{glue} benchmark, whose task is binary sentiment classification for sentences. We also split these sentences into 5 clients by LDA. In addition, we use the official train/valid/test splits for SST-2.

\noindent\textbf{CoLA} is also a dataset from GLUE benchmark, whose task is binary classification for sentences---whether it is a grammatical English sentence. We exactly follow the experimental setup in SST-2.

\noindent\textbf{Cora \& CiteSeer \& PubMed}~{\cite{citation1,citation2}} are three widely adopted graph datasets, whose tasks are node classification. Following FS-G~\cite{fsg}, a community splitter is applied to each graph to generate five subgraphs for each client. We also split the nodes into train/valid/test sets, where the ratio is 60\%:20\%:20\%.

\noindent\textbf{Tabular datasets} are consist of 7 tabular datasets from OpenML~\cite{openml2}, whose task ids (name of source data) are 31 (\textbf{credit-g}), 53 (\textbf{vehicle}), 3917 (\textbf{kc1}), 10101 (\textbf{blood-transfusion-service-center}), 146818 (\textbf{Australian}), 146821 (\textbf{car}) and 146822 (\textbf{segment}). We split each dataset into 5 clients by LDA, respectively. In addition, we set the ratio of train/valid/test splits to 80\%:10\%:10\%.

\section{Proof of Proposition 1}
\label{sec:proof}
What we need to calculate is the expected maximum of i.i.d. exponential random variable.
Proposition~\ref{prop:1} states that, for $N$ exponential variables independently drawn from $\text{Exp}(\cdot|\frac{1}{c})$, the expectation is $\sum_{i=1}^{N}\frac{c}{i}$.
There are many ways to prove this useful proposition, and we provide a proof starting from studying the minimum of the exponential random variables.
\begin{proof}
According to the \href{https://en.wikipedia.org/wiki/Exponential_distribution#Distribution_of_the_minimum_of_exponential_random_variables}{Wikipedia page about exponential distribution}, the minimum of them obeys $\text{Exp}(\cdot|\frac{N}{c})$.
Denoting the $i$-th minimum of them by $T_i$, $T_1 \sim \text{Exp}(\cdot|\frac{N}{c})$ and $T_N$ is what we are interested in.
Meanwhile, it is well known that exponential distribution is memoryless, namely, $\Pr(X>s+t|X>s)=\Pr(X>t)$.
Thus, $T_2 - T_1$ obeys the same distribution as the minimum of $N-1$ such random variables, that is to say, $T_2 - T_1$ is a random variable drawn from $\text{Exp}(\cdot|\frac{N-1}{c})$.
Similarly, $T_{i+1} - T_i ~\text{Exp}(\cdot|\frac{N-i}{c}), i=1,\ldots,N-1$. Thus, we have:\\
\begin{equation}
    \mathbb{E}[T_N] = \mathbb{E}[T_1 + \sum_{i=1}^{N-1}(T_{i+1}-T_i)] =\frac{c}{N} + \sum_{i=1}^{N-1}\frac{c}{N-i} = \sum_{i=1}^{N}\frac{c}{i},
\end{equation}
which concludes this proof.
\end{proof}

\section{Details of the Study about New Fidelity}
\label{sec:new_fidelity}
We compare the performance of \textit{HB} with different \textit{sample\_rate}s to learn a 2-layer CNN with 2,048 hidden units on FEMNIST.
To simulate a system condition with bad network status, we set the upload bandwidth $B_{\text{up}}^{(\text{server})}$ and $B_{\text{up}}^{(\text{client})}$ to 0.25MB/second and the download bandwidth $B_{\text{down}^{(\text{client})}}$ to 0.75MB/second~\cite{fedoptimizationsurvey}. As for good network status, we set the upload bandwidth $B_{\text{up}}^{(\text{server})}$ and $B_{\text{up}}^{(\text{client})}$ to 0.25GB/second and the download bandwidth $B_{(\text{down})}^{(\text{server})}$ to 0.75GB/second.
In both cases, we fix the computation overhead so that it is negligible and significant, respectively.
As for the rest settings, we largely follow that in Section~\ref{subsec:exp1}.
In both cases, as shown in Figure~\ref{fig:exp3}, the severity of the straggler issue raises with the \textit{sample\_rate} increases.
As a limited time budget, the FL procedure with a lower \textit{sample\_rate} achieves a more competitive result in the bad network status. In comparison, that with a higher \textit{sample\_rate} achieves a more competitive result in the good network status.
In conclusion, this study suggests a best practice that we should carefully balance these two fidelity dimensions, \textit{\#round} and \textit{sample\_rate}, w.r.t. the system condition. Better choices tend to achieve more economical accuracy-efficiency trade-offs for \subj.

\section{Details on \ours Benchmarks}
\label{sec:benchmarks}
\ours consists of five categories of benchmarks on the different datasets (see Appendix~\ref{sec:datasets}) with three modes.
In this part, we provide more details about how we construct the \subj problems provided by current \ours and the three modes to interact with them.

\begin{table}[htbp]
\centering
\caption{The search space of our benchmarks, where continuous search spaces are discretized into several bins under the tabular mode.}
\label{tab:hpo_space}
\begin{tabular}{ccccccc}
\toprule
Benchmark                  &  & Name                 & Type  & Log & \#Bins          & Range           \\ \midrule
\multirow{9}{*}{CNN}       & \multirow{5}{*}{Client} & batch\_size    & int   & $\times$   & -   & \{16, 32, 64\}  \\
                           &                         & weight\_decay  & float & $\times$ & 4 & {[}0, 0.001{]}  \\
                           &                         & dropout        & float & $\times$ & 2     & {[}0, 0.5{]}    \\
                           &                         & step\_size     & int   & $\times$ & 4      & {[}1, 4{]}      \\
                           &                         & learning\_rate & float & $\checkmark$ & 10 & {[}0.01, 1.0{]} \\ \cline{2-7} 
                           & \multirow{2}{*}{Server} & momentum   & float & $\times$ & 2     & {[}0.0, 0.9{]}  \\ 
                           &                         & learning\_rate   & float & $\times$ & 3     & {[}0.1, 1.0{]}  \\ \cline{2-7} 
                           & \multirow{2}{*}{Fidelity} & sample\_rate   & float & $\times$ & 5     & {[}0.2, 1.0{]}  \\ 
                           &                         & round   & int & $\times$ & 250     & {[}1, 500{]}  \\ \hline
\multirow{9}{*}{BERT} & \multirow{5}{*}{Client} & batch\_size    & int   & $\times$ & -     & \{8, 16, 32, 64, 128\}  \\
                           &                         & weight\_decay  & float & $\times$ & 4 & {[}0, 0.001{]}  \\
                           &                         & dropout        & float & $\times$ & 2     & {[}0, 0.5{]}    \\
                           &                         & step\_size     & int   & $\times$ & 4     & {[}1, 4{]}     \\
                           &                         & learning\_rate & float & $\checkmark$& 10  & {[}0.01, 1.0{]} \\ \cline{2-7} 
                           & \multirow{2}{*}{Server} & momentum   & float & $\times$ & 2     & {[}0.0, 0.9{]}  \\ 
                           &                         & learning\_rate   & float & $\times$ & 3     & {[}0.1, 1.0{]}  \\ \cline{2-7} 
                           & \multirow{2}{*}{Fidelity} & sample\_rate   & float & $\times$ & 5     & {[}0.2, 1.0{]}  \\ 
                           &                         & round   & int & $\times$ & 40     & {[}1, 40{]}  \\ \hline
\multirow{8}{*}{GNN} & \multirow{4}{*}{Client} & weight\_decay & float & $\times$ & 4 & {[}0, 0.001{]} \\
                           &                         & dropout        & float & $\times$ & 2     & {[}0, 0.5{]}    \\
                           &                         & step\_size     & int   & $\times$ & 8     & {[}1, 8{]}      \\
                           &                         & learning\_rate & float & $\checkmark$ & 10 & {[}0.01, 1.0{]} \\ \cline{2-7} 
                          & \multirow{2}{*}{Server} & momentum   & float & $\times$ & 2     & {[}0.0, 0.9{]}  \\ 
                           &                         & learning\_rate   & float & $\times$ & 3     & {[}0.1, 1.0{]}  \\ \cline{2-7} 
                           & \multirow{2}{*}{Fidelity} & sample\_rate   & float & $\times$ & 5     & {[}0.2, 1.0{]}  \\ 
                           &                         & round   & int & $\times$ & 500     & {[}1, 500{]}  \\ \hline
\multirow{8}{*}{LR}        & \multirow{4}{*}{Client} & batch\_size    & int   & $\checkmark$ & 7  & {[}4, 256{]}  \\
                           &                         & weight\_decay  & float & $\times$ & 4 & {[}0, 0.001{]}  \\
                           &                         & step\_size     & int   & $\times$ & 4     & {[}1, 4{]}      \\
                           &                         & learning\_rate & float & $\checkmark$ & 6 & {[}0.00001, 1.0{]} \\ \cline{2-7} 
                           & \multirow{2}{*}{Server} & momentum   & float & $\times$  & 2    & {[}0.0, 0.9{]}  \\ 
                           &                         & learning\_rate   & float & $\times$ & 3     & {[}0.1, 1.0{]}  \\ \cline{2-7} 
                           & \multirow{2}{*}{Fidelity} & sample\_rate   & float & $\times$ & 5     & {[}0.2, 1.0{]}  \\ 
                           &                         & round   & int & $\times$ & 500     & {[}1, 500{]}  \\ \hline
\multirow{10}{*}{MLP}       & \multirow{6}{*}{Client} & batch\_size    & int   & $\checkmark$ & 7 & {[}4, 256{]}  \\
                           &                         & weight\_decay  & float & $\times$ & 4 & {[}0, 0.001{]}  \\
                           &                         & step\_size     & int   & $\times$     & 4 & {[}1, 4{]}      \\
                           &                         & learning\_rate & float & $\checkmark$ & 6 & {[}0.00001, 1.0{]} \\
                           &                         & depth          & int   & $\times$     & 3 & {[}1, 3{]} \\
                           &                         & width          & int   & $\checkmark$ & 7 & {[}16, 1024{]} \\
                           \cline{2-7} 
                           & \multirow{2}{*}{Server} & momentum   & float & $\times$     & 2 & {[}0.0, 0.9{]}  \\ 
                           &                         & learning\_rate   & float & $\times$     & 3 & {[}0.1, 1.0{]}  \\ \cline{2-7} 
                           & \multirow{2}{*}{Fidelity} & sample\_rate   & float & $\times$     & 5 & {[}0.2, 1.0{]}  \\ 
                           &                         & round   & int & $\times$     & 500 & {[}1, 500{]}  \\ 
\bottomrule
\end{tabular}%
\end{table}

\subsection{Category}
\label{subsec:bench_category}
We categorize our benchmarks by model types. Each benchmark is designed to solve specific FL HPO problems on its data domain, wherein CNN benchmark on CV, BERT benchmark on NLP, GNN benchmark on the graphs, and LR \& MLP benchmark on tabular data. All benchmarks have several hyperparameters on configuration space and two on fidelity space, namely sample rate of FL and FL round. And the benchmarks support several FL algorithms, such as FedAvg and FedOPT.

\noindent\textbf{CNN} benchmark learns a two-layer CNN with 2048 hidden units on FEMNIST and 128 hidden units on CIFAR-10 with five hyperparameters on configuration space that tune the batch size of the dataloader, the weight decay, the learning rate, the dropout of the CNN models, and the step size of local training round in client each FL communication round. The tabular and surrogate mode of the CNN benchmark only supports FedAvg due to our limitations in computing resources for now, but we will update \ours with more results as soon as possible.

\noindent\textbf{BERT} benchmark fine-tunes a pre-trained language model, BERT-Tiny, which has two layers and 128 hidden units, on CoLA and SST-2. The BERT benchmark also has five hyperparameters on configuration space which is the same as CNN benchmark. In addition, the BERT benchmark support FedAvg and FedOPT with all three mode.

\noindent\textbf{GNN} benchmark learns a two-layer GCN with 64 hidden units on Cora, CiteSeer and PubMed. The GNN benchmark has four on hyperparameters configuration space that tune the weight decay, the learning rate, the dropout of the GNN models, and the step size of local training round in client each FL communication round. The GNN benchmark support FedAvg and FedOPT with all three mode.

\noindent\textbf{LR} benchmark learns a lr on seven tasks from OpenML, see Appendix~\ref{sec:datasets} for details. The LR benchmark has four on hyperparameters configuration space that tune the batch size of the dataloader, the weight decay, the learning rate, and the step size of local training round in client each FL communication round. The LR benchmark support FedAvg and FedOPT with all three mode.

\noindent\textbf{MLP} benchmark's the vast majority of settings are the same as LR benchmark. But in particular, we add depth and width of the MLP to search space in terms of model architecture. The MLP benchmark also support FedAvg and FedOPT with all three mode.

\subsection{Mode}
\label{subsec:bench_mode}
Following HPOBench~\cite{hpobench}, \ours provides three different modes for function evaluation: the tabular mode, the surrogate mode, and the raw mode. The valid input hyperparameter configurations and the speed of acquiring feedback vary from mode to mode. Users can choose the desired mode according to the purposes of their experiments.

\noindent\textbf{Tabular mode}.
The idea is to evaluate the performance of many different hyperparameter configurations in advance so that users can acquire their results immediately. To this end, we firstly attain a grid search space from our original search space (see Table~\ref{tab:hpo_space}). For hyperparameters whose original search space is discrete, we just preserve its original one. As for continuous ones, we discretize them into several bins (see Table~\ref{tab:hpo_space} for details). Then we evaluate each configuration in the grid search space. To ensure that the results are reproducible, we execute the FL procedure in the Docker container environment. Evaluation for each specific configuration is repeated three times with different random seeds. We record the averaged loss, accuracy, and F1 score for each train/valid/test split. Users can choose the desired metric as the output of the black-box function via \ours's APIs.
We plan to augment these tables (i.e., with a denser grid) in the next version of \ours.

\noindent\textbf{Surrogate mode}.
As the valid input configurations of tabular mode are those on the grid, the tabular mode cannot be used for comparing the optimization of real-valued hyperparameters. Thus, we train a surrogate model on the lookup table of tabular mode, which enables us to evaluate arbitrary configuration by model inference. Specifically, we conduct 10-fold cross-validation to train and evaluate random forest models (implemented in scikit-learn~\cite{sklearn}) on the tabular data. Meanwhile, we search for suitable hyperparameters for the random forest models with the number of trees in \{10, 20\} and the max depth in \{10, 15, 20\}. The mean absolute error (MAE) of the surrogate model w.r.t. the true value is within an acceptable threshold. For example, in predicting the true average loss on the CNN benchmark, the surrogate model has a training error of 0.0061 and a testing error of 0.0077. In addition to the off-the-shelf surrogate models we provide, \ours offers tools for users to build brand-new surrogate models.
Meanwhile, we notice the recent successes of neural network-based surrogate, e.g., \href{https://github.com/slds-lmu/yahpo_gym}{YAHPO Gym}~\cite{yahpo}, and we will also try it in the next version of \ours.

\noindent\textbf{Raw mode}. Both of the above modes, although they can respond quickly, are limited to pre-designed search space.
Thus, we introduce raw mode to \ours, where user-defined search spaces are allowed.
Once \ours's APIs are called with specific hyperparameters, a containerized and standalone FL procedure (supported by FS) will be launched. It is worth noting that although we use standalone simulation to eliminate the communication cost, raw mode still consumes much more computation cost than tabular and surrogate modes.

\section{More Results}
\label{sec:more_results}
In this section, we show the detailed experimental results of the optimizers on \ours benchmarks under different modes. 
We firstly report the averaged best-seen validation loss, from which the mean rank over time for all optimizers can be deduced.
Due to time and computing resource constraints, we do not have a complete experimental result of the raw mode, which we will supplement as soon as possible.

\subsection{Tabular mode}
\begin{figure}[htbp]
	\centering

	\begin{subfigure}{0.25\linewidth}
		\centering
		\includegraphics[width=0.9\linewidth]{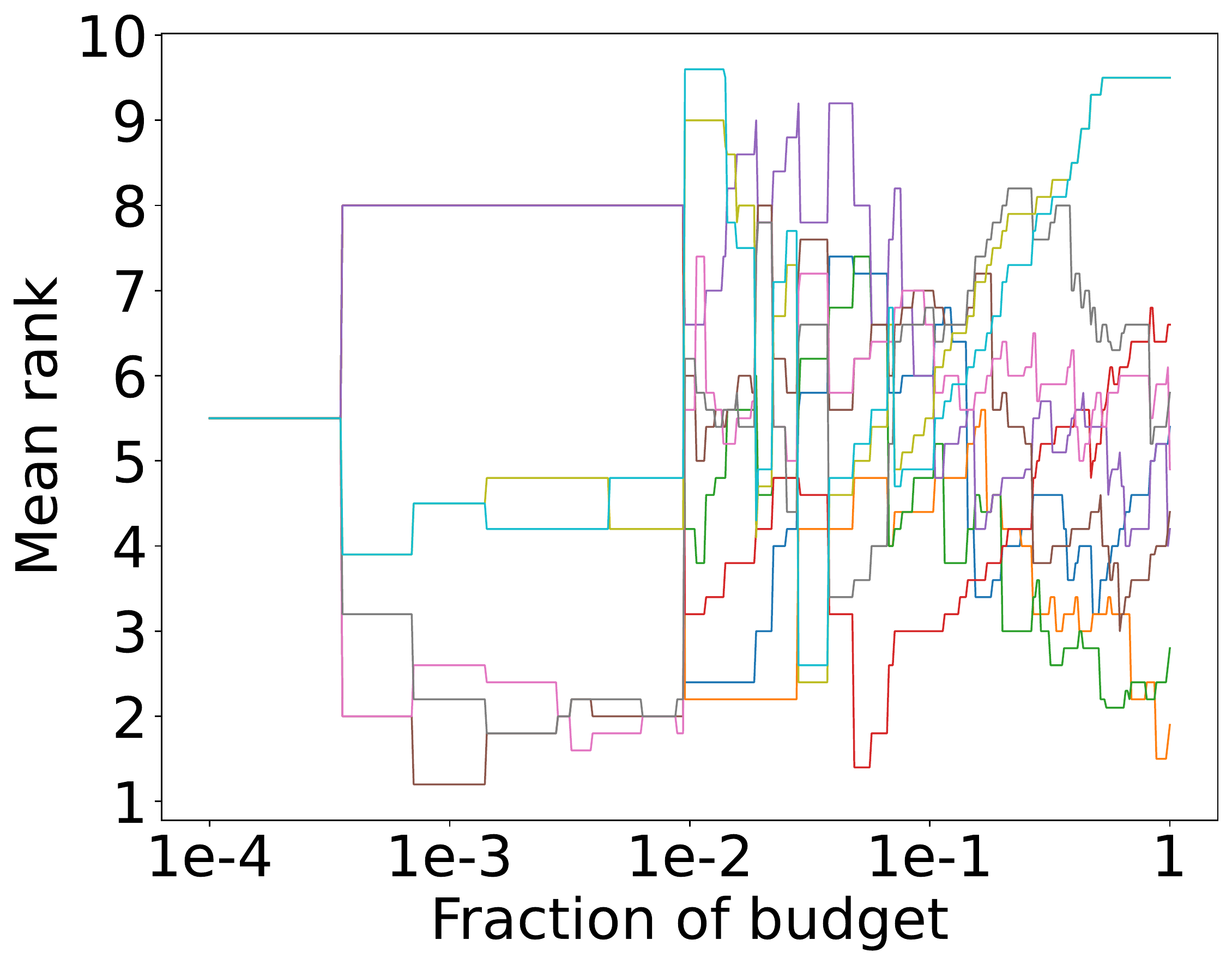}
		\caption{$\text{ALL}_{\text{FEMNIST}}$}
		\label{fig:All_FEMNIST_avg}
	\end{subfigure}
	\begin{subfigure}{0.25\linewidth}
		\centering
		\includegraphics[width=0.9\linewidth]{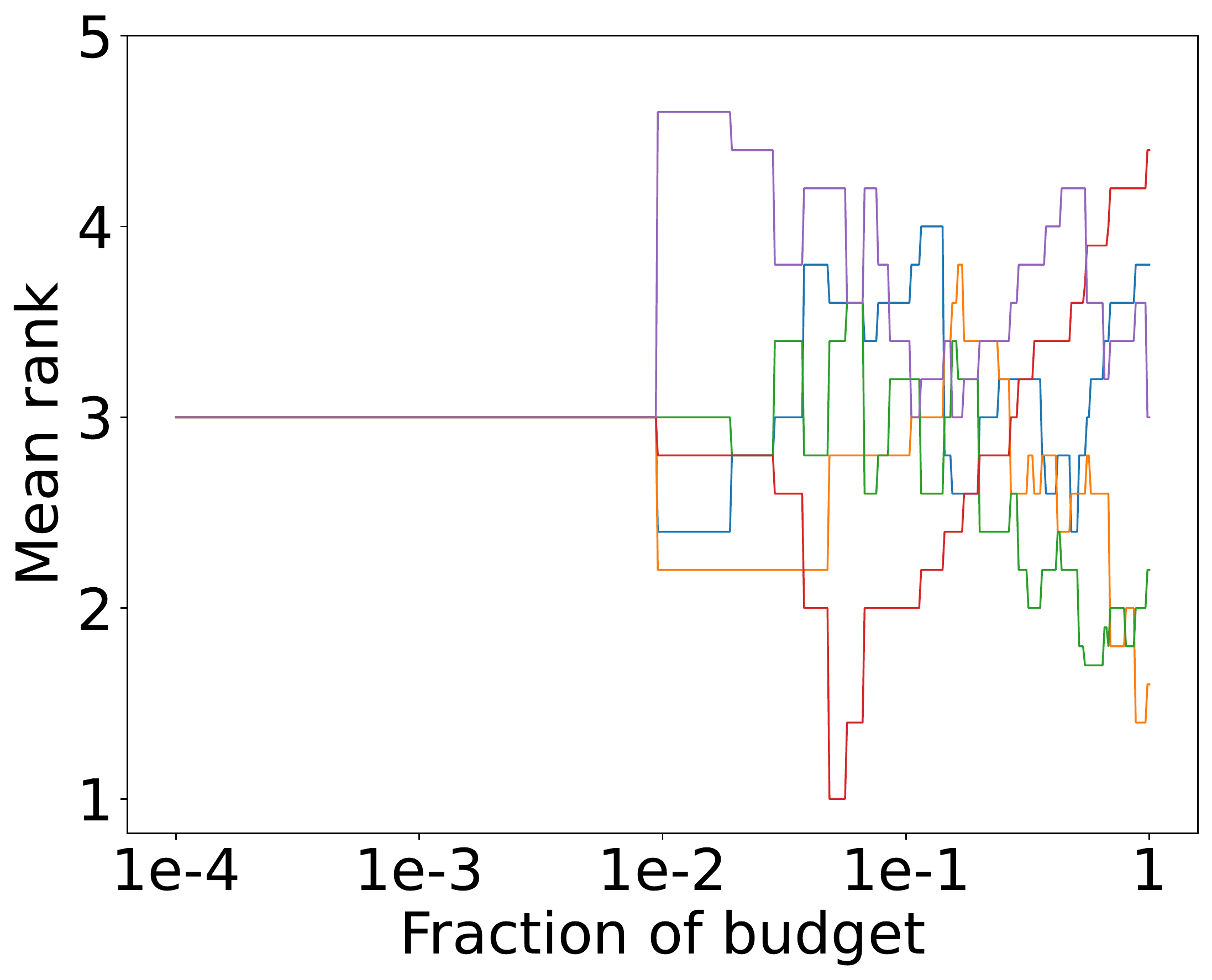}
		\caption{$\textit{BBO}_{\text{FEMNIST}}$}
		\label{fig:BBO_FEMNIST_avg}
	\end{subfigure}
	\begin{subfigure}{0.25\linewidth}
		\centering
		\includegraphics[width=0.9\linewidth]{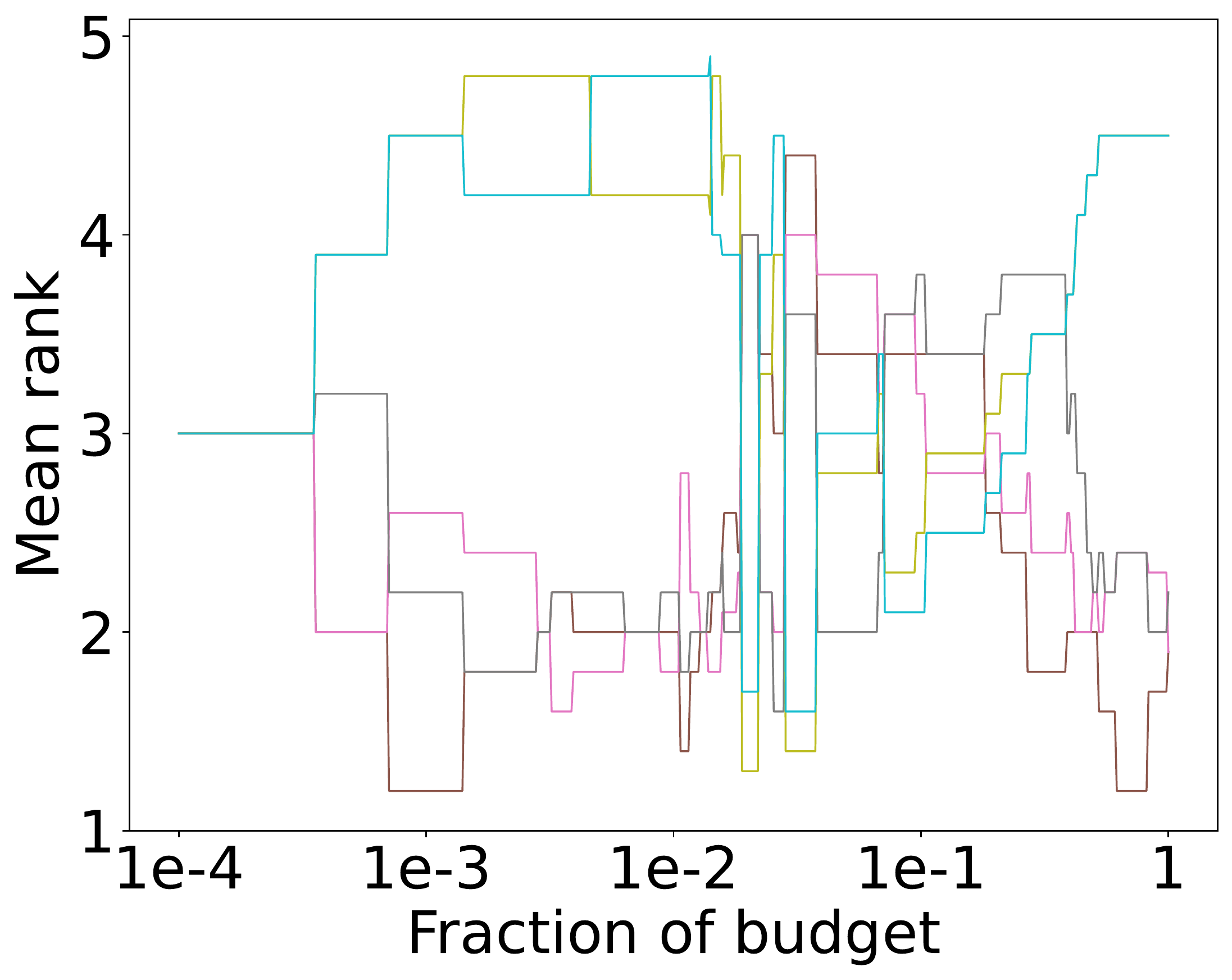}
		\caption{$\textit{MF}_{\text{FEMNIST}}$}
		\label{fig:MF_FEMNIST_avg}
	\end{subfigure}
	\centering
	\hspace*{1.2cm}\begin{subfigure}{1.0\linewidth}
		\centering
		\includegraphics[width=0.95\linewidth]{materials/legend_rank_new.pdf}
	\end{subfigure}
	\vspace{-0.1in}

	\caption{Mean rank over time on CNN benchmark (FedAvg).}
	\label{fig:entire_cnn_tabular_avg_rank}
\end{figure}

\begin{figure}[htbp]
	\centering
	\begin{subfigure}{0.25\linewidth}
		\centering
		\includegraphics[width=0.9\linewidth]{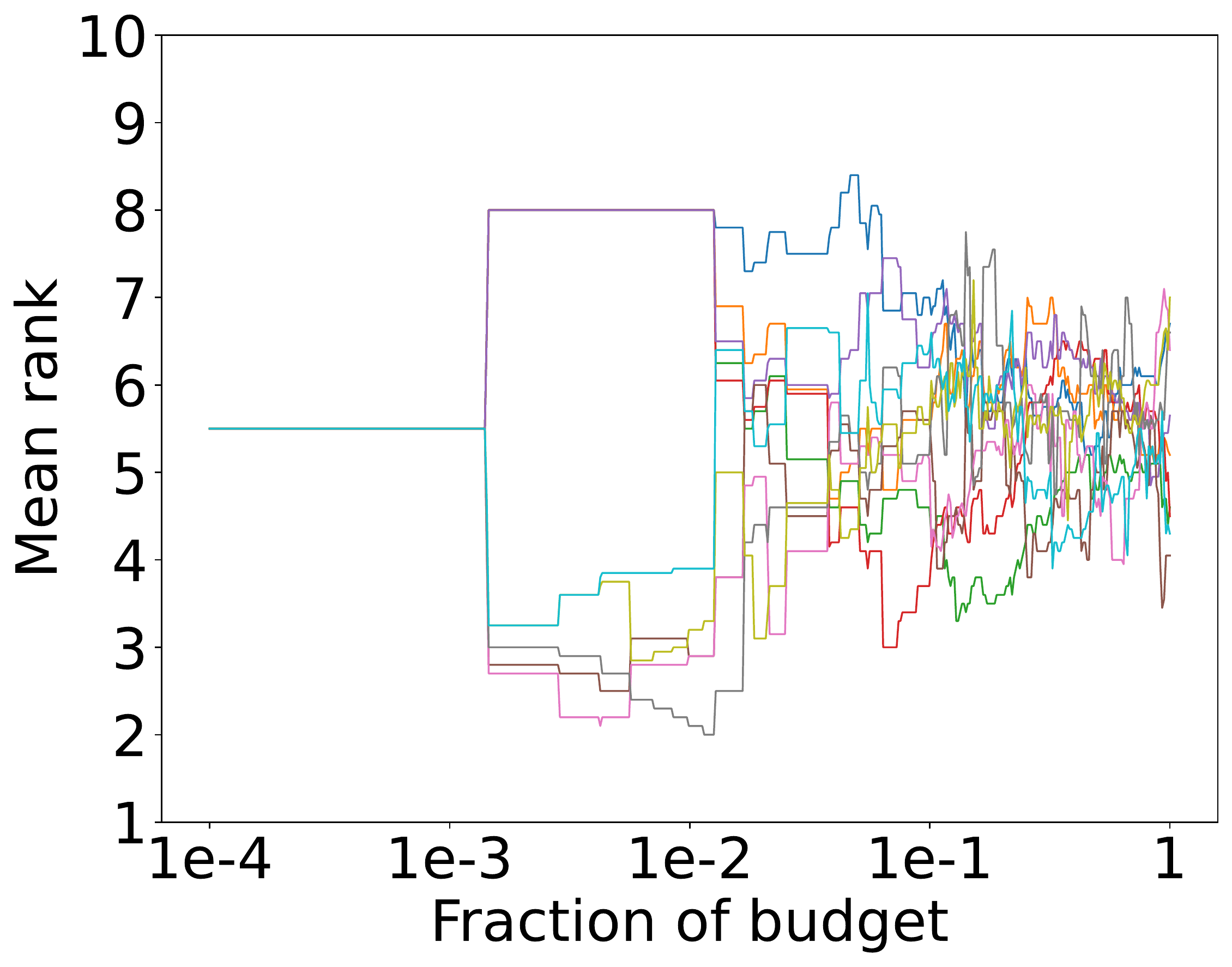}
		\caption{$\text{ALL}_\text{{BERT}}$}
		\label{fig:All_BERT_avg}
	\end{subfigure}
	\begin{subfigure}{0.25\linewidth}
		\centering
		\includegraphics[width=0.9\linewidth]{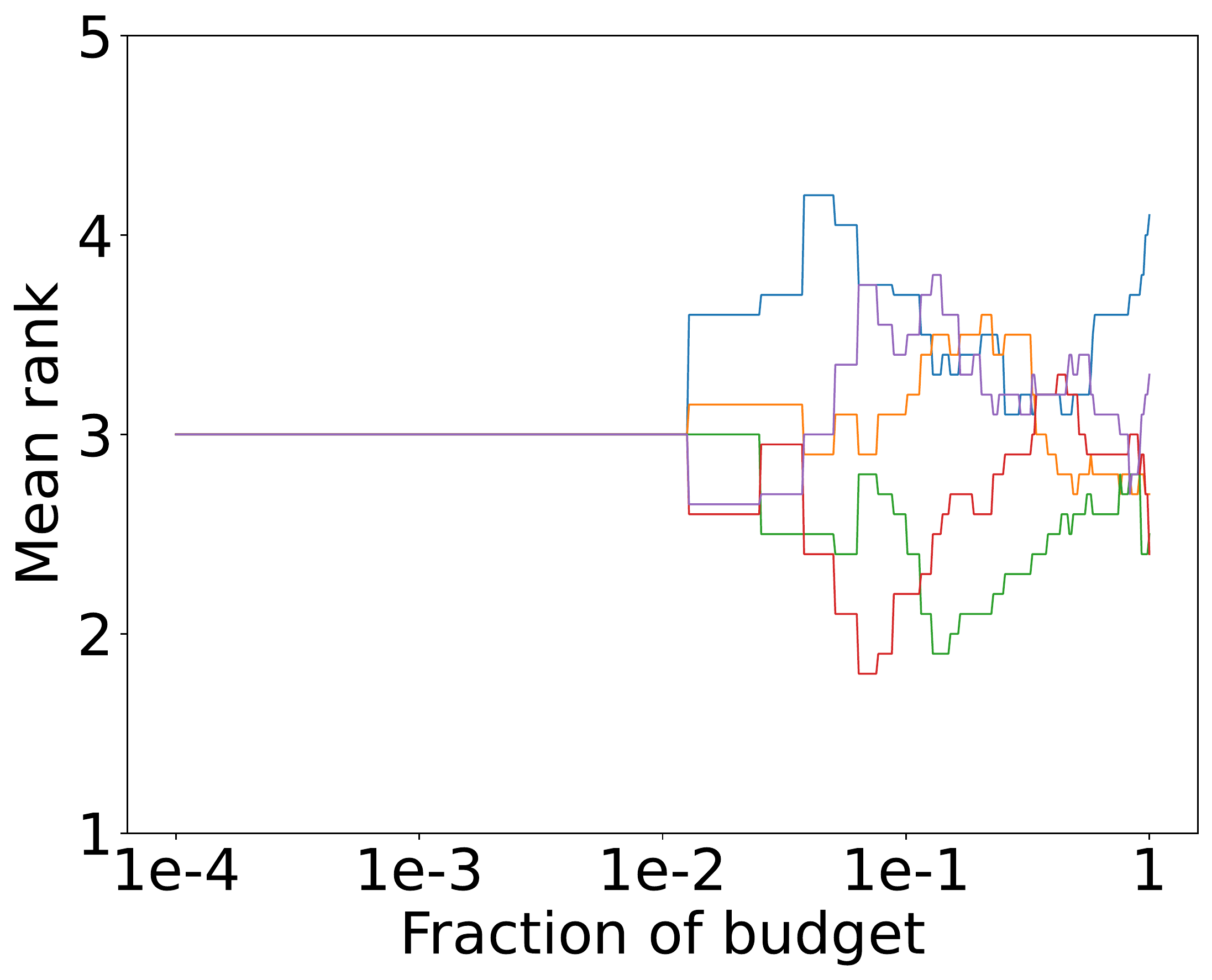}
		\caption{$\textit{BBO}_\text{{BERT}}$}
		\label{fig:BBO_BERT_avg}
	\end{subfigure}
	\begin{subfigure}{0.25\linewidth}
		\centering
		\includegraphics[width=0.9\linewidth]{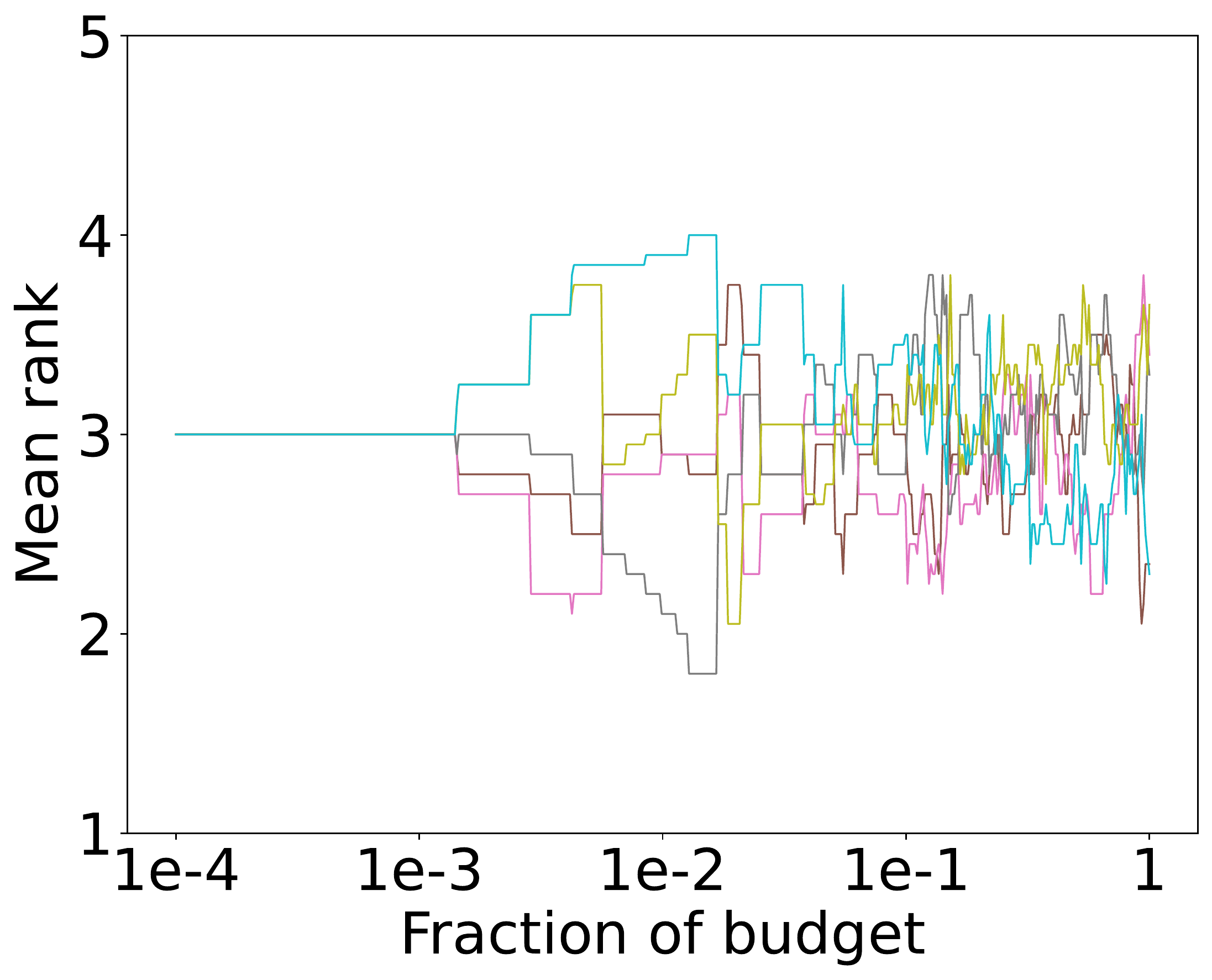}
		\caption{$\textit{MF}_\text{{BERT}}$}
		\label{fig:MF_BERT_avg}
	\end{subfigure}

	\begin{subfigure}{0.25\linewidth}
		\centering
		\includegraphics[width=0.9\linewidth]{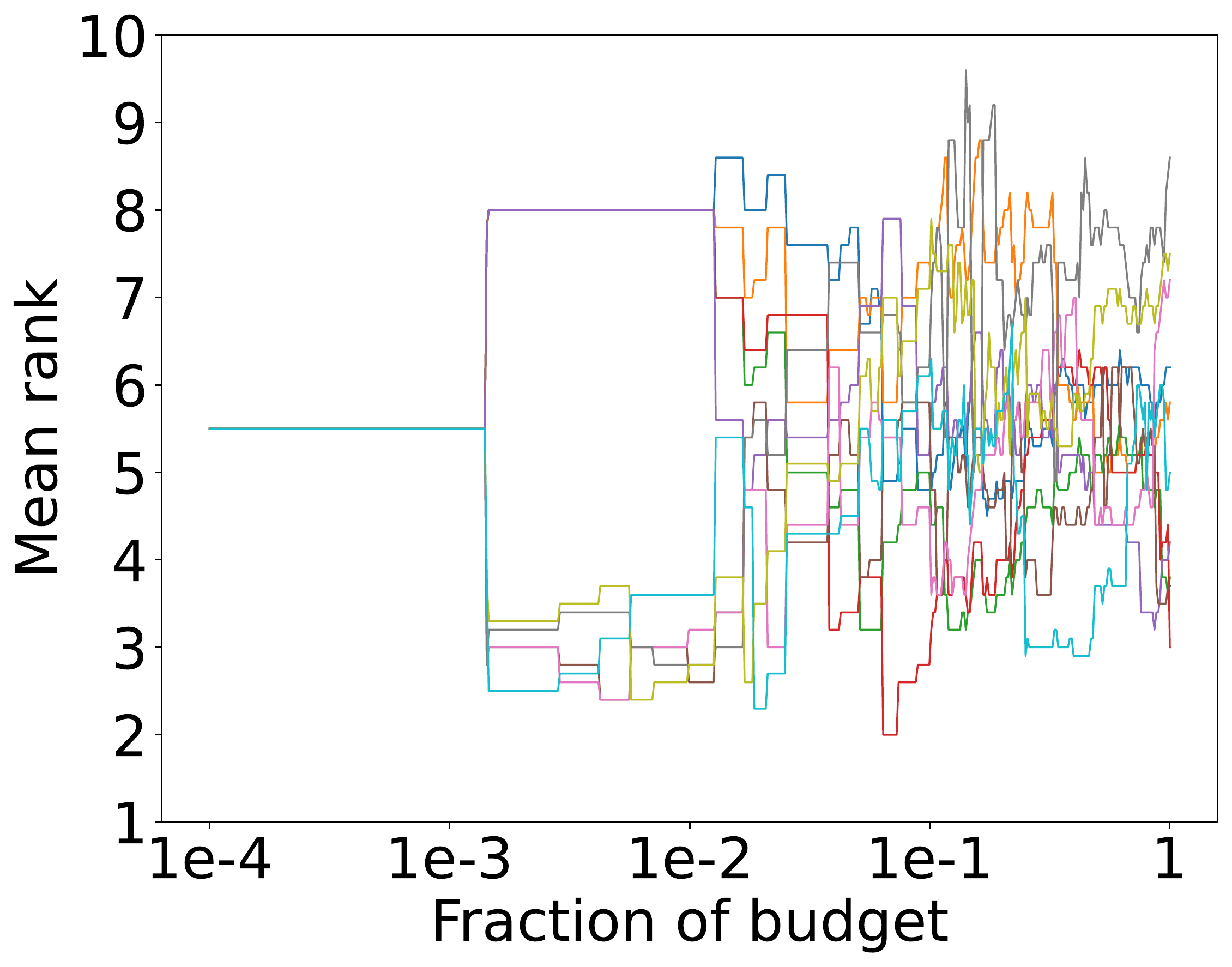}
		\caption{$\text{ALL}_{\text{CoLA}}$}
		\label{fig:All_CoLA_avg}
	\end{subfigure}
	\begin{subfigure}{0.25\linewidth}
		\centering
		\includegraphics[width=0.9\linewidth]{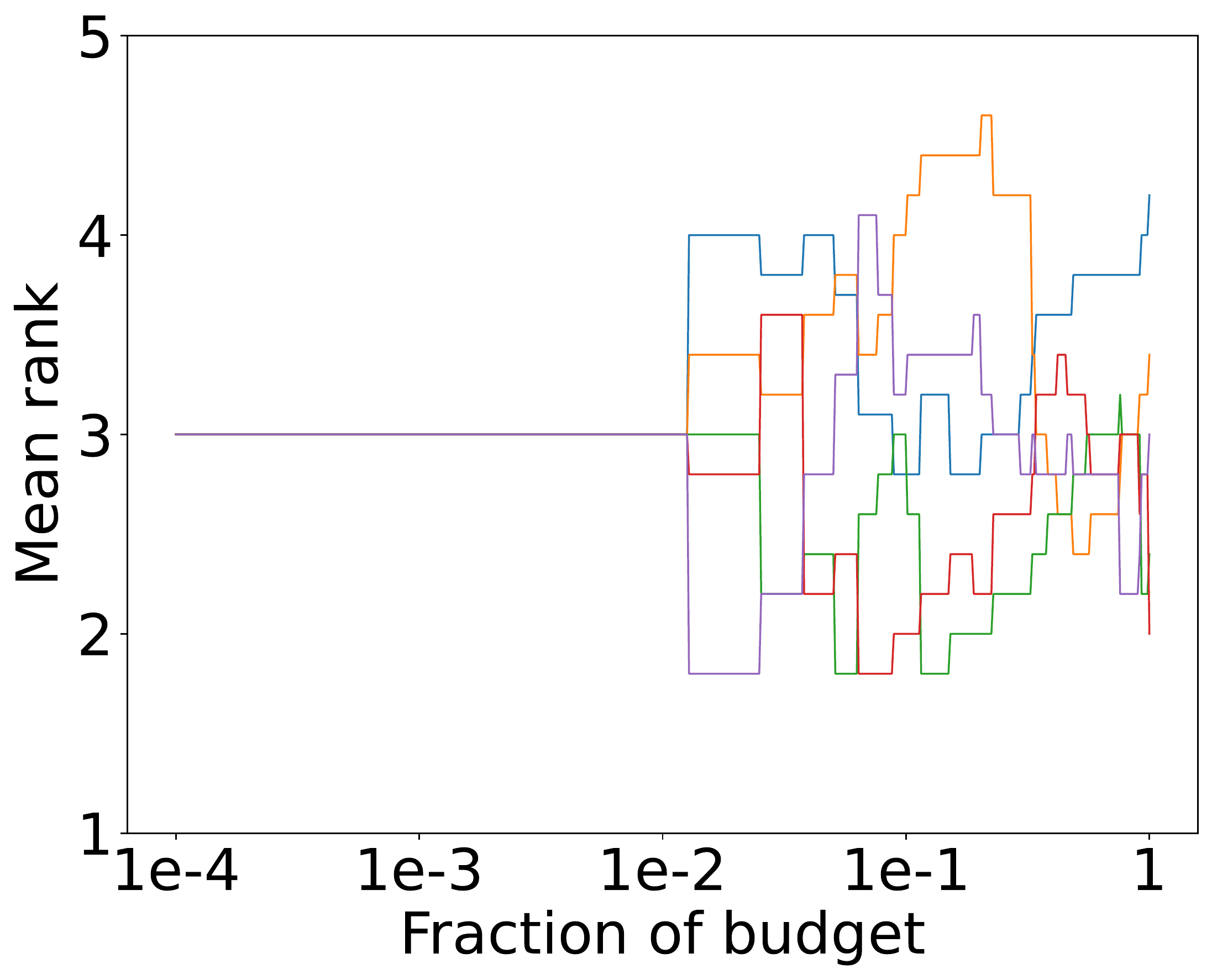}
		\caption{$\textit{BBO}_{\text{CoLA}}$}
		\label{fig:BBO_CoLA_avg}
	\end{subfigure}
	\begin{subfigure}{0.25\linewidth}
		\centering
		\includegraphics[width=0.9\linewidth]{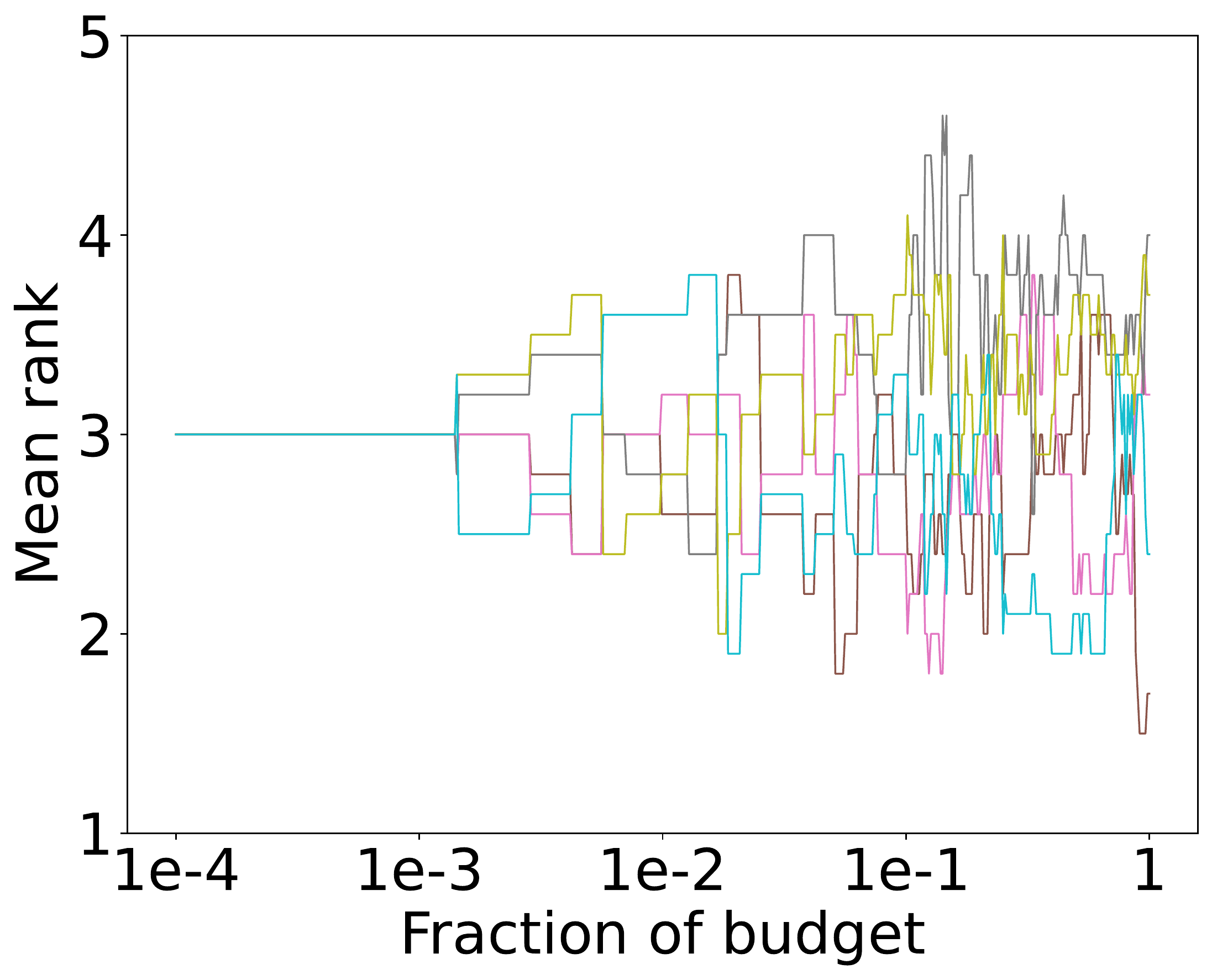}
		\caption{$\textit{MF}_{\text{CoLA}}$}
		\label{fig:MF_CoLA_avg}
	\end{subfigure}
	
	\begin{subfigure}{0.25\linewidth}
		\centering
		\includegraphics[width=0.9\linewidth]{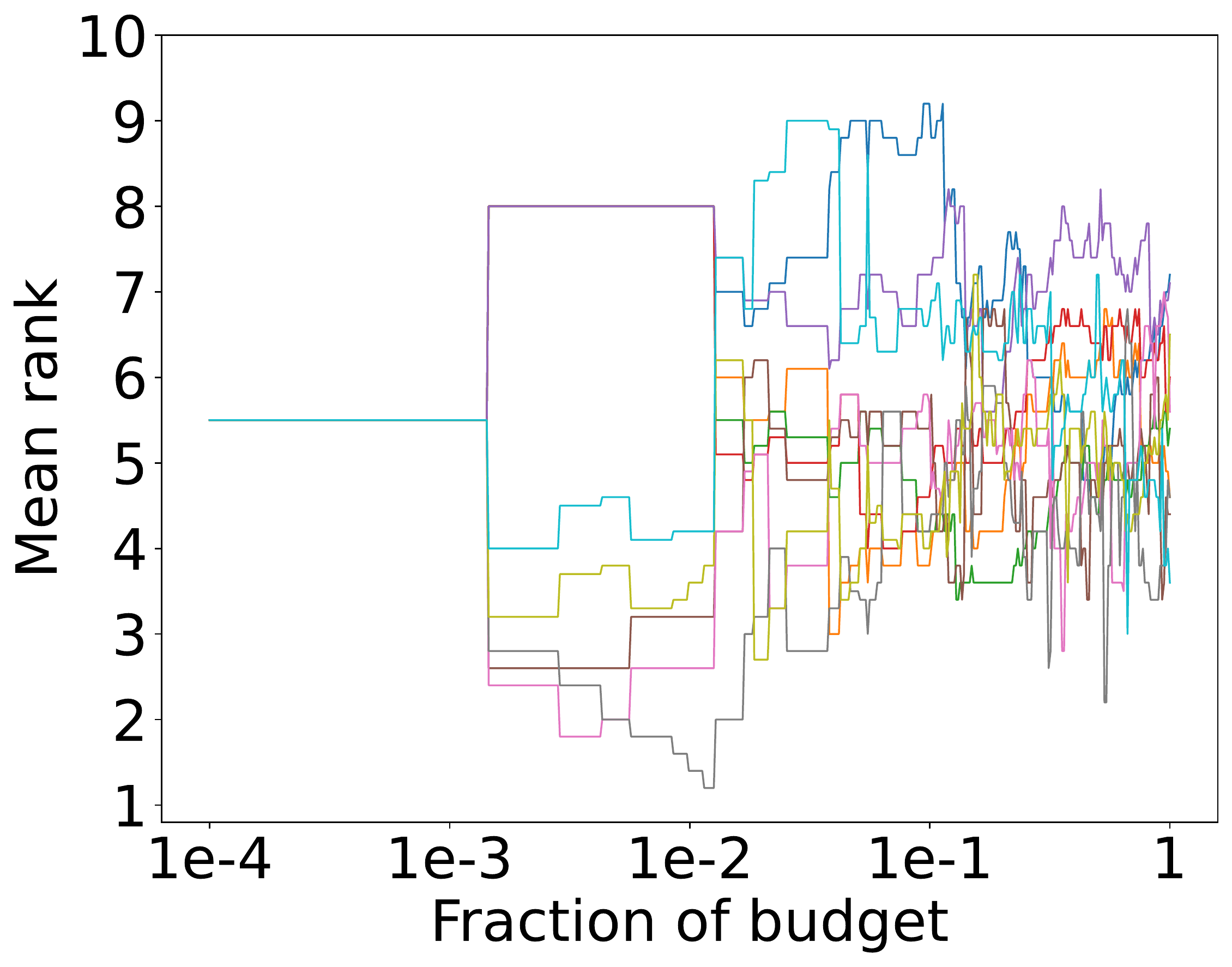}
		\caption{$\text{ALL}_{\text{SST-2}}$}
		\label{fig:All_SST-2_avg}
	\end{subfigure}
	\begin{subfigure}{0.25\linewidth}
		\centering
		\includegraphics[width=0.9\linewidth]{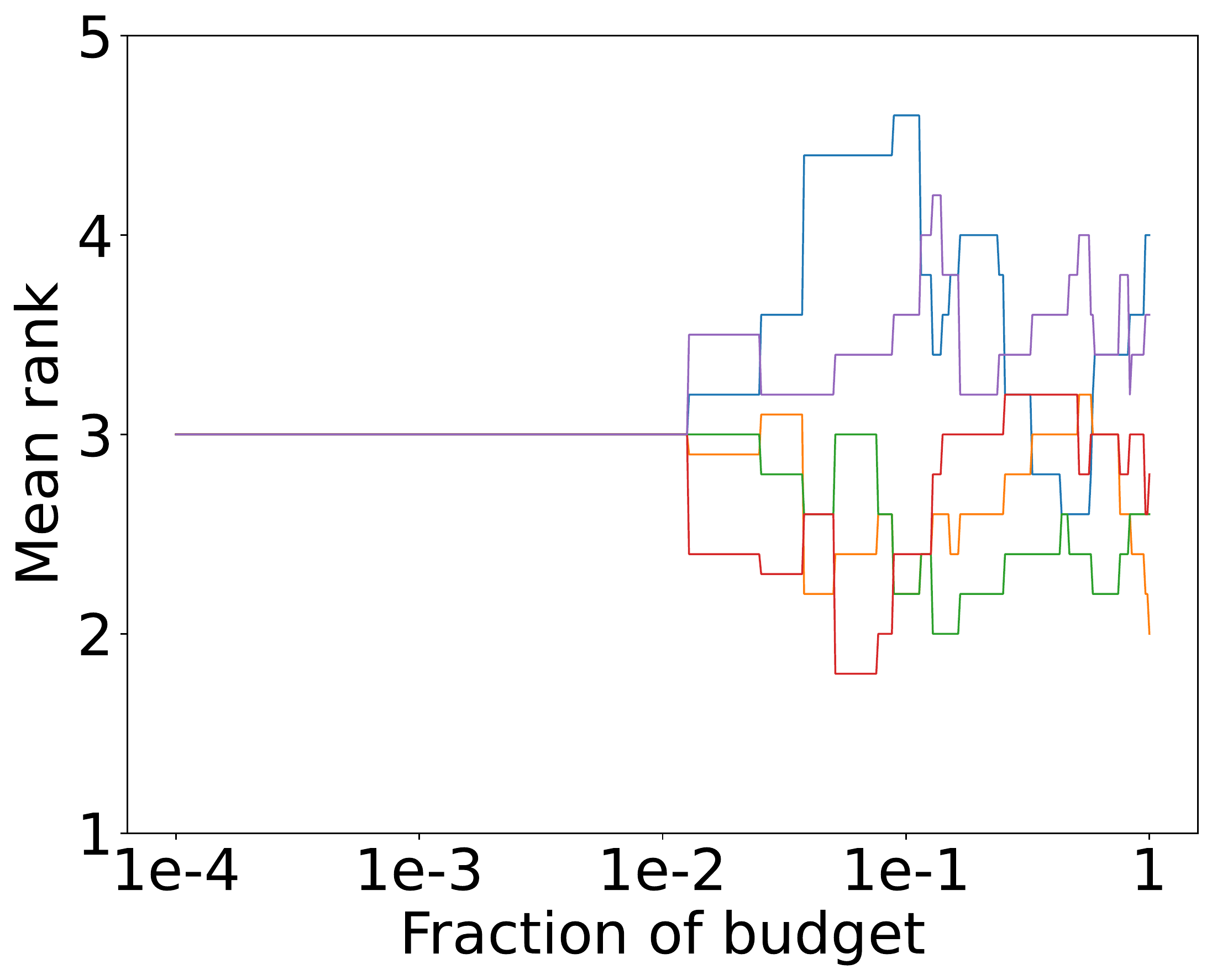}
		\caption{$\textit{BBO}_{\text{SST-2}}$}
		\label{fig:BBO_SST-2_avg}
	\end{subfigure}
	\begin{subfigure}{0.25\linewidth}
		\centering
		\includegraphics[width=0.9\linewidth]{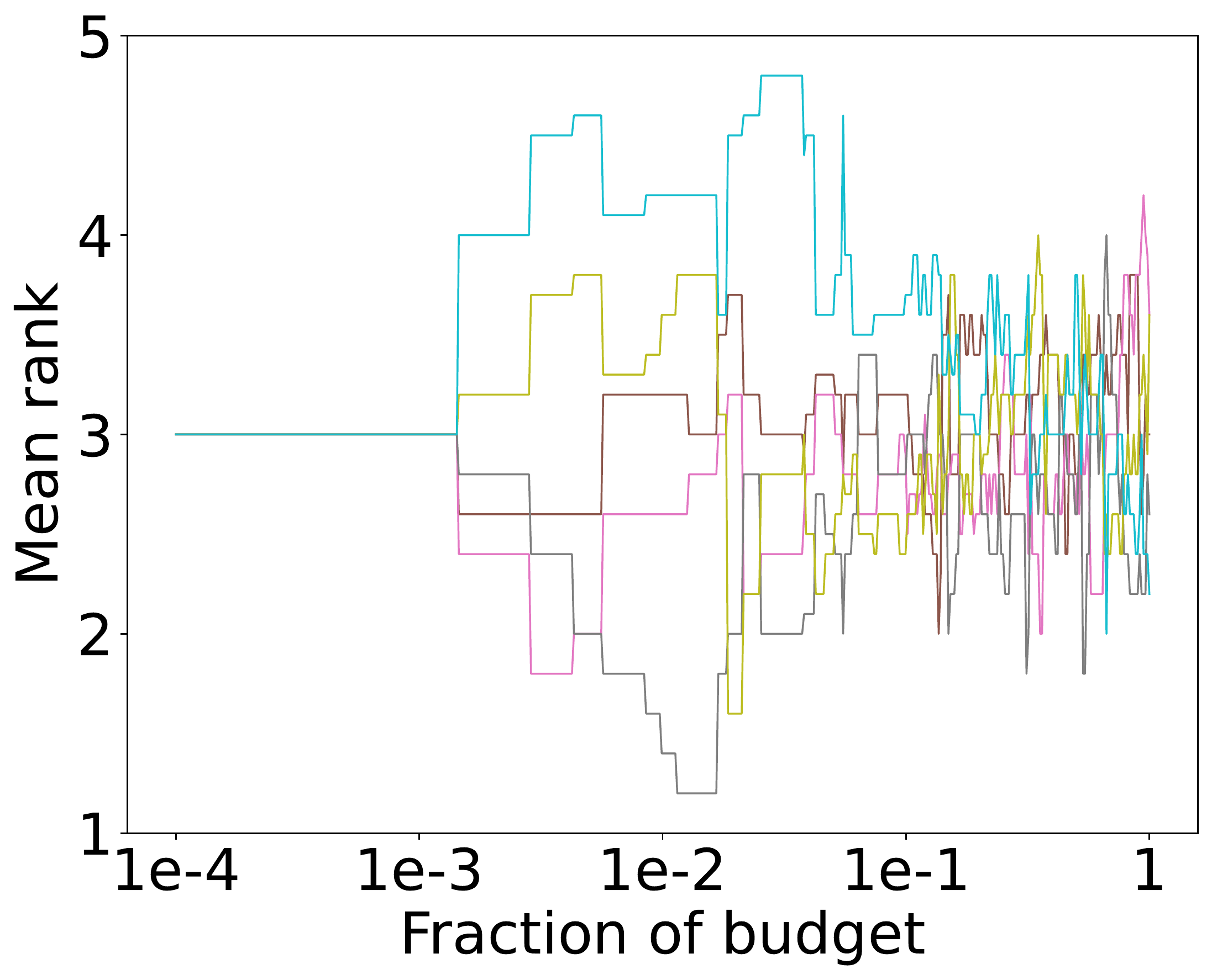}
		\caption{$\textit{MF}_{\text{SST-2}}$}
		\label{fig:MF_SST-2_avg}
	\end{subfigure}
	
	\centering
	\hspace*{1.2cm}\begin{subfigure}{1.0\linewidth}
		\centering
		\includegraphics[width=0.95\linewidth]{materials/legend_rank_new.pdf}
	\end{subfigure}
	\vspace{-0.1in}
	
	\caption{Mean rank over time on BERT benchmark (FedAvg).}
	\label{fig:entire_bert_tabular_avg_rank}
\end{figure}

\begin{figure}[htbp]
	\centering
	\begin{subfigure}{0.25\linewidth}
		\centering
		\includegraphics[width=0.9\linewidth]{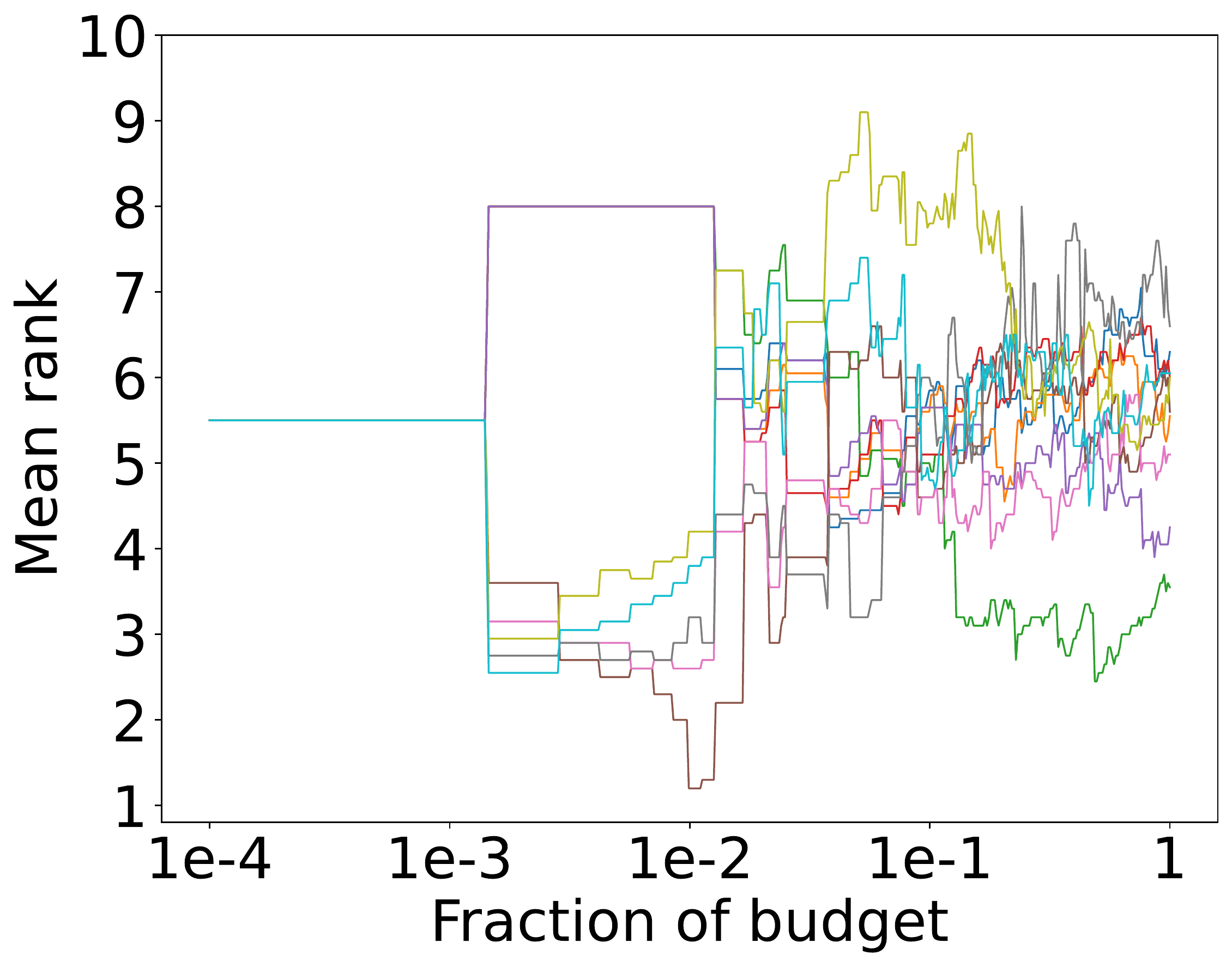}
		\caption{$\text{ALL}_\text{{BERT}}$}
		\label{fig:All_BERT_opt}
	\end{subfigure}
	\begin{subfigure}{0.25\linewidth}
		\centering
		\includegraphics[width=0.9\linewidth]{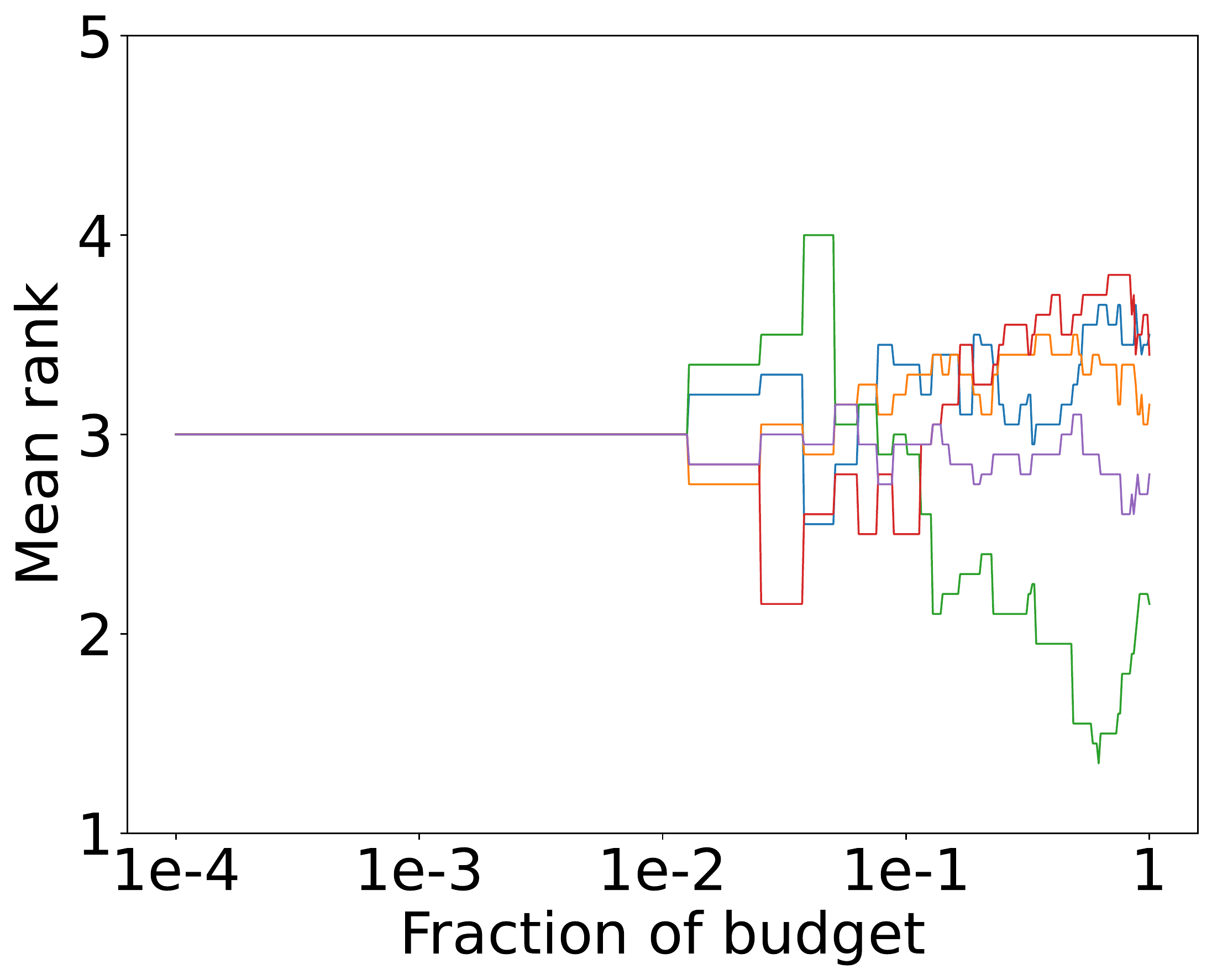}
		\caption{$\textit{BBO}_\text{{BERT}}$}
		\label{fig:BBO_BERT_opt}
	\end{subfigure}
	\begin{subfigure}{0.25\linewidth}
		\centering
		\includegraphics[width=0.9\linewidth]{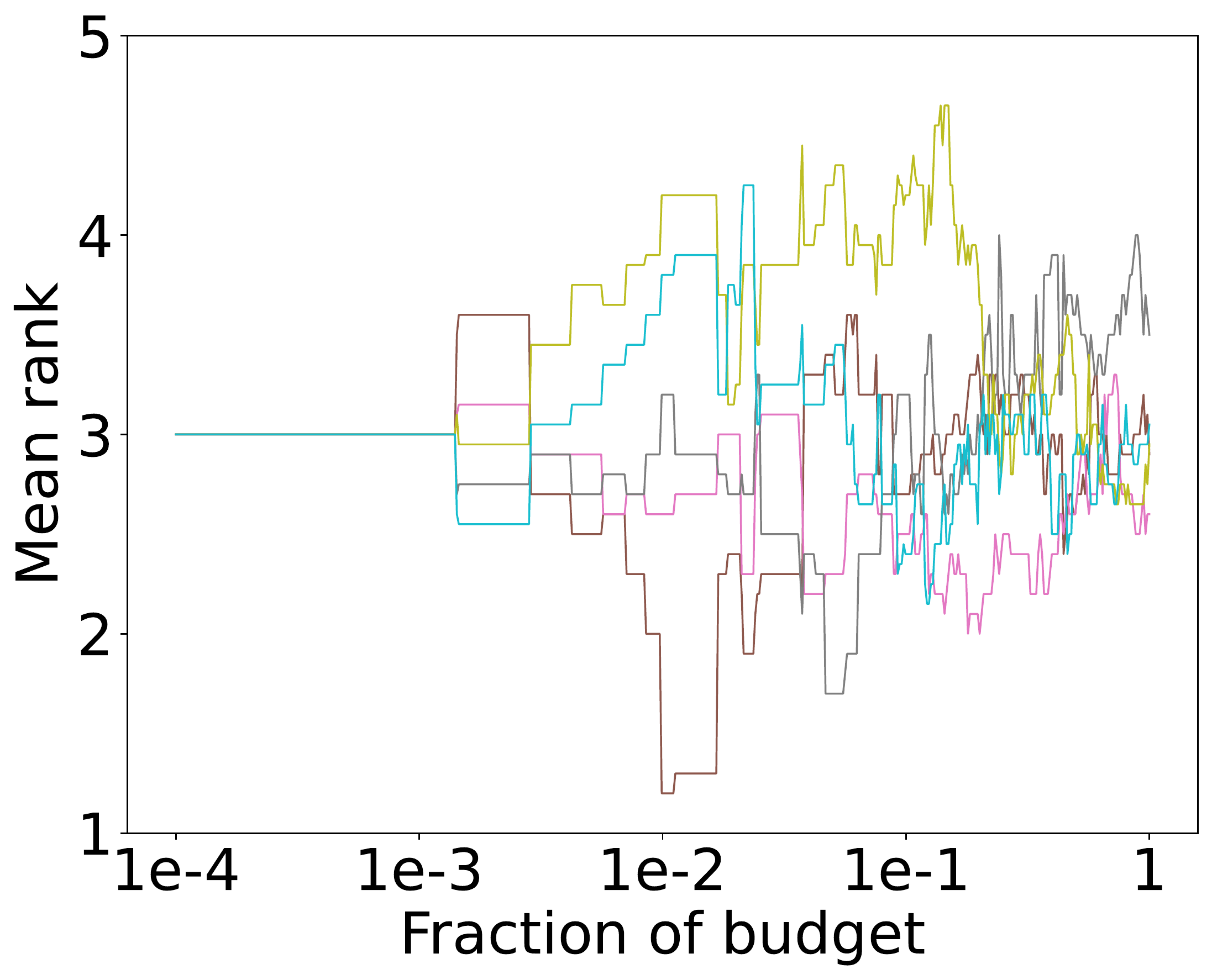}
		\caption{$\textit{MF}_\text{{BERT}}$}
		\label{fig:MF_BERT_opt}
	\end{subfigure}

	\begin{subfigure}{0.25\linewidth}
		\centering
		\includegraphics[width=0.9\linewidth]{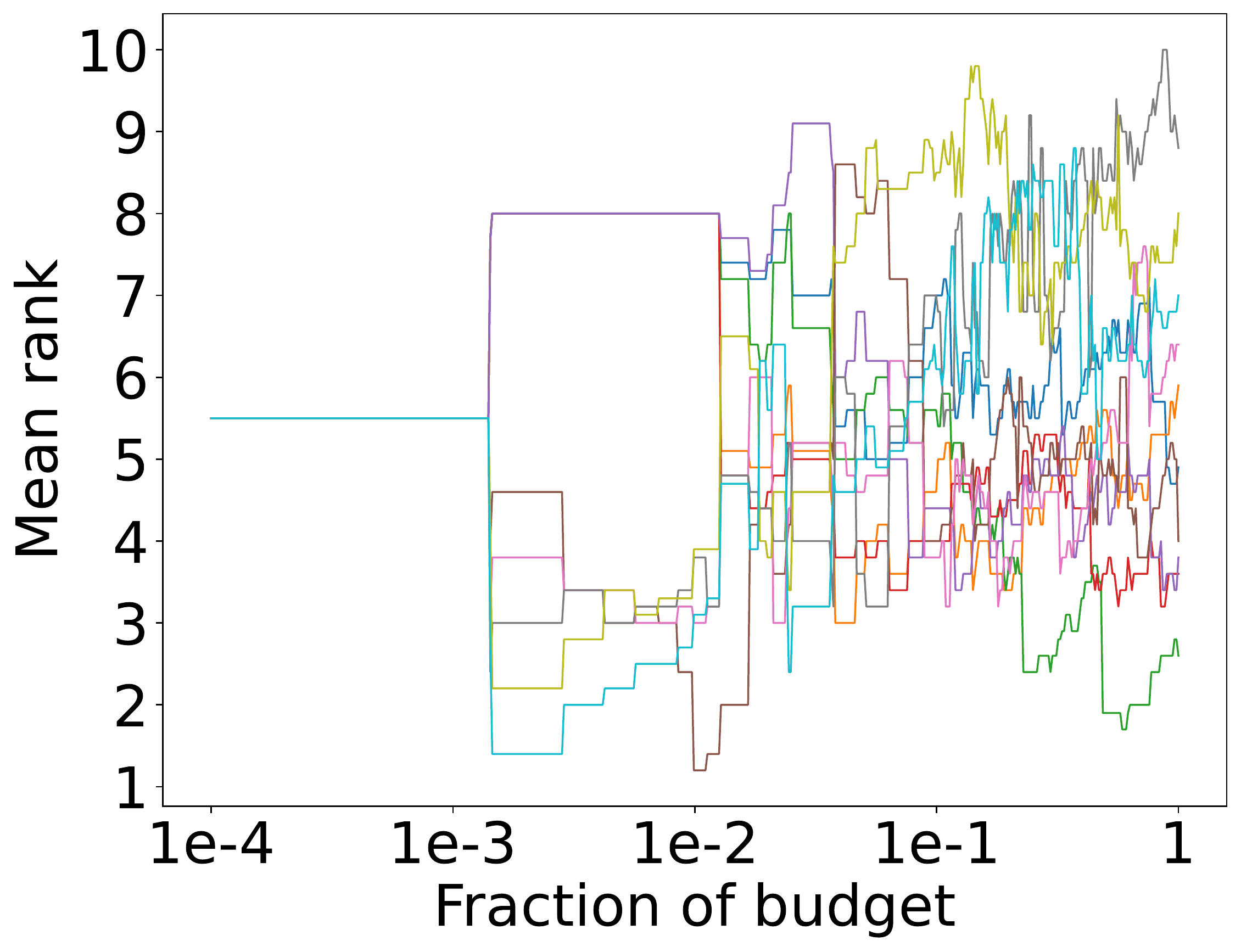}
		\caption{$\text{ALL}_{\text{CoLA}}$}
		\label{fig:All_CoLA_opt}
	\end{subfigure}
	\begin{subfigure}{0.25\linewidth}
		\centering
		\includegraphics[width=0.9\linewidth]{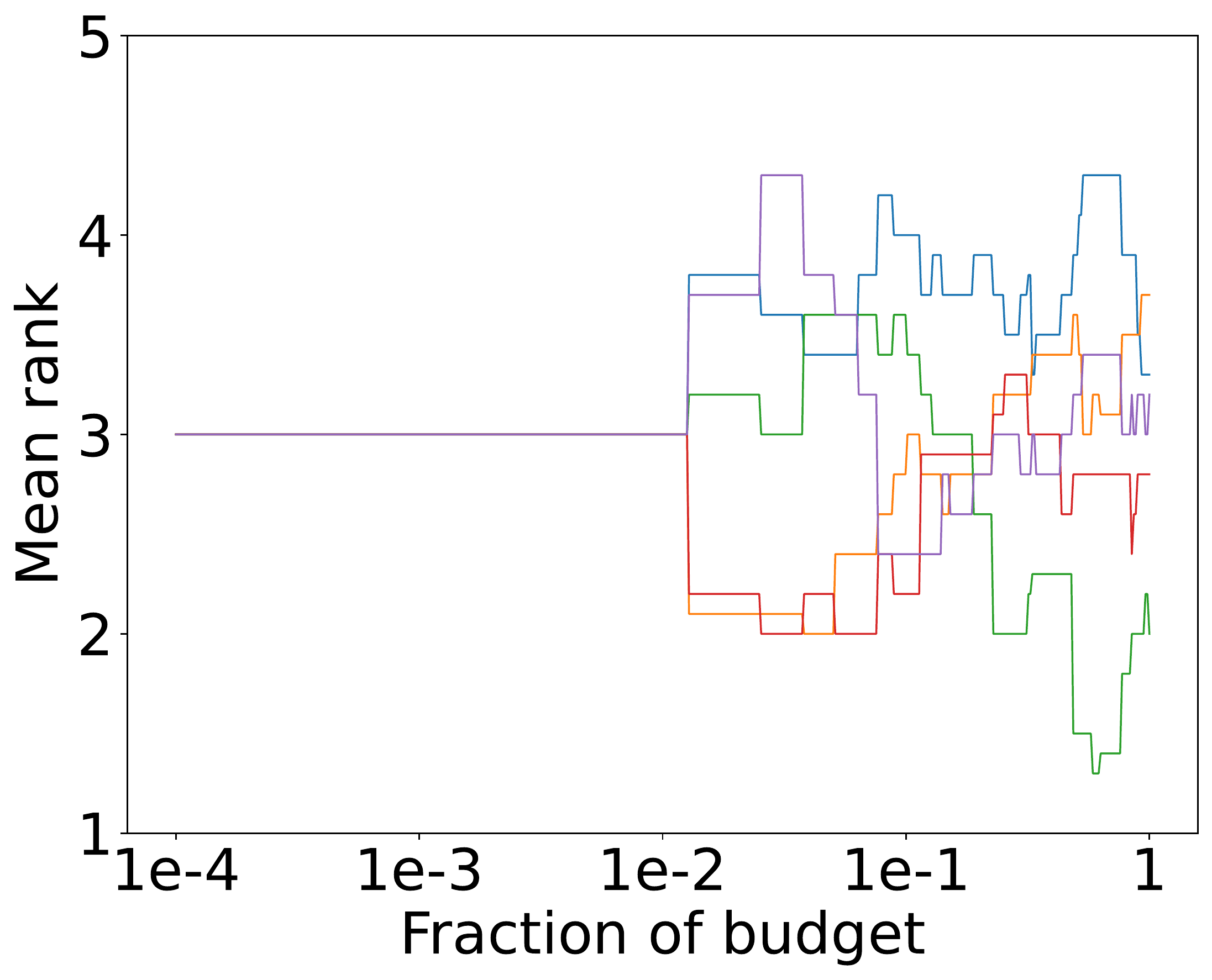}
		\caption{$\textit{BBO}_{\text{CoLA}}$}
		\label{fig:BBO_CoLA_opt}
	\end{subfigure}
	\begin{subfigure}{0.25\linewidth}
		\centering
		\includegraphics[width=0.9\linewidth]{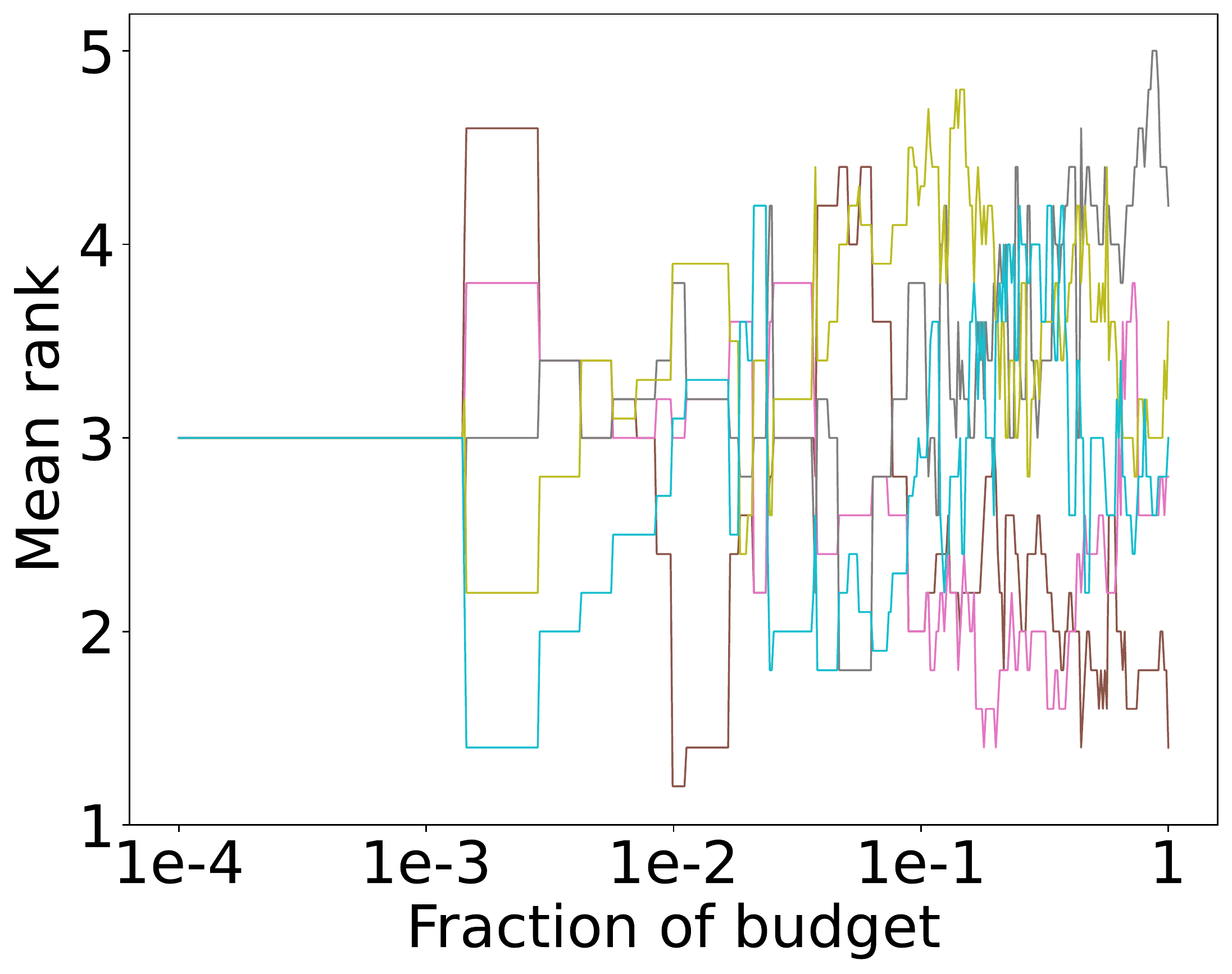}
		\caption{$\textit{MF}_{\text{CoLA}}$}
		\label{fig:MF_CoLA_opt}
	\end{subfigure}
	
	\begin{subfigure}{0.25\linewidth}
		\centering
		\includegraphics[width=0.9\linewidth]{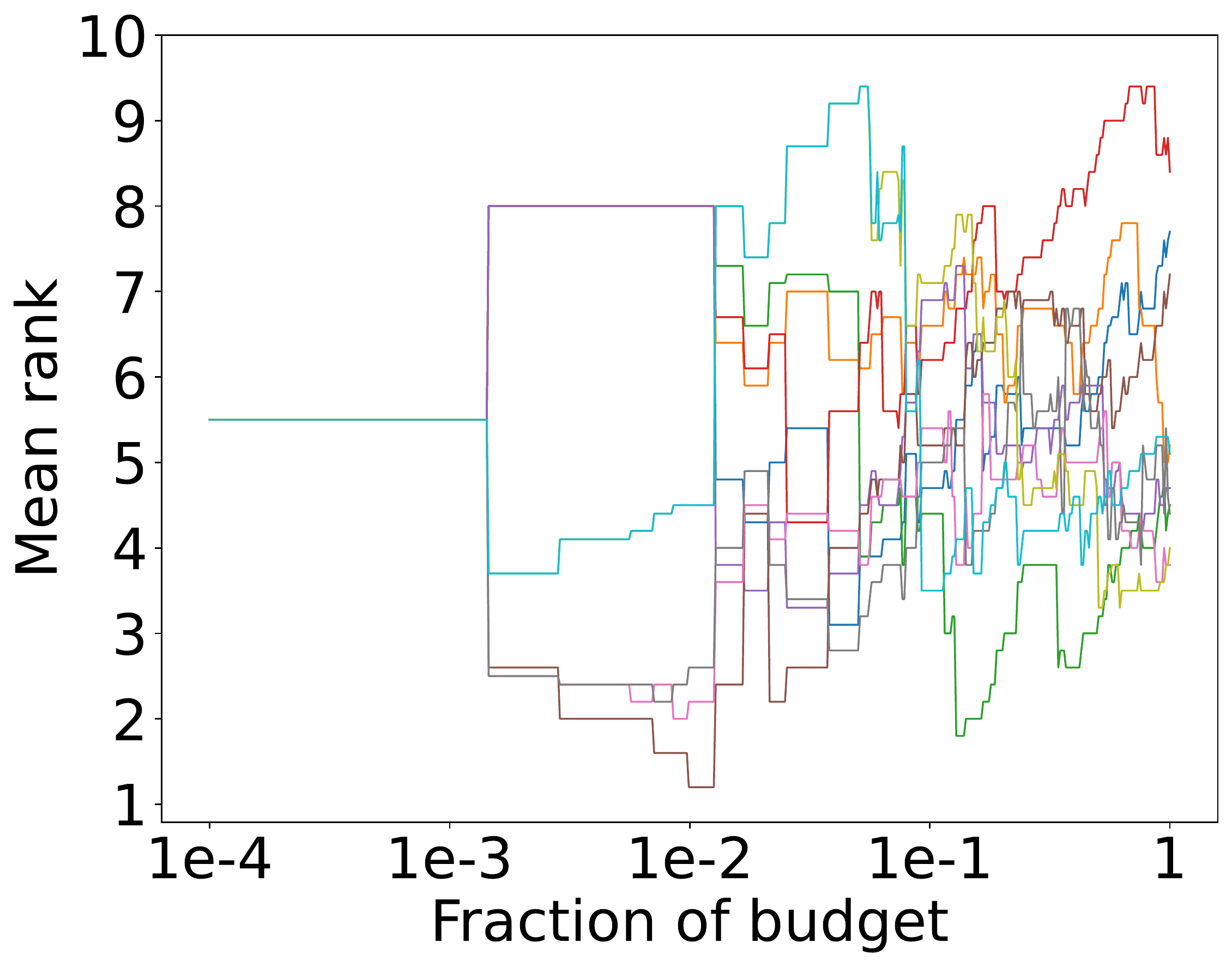}
		\caption{$\text{ALL}_{\text{SST-2}}$}
		\label{fig:All_SST-2_opt}
	\end{subfigure}
	\begin{subfigure}{0.25\linewidth}
		\centering
		\includegraphics[width=0.9\linewidth]{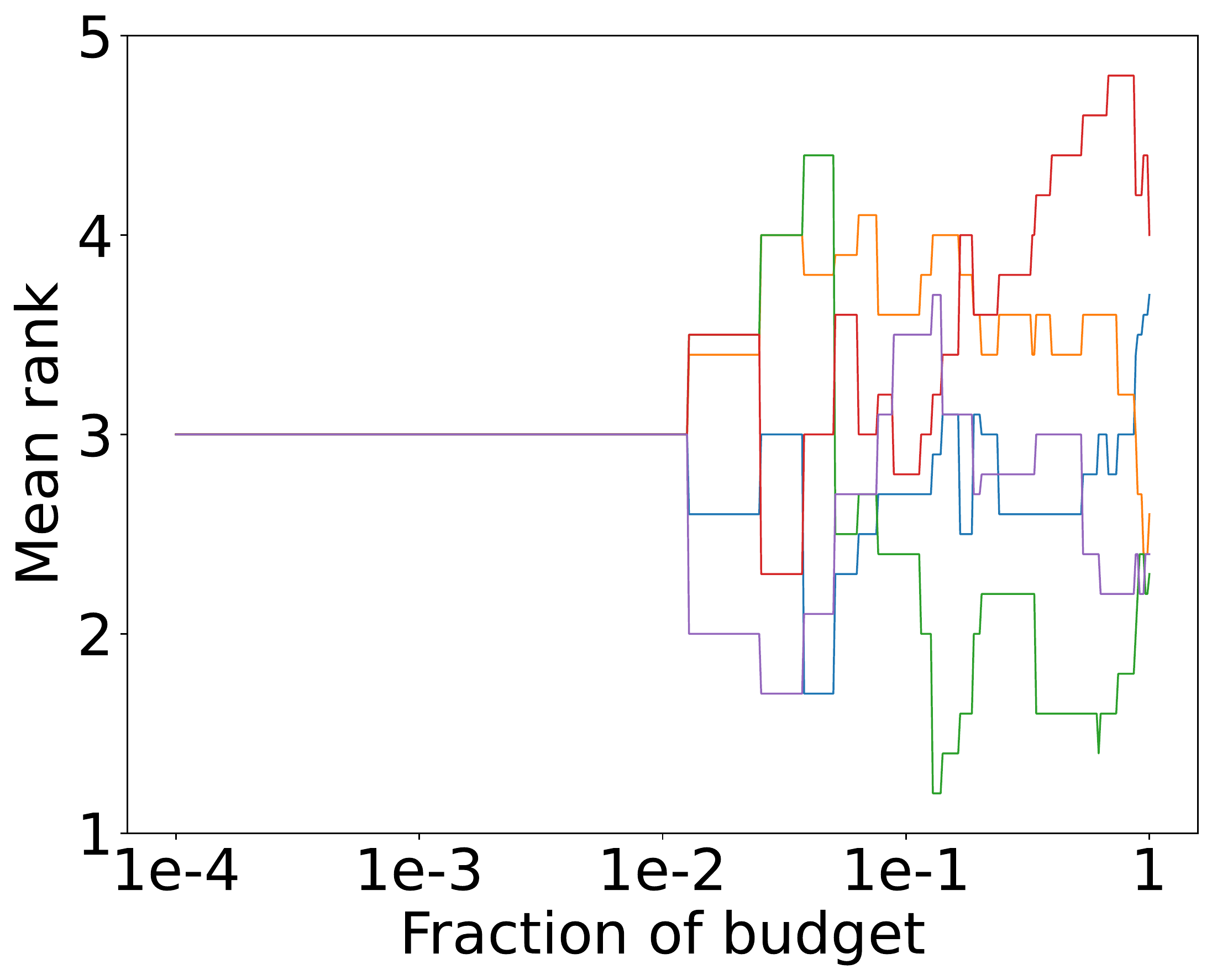}
		\caption{$\textit{BBO}_{\text{SST-2}}$}
		\label{fig:BBO_SST-2_opt}
	\end{subfigure}
	\begin{subfigure}{0.25\linewidth}
		\centering
		\includegraphics[width=0.9\linewidth]{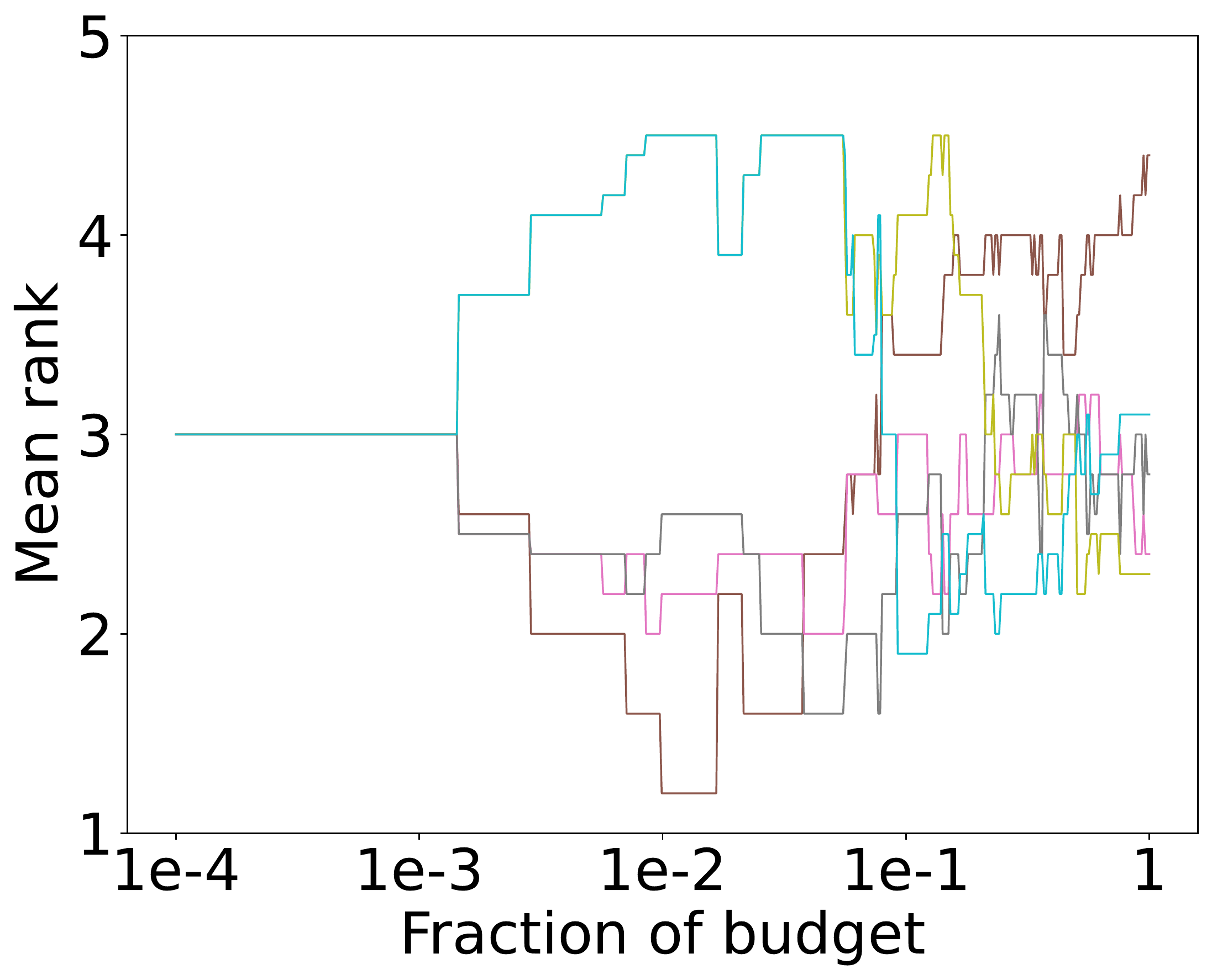}
		\caption{$\textit{MF}_{\text{SST-2}}$}
		\label{fig:MF_SST-2_opt}
	\end{subfigure}
	
	\centering
	\hspace*{1.2cm}\begin{subfigure}{1.0\linewidth}
		\centering
		\includegraphics[width=0.95\linewidth]{materials/legend_rank_new.pdf}
	\end{subfigure}
	\vspace{-0.1in}
	
	\caption{Mean rank over time on BERT benchmark (FedOPT).}
	\label{fig:entire_bert_tabular_opt_rank}
\end{figure}

\begin{figure}[htbp]
	\centering
	\begin{subfigure}{0.25\linewidth}
		\centering
		\includegraphics[width=0.9\linewidth]{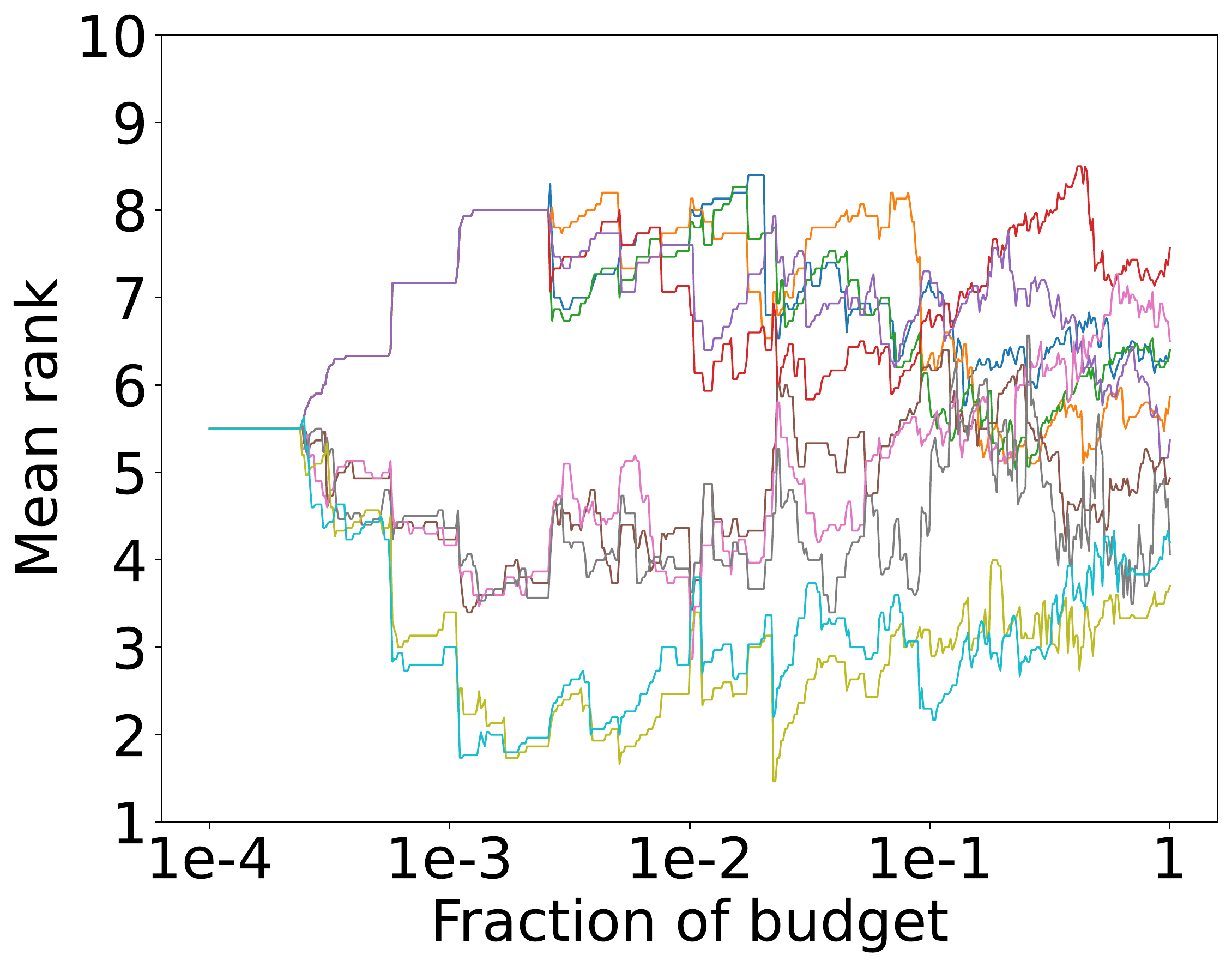}
		\caption{$\text{ALL}_\text{{GNN}}$}
		\label{fig:All_GNN_avg}
	\end{subfigure}
	\begin{subfigure}{0.25\linewidth}
		\centering
		\includegraphics[width=0.9\linewidth]{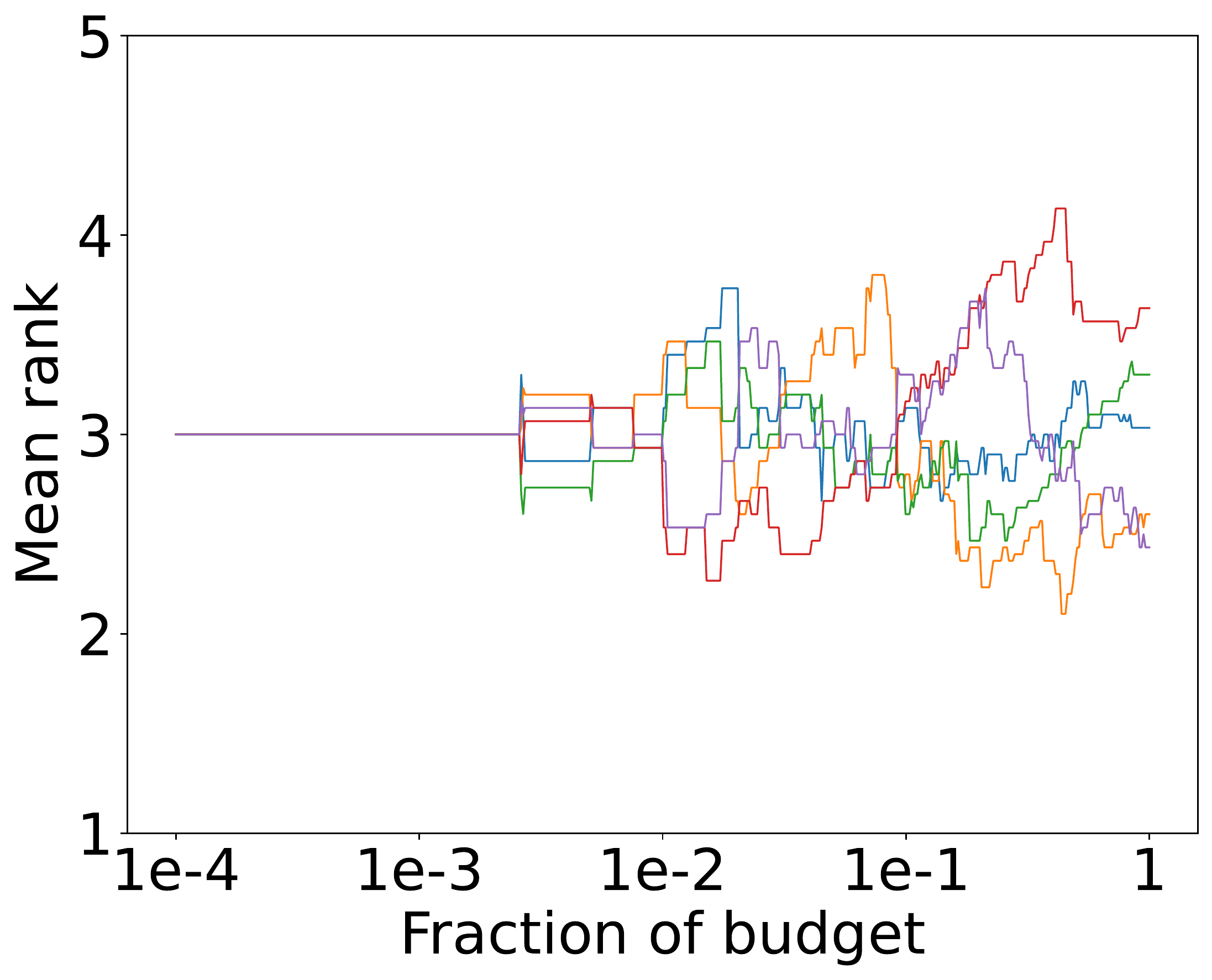}
		\caption{$\textit{BBO}_\text{{GNN}}$}
		\label{fig:BBO_GNN_avg}
	\end{subfigure}
	\begin{subfigure}{0.25\linewidth}
		\centering
		\includegraphics[width=0.9\linewidth]{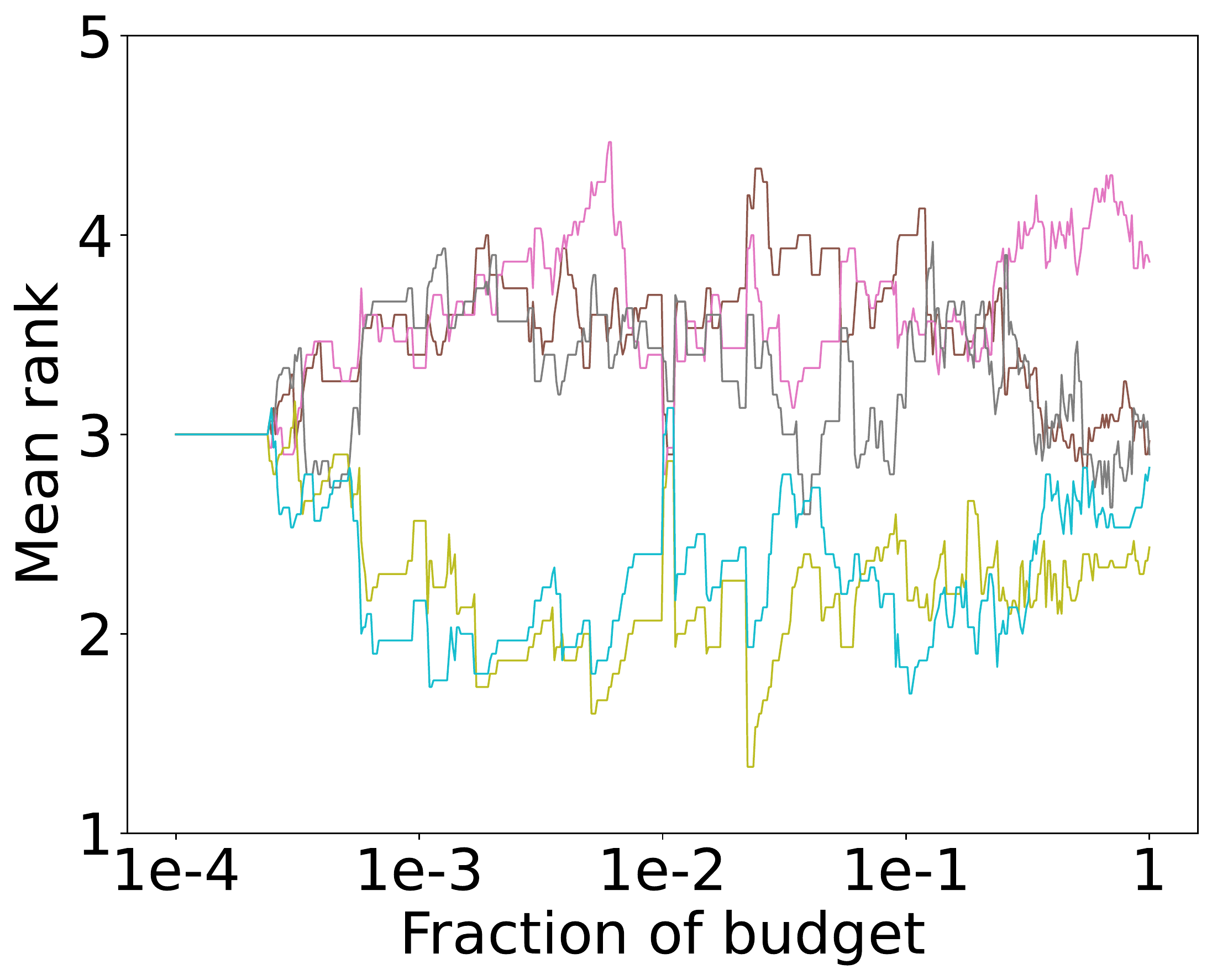}
		\caption{$\textit{MF}_\text{{GNN}}$}
		\label{fig:MF_GNN_avg}
	\end{subfigure}

	\begin{subfigure}{0.25\linewidth}
		\centering
		\includegraphics[width=0.9\linewidth]{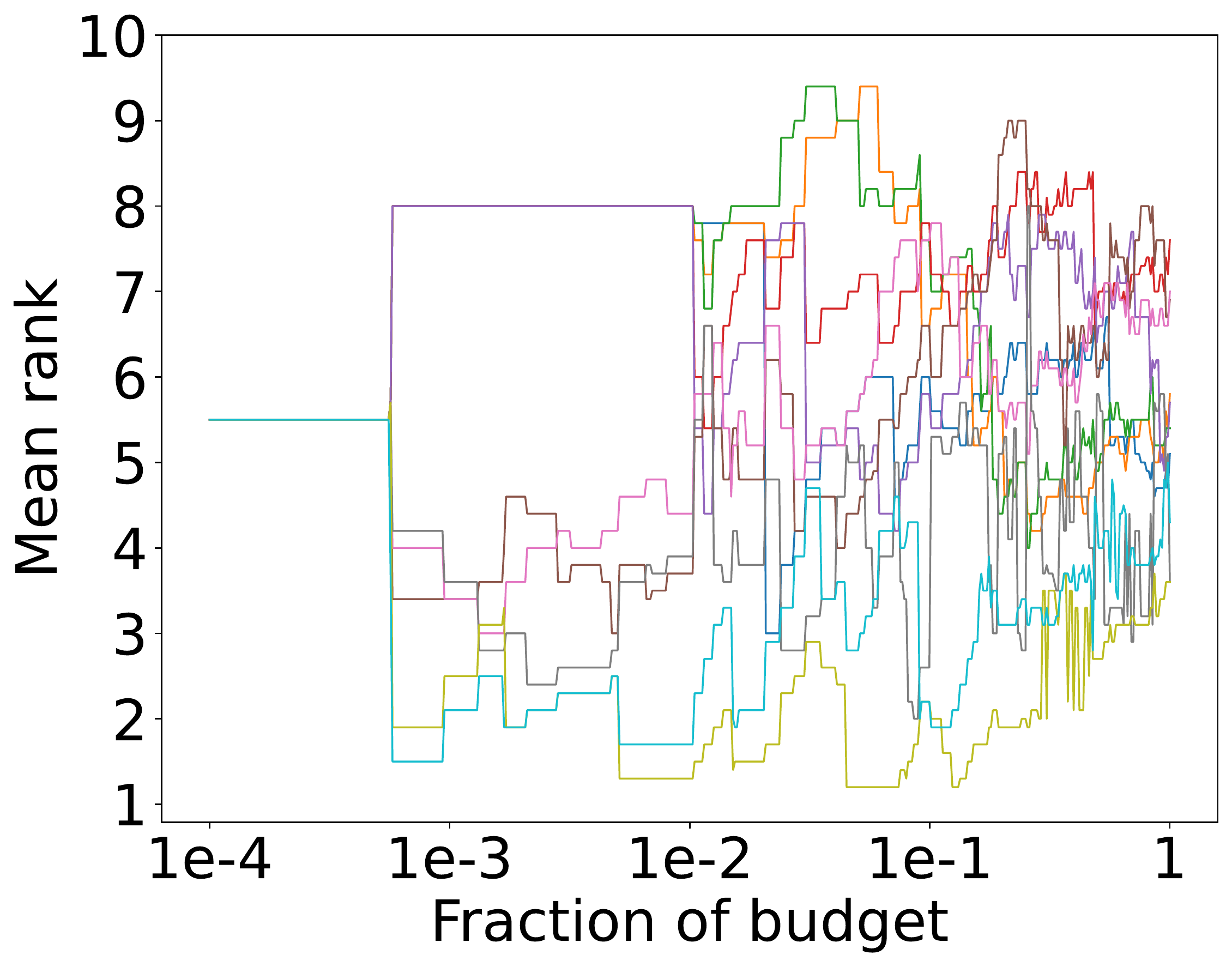}
		\caption{$\text{ALL}_{\text{Cora}}$}
		\label{fig:All_Cora_avg}
	\end{subfigure}
	\begin{subfigure}{0.25\linewidth}
		\centering
		\includegraphics[width=0.9\linewidth]{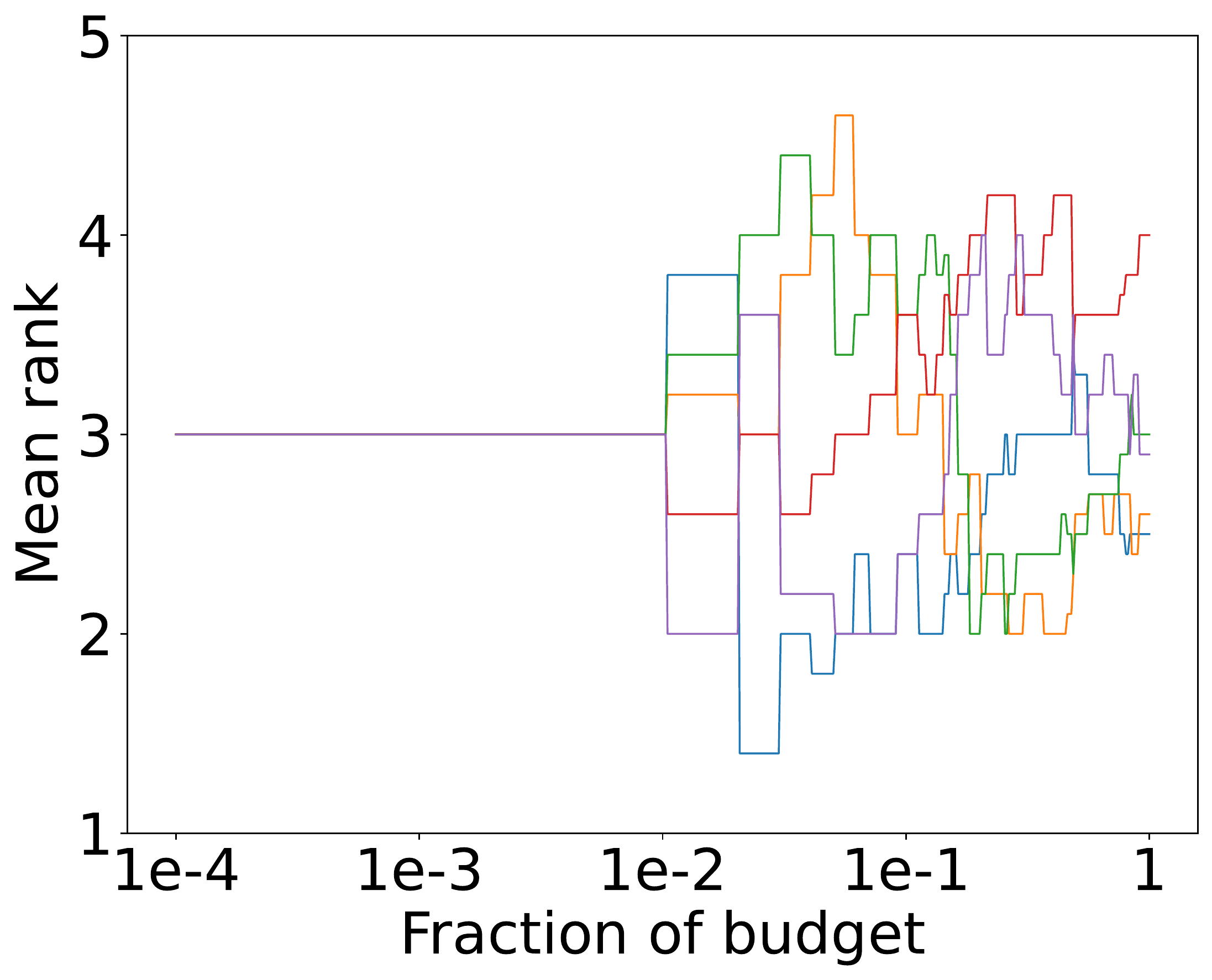}
		\caption{$\textit{BBO}_{\text{Cora}}$}
		\label{fig:BBO_Cora_avg}
	\end{subfigure}
	\begin{subfigure}{0.25\linewidth}
		\centering
		\includegraphics[width=0.9\linewidth]{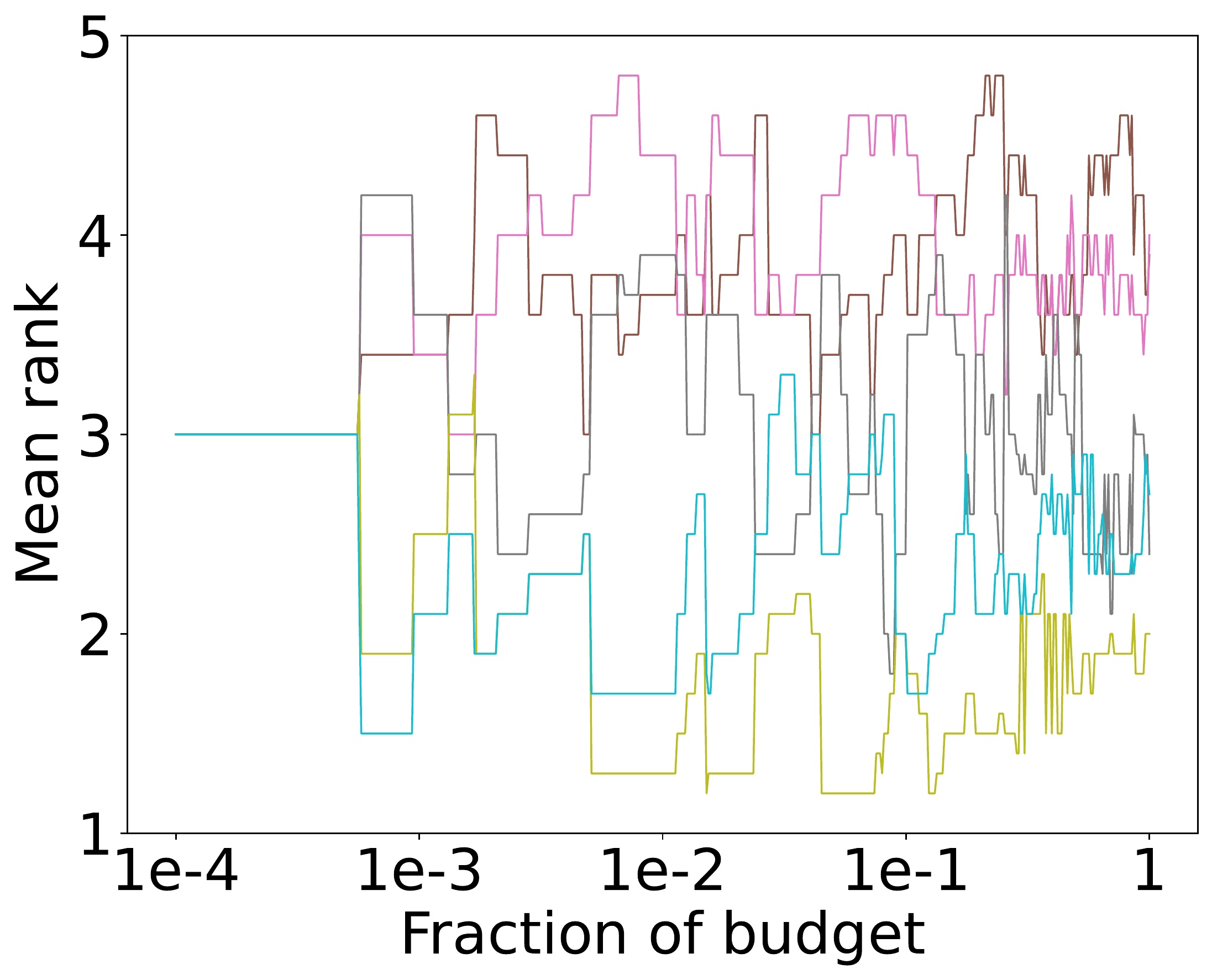}
		\caption{$\textit{MF}_{\text{Cora}}$}
		\label{fig:MF_Cora_avg}
	\end{subfigure}
	
	\begin{subfigure}{0.25\linewidth}
		\centering
		\includegraphics[width=0.9\linewidth]{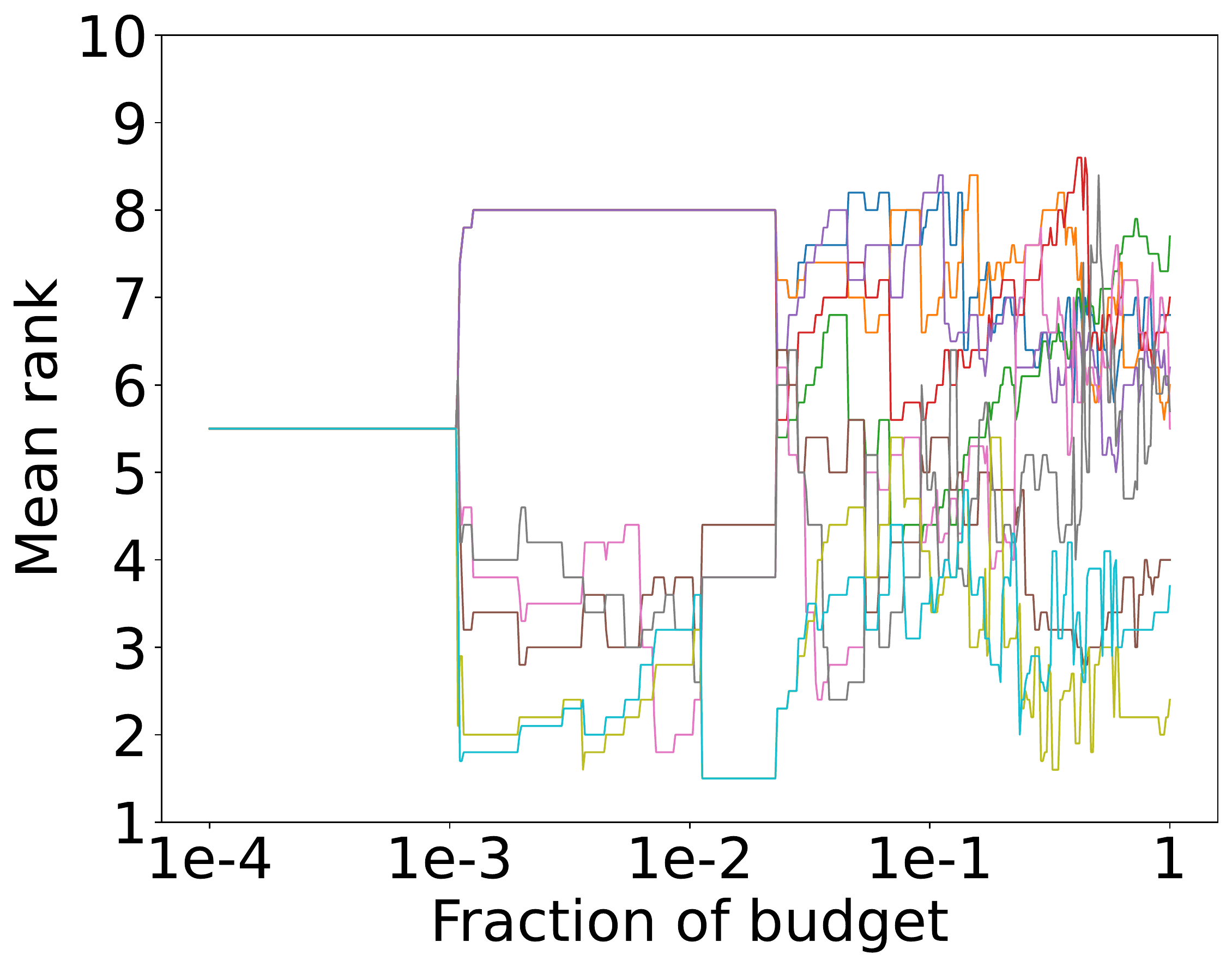}
		\caption{$\text{ALL}_{\text{CiteSeer}}$}
		\label{fig:All_Citeseer_avg}
	\end{subfigure}
	\begin{subfigure}{0.25\linewidth}
		\centering
		\includegraphics[width=0.9\linewidth]{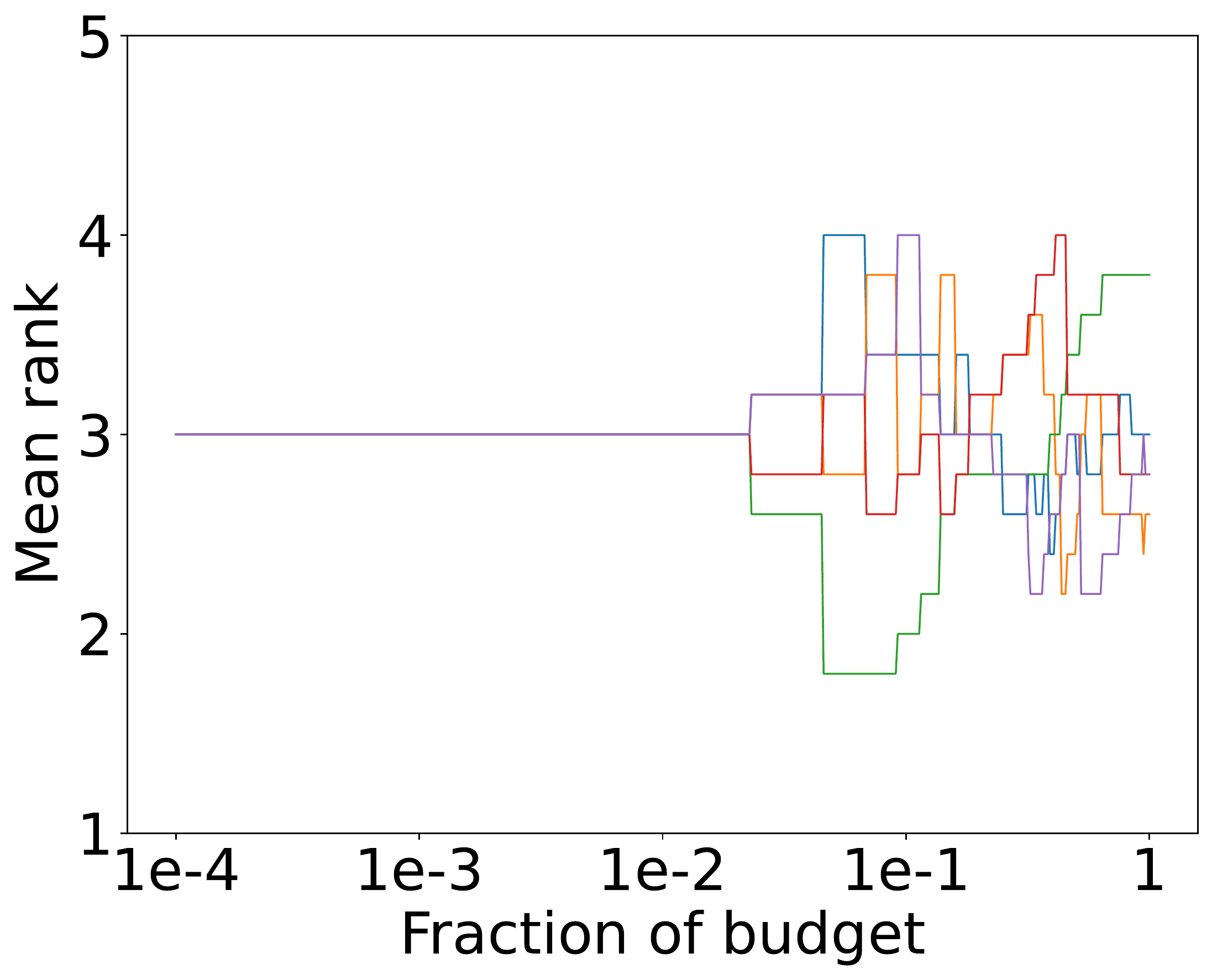}
		\caption{$\textit{BBO}_{\text{CiteSeer}}$}
		\label{fig:BBO_Citeseer_avg}
	\end{subfigure}
	\begin{subfigure}{0.25\linewidth}
		\centering
		\includegraphics[width=0.9\linewidth]{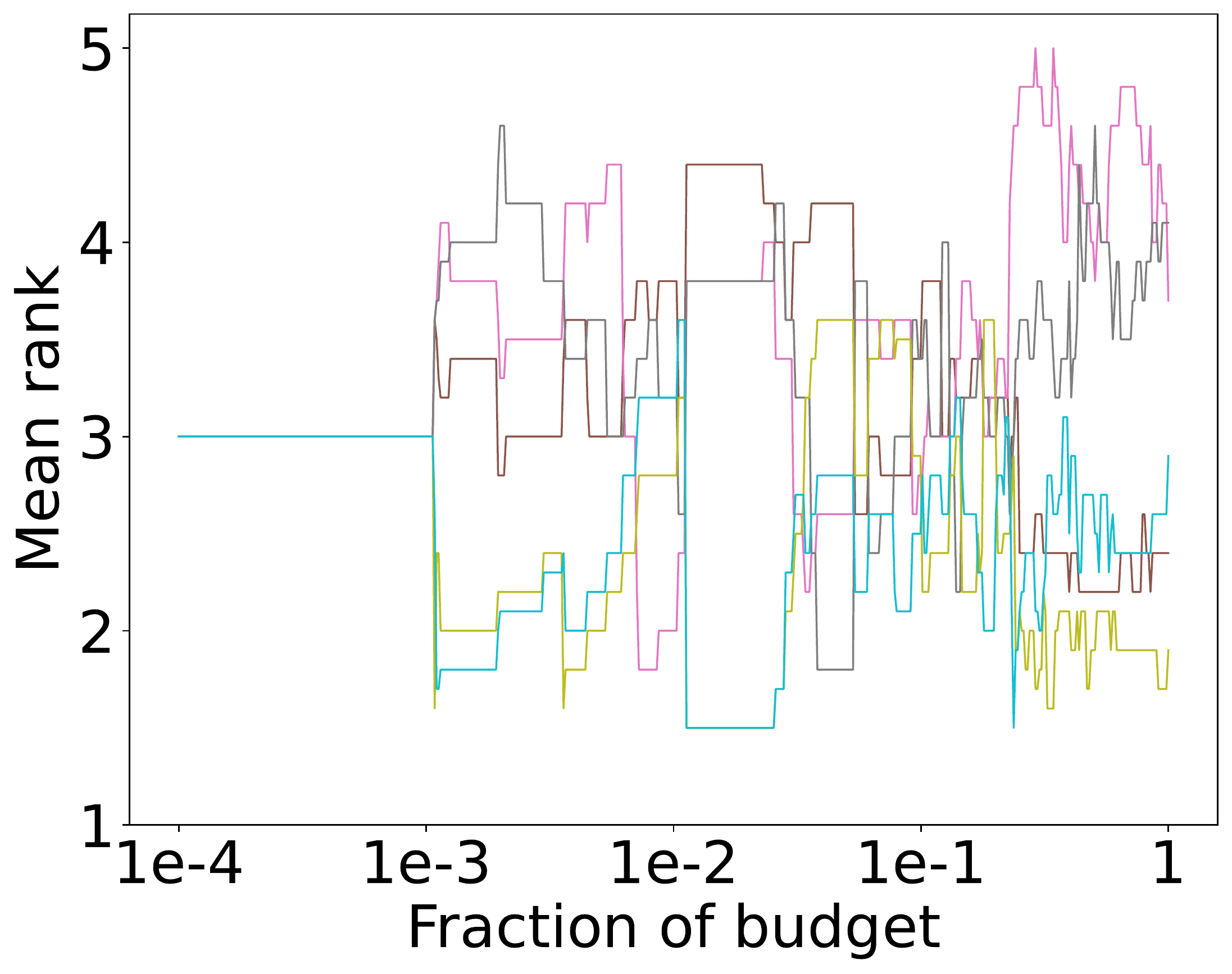}
		\caption{$\textit{MF}_{\text{CiteSeer}}$}
		\label{fig:MF_Citeseer_avg}
	\end{subfigure}
	
	\begin{subfigure}{0.25\linewidth}
		\centering
		\includegraphics[width=0.9\linewidth]{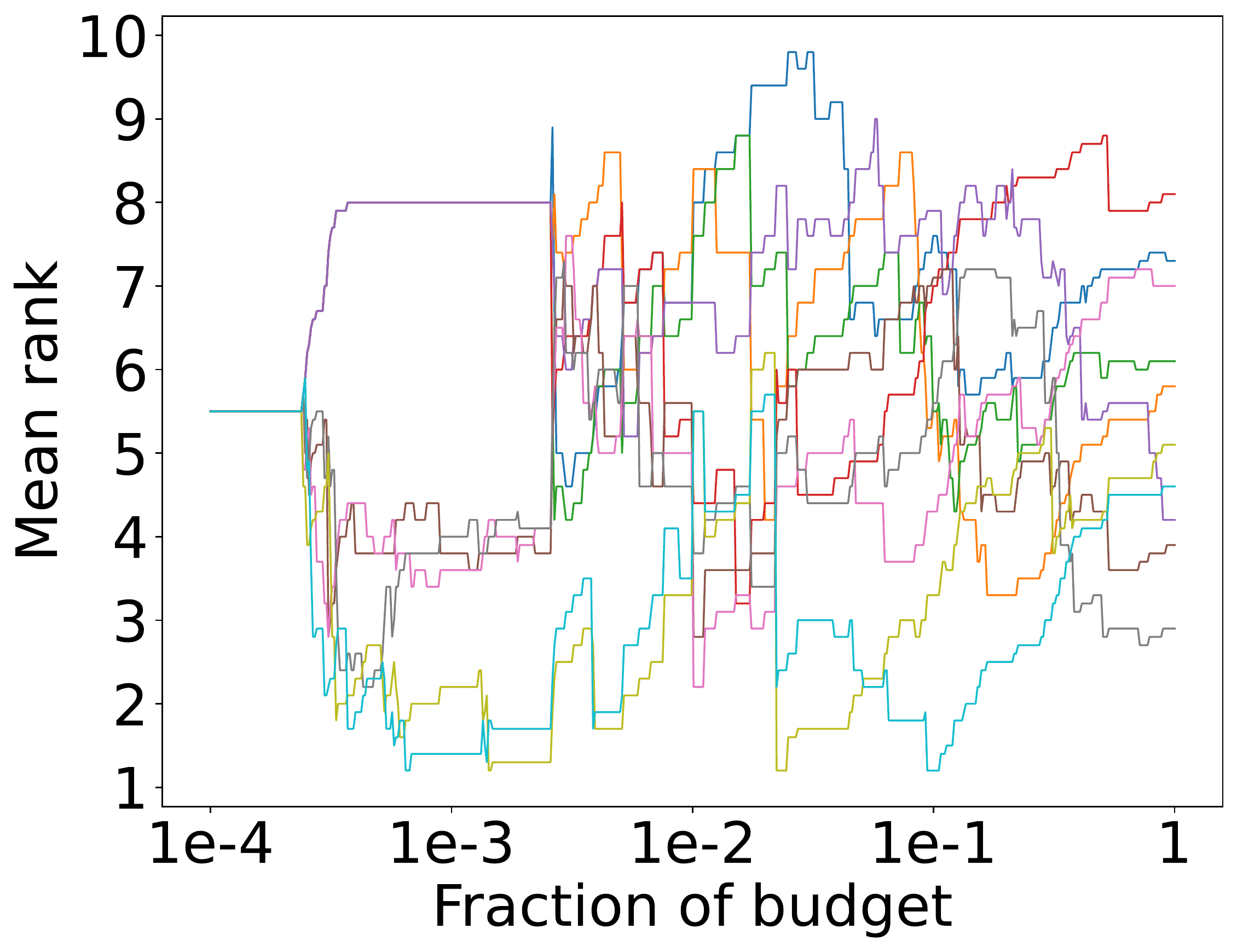}
		\caption{$\text{ALL}_{\text{PubMed}}$}
		\label{fig:All_Pubmed_avg}
	\end{subfigure}
	\begin{subfigure}{0.25\linewidth}
		\centering
		\includegraphics[width=0.9\linewidth]{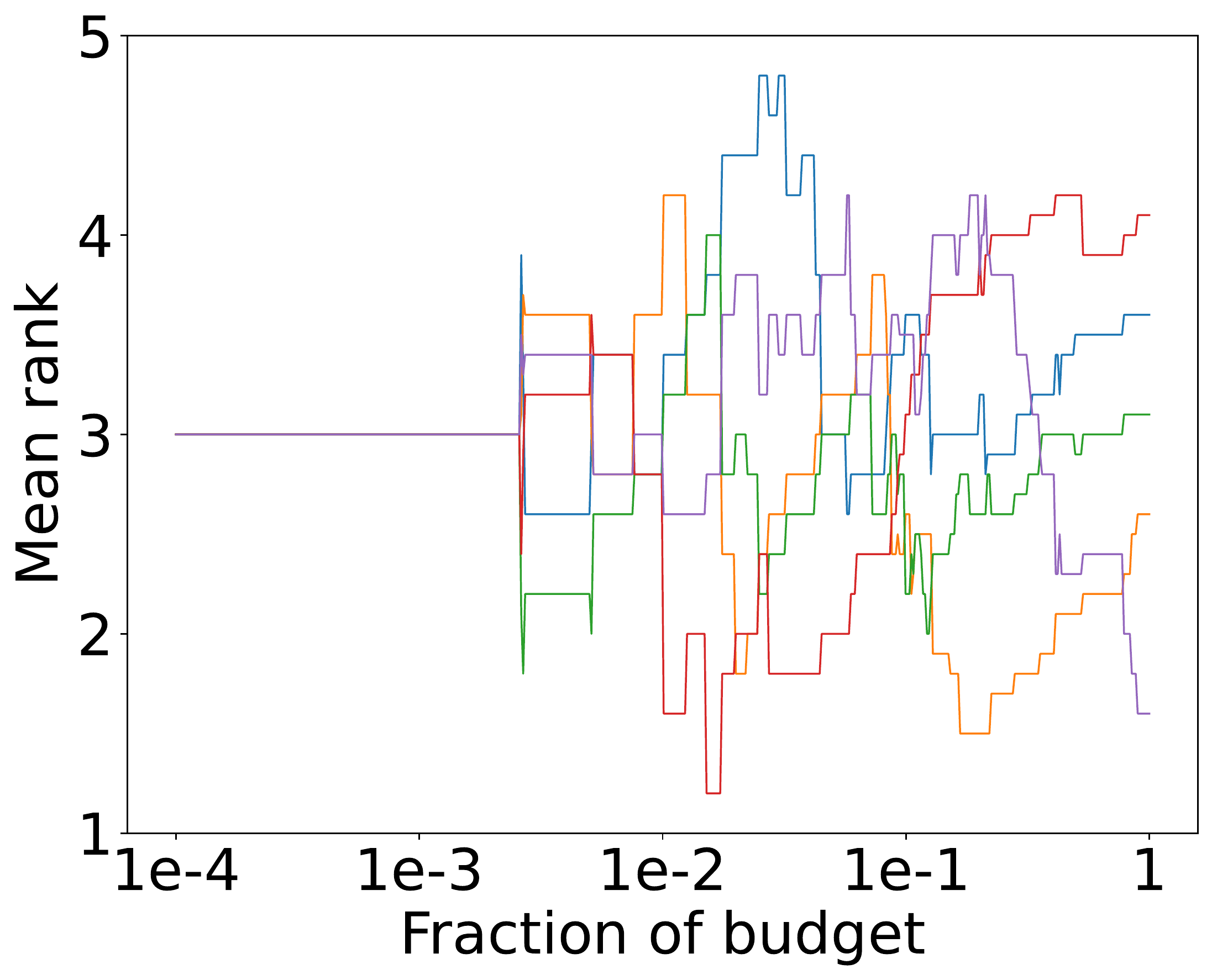}
		\caption{$\textit{BBO}_{\text{PubMed}}$}
		\label{fig:BBO_Pubmed_avg}
	\end{subfigure}
	\begin{subfigure}{0.25\linewidth}
		\centering
		\includegraphics[width=0.9\linewidth]{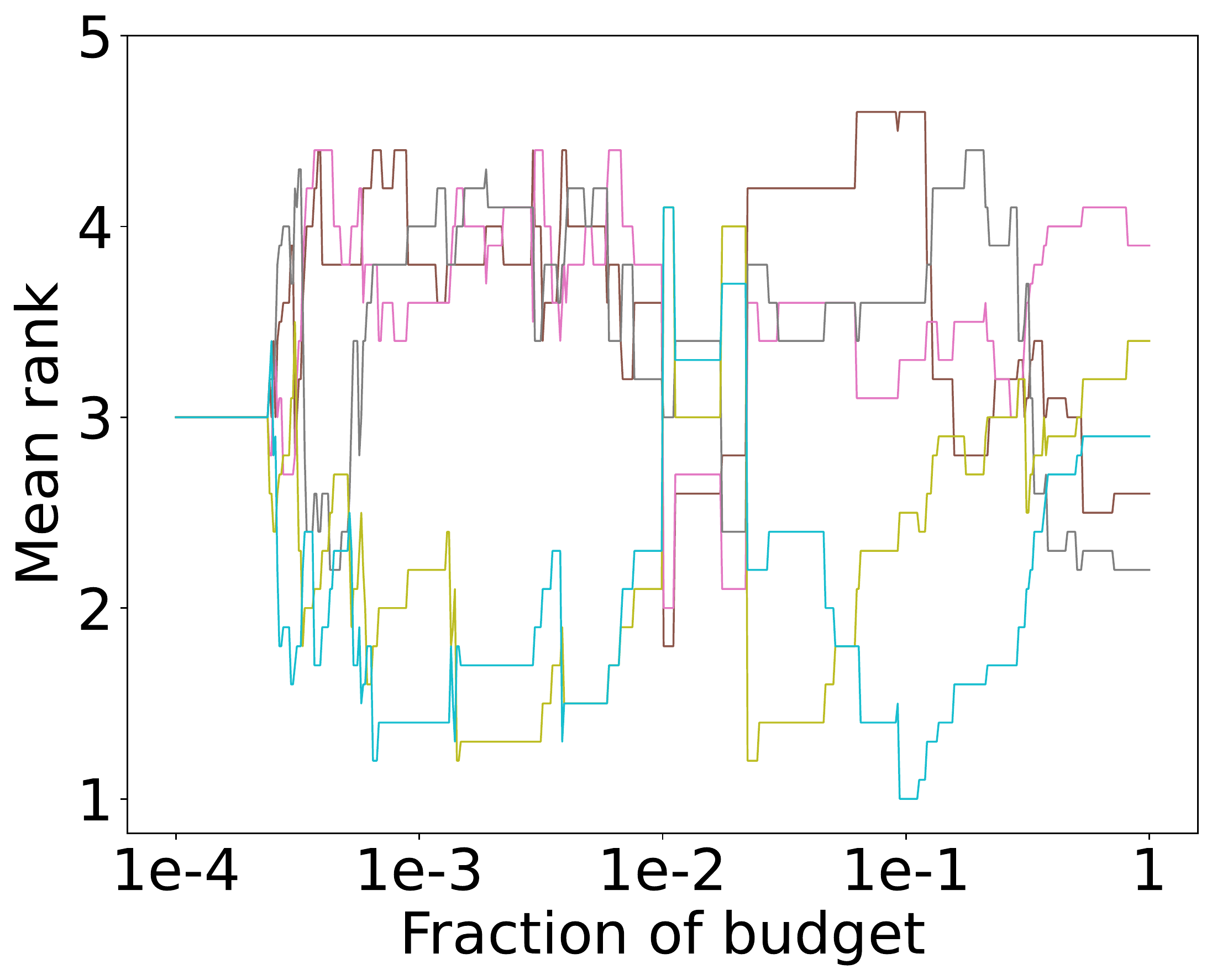}
		\caption{$\textit{MF}_{\text{PubMed}}$}
		\label{fig:MF_Pubmed_avg}
	\end{subfigure}
	
	\centering
	\hspace*{1.2cm}\begin{subfigure}{1.0\linewidth}
		\centering
		\includegraphics[width=0.95\linewidth]{materials/legend_rank_new.pdf}
	\end{subfigure}
	\vspace{-0.1in}
	
	\caption{Mean rank over time on GNN benchmark (FedAvg).}
	\label{fig:entire_gcn_tabular_avg_rank}
\end{figure}

\begin{figure}[htbp]
	\centering
	\begin{subfigure}{0.25\linewidth}
		\centering
		\includegraphics[width=0.9\linewidth]{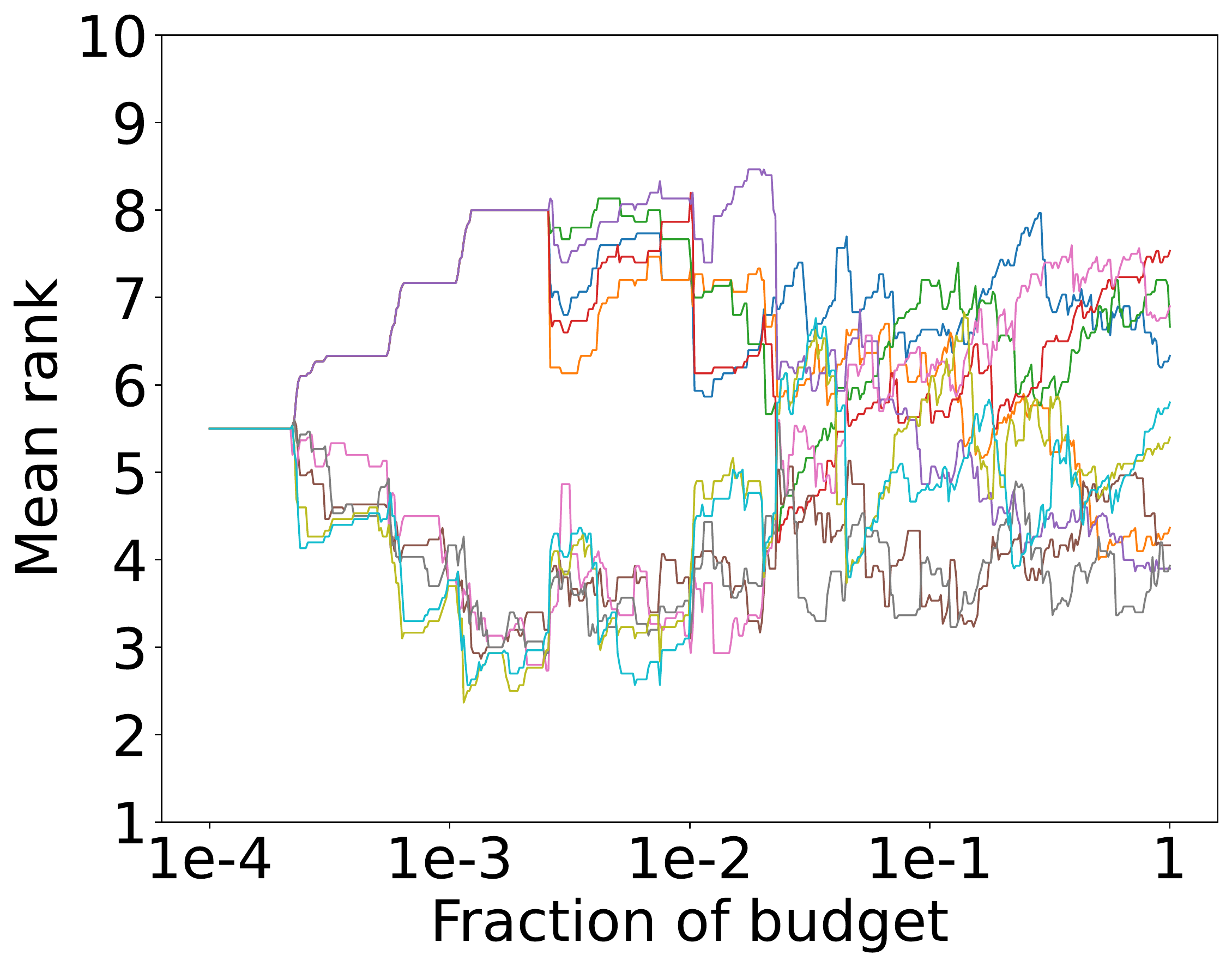}
		\caption{$\text{ALL}_\text{{GNN}}$}
		\label{fig:All_GNN_opt}
	\end{subfigure}
	\begin{subfigure}{0.25\linewidth}
		\centering
		\includegraphics[width=0.9\linewidth]{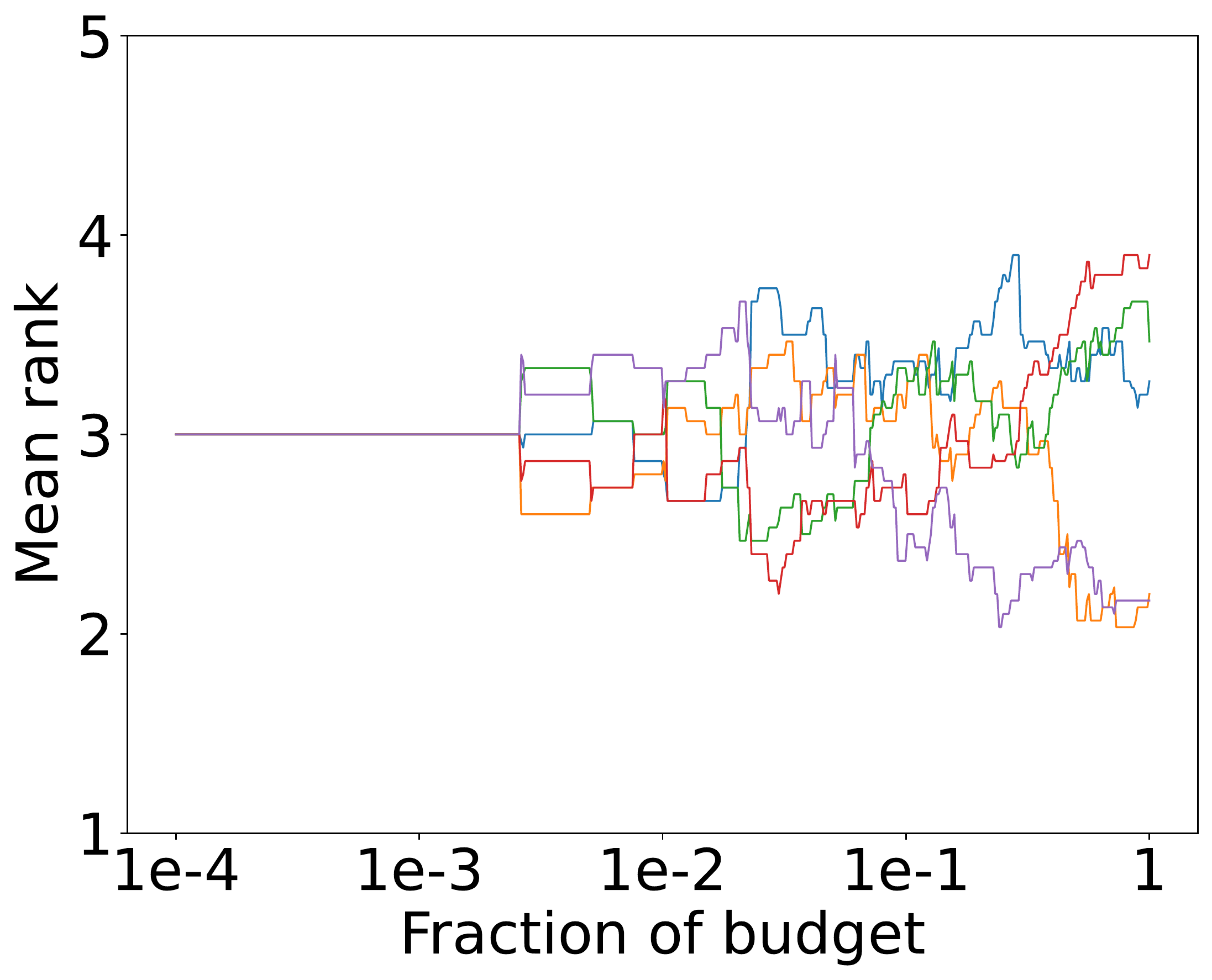}
		\caption{$\textit{BBO}_\text{{GNN}}$}
		\label{fig:BBO_GNN_opt}
	\end{subfigure}
	\begin{subfigure}{0.25\linewidth}
		\centering
		\includegraphics[width=0.9\linewidth]{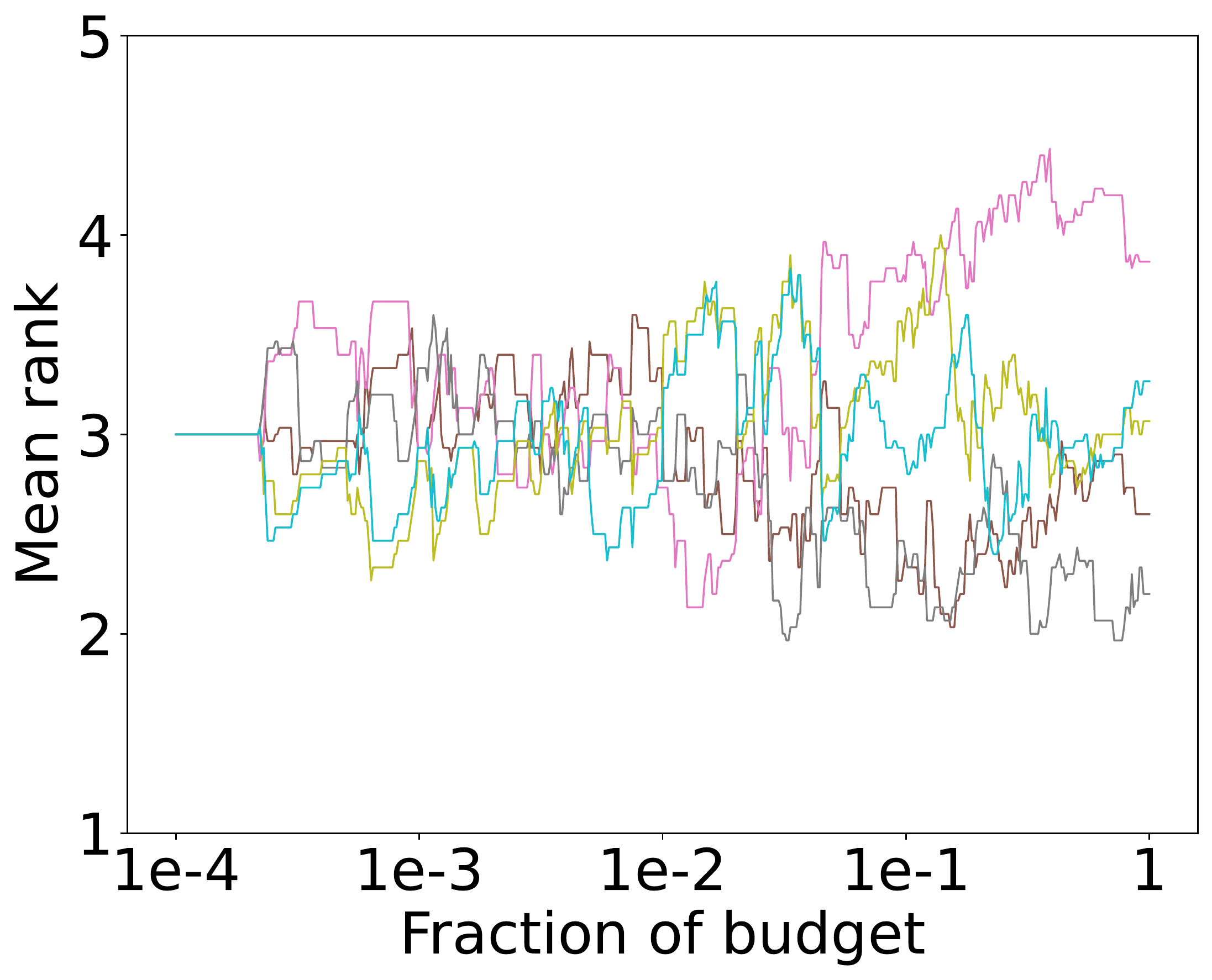}
		\caption{$\textit{MF}_\text{{GNN}}$}
		\label{fig:MF_GNN_opt}
	\end{subfigure}

	\begin{subfigure}{0.25\linewidth}
		\centering
		\includegraphics[width=0.9\linewidth]{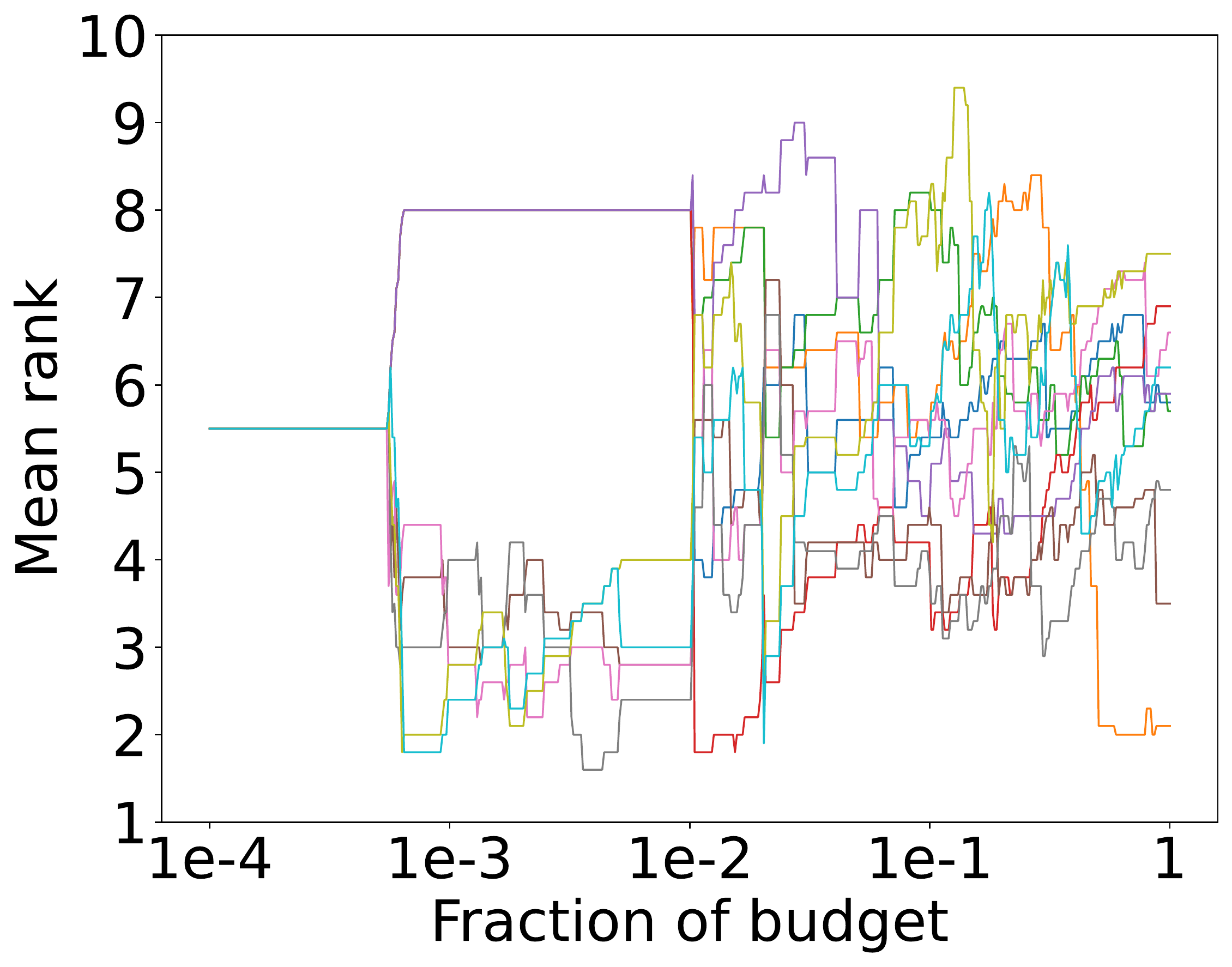}
		\caption{$\text{ALL}_{\text{Cora}}$}
		\label{fig:All_Cora_opt}
	\end{subfigure}
	\begin{subfigure}{0.25\linewidth}
		\centering
		\includegraphics[width=0.9\linewidth]{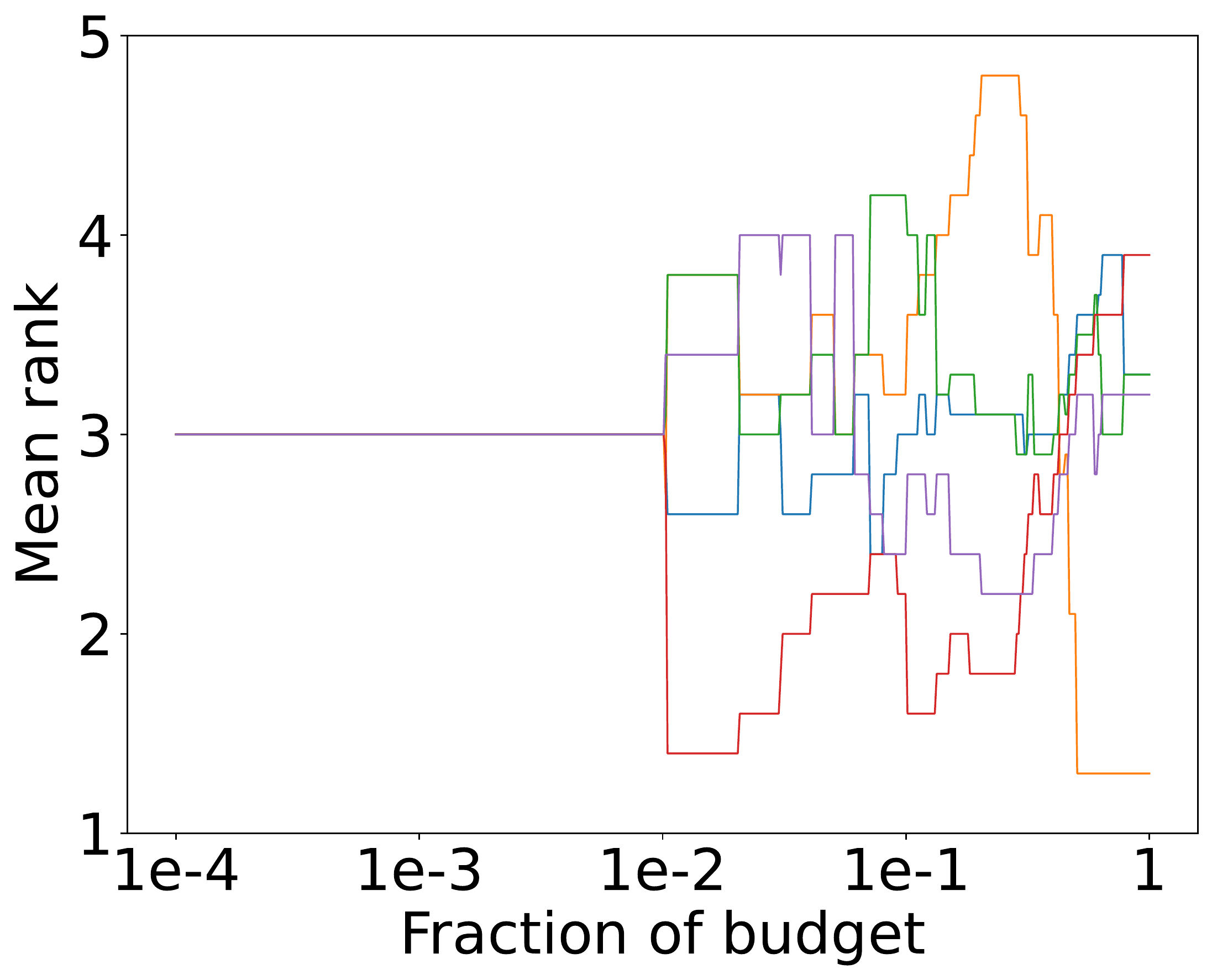}
		\caption{$\textit{BBO}_{\text{Cora}}$}
		\label{fig:BBO_Cora_opt}
	\end{subfigure}
	\begin{subfigure}{0.25\linewidth}
		\centering
		\includegraphics[width=0.9\linewidth]{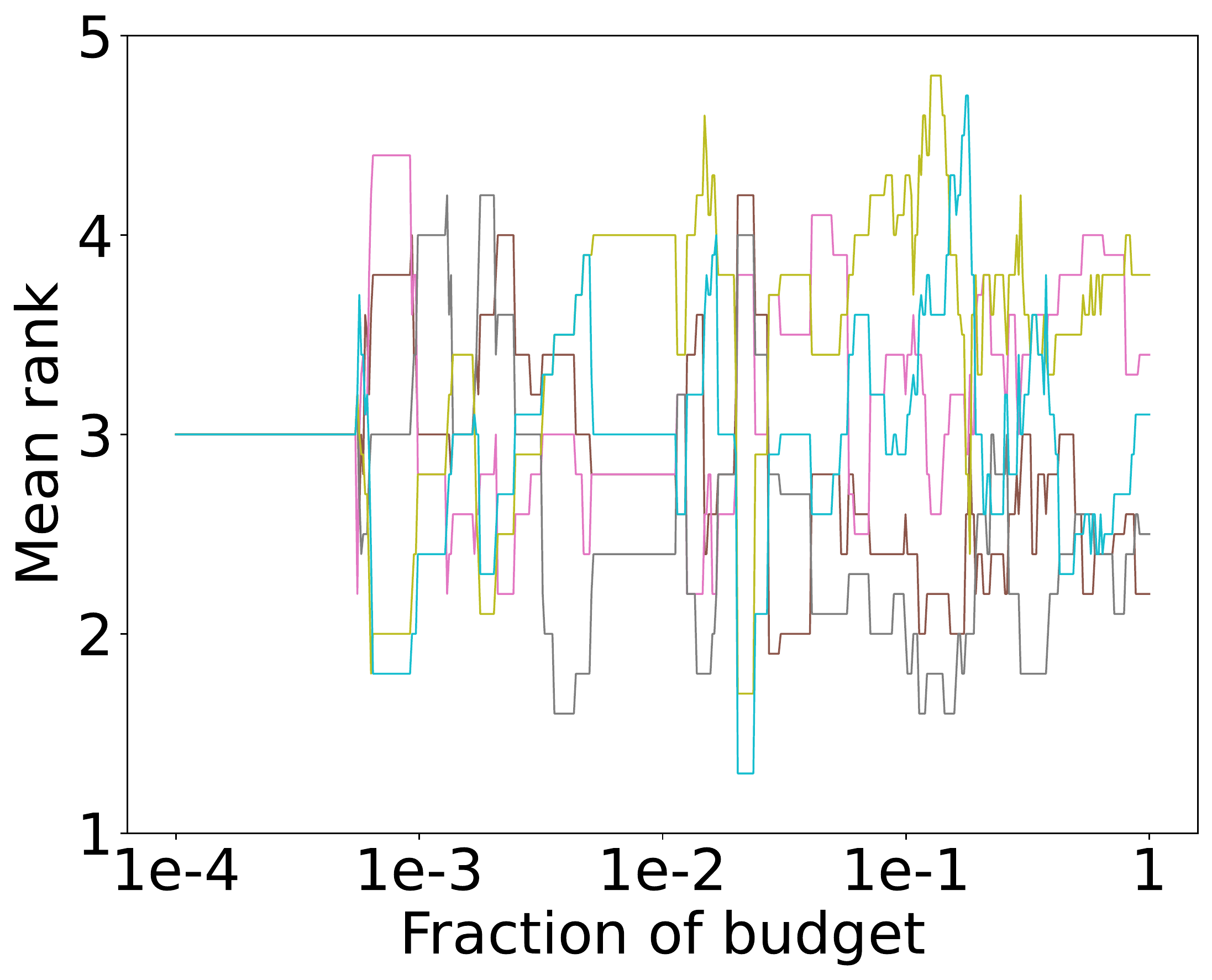}
		\caption{$\textit{MF}_{\text{Cora}}$}
		\label{fig:MF_Cora_opt}
	\end{subfigure}
	
	\begin{subfigure}{0.25\linewidth}
		\centering
		\includegraphics[width=0.9\linewidth]{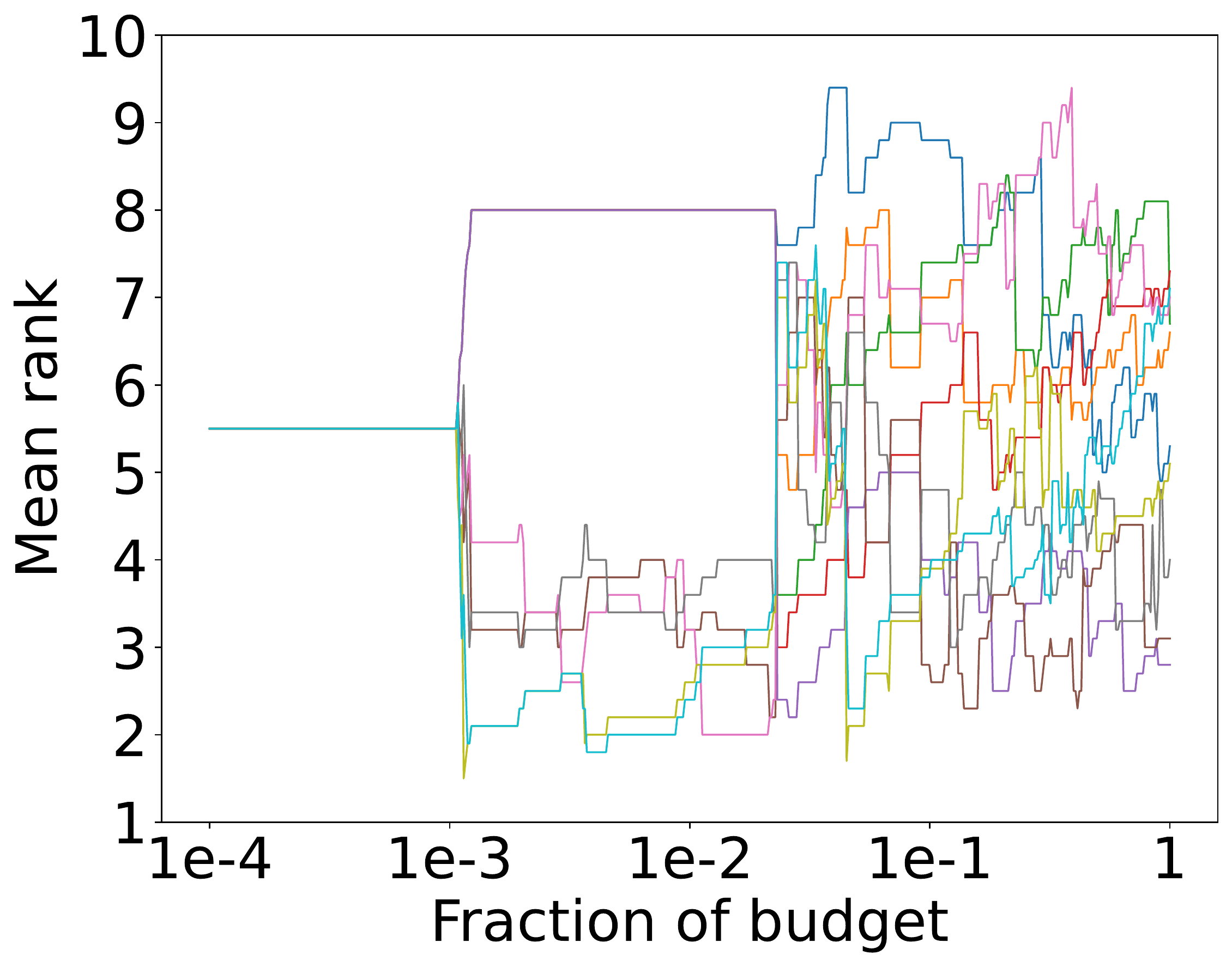}
		\caption{$\text{ALL}_{\text{CiteSeer}}$}
		\label{fig:All_Citeseer_opt}
	\end{subfigure}
	\begin{subfigure}{0.25\linewidth}
		\centering
		\includegraphics[width=0.9\linewidth]{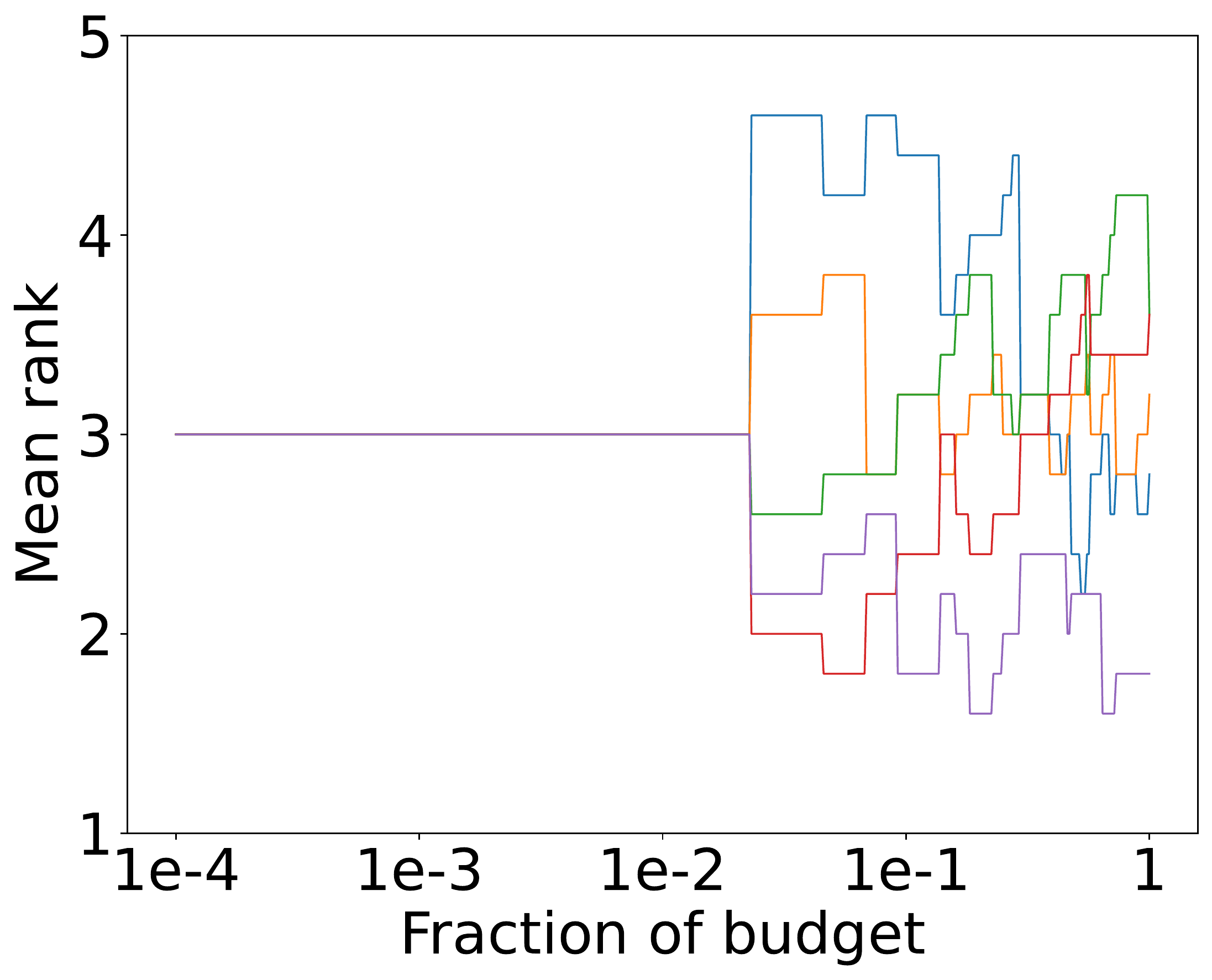}
		\caption{$\textit{BBO}_{\text{CiteSeer}}$}
		\label{fig:BBO_Citeseer_opt}
	\end{subfigure}
	\begin{subfigure}{0.25\linewidth}
		\centering
		\includegraphics[width=0.9\linewidth]{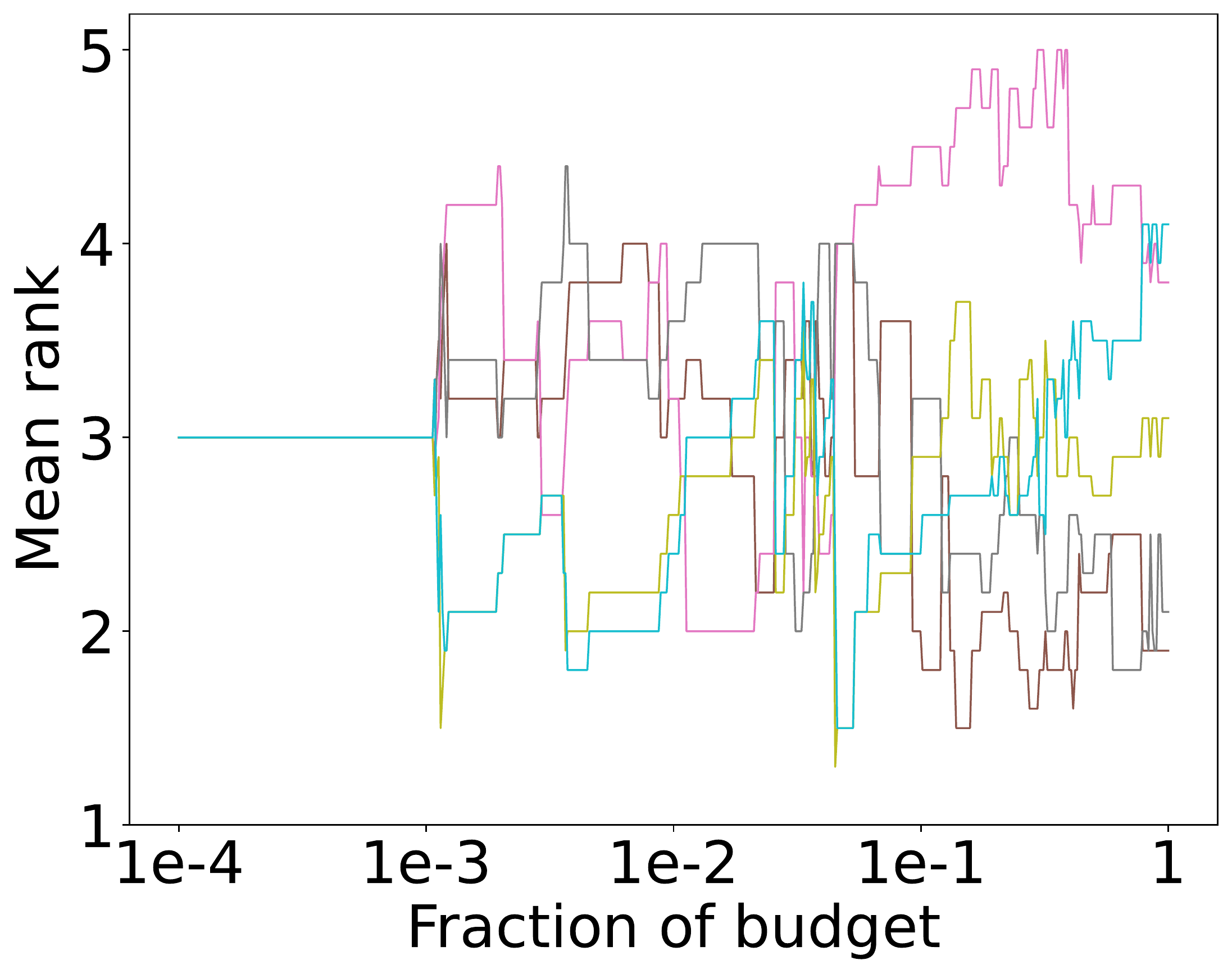}
		\caption{$\textit{MF}_{\text{CiteSeer}}$}
		\label{fig:MF_Citeseer_opt}
	\end{subfigure}
	
	\begin{subfigure}{0.25\linewidth}
		\centering
		\includegraphics[width=0.9\linewidth]{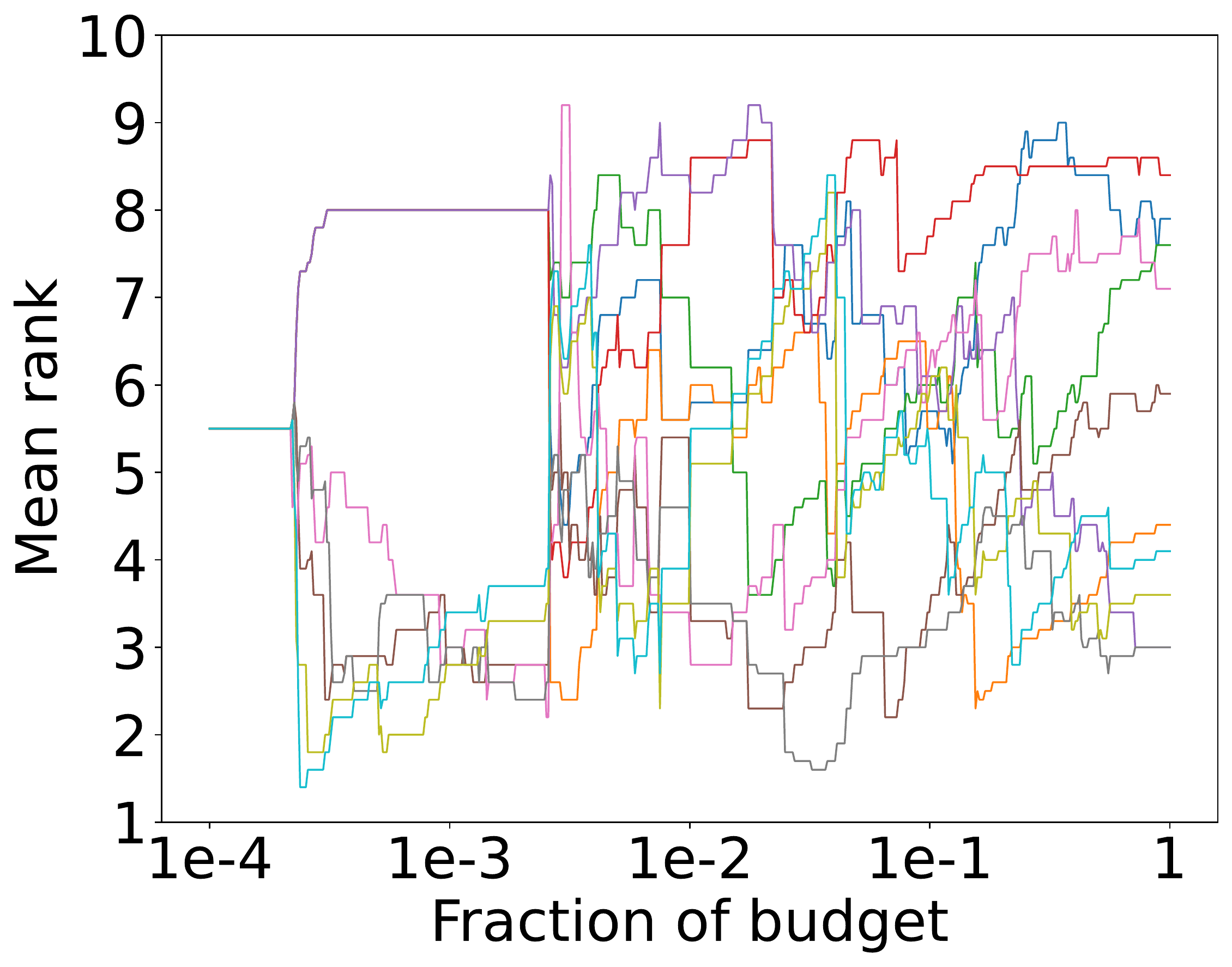}
		\caption{$\text{ALL}_{\text{PubMed}}$}
		\label{fig:All_Pubmed_opt}
	\end{subfigure}
	\begin{subfigure}{0.25\linewidth}
		\centering
		\includegraphics[width=0.9\linewidth]{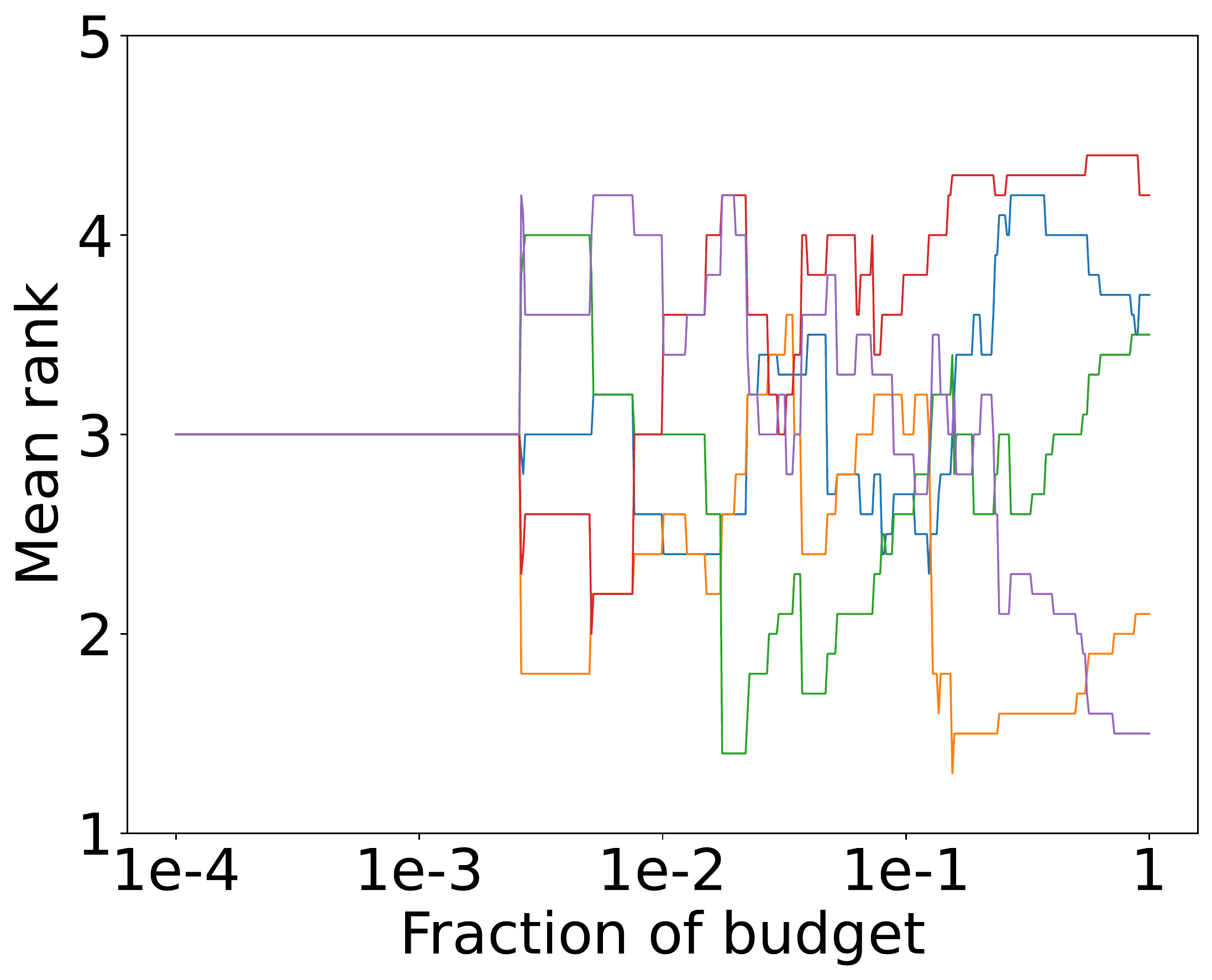}
		\caption{$\textit{BBO}_{\text{PubMed}}$}
		\label{fig:BBO_Pubmed_opt}
	\end{subfigure}
	\begin{subfigure}{0.25\linewidth}
		\centering
		\includegraphics[width=0.9\linewidth]{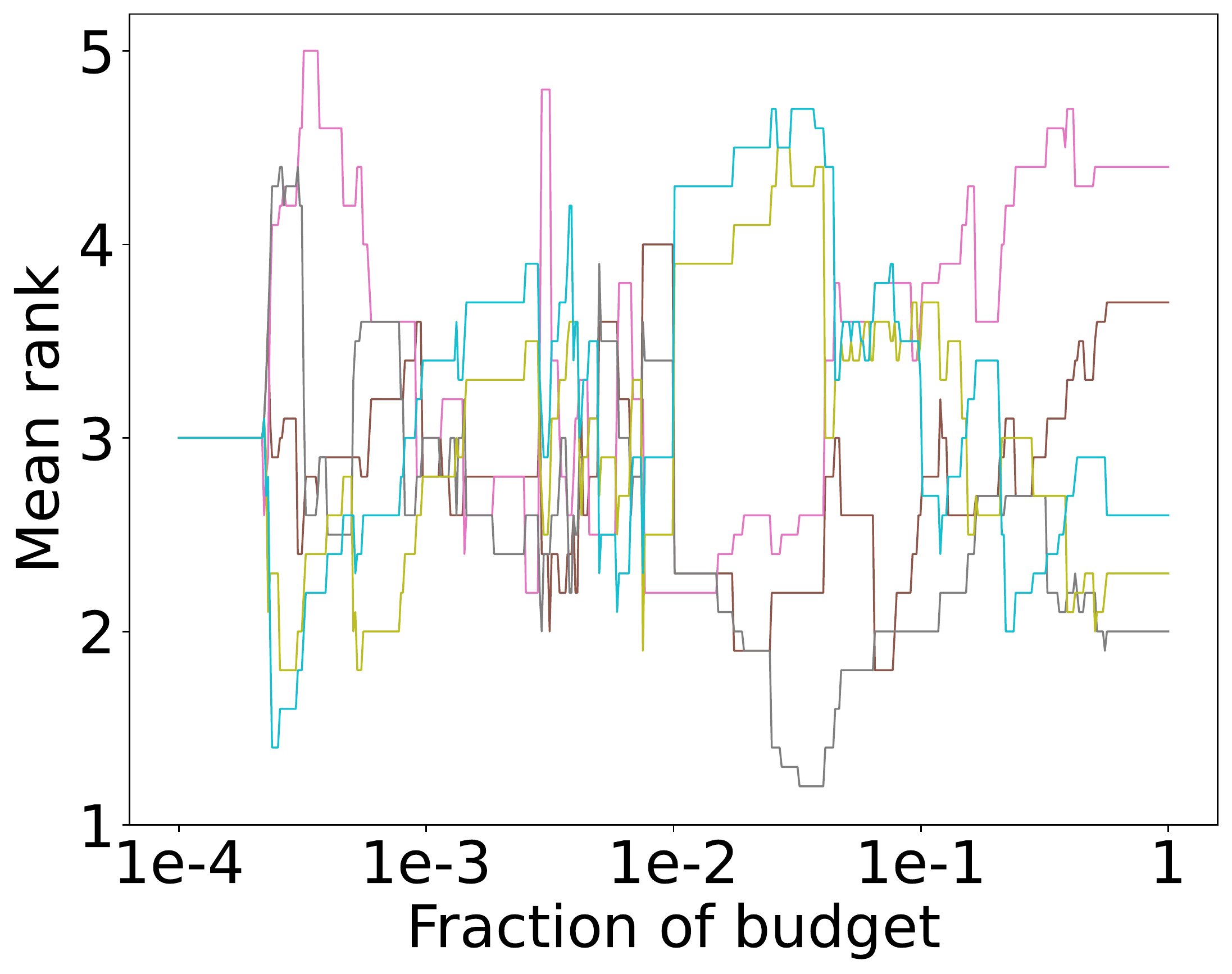}
		\caption{$\textit{MF}_{\text{PubMed}}$}
		\label{fig:MF_Pubmed_opt}
	\end{subfigure}
	
	\centering
	\hspace*{1.2cm}\begin{subfigure}{1.0\linewidth}
		\centering
		\includegraphics[width=0.95\linewidth]{materials/legend_rank_new.pdf}
	\end{subfigure}
	\vspace{-0.1in}
	
	\caption{Mean rank over time on GNN benchmark (FedOPT).}
	\label{fig:entire_gcn_tabular_opt_rank}
\end{figure}

\begin{figure}[htbp]
	\centering
	\begin{subfigure}{0.25\linewidth}
		\centering
		\includegraphics[width=0.9\linewidth]{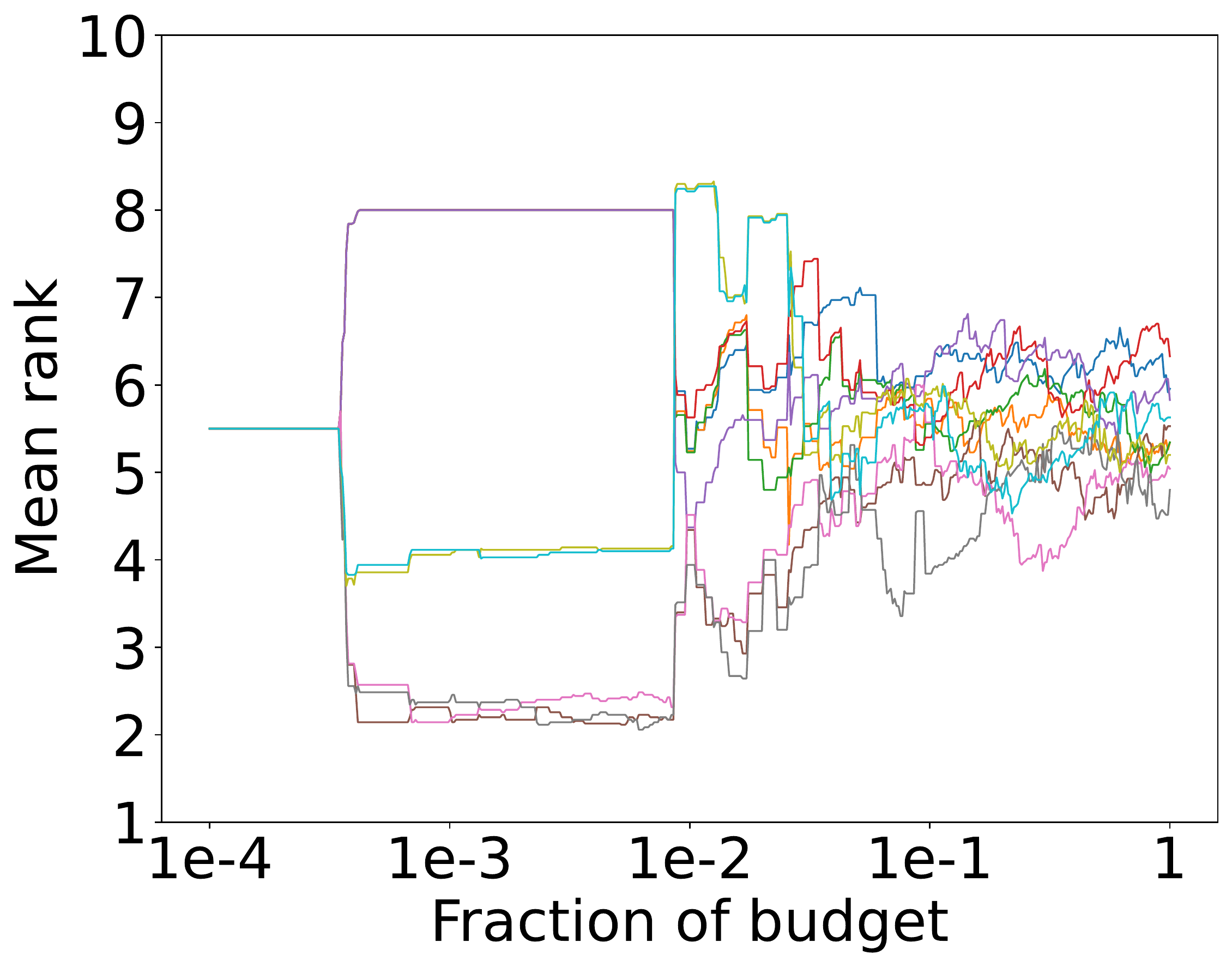}
		\caption{$\text{ALL}_{LR}$}
		\label{fig:All_LR_avg}
	\end{subfigure}
	\begin{subfigure}{0.25\linewidth}
		\centering
		\includegraphics[width=0.9\linewidth]{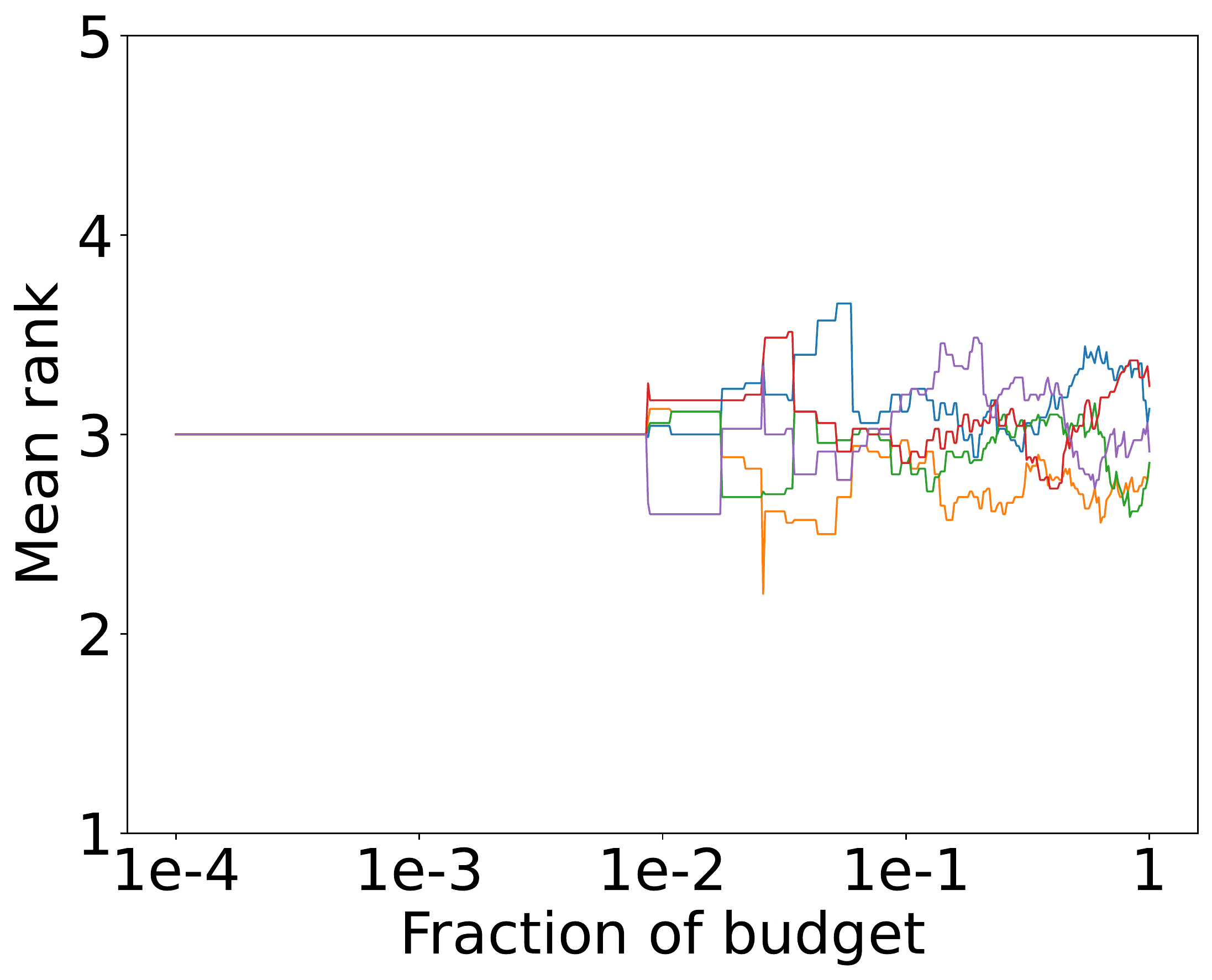}
		\caption{$\textit{BBO}_{LR}$}
		\label{fig:BBO_LR_avg}
	\end{subfigure}
	\begin{subfigure}{0.25\linewidth}
		\centering
		\includegraphics[width=0.9\linewidth]{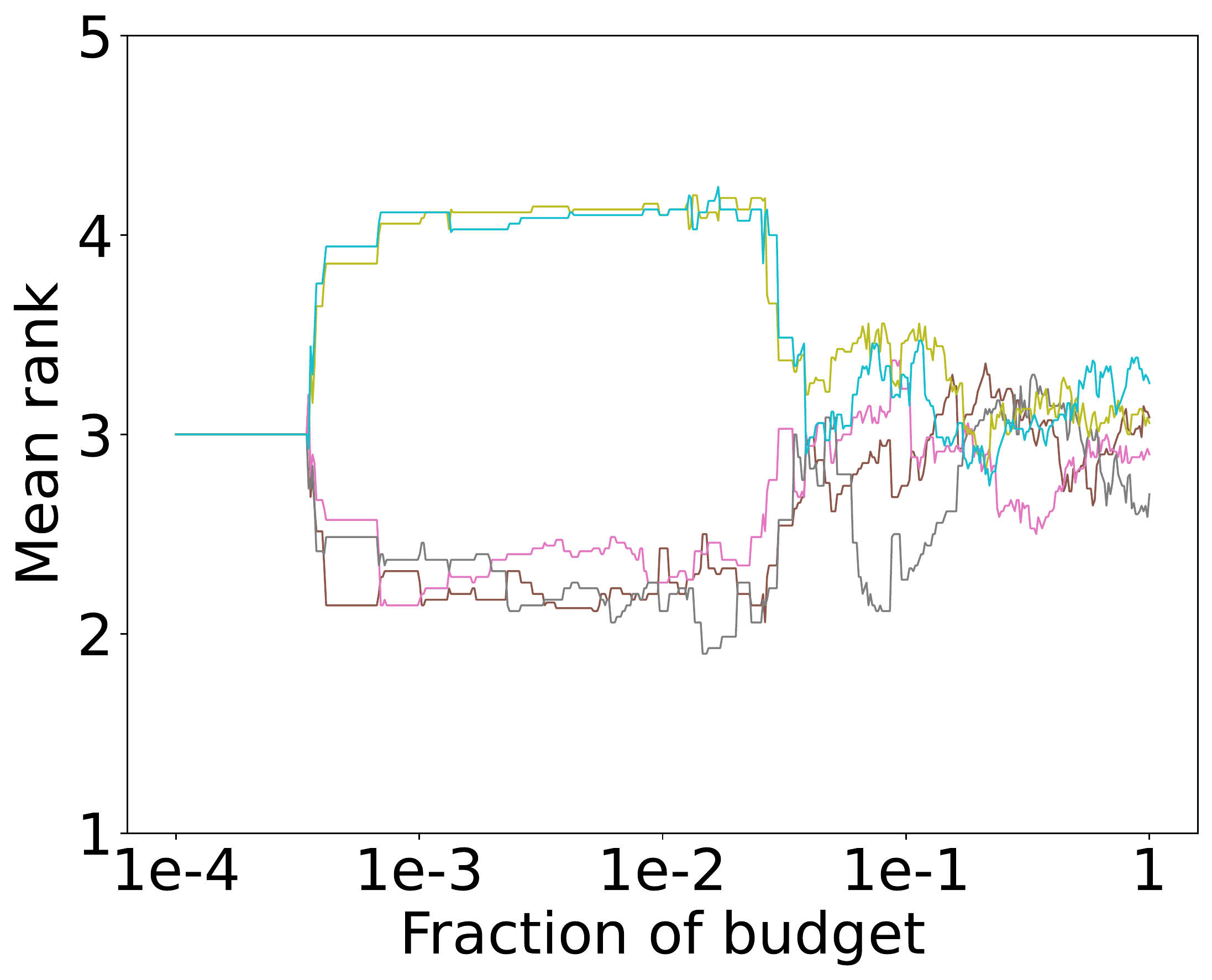}
		\caption{$\textit{MF}_{LR}$}
		\label{fig:MF_LR_avg}
	\end{subfigure}

	\begin{subfigure}{0.25\linewidth}
		\centering
		\includegraphics[width=0.9\linewidth]{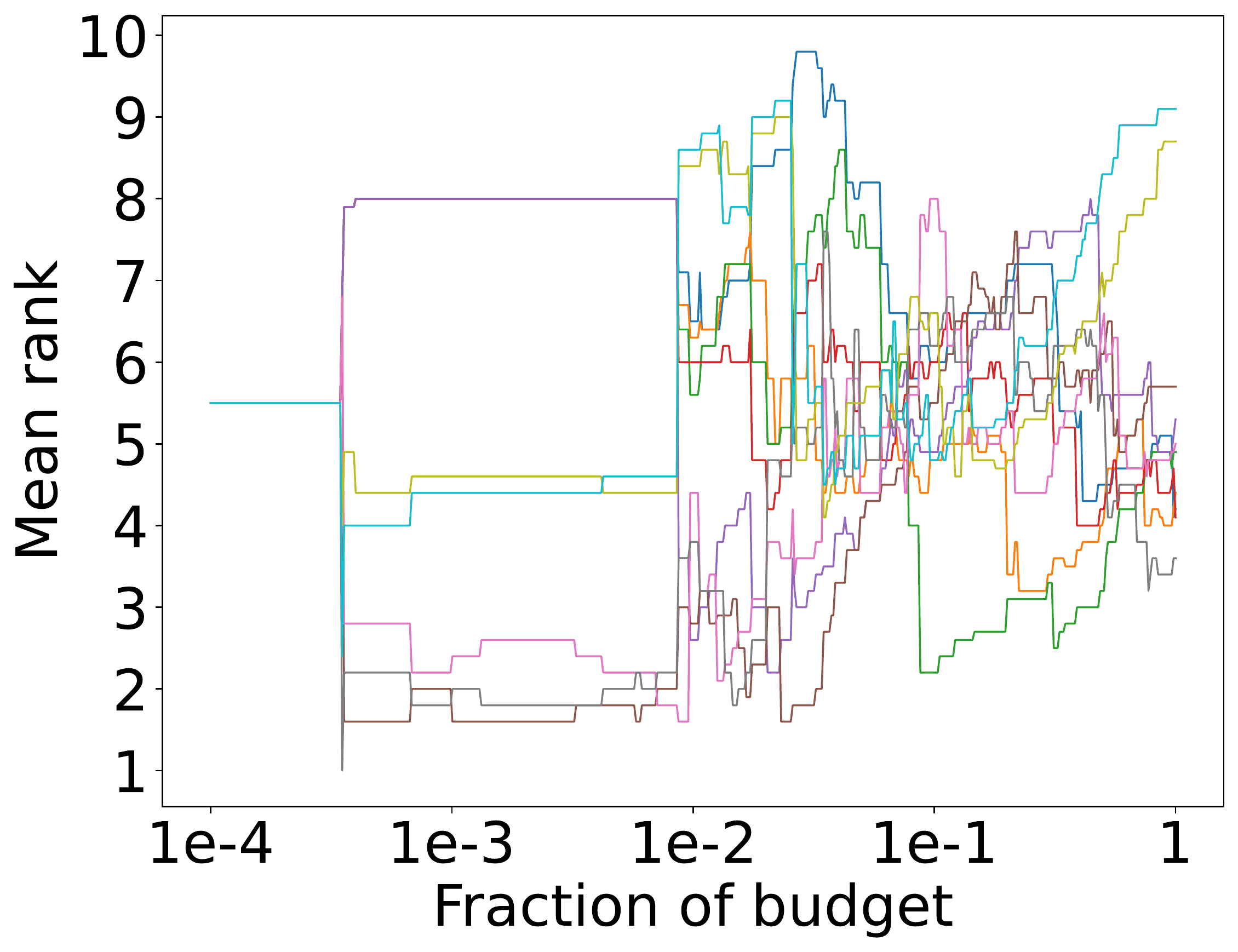}
		\caption{$\text{ALL}_{31\text{OpenML}}$}
		\label{fig:All_31_openml_avg}
	\end{subfigure}
	\begin{subfigure}{0.25\linewidth}
		\centering
		\includegraphics[width=0.9\linewidth]{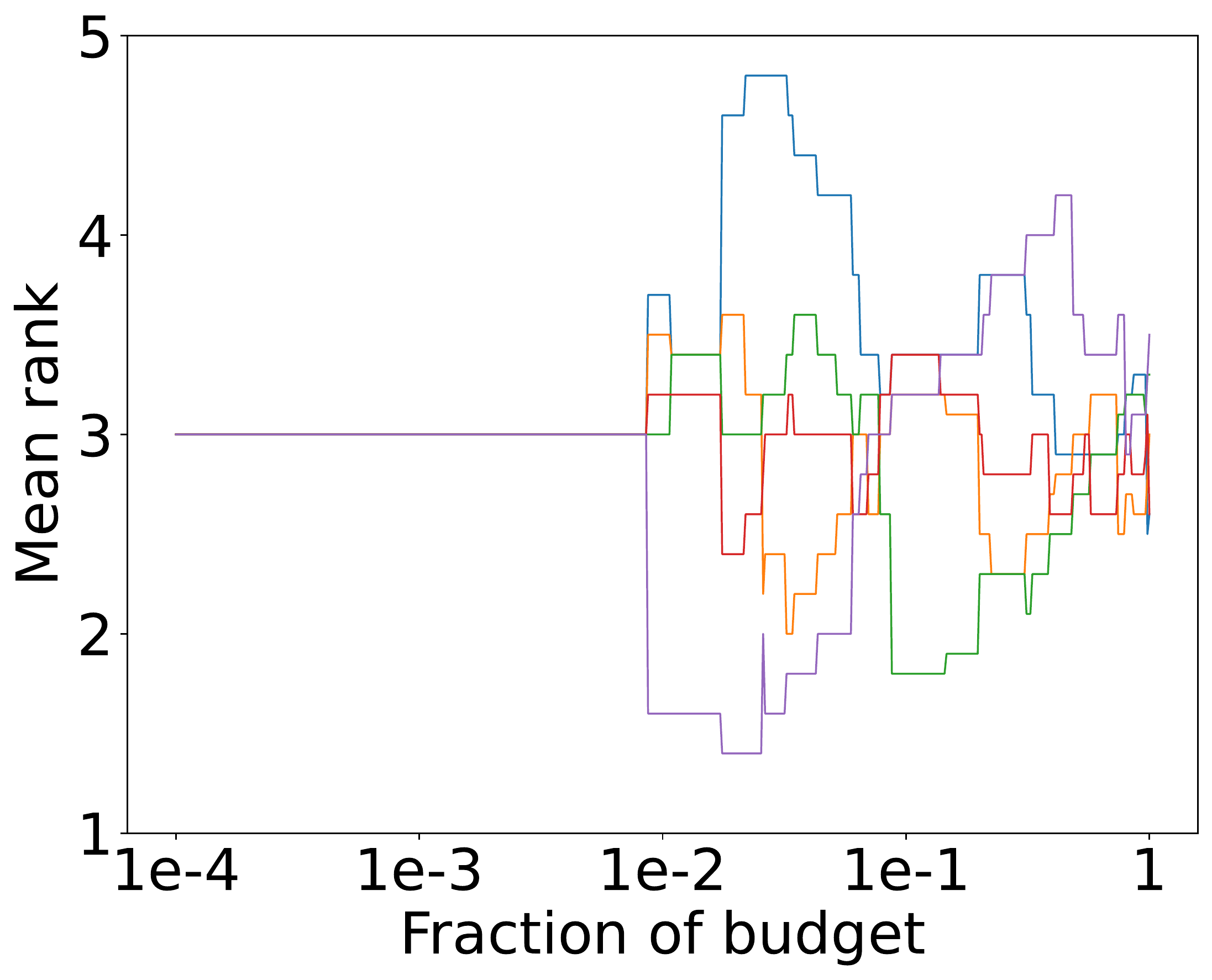}
		\caption{$\textit{BBO}_{31\text{OpenML}}$}
		\label{fig:BBO_31_openml_avg}
	\end{subfigure}
	\begin{subfigure}{0.25\linewidth}
		\centering
		\includegraphics[width=0.9\linewidth]{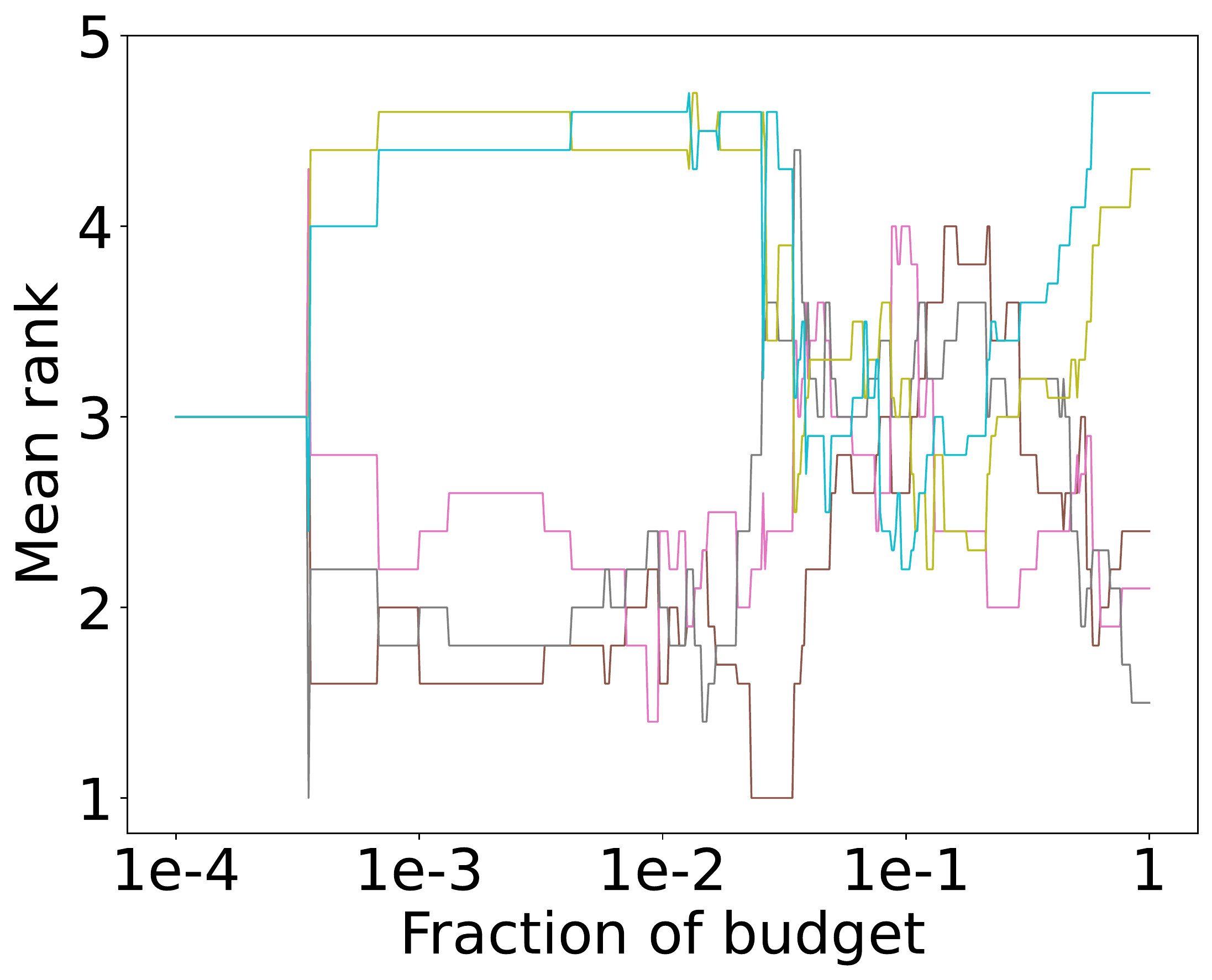}
		\caption{$\textit{MF}_{31\text{OpenML}}$}
		\label{fig:MF_31_openml_avg}
	\end{subfigure}
	
	\begin{subfigure}{0.25\linewidth}
		\centering
		\includegraphics[width=0.9\linewidth]{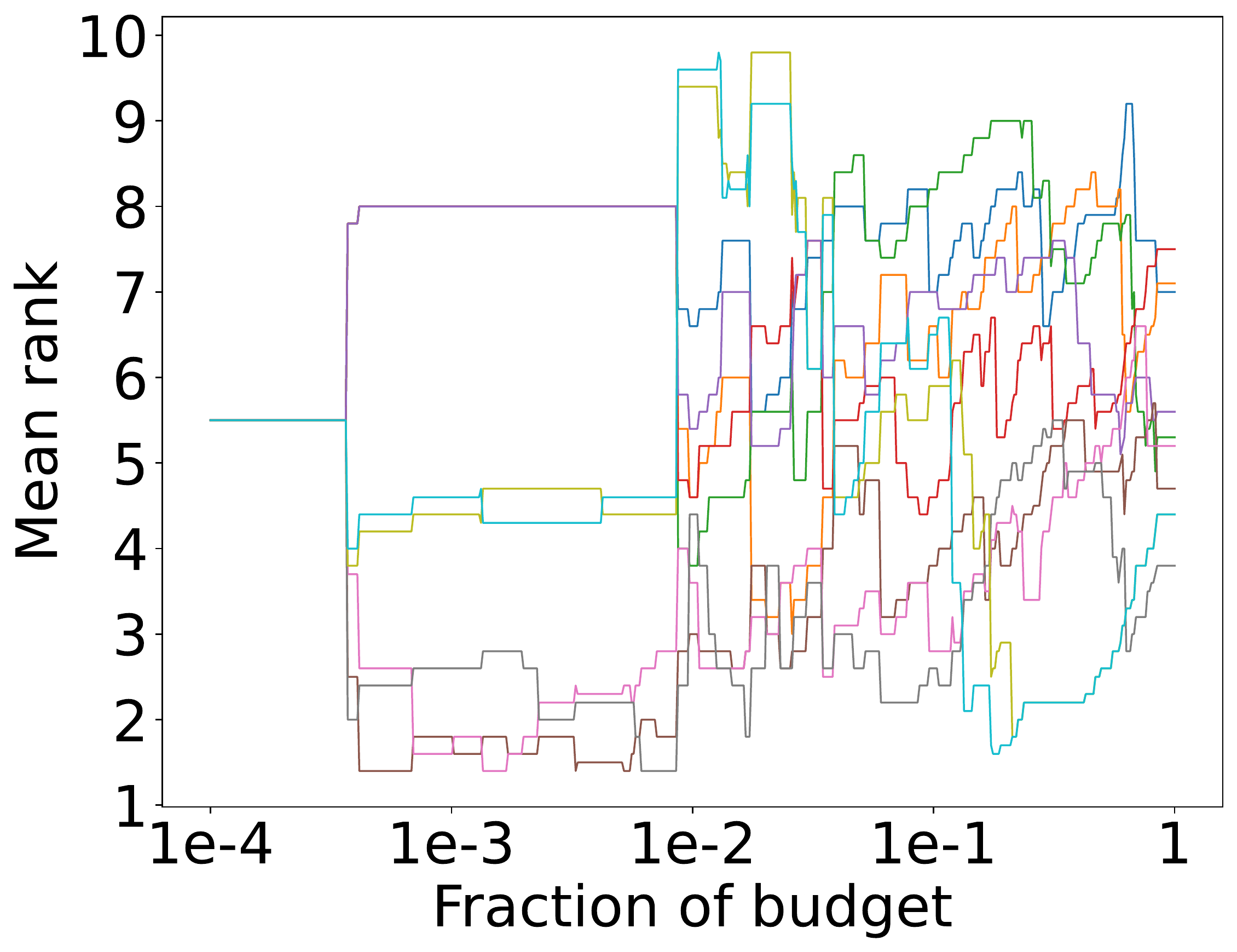}
		\caption{$\text{ALL}_{53\text{OpenML}}$}
		\label{fig:All_53_openml_avg}
	\end{subfigure}
	\begin{subfigure}{0.25\linewidth}
		\centering
		\includegraphics[width=0.9\linewidth]{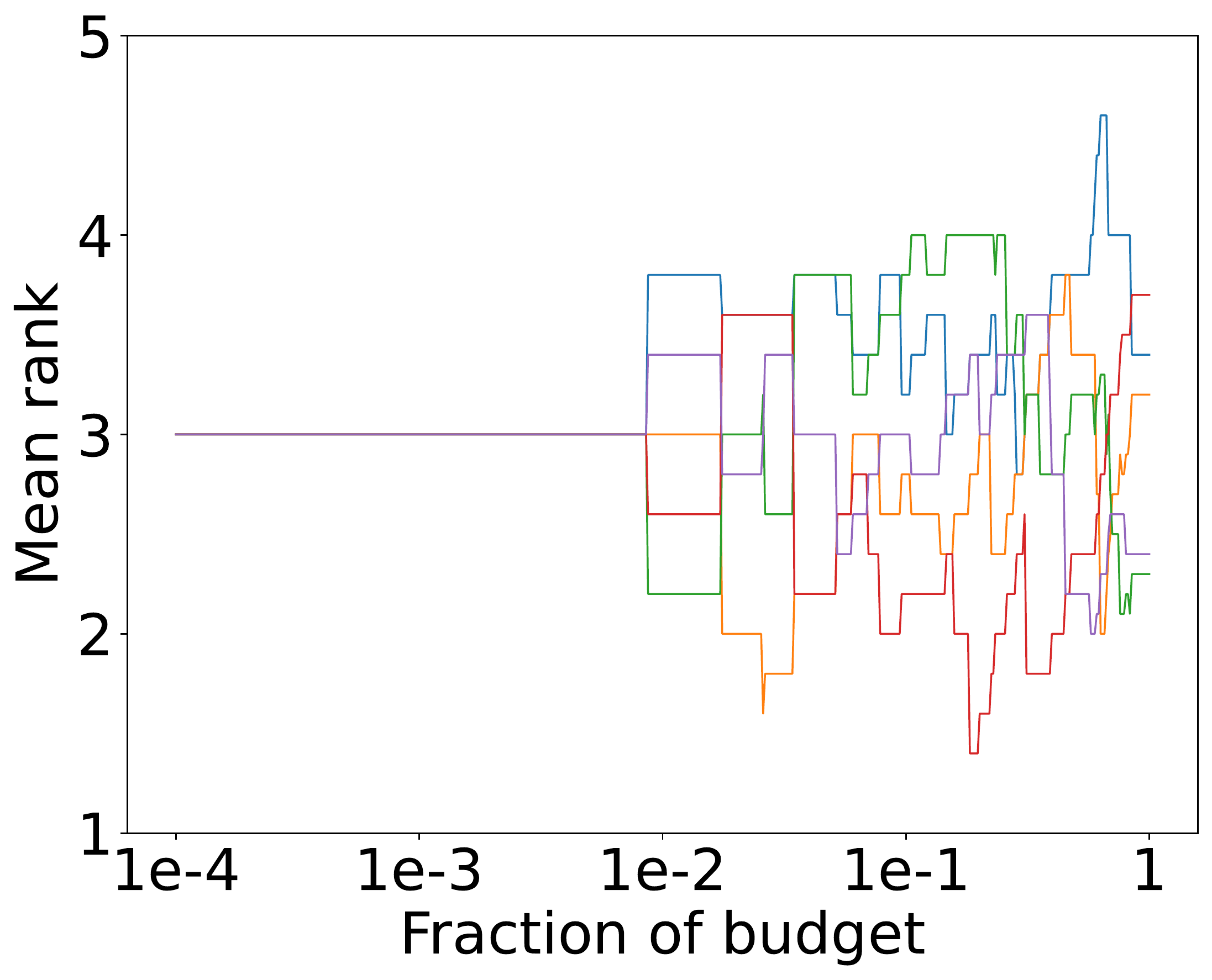}
		\caption{$\textit{BBO}_{53\text{OpenML}}$}
		\label{fig:BBO_53_openml_avg}
	\end{subfigure}
	\begin{subfigure}{0.25\linewidth}
		\centering
		\includegraphics[width=0.9\linewidth]{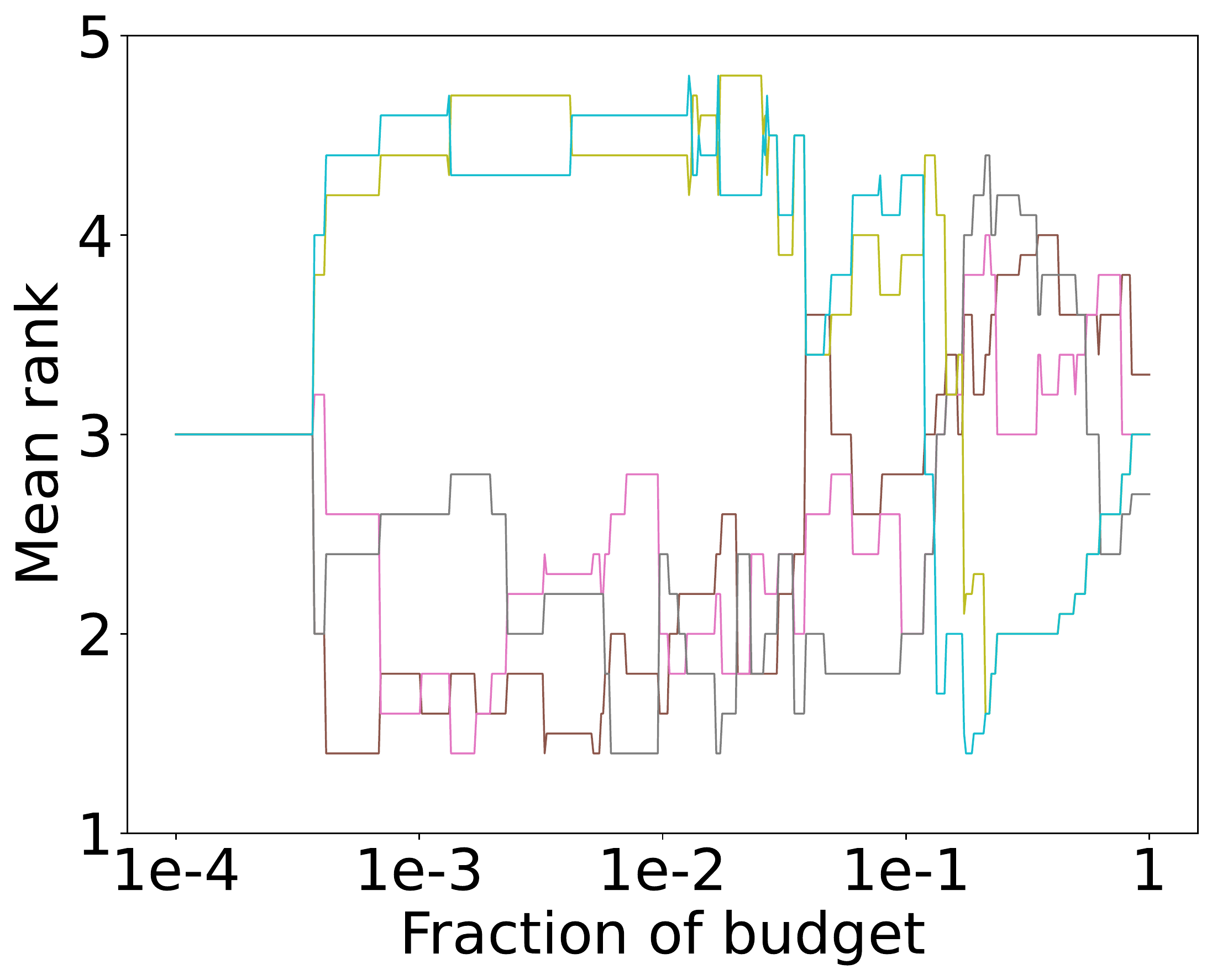}
		\caption{$\textit{MF}_{53\text{OpenML}}$}
		\label{fig:MF_53_openml_avg}
	\end{subfigure}
	
	\begin{subfigure}{0.25\linewidth}
		\centering
		\includegraphics[width=0.9\linewidth]{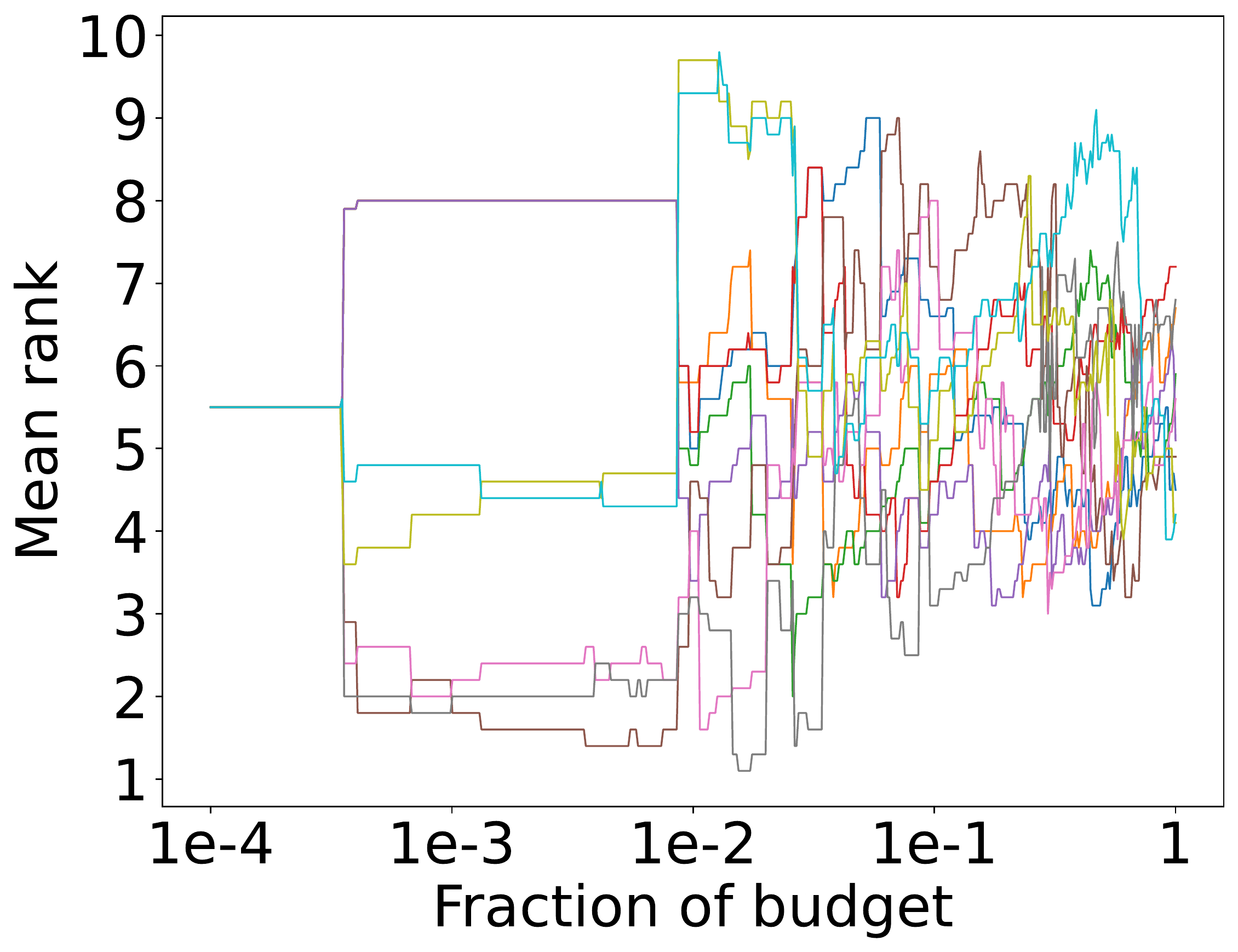}
		\caption{$\text{ALL}_{10101\text{OpenML}}$}
		\label{fig:All_10101_openml_avg}
	\end{subfigure}
	\begin{subfigure}{0.25\linewidth}
		\centering
		\includegraphics[width=0.9\linewidth]{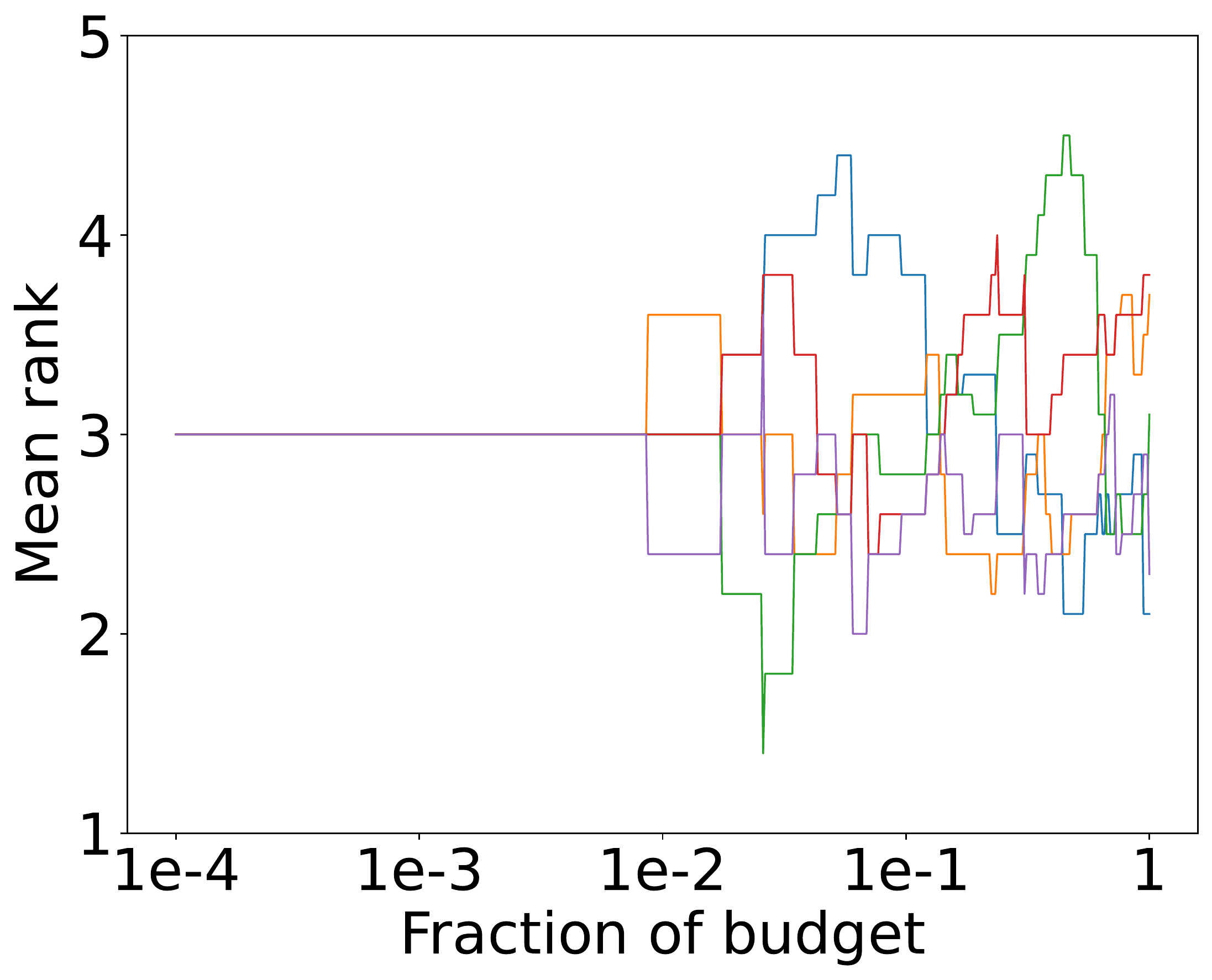}
		\caption{$\textit{BBO}_{10101\text{OpenML}}$}
		\label{fig:BBO_10101_openml_avg}
	\end{subfigure}
	\begin{subfigure}{0.25\linewidth}
		\centering
		\includegraphics[width=0.9\linewidth]{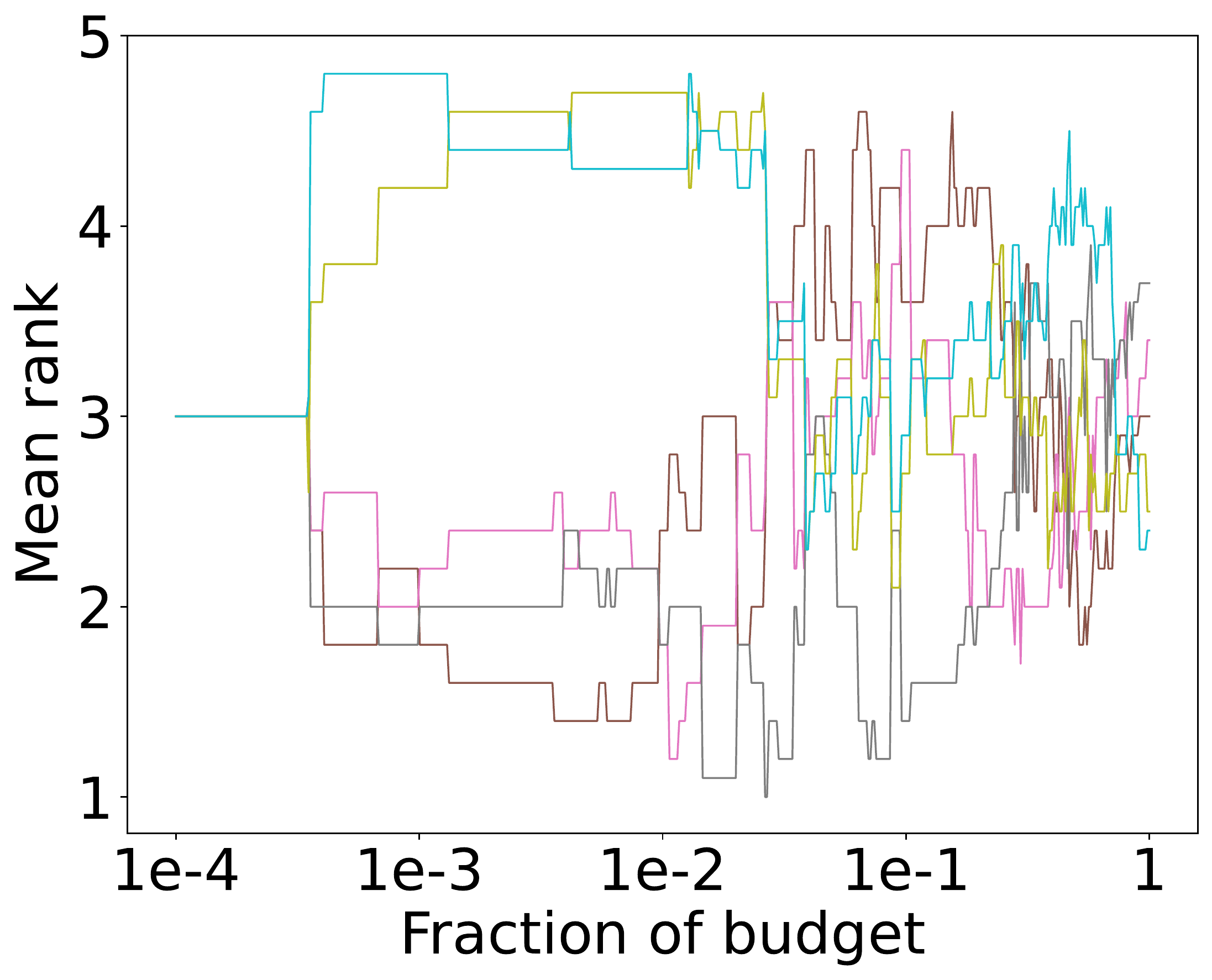}
		\caption{$\textit{MF}_{10101\text{OpenML}}$}
		\label{fig:MF_10101_openml_avg}
	\end{subfigure}
	
	\begin{subfigure}{0.25\linewidth}
		\centering
		\includegraphics[width=0.9\linewidth]{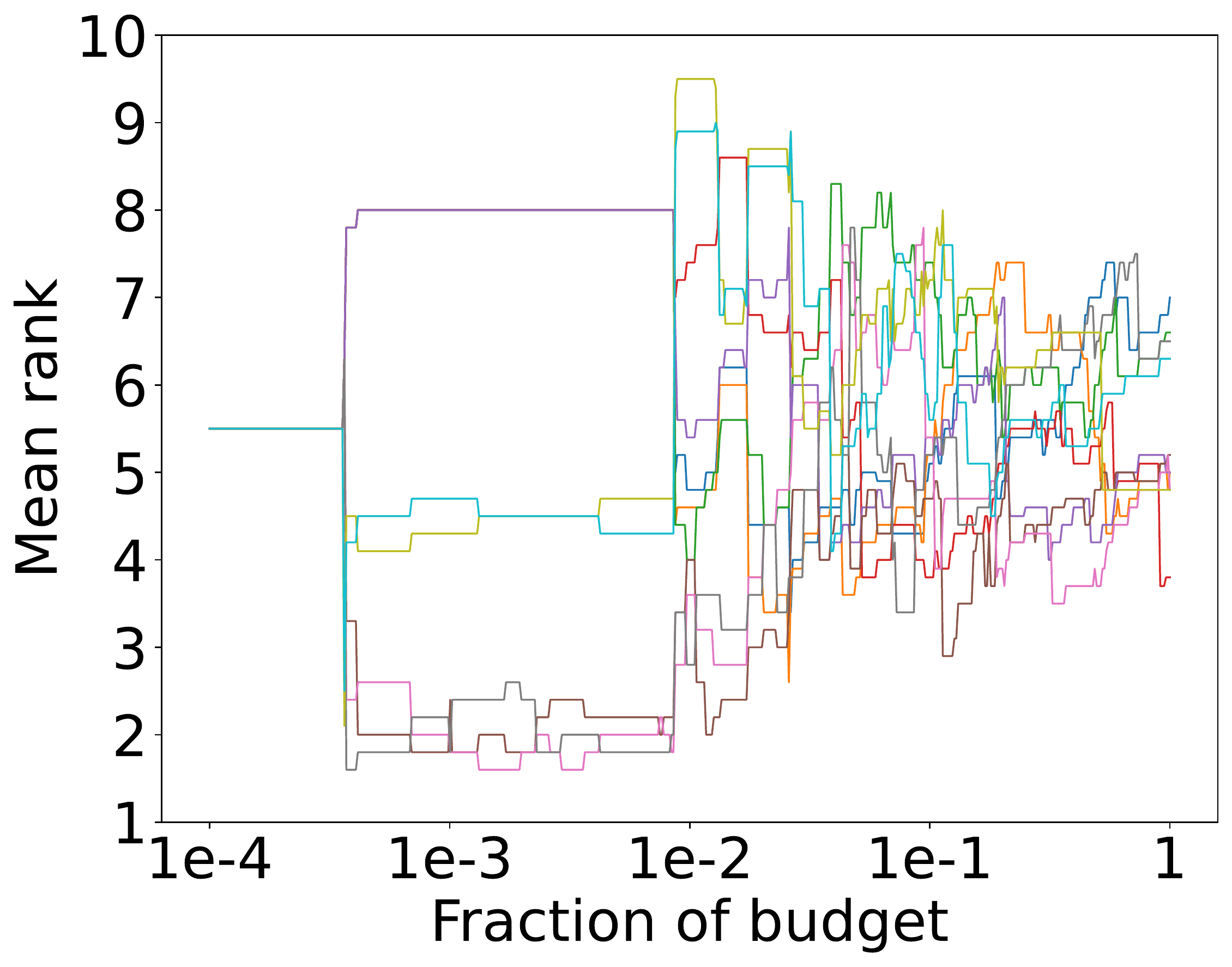}
		\caption{$\text{ALL}_{146818\text{OpenML}}$}
		\label{fig:All_146818_openml_avg}
	\end{subfigure}
	\begin{subfigure}{0.25\linewidth}
		\centering
		\includegraphics[width=0.9\linewidth]{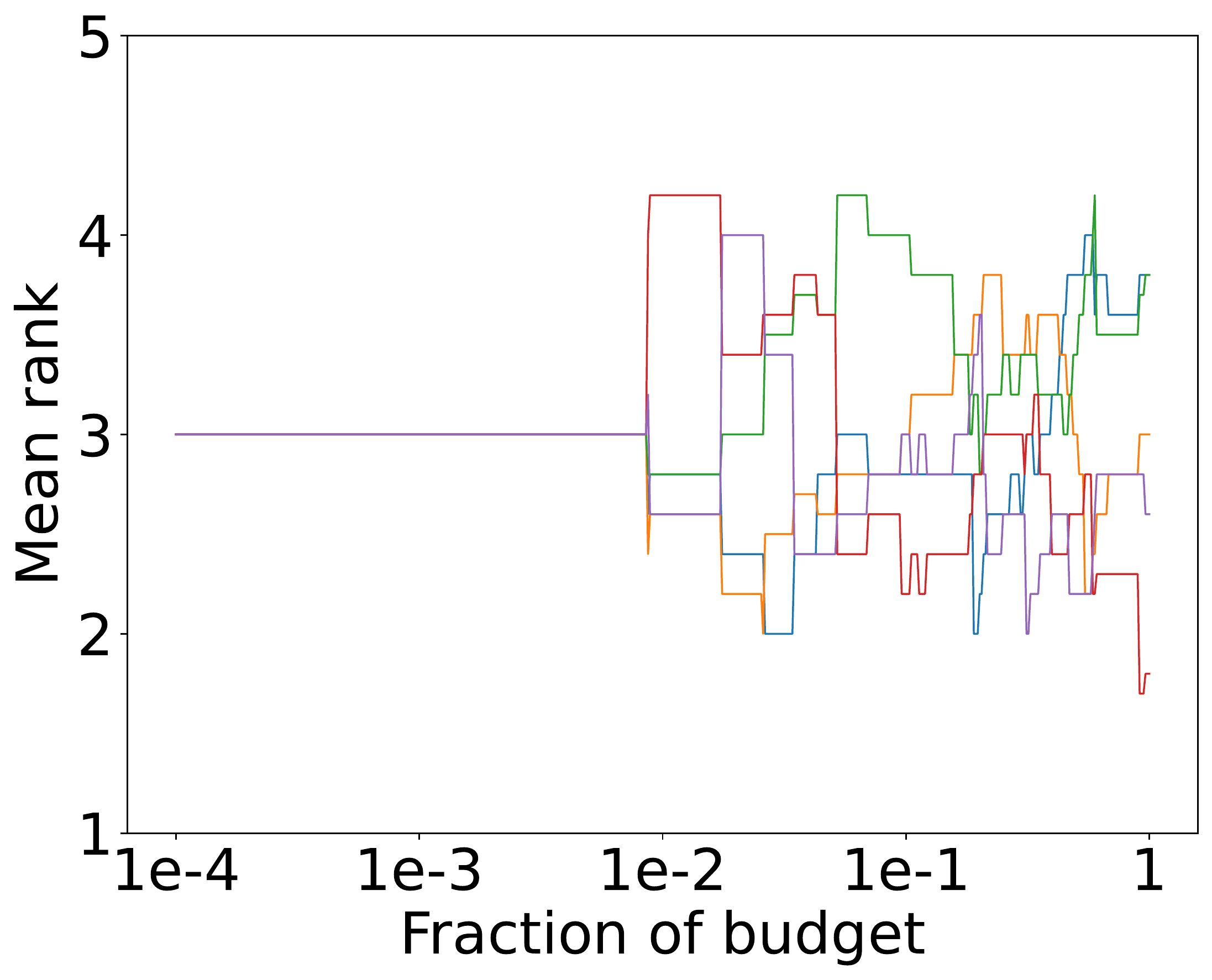}
		\caption{$\textit{BBO}_{146818\text{OpenML}}$}
		\label{fig:BBO_146818_openml_avg}
	\end{subfigure}
	\begin{subfigure}{0.25\linewidth}
		\centering
		\includegraphics[width=0.9\linewidth]{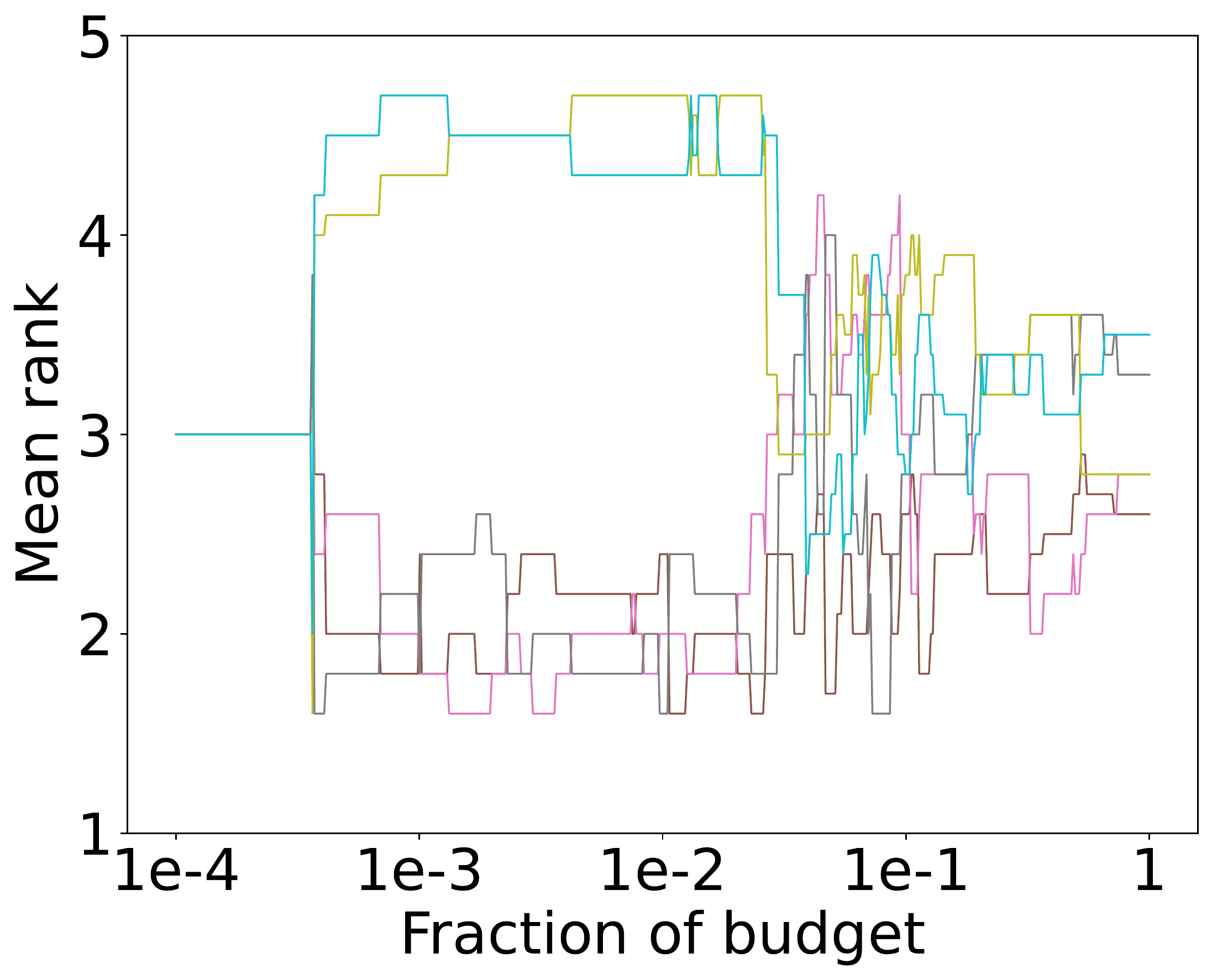}
		\caption{$\textit{MF}_{146818\text{OpenML}}$}
		\label{fig:MF_146818_openml_avg}
	\end{subfigure}
	
	\begin{subfigure}{0.25\linewidth}
		\centering
		\includegraphics[width=0.9\linewidth]{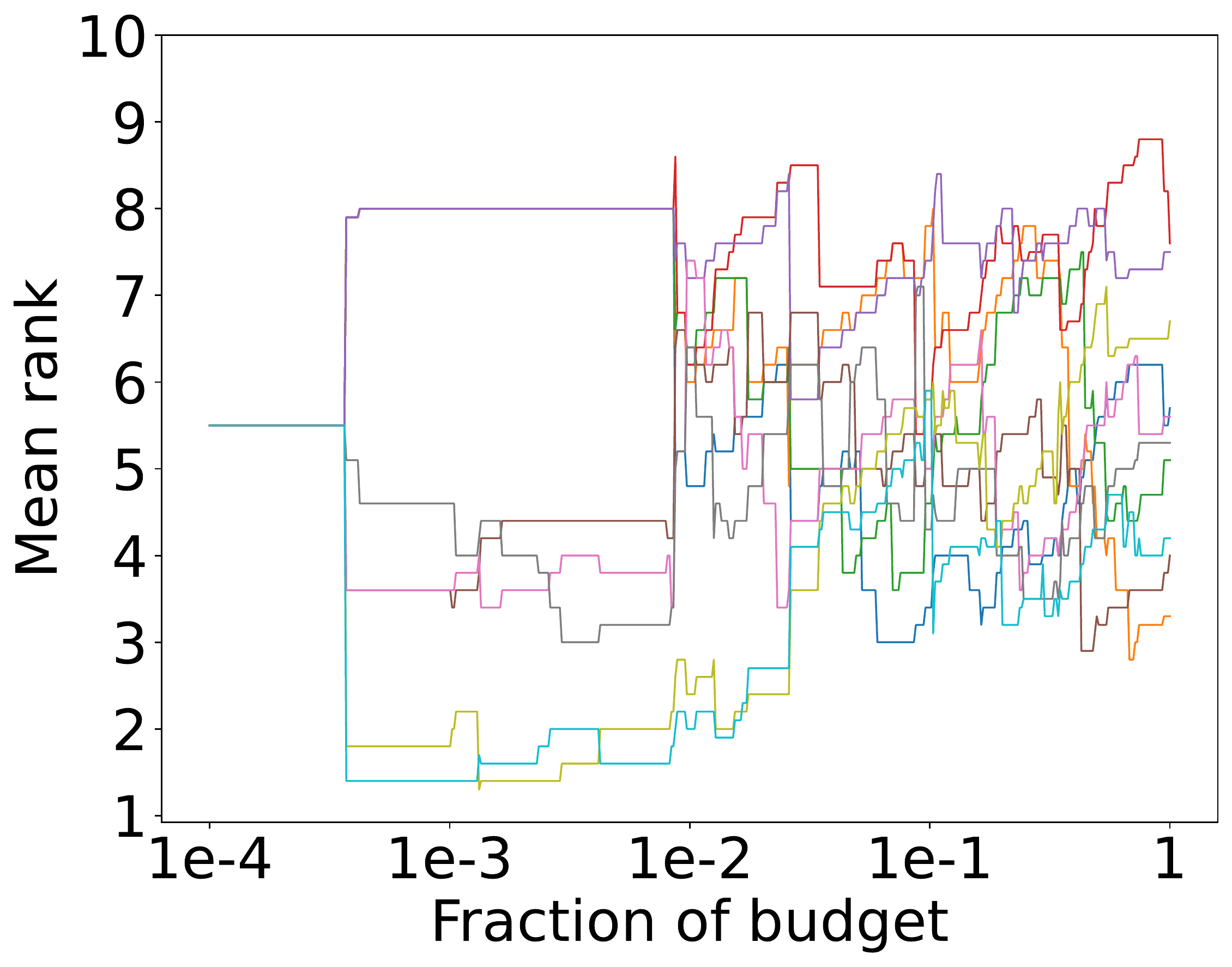}
		\caption{$\text{ALL}_{146821\text{OpenML}}$}
		\label{fig:All_146821_openml_avg}
	\end{subfigure}
	\begin{subfigure}{0.25\linewidth}
		\centering
		\includegraphics[width=0.9\linewidth]{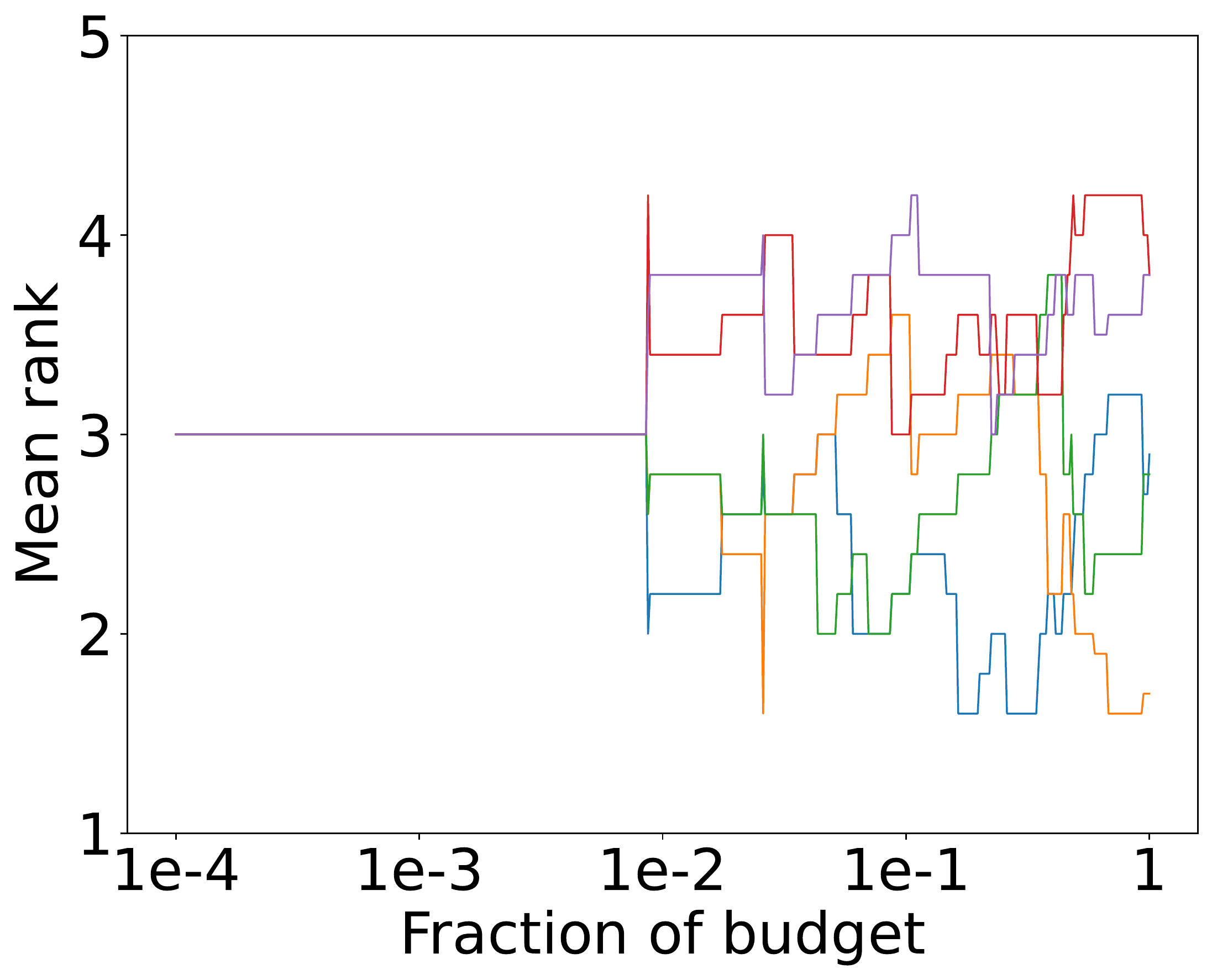}
		\caption{$\textit{BBO}_{146821\text{OpenML}}$}
		\label{fig:BBO_146821_openml_avg}
	\end{subfigure}
	\begin{subfigure}{0.25\linewidth}
		\centering
		\includegraphics[width=0.9\linewidth]{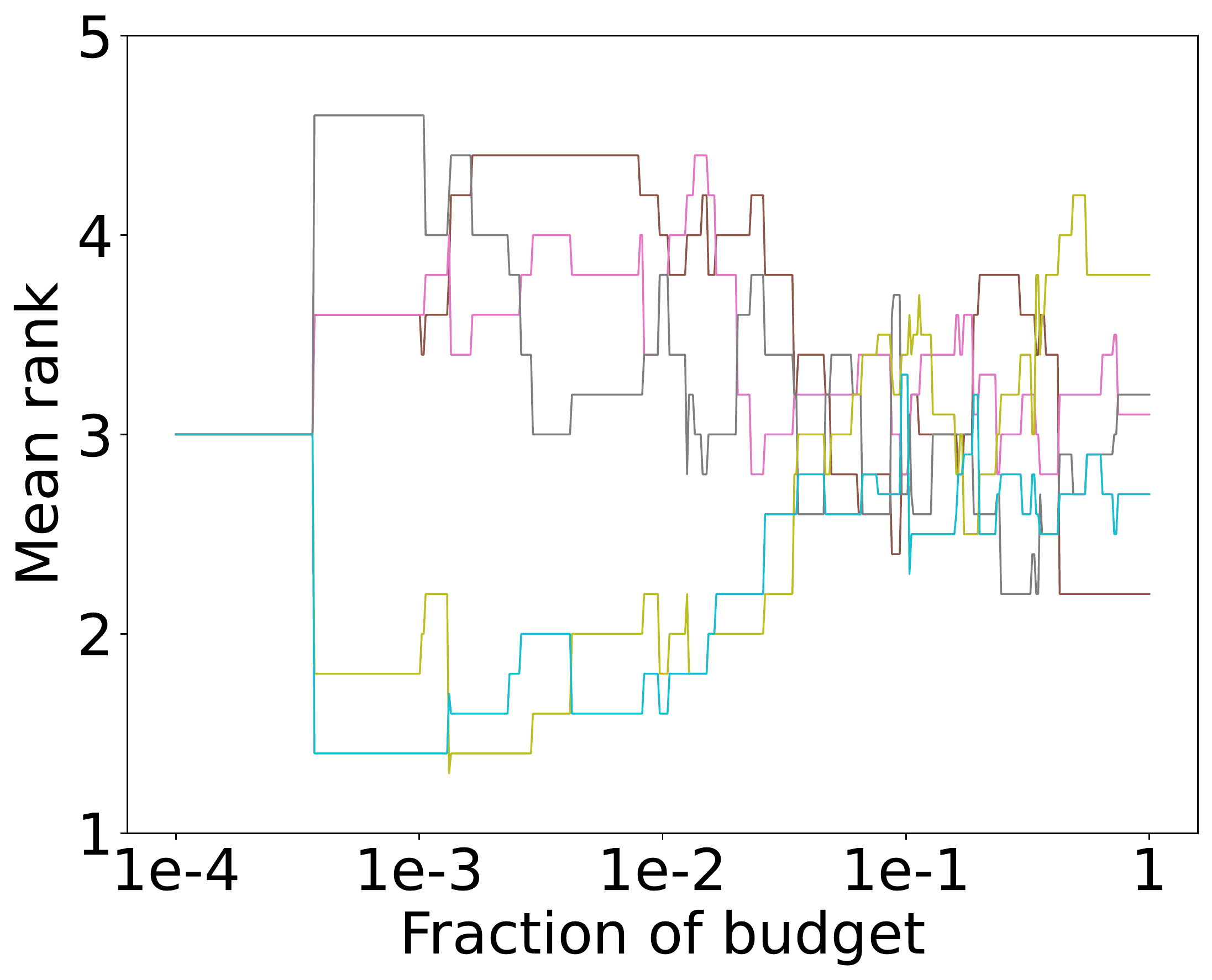}
		\caption{$\textit{MF}_{146821\text{OpenML}}$}
		\label{fig:MF_146821_openml_avg}
	\end{subfigure}
	
	\begin{subfigure}{0.25\linewidth}
		\centering
		\includegraphics[width=0.9\linewidth]{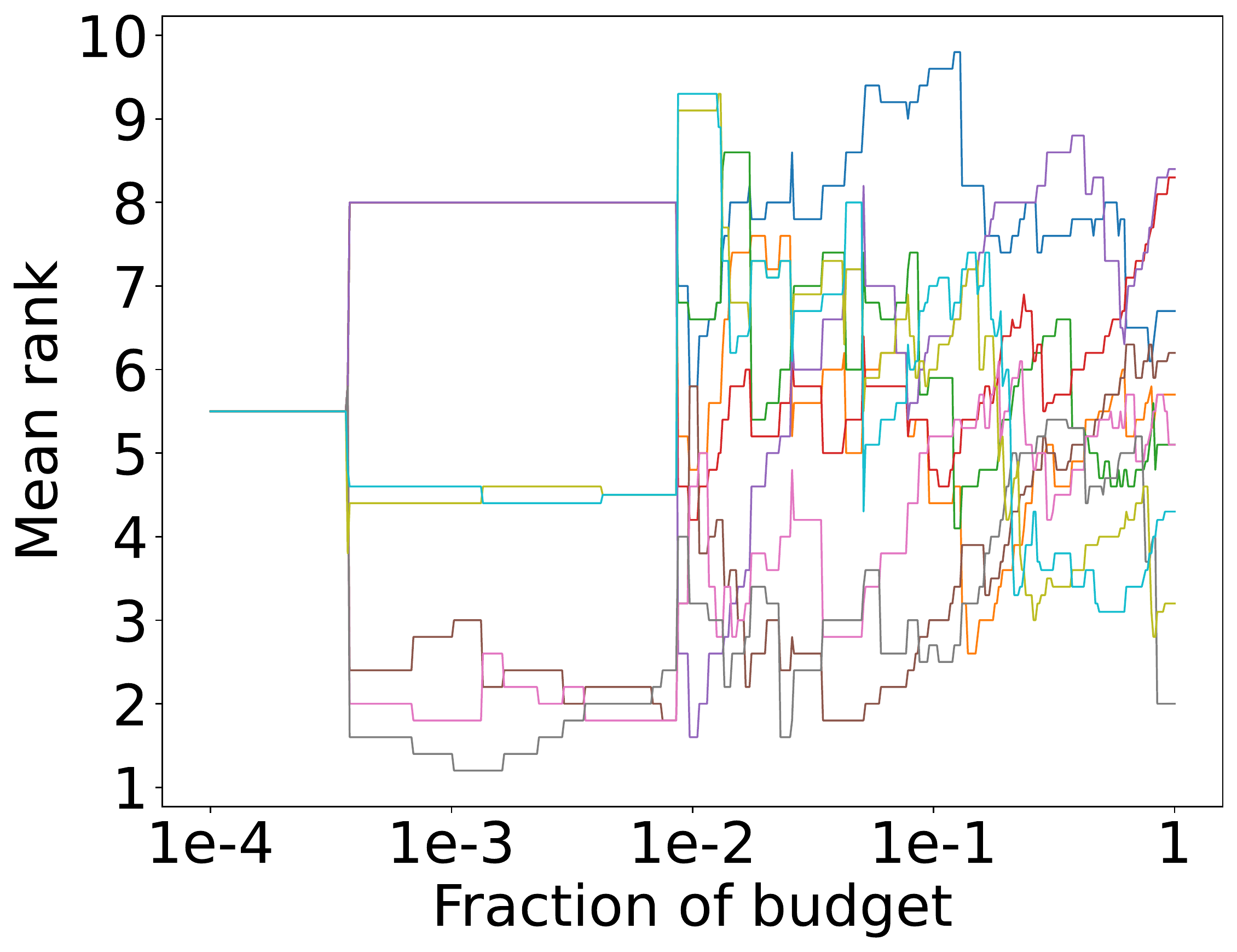}
		\caption{$\text{ALL}_{146822\text{OpenML}}$}
		\label{fig:All_146822_openml_avg}
	\end{subfigure}
	\begin{subfigure}{0.25\linewidth}
		\centering
		\includegraphics[width=0.9\linewidth]{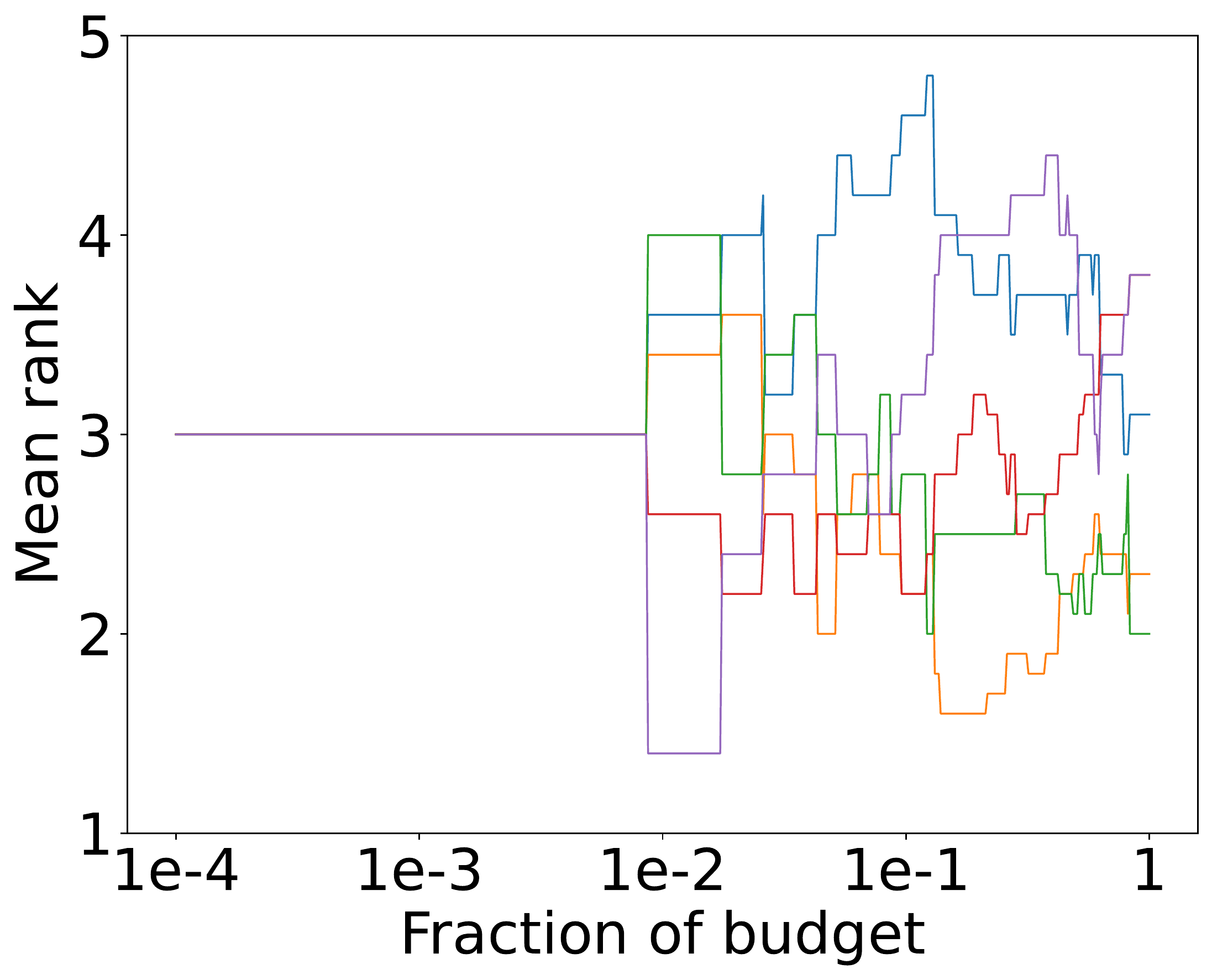}
		\caption{$\textit{BBO}_{146822\text{OpenML}}$}
		\label{fig:BBO_146822_openml_avg}
	\end{subfigure}
	\begin{subfigure}{0.25\linewidth}
		\centering
		\includegraphics[width=0.9\linewidth]{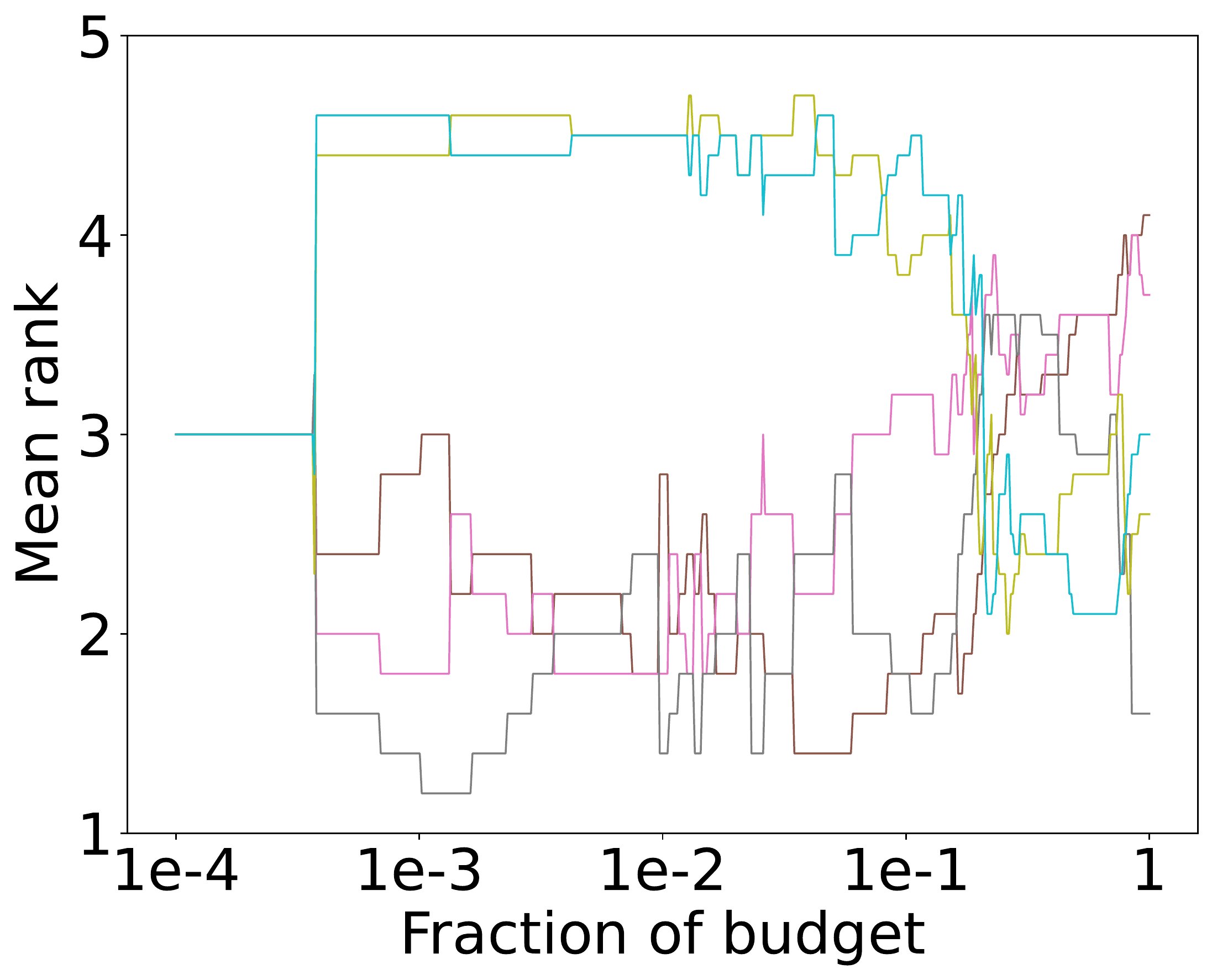}
		\caption{$\textit{MF}_{146822\text{OpenML}}$}
		\label{fig:MF_146822_openml_avg}
	\end{subfigure}
	
	\centering
	\hspace*{1.2cm}\begin{subfigure}{1.0\linewidth}
		\centering
		\includegraphics[width=0.95\linewidth]{materials/legend_rank_new.pdf}
	\end{subfigure}
	\vspace{-0.1in}
	
	\caption{Mean rank over time on LR benchmark (FedAvg).}
	\label{fig:entire_lr_tabular_avg_rank}
\end{figure}

\begin{figure}[htbp]
	\centering
	\begin{subfigure}{0.25\linewidth}
		\centering
		\includegraphics[width=0.9\linewidth]{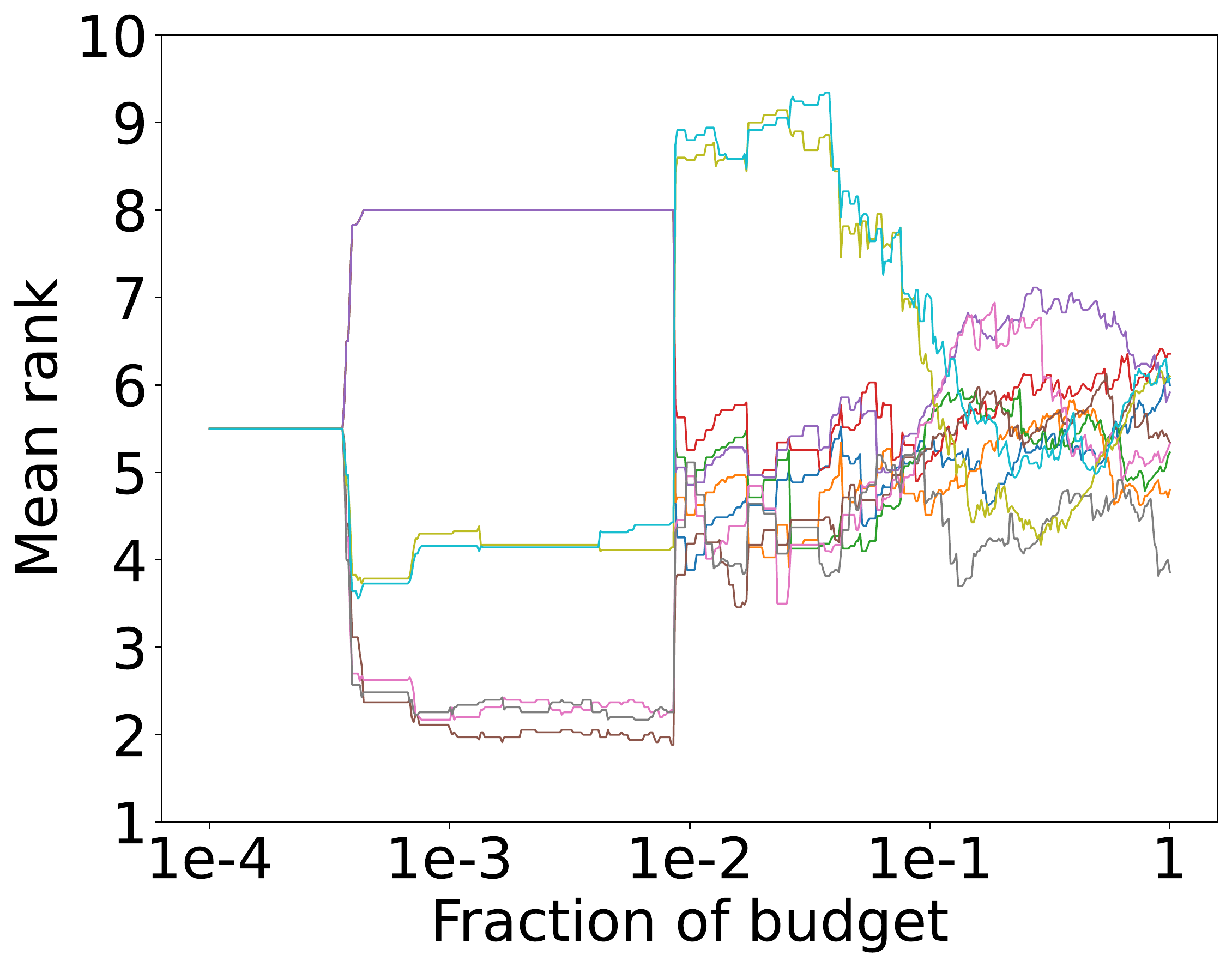}
		\caption{$\text{ALL}_{LR}$}
		\label{fig:All_LR_opt}
	\end{subfigure}
	\begin{subfigure}{0.25\linewidth}
		\centering
		\includegraphics[width=0.9\linewidth]{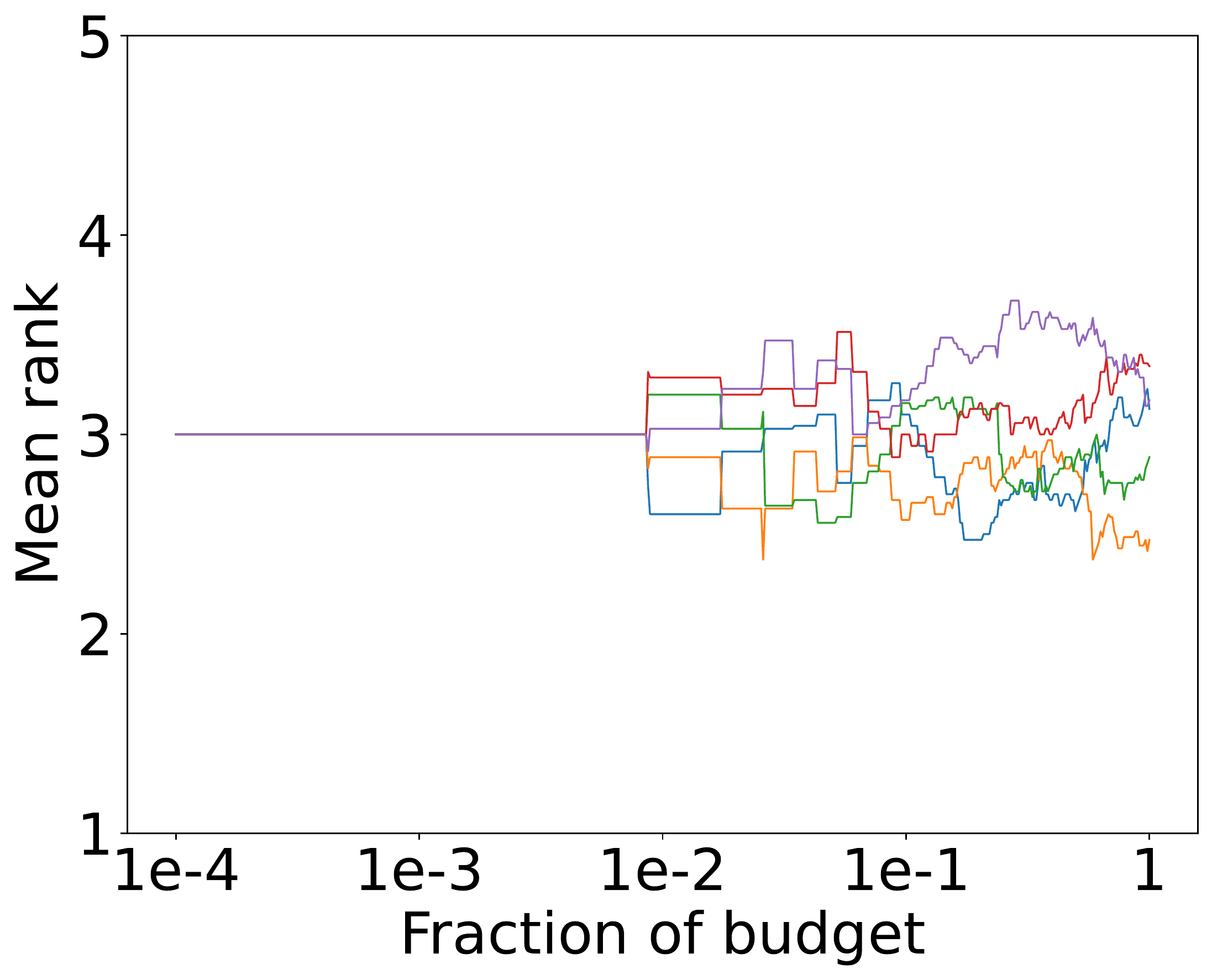}
		\caption{$\textit{BBO}_{LR}$}
		\label{fig:BBO_LR_opt}
	\end{subfigure}
	\begin{subfigure}{0.25\linewidth}
		\centering
		\includegraphics[width=0.9\linewidth]{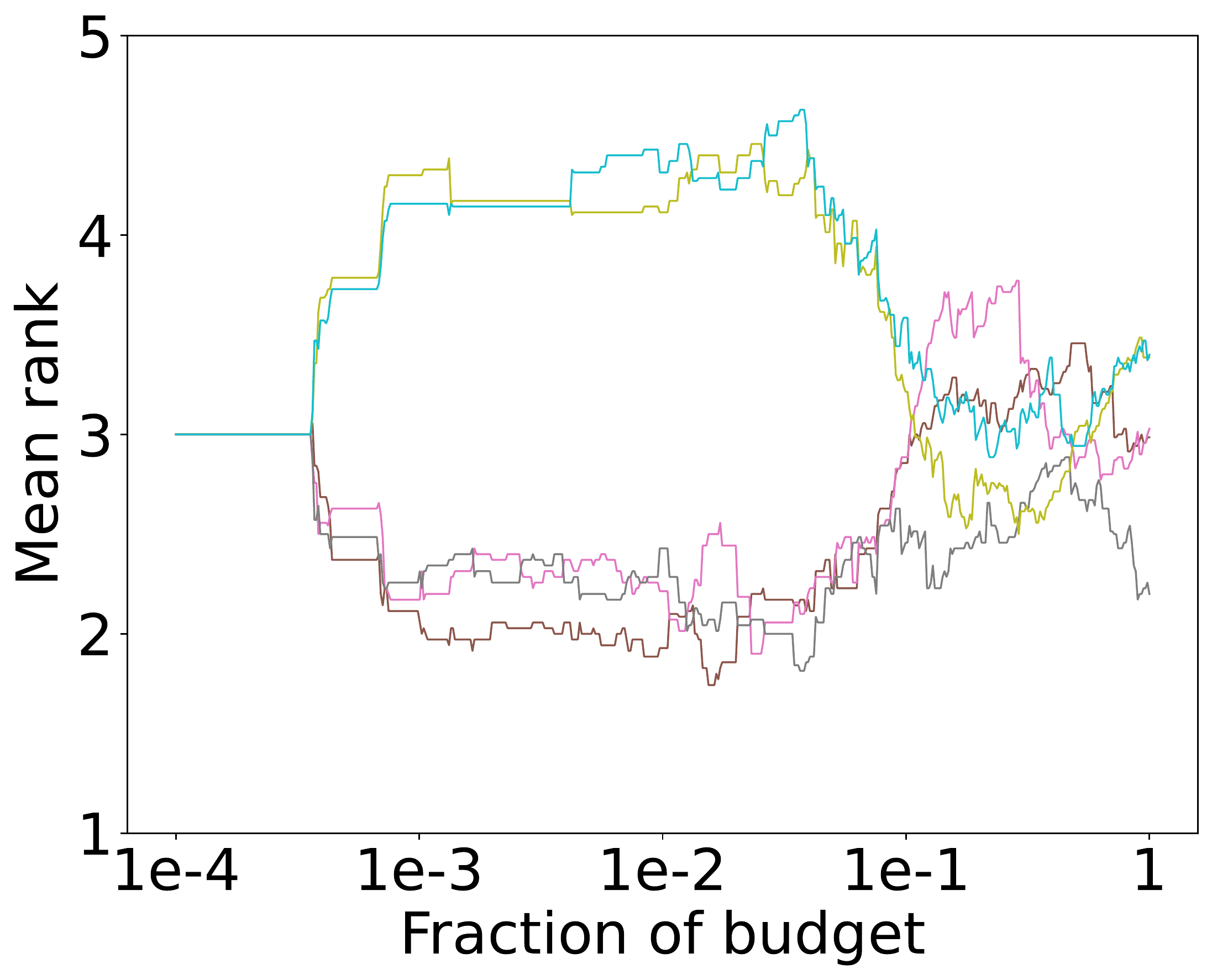}
		\caption{$\textit{MF}_{LR}$}
		\label{fig:MF_LR_opt}
	\end{subfigure}
	\qquad

	\begin{subfigure}{0.25\linewidth}
		\centering
		\includegraphics[width=0.9\linewidth]{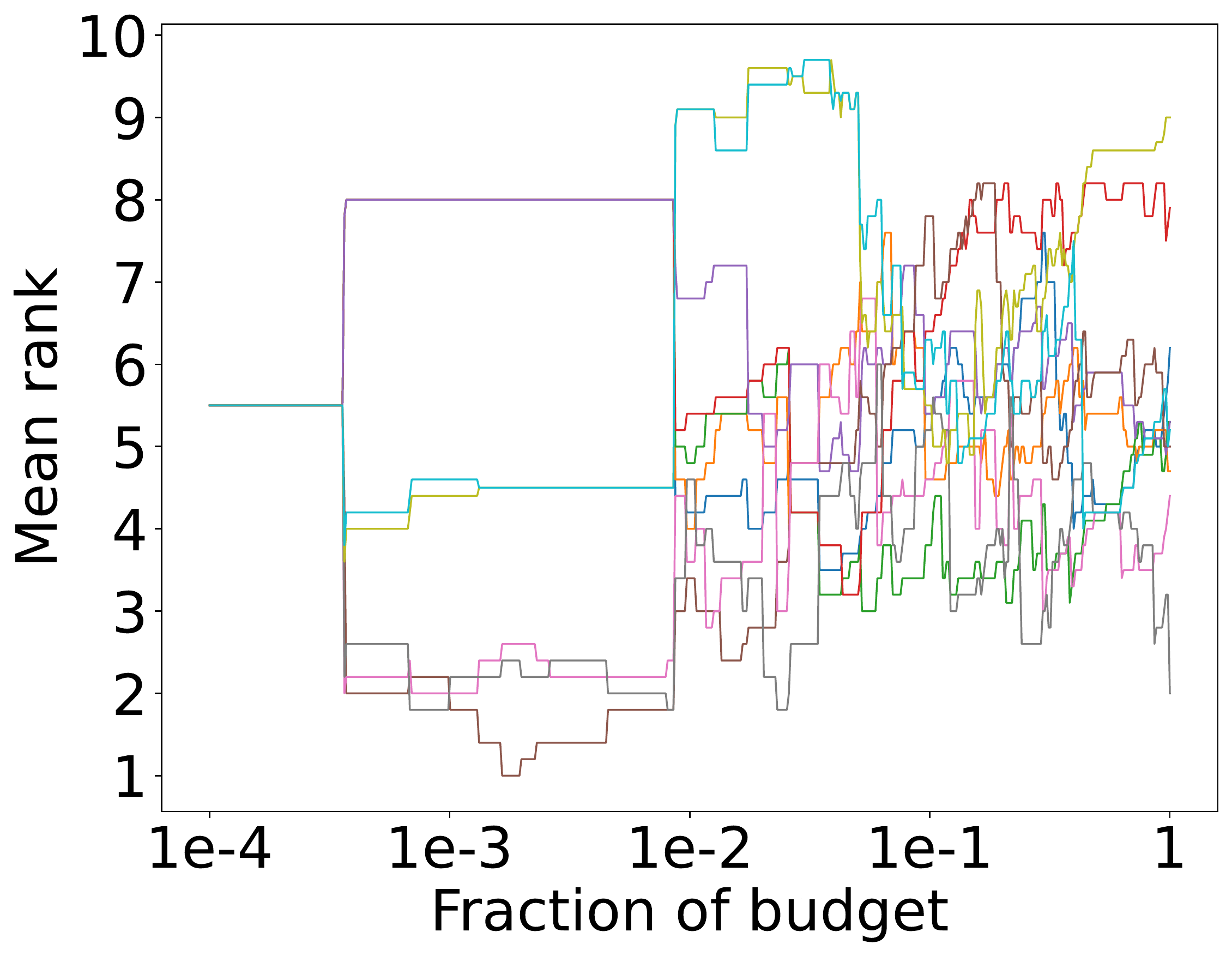}
		\caption{$\text{ALL}_{31\text{OpenML}}$}
		\label{fig:All_31_openml_opt}
	\end{subfigure}
	\begin{subfigure}{0.25\linewidth}
		\centering
		\includegraphics[width=0.9\linewidth]{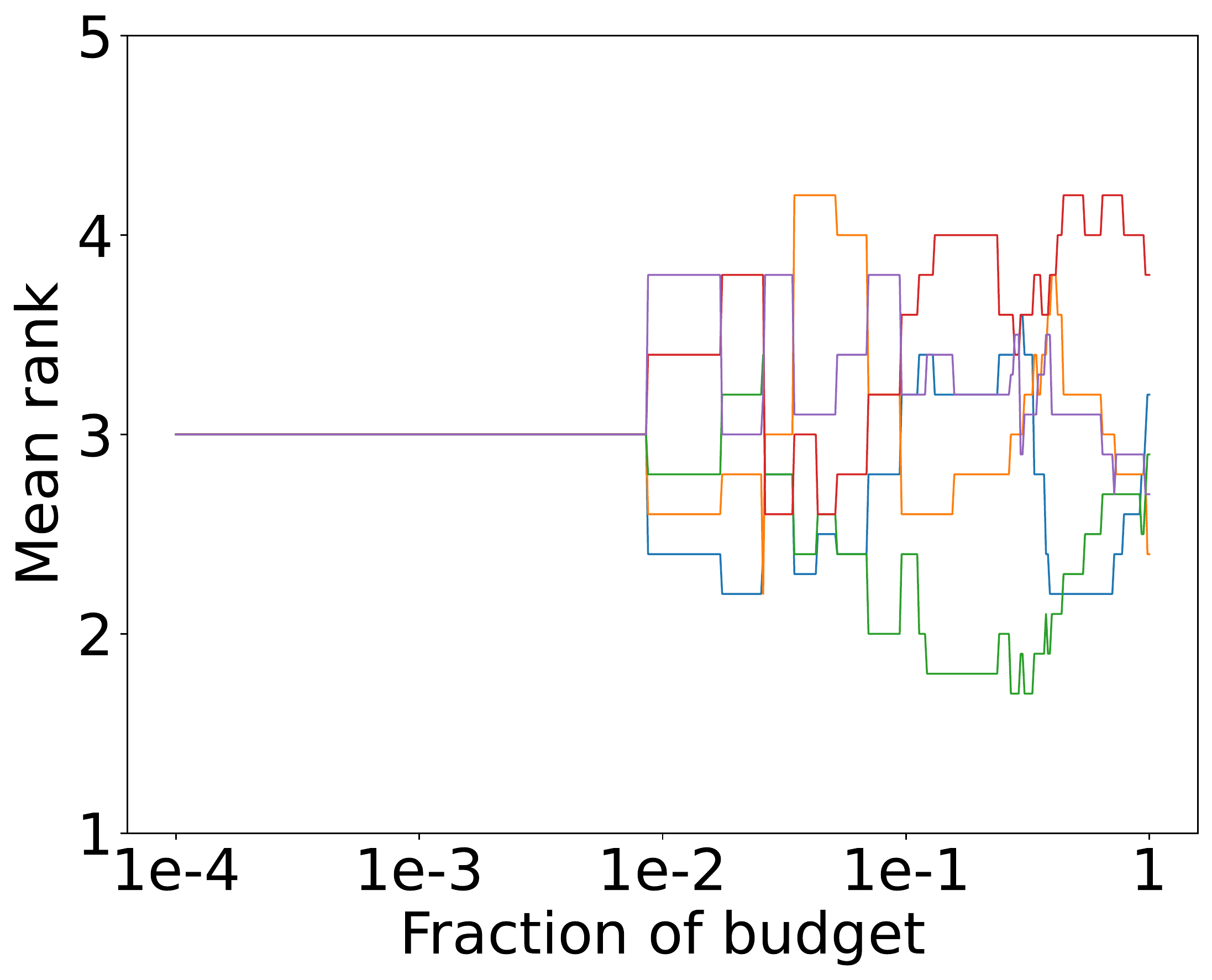}
		\caption{$\textit{BBO}_{31\text{OpenML}}$}
		\label{fig:BBO_31_openml_opt}
	\end{subfigure}
	\begin{subfigure}{0.25\linewidth}
		\centering
		\includegraphics[width=0.9\linewidth]{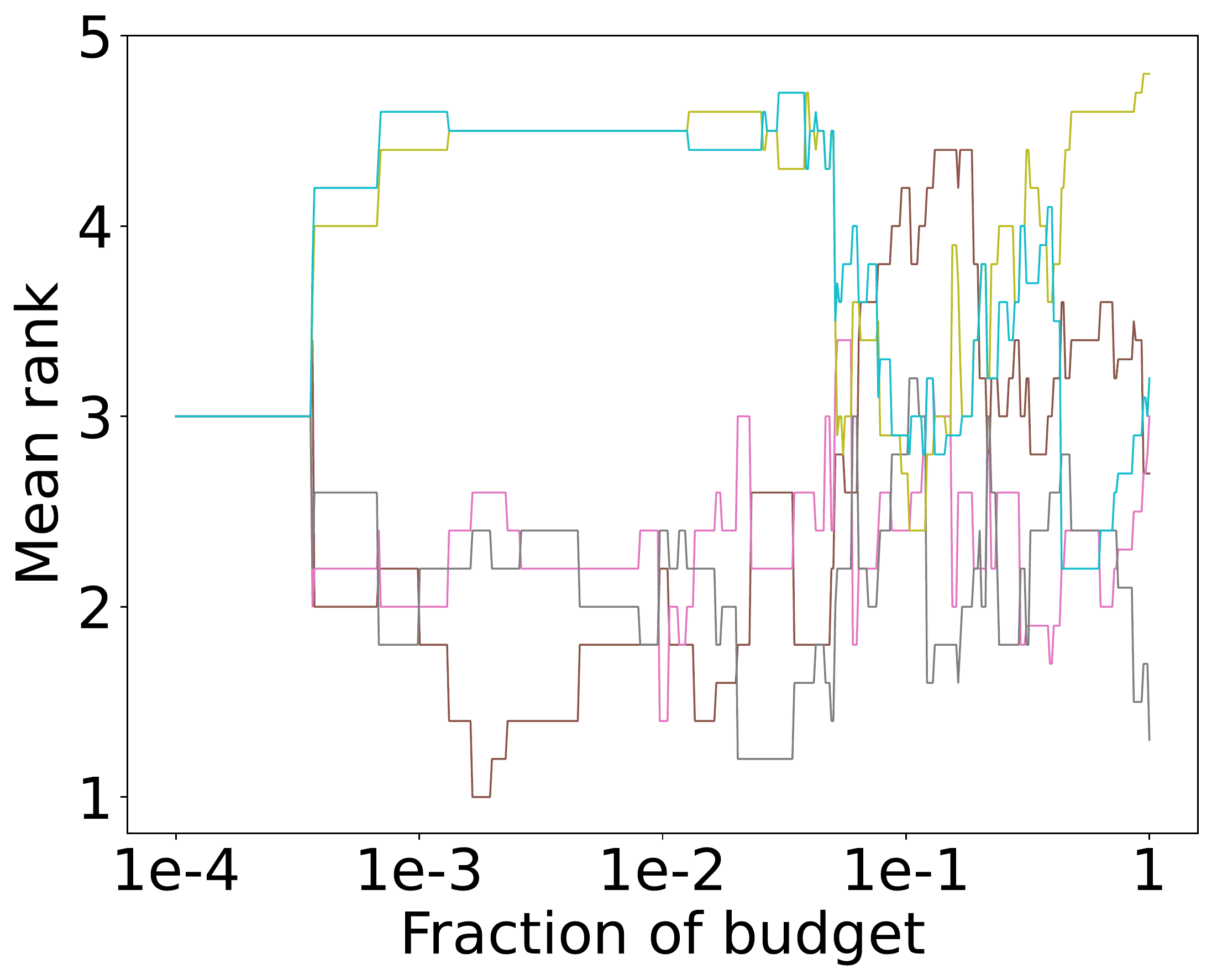}
		\caption{$\textit{MF}_{31\text{OpenML}}$}
		\label{fig:MF_31_openml_opt}
	\end{subfigure}
	
	\begin{subfigure}{0.25\linewidth}
		\centering
		\includegraphics[width=0.9\linewidth]{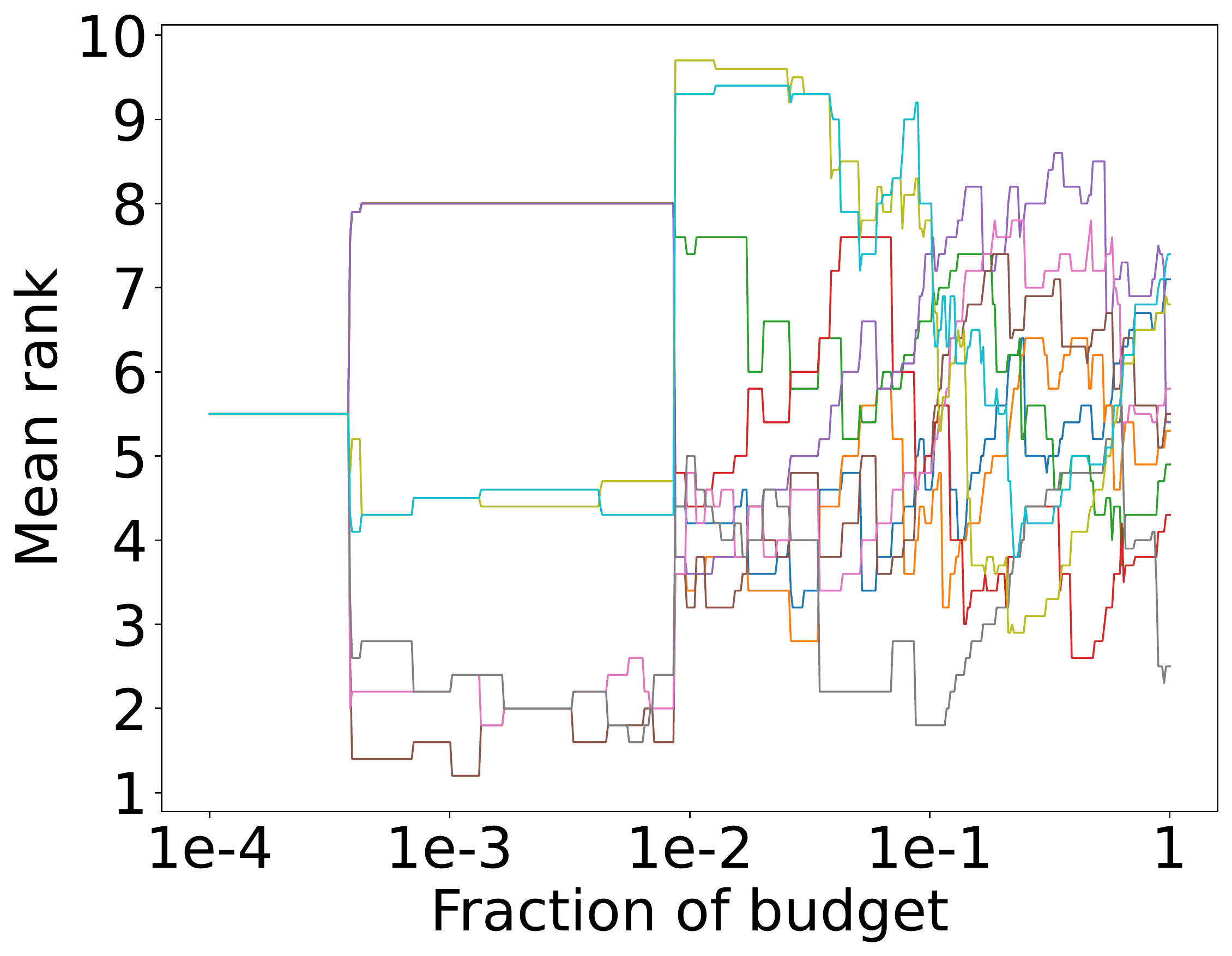}
		\caption{$\text{ALL}_{53\text{OpenML}}$}
		\label{fig:All_53_openml_opt}
	\end{subfigure}
	\begin{subfigure}{0.25\linewidth}
		\centering
		\includegraphics[width=0.9\linewidth]{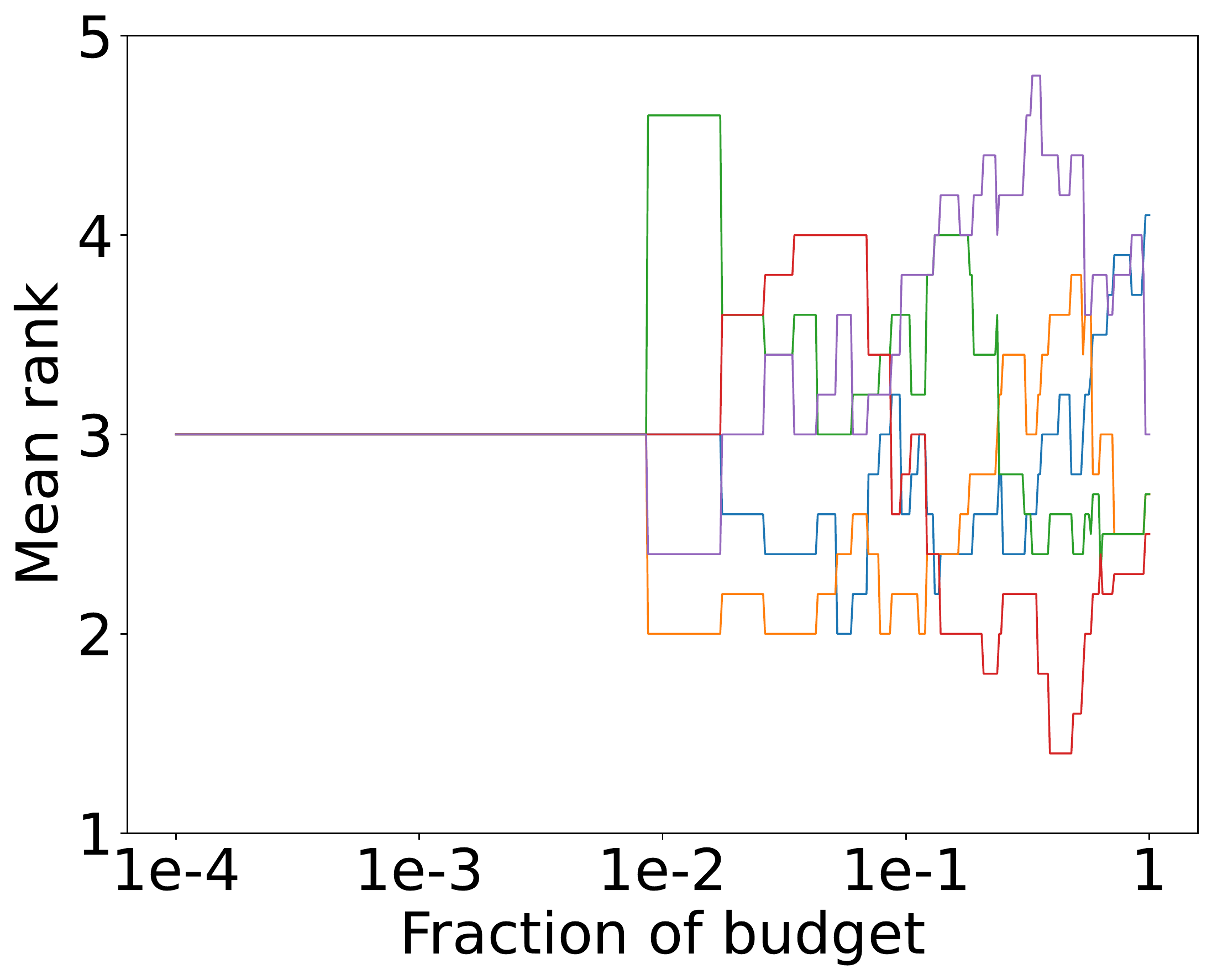}
		\caption{$\textit{BBO}_{53\text{OpenML}}$}
		\label{fig:BBO_53_openml_opt}
	\end{subfigure}
	\begin{subfigure}{0.25\linewidth}
		\centering
		\includegraphics[width=0.9\linewidth]{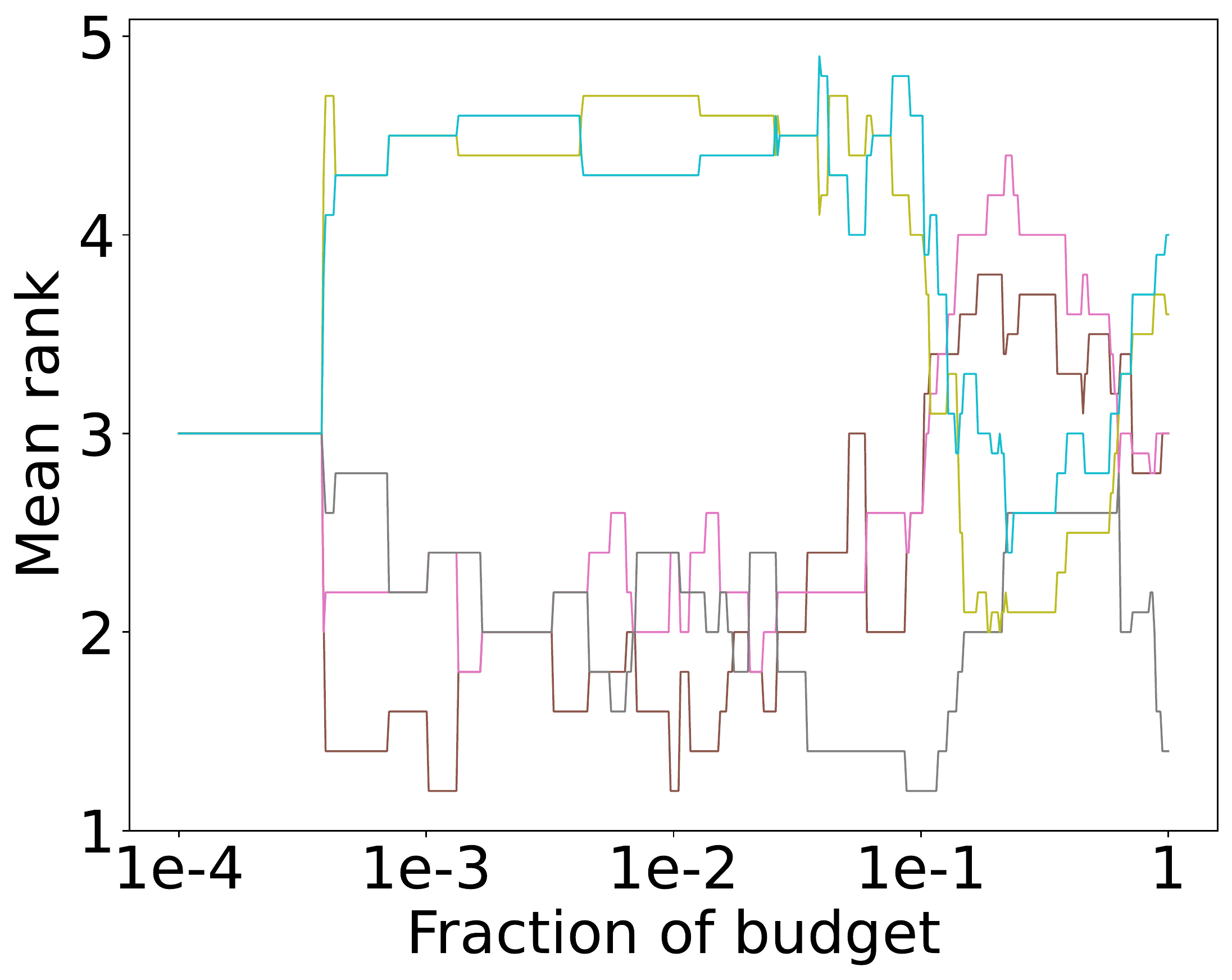}
		\caption{$\textit{MF}_{53\text{OpenML}}$}
		\label{fig:MF_53_openml_opt}
	\end{subfigure}
	
	\begin{subfigure}{0.25\linewidth}
		\centering
		\includegraphics[width=0.9\linewidth]{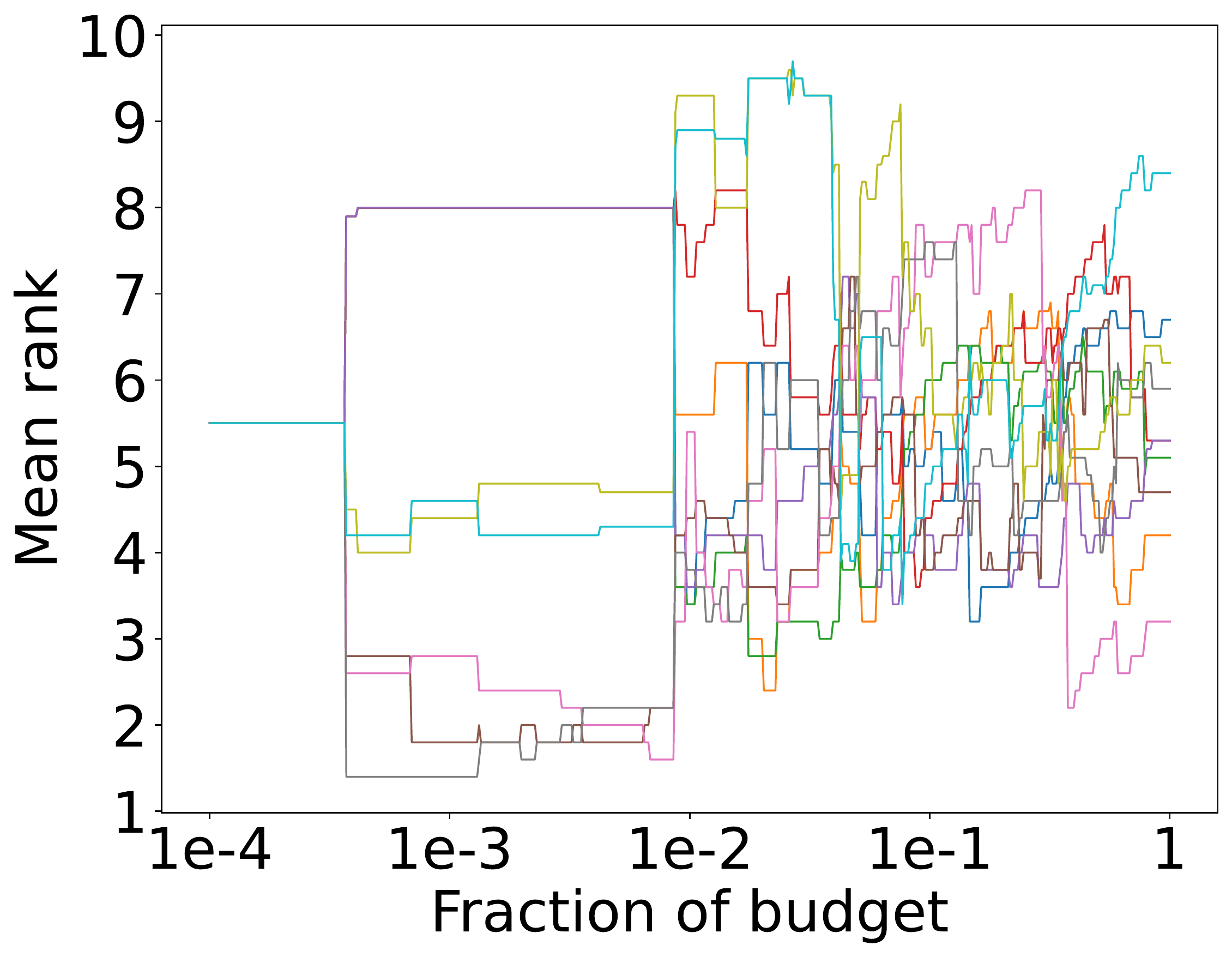}
		\caption{$\text{ALL}_{10101\text{OpenML}}$}
		\label{fig:All_10101_openml_opt}
	\end{subfigure}
	\begin{subfigure}{0.25\linewidth}
		\centering
		\includegraphics[width=0.9\linewidth]{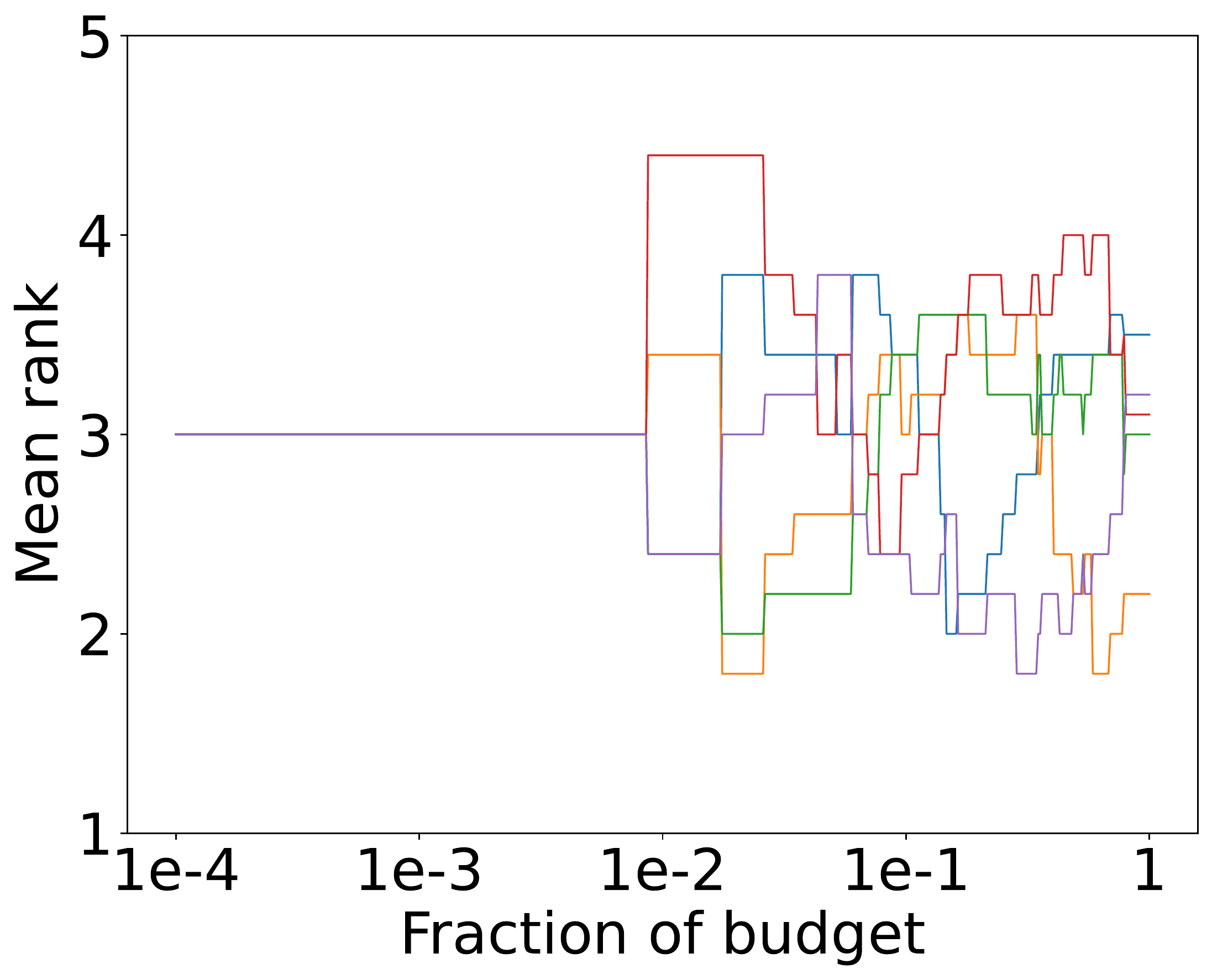}
		\caption{$\textit{BBO}_{10101\text{OpenML}}$}
		\label{fig:BBO_10101_openml_opt}
	\end{subfigure}
	\begin{subfigure}{0.25\linewidth}
		\centering
		\includegraphics[width=0.9\linewidth]{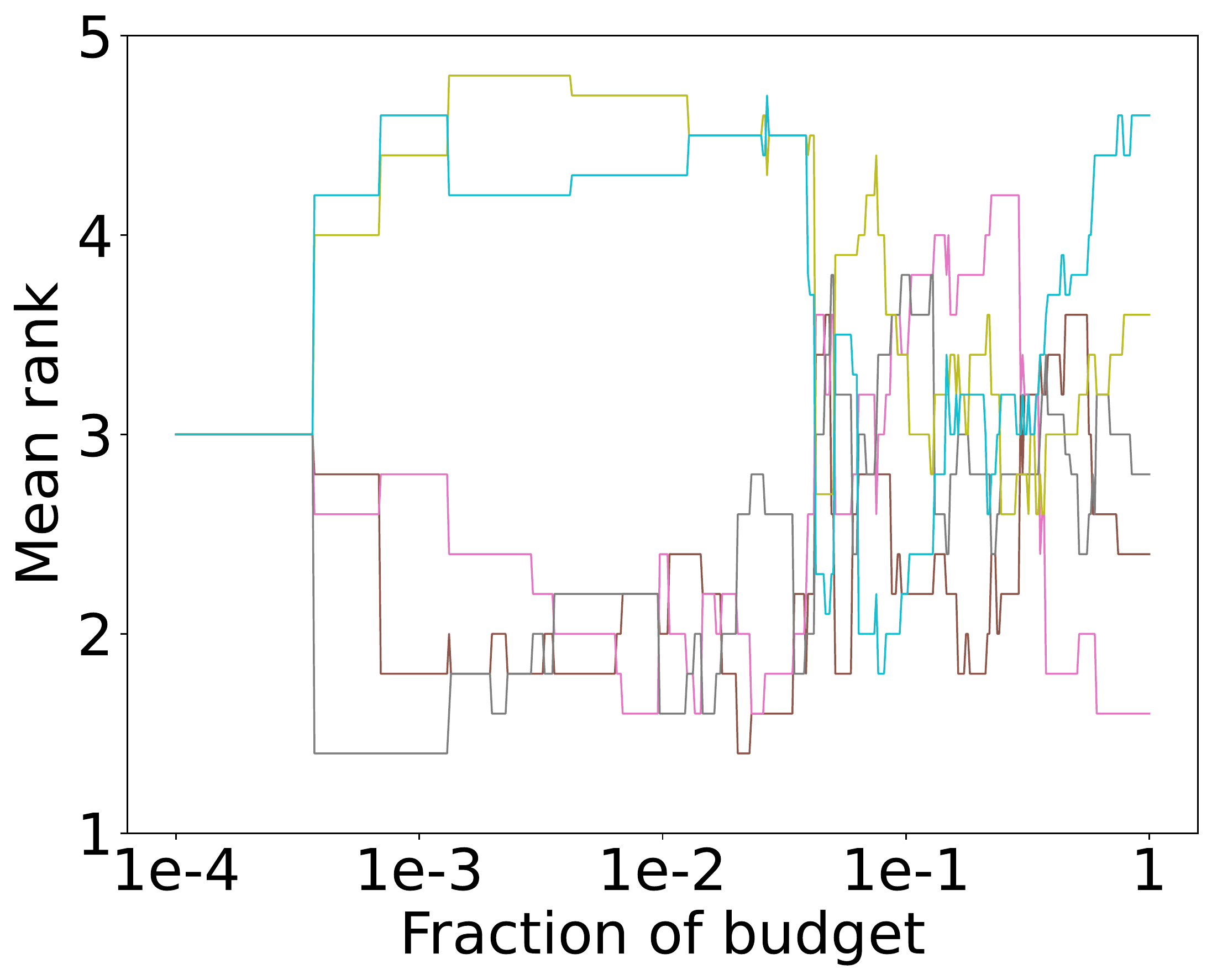}
		\caption{$\textit{MF}_{10101\text{OpenML}}$}
		\label{fig:MF_10101_openml_opt}
	\end{subfigure}
	
	\begin{subfigure}{0.25\linewidth}
		\centering
		\includegraphics[width=0.9\linewidth]{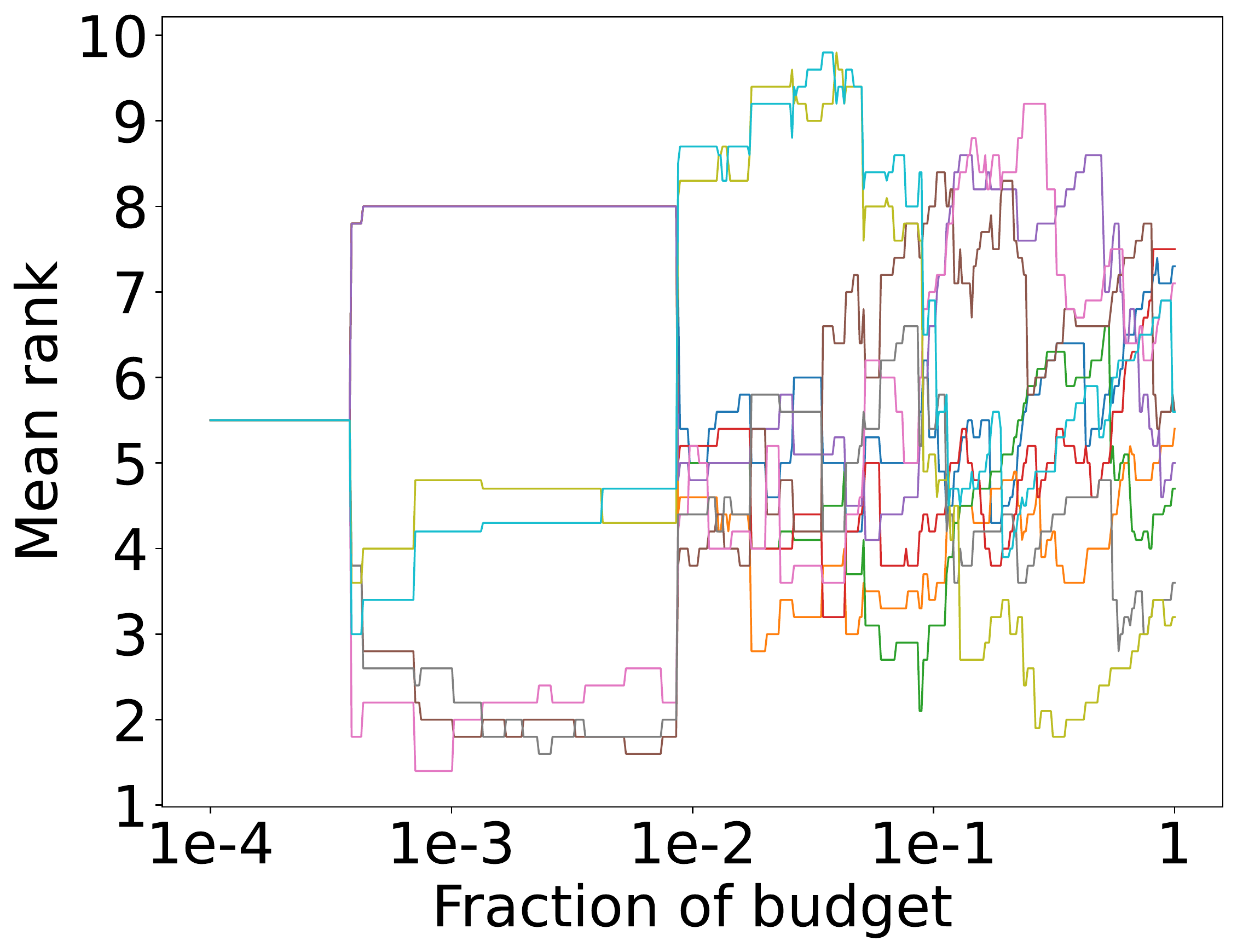}
		\caption{$\text{ALL}_{146818\text{OpenML}}$}
		\label{fig:All_146818_openml_opt}
	\end{subfigure}
	\begin{subfigure}{0.25\linewidth}
		\centering
		\includegraphics[width=0.9\linewidth]{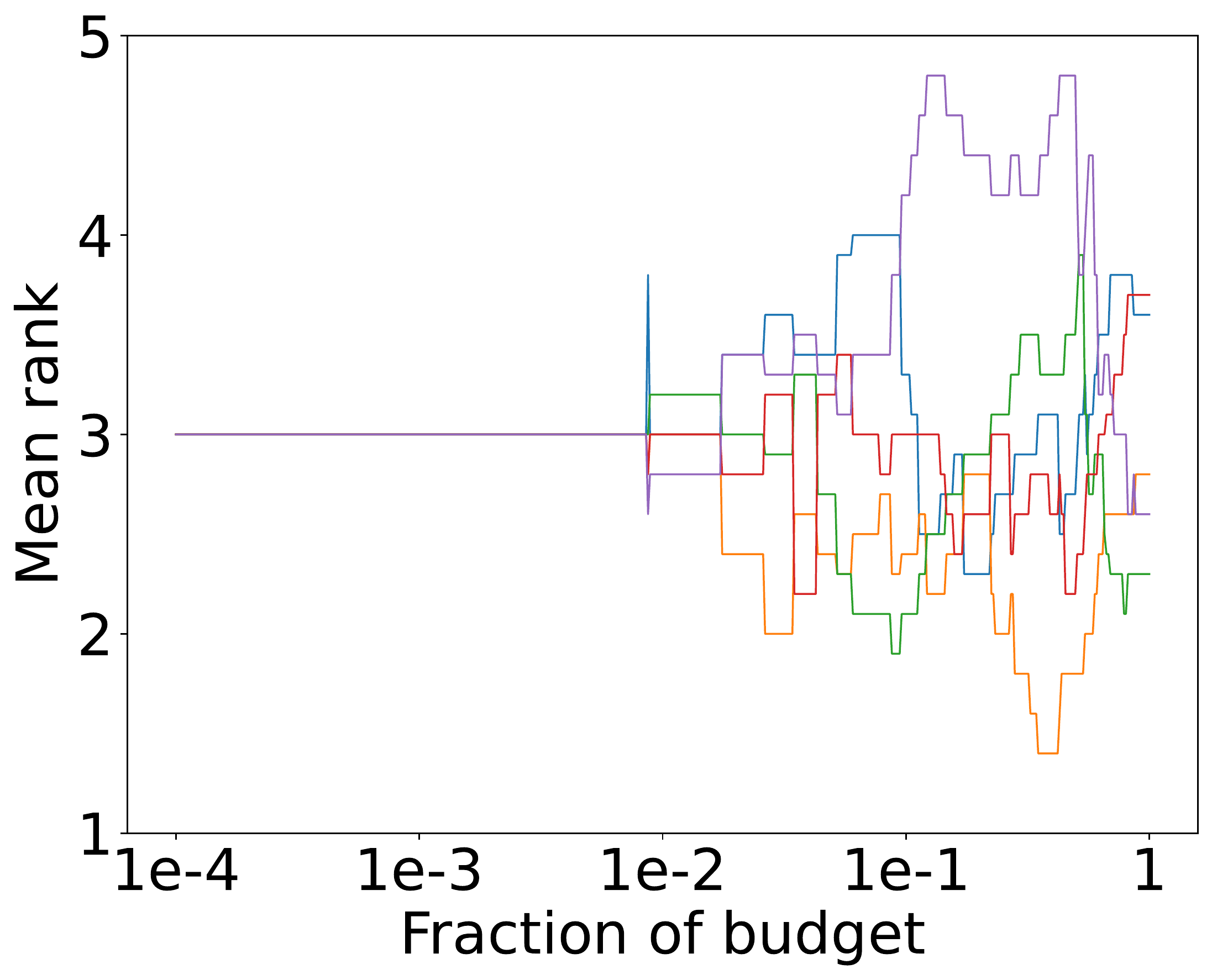}
		\caption{$\textit{BBO}_{146818\text{OpenML}}$}
		\label{fig:BBO_146818_openml_opt}
	\end{subfigure}
	\begin{subfigure}{0.25\linewidth}
		\centering
		\includegraphics[width=0.9\linewidth]{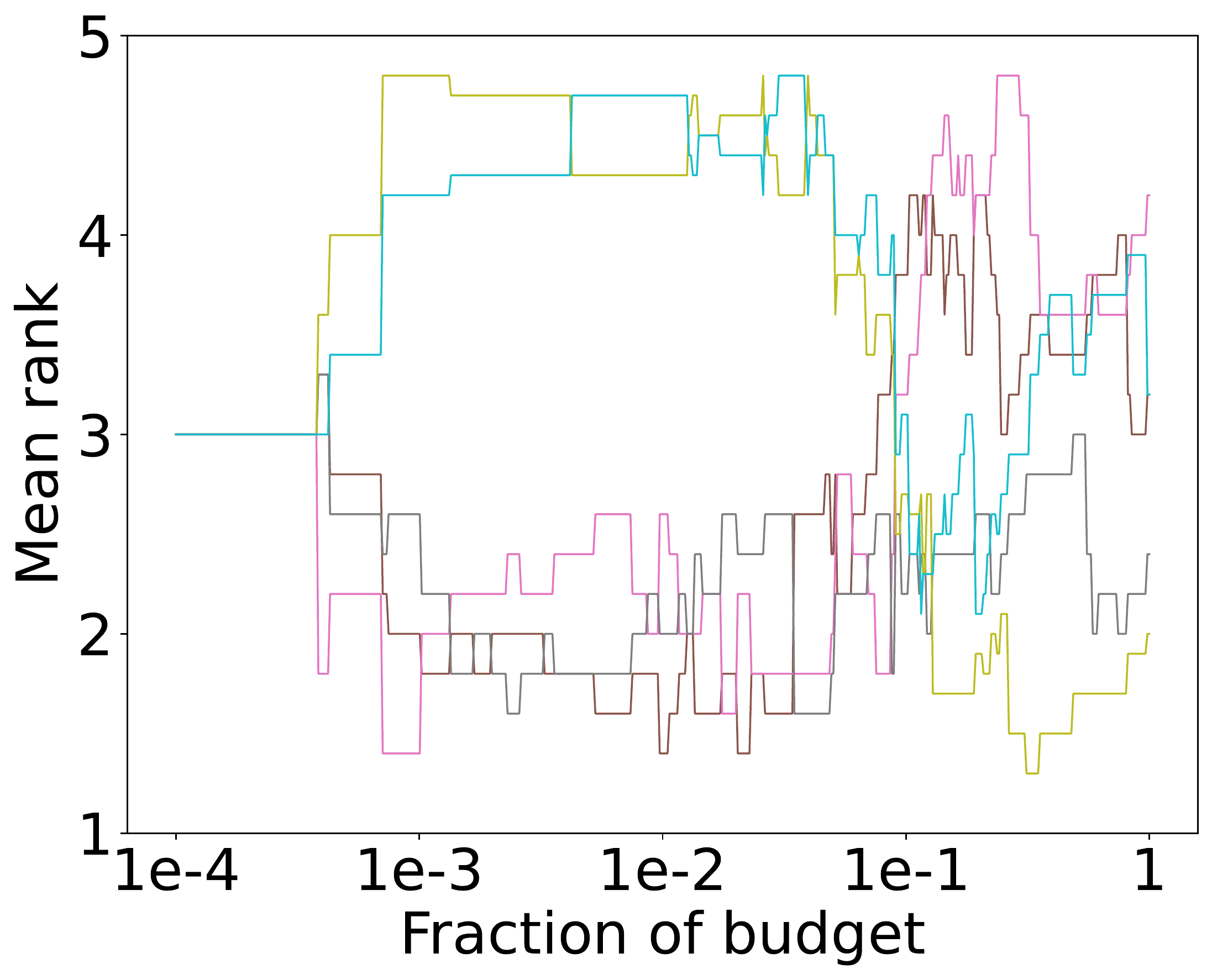}
		\caption{$\textit{MF}_{146818\text{OpenML}}$}
		\label{fig:MF_146818_openml_opt}
	\end{subfigure}
	
	\begin{subfigure}{0.25\linewidth}
		\centering
		\includegraphics[width=0.9\linewidth]{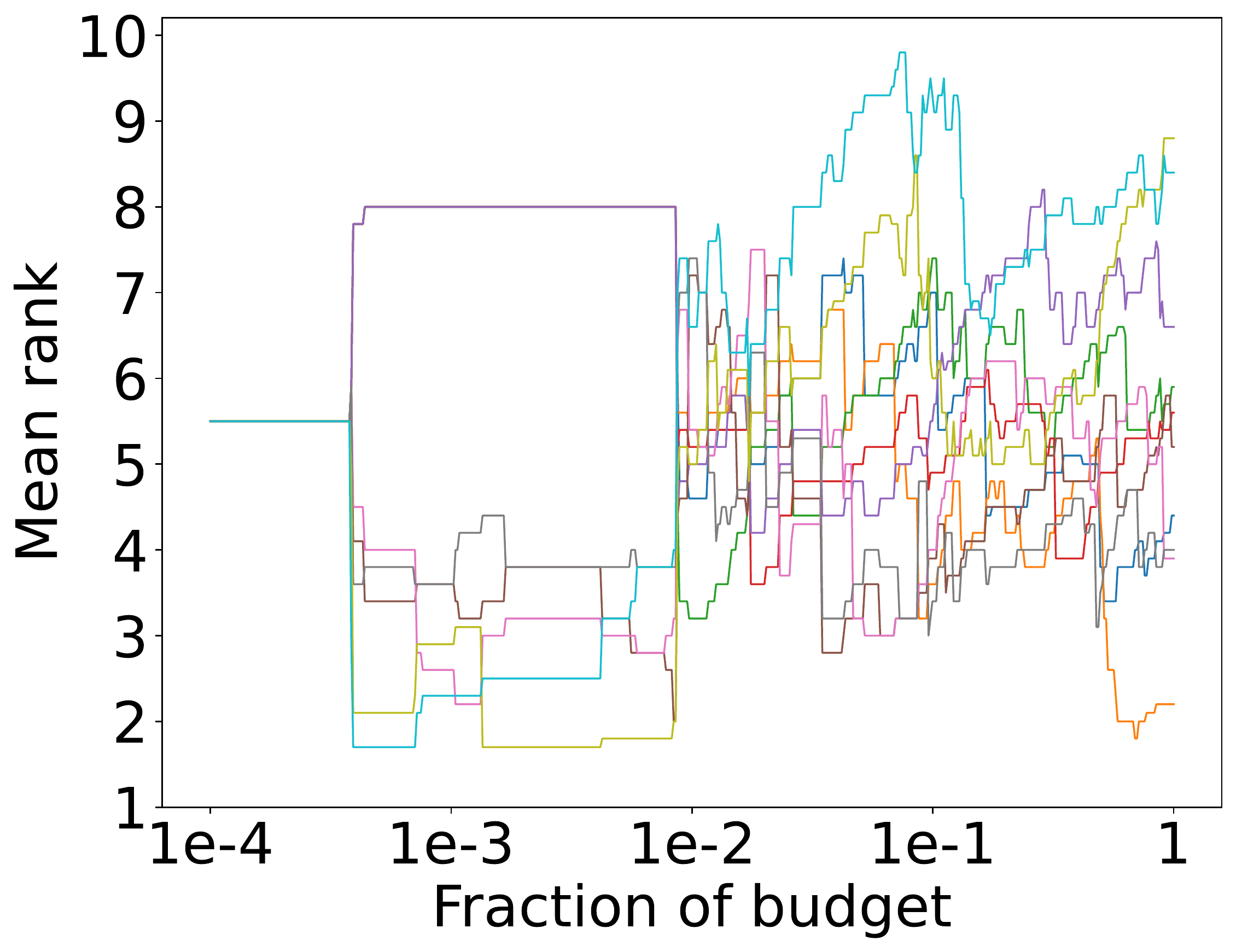}
		\caption{$\text{ALL}_{146821\text{OpenML}}$}
		\label{fig:All_146821_openml_opt}
	\end{subfigure}
	\begin{subfigure}{0.25\linewidth}
		\centering
		\includegraphics[width=0.9\linewidth]{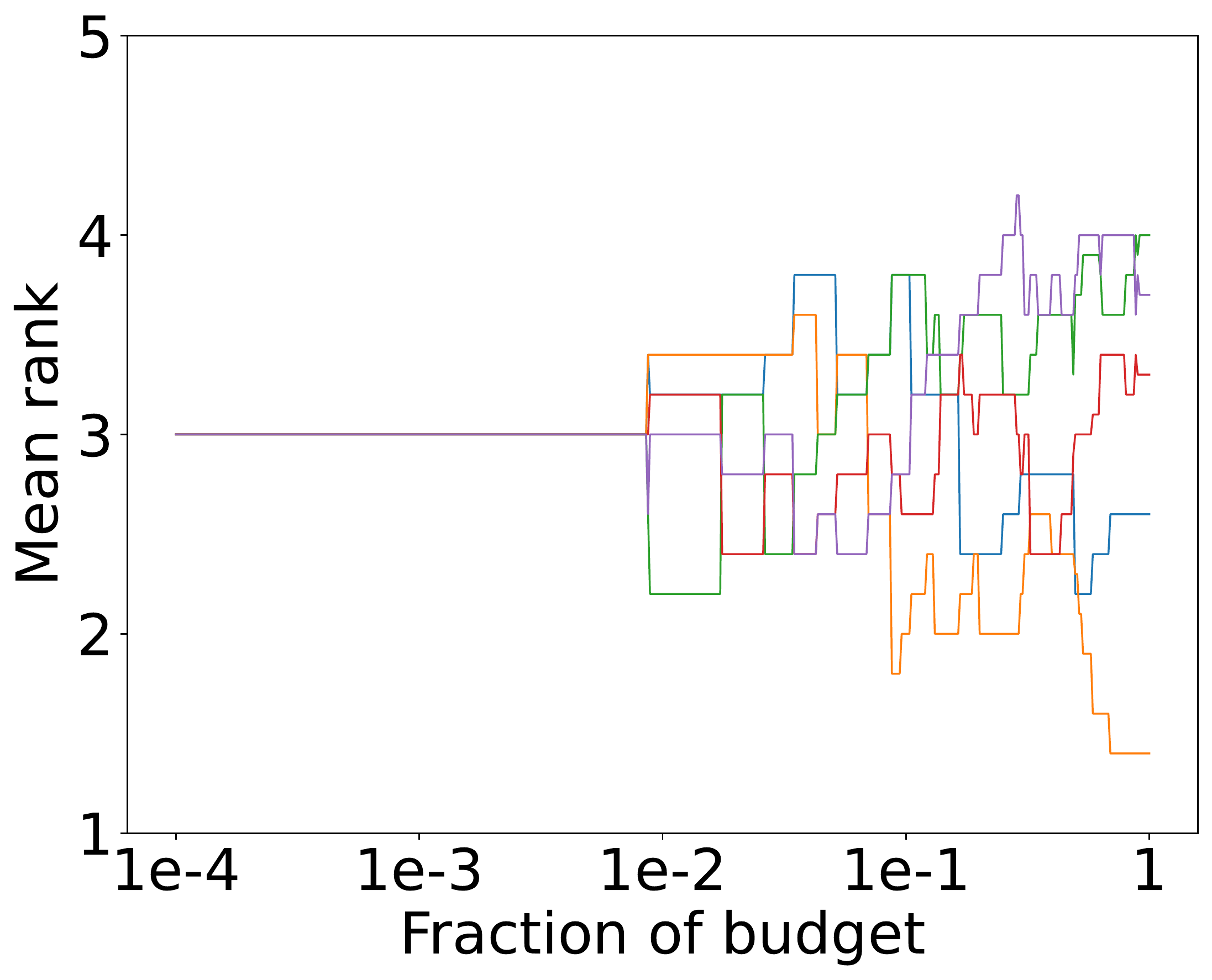}
		\caption{$\textit{BBO}_{146821\text{OpenML}}$}
		\label{fig:BBO_146821_openml_opt}
	\end{subfigure}
	\begin{subfigure}{0.25\linewidth}
		\centering
		\includegraphics[width=0.9\linewidth]{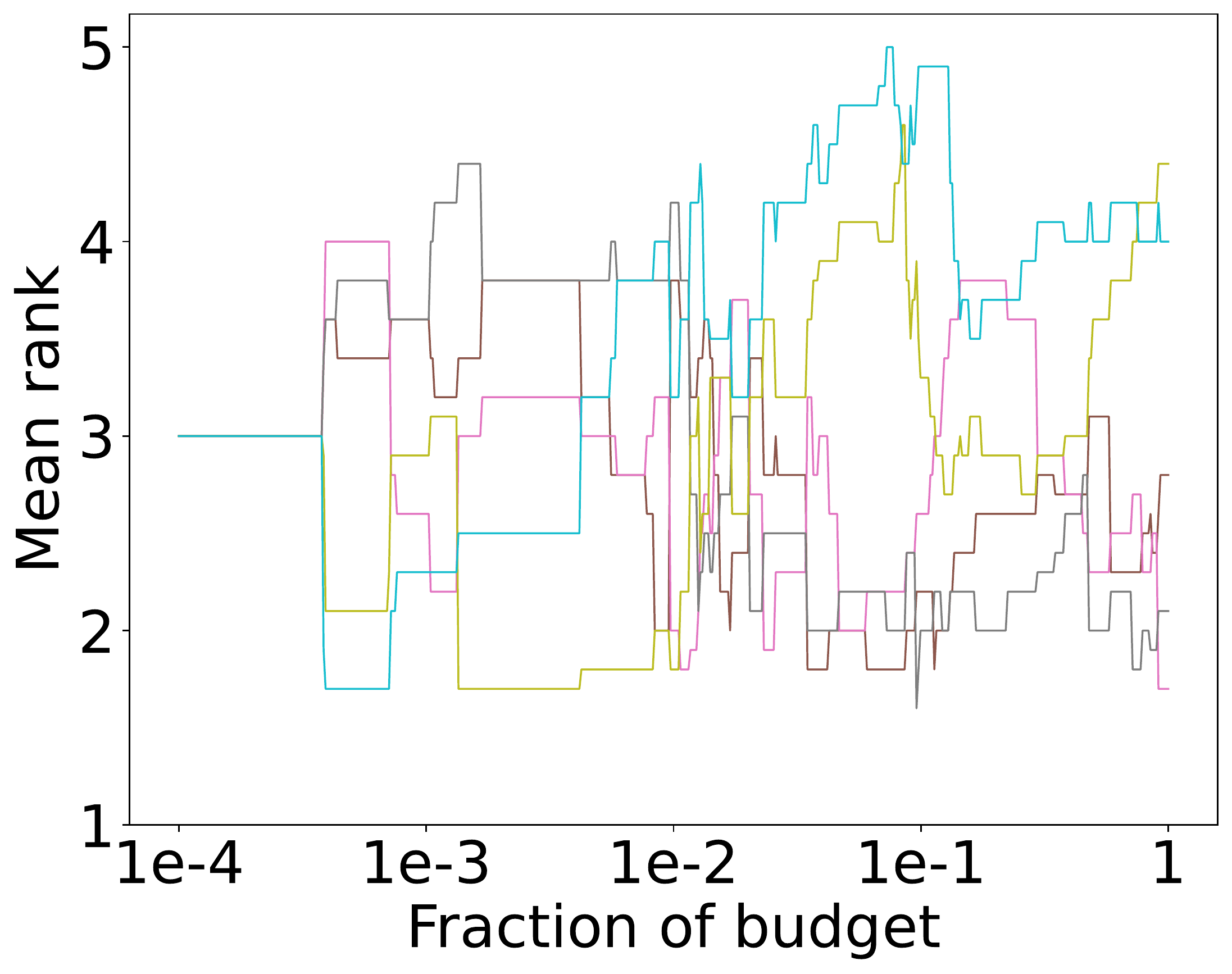}
		\caption{$\textit{MF}_{146821\text{OpenML}}$}
		\label{fig:MF_146821_openml_opt}
	\end{subfigure}
	
	\begin{subfigure}{0.25\linewidth}
		\centering
		\includegraphics[width=0.9\linewidth]{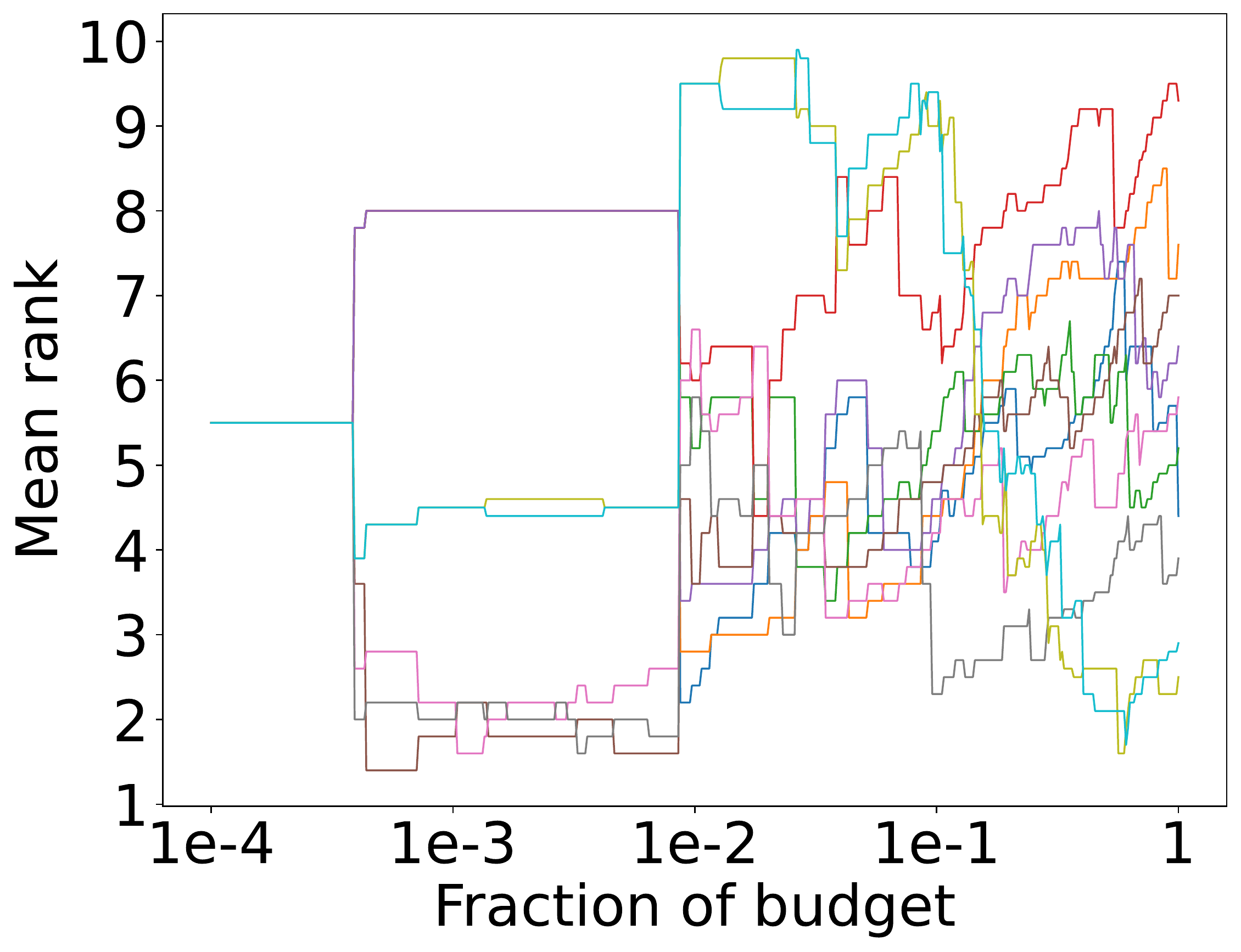}
		\caption{$\text{ALL}_{146822\text{OpenML}}$}
		\label{fig:All_146822_openml_opt}
	\end{subfigure}
	\begin{subfigure}{0.25\linewidth}
		\centering
		\includegraphics[width=0.9\linewidth]{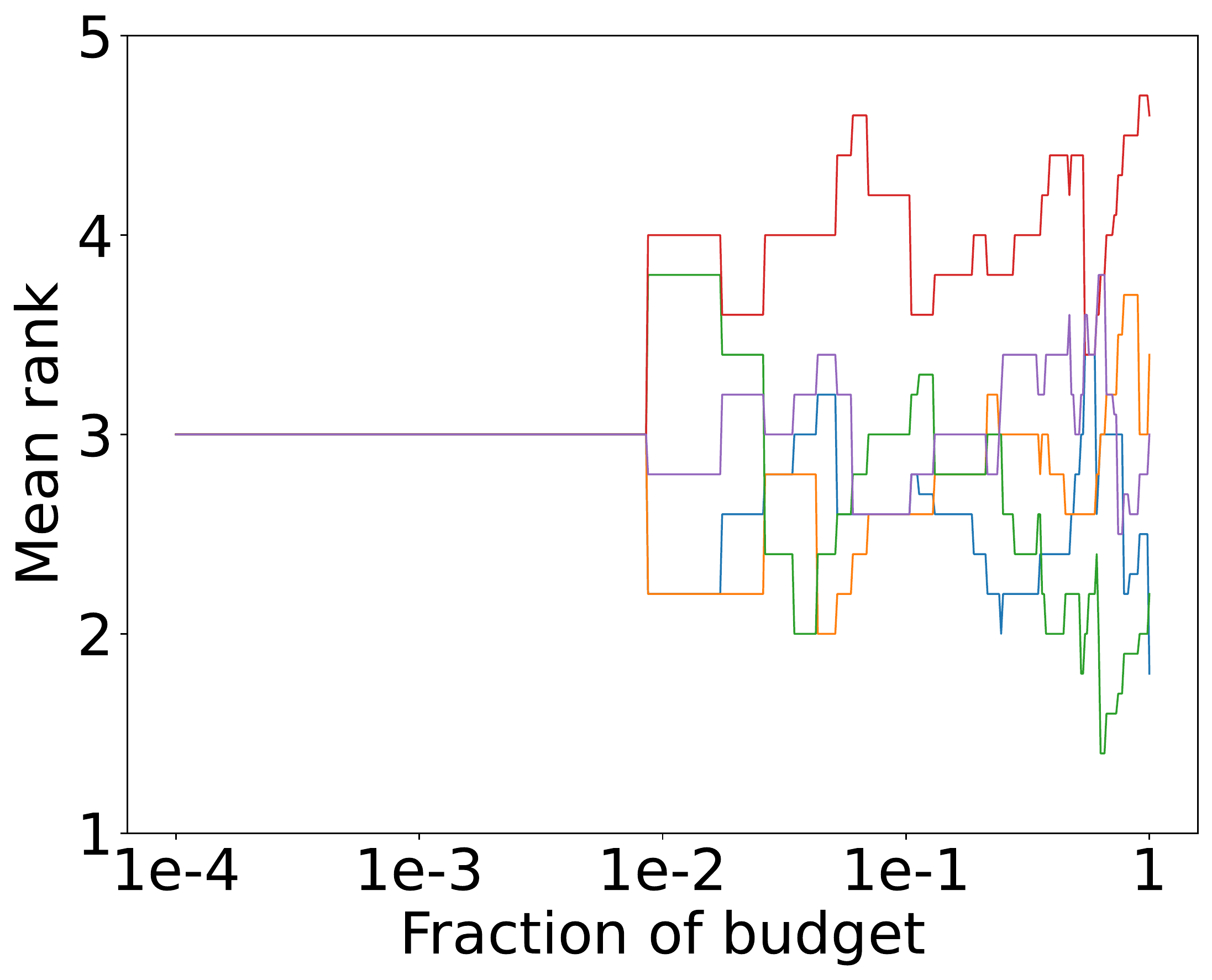}
		\caption{$\textit{BBO}_{146822\text{OpenML}}$}
		\label{fig:BBO_146822_openml_opt}
	\end{subfigure}
	\begin{subfigure}{0.25\linewidth}
		\centering
		\includegraphics[width=0.9\linewidth]{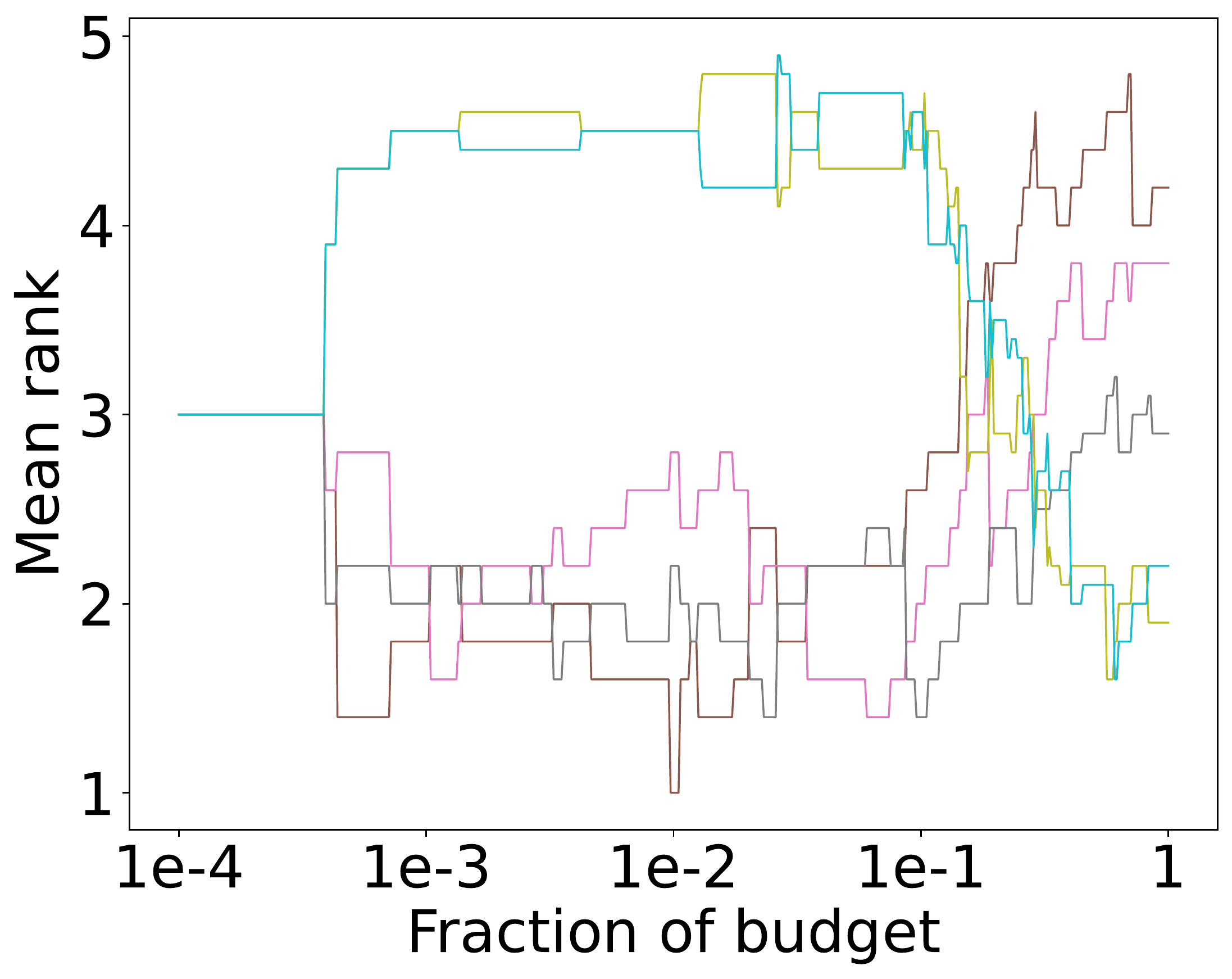}
		\caption{$\textit{MF}_{146822\text{OpenML}}$}
		\label{fig:MF_146822_openml_opt}
	\end{subfigure}
	
	\centering
	\hspace*{1.2cm}\begin{subfigure}{1.0\linewidth}
		\centering
		\includegraphics[width=0.95\linewidth]{materials/legend_rank_new.pdf}
	\end{subfigure}
	\vspace{-0.1in}
	
	\caption{Mean rank over time on LR benchmark (FedOPT).}
	\label{fig:entire_lr_tabular_opt_rank}
\end{figure}

\begin{figure}[htbp]
	\centering
	\begin{subfigure}{0.25\linewidth}
		\centering
		\includegraphics[width=0.9\linewidth]{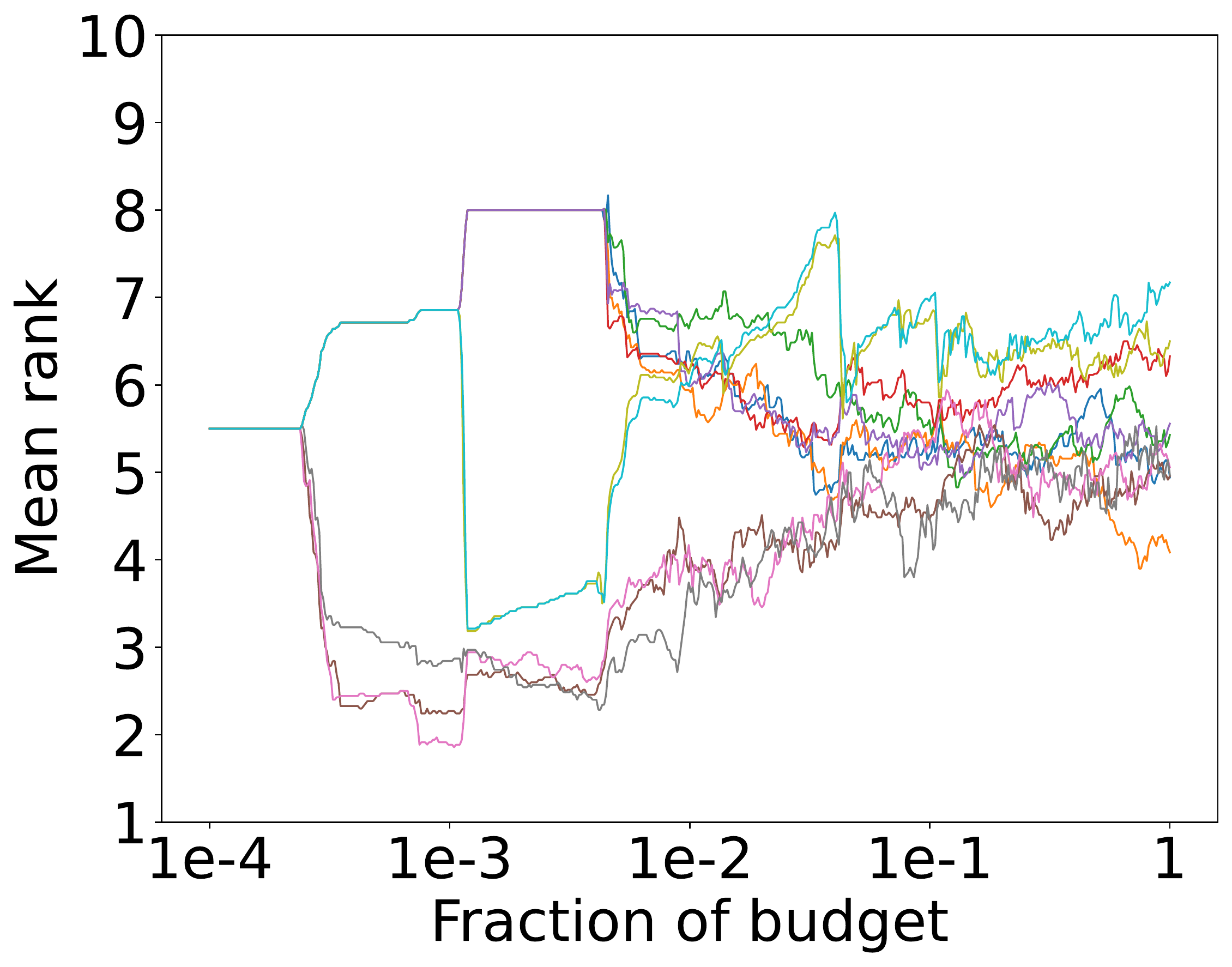}
		\caption{$\text{ALL}_{MLP}$}
		\label{fig:All_MLP_avg}
	\end{subfigure}
	\begin{subfigure}{0.25\linewidth}
		\centering
		\includegraphics[width=0.9\linewidth]{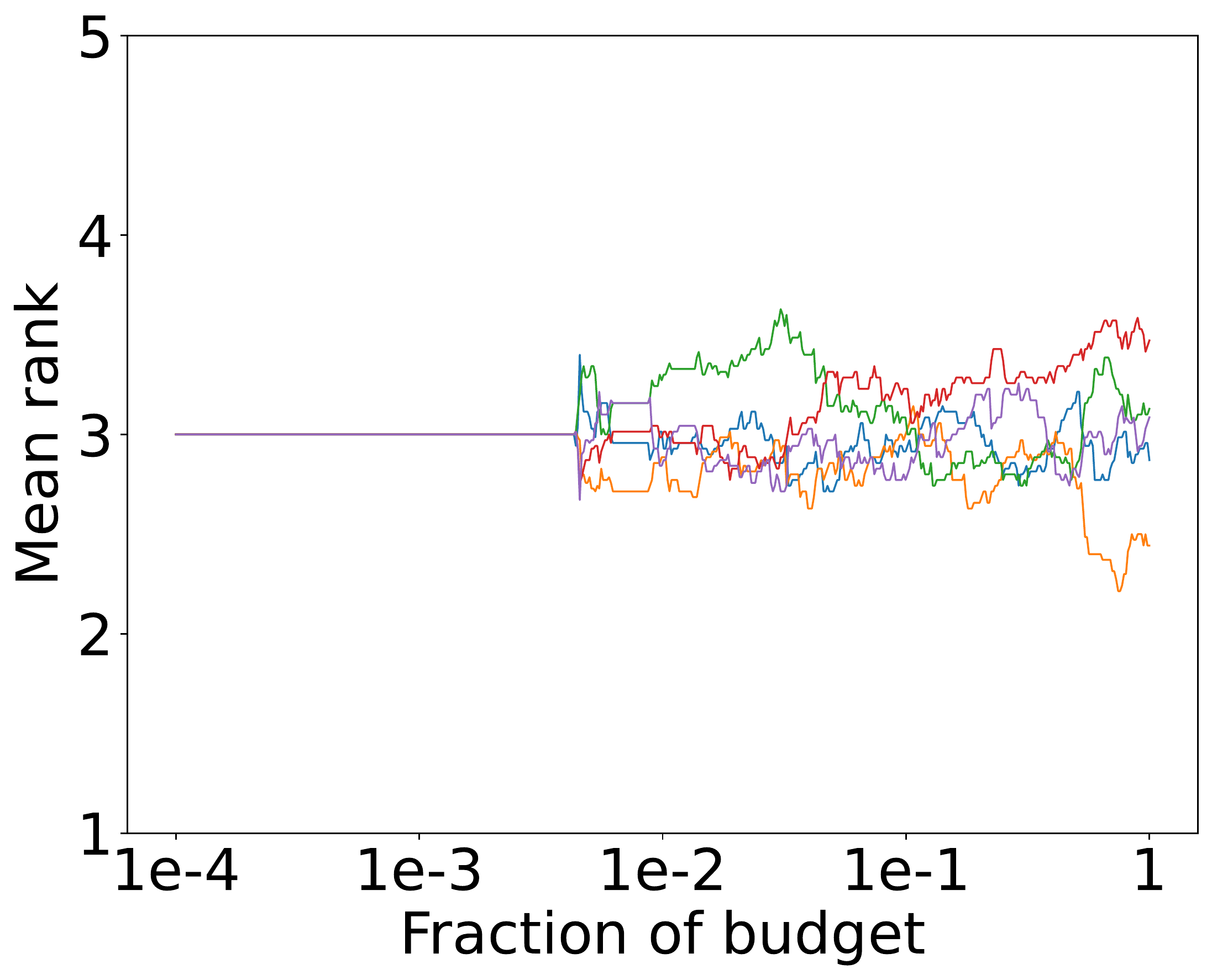}
		\caption{$\textit{BBO}_{MLP}$}
		\label{fig:BBO_MLP_avg}
	\end{subfigure}
	\begin{subfigure}{0.25\linewidth}
		\centering
		\includegraphics[width=0.9\linewidth]{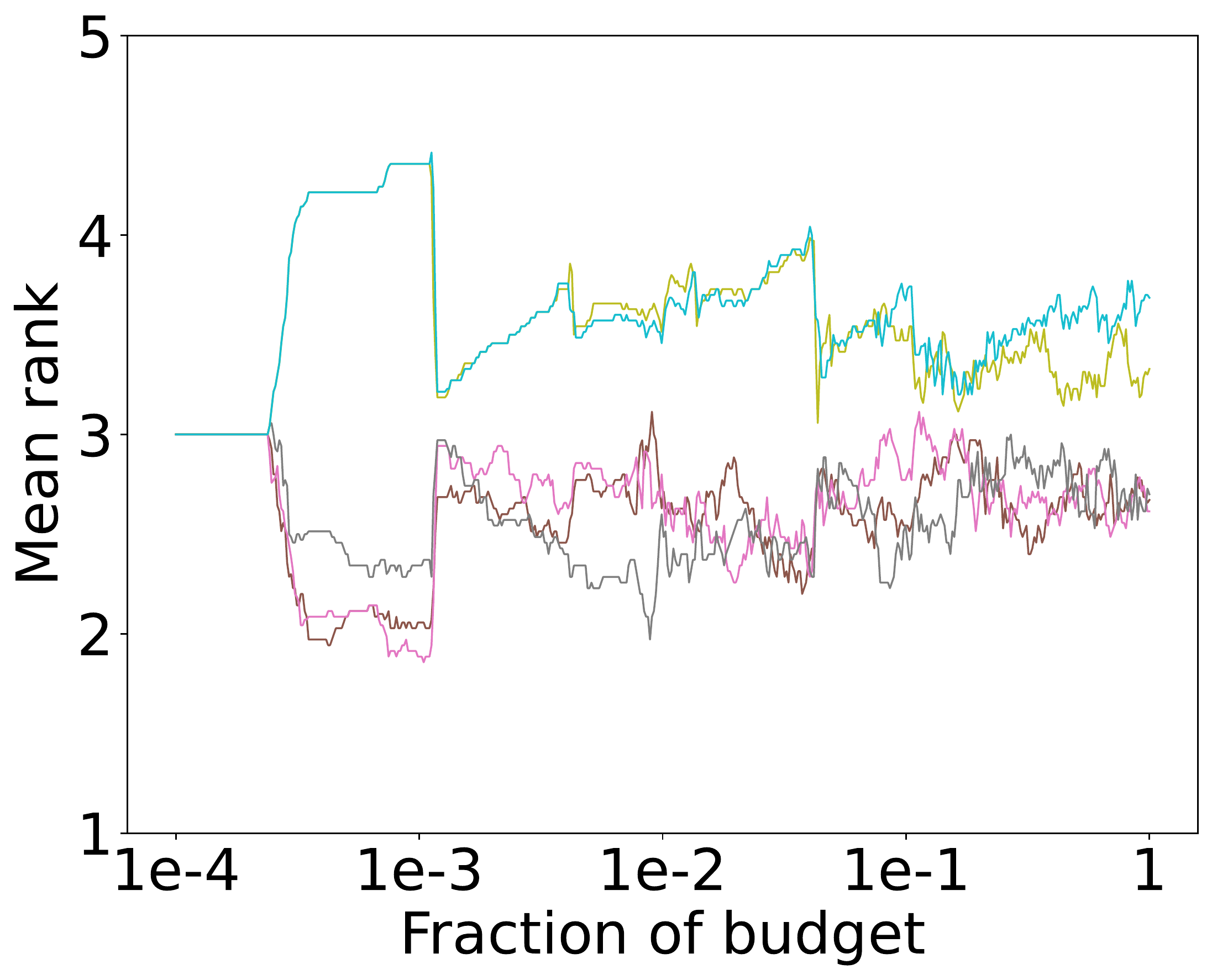}
		\caption{$\textit{MF}_{MLP}$}
		\label{fig:MF_MLP_avg}
	\end{subfigure}

	\begin{subfigure}{0.25\linewidth}
		\centering
		\includegraphics[width=0.9\linewidth]{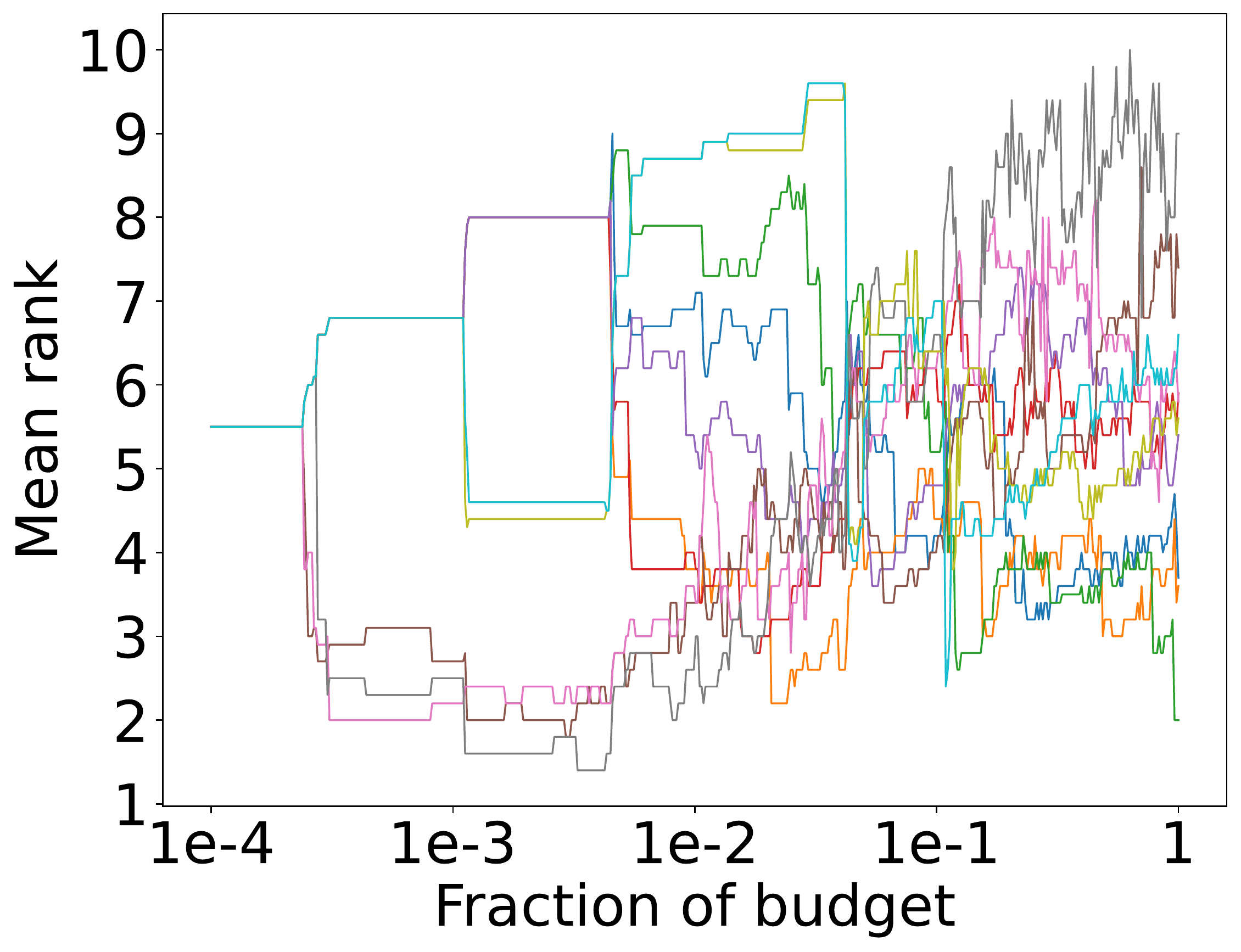}
		\caption{$\text{ALL}_{31\text{OpenML}}$}
		\label{fig:All_31_openml_avg_MLP}
	\end{subfigure}
	\begin{subfigure}{0.25\linewidth}
		\centering
		\includegraphics[width=0.9\linewidth]{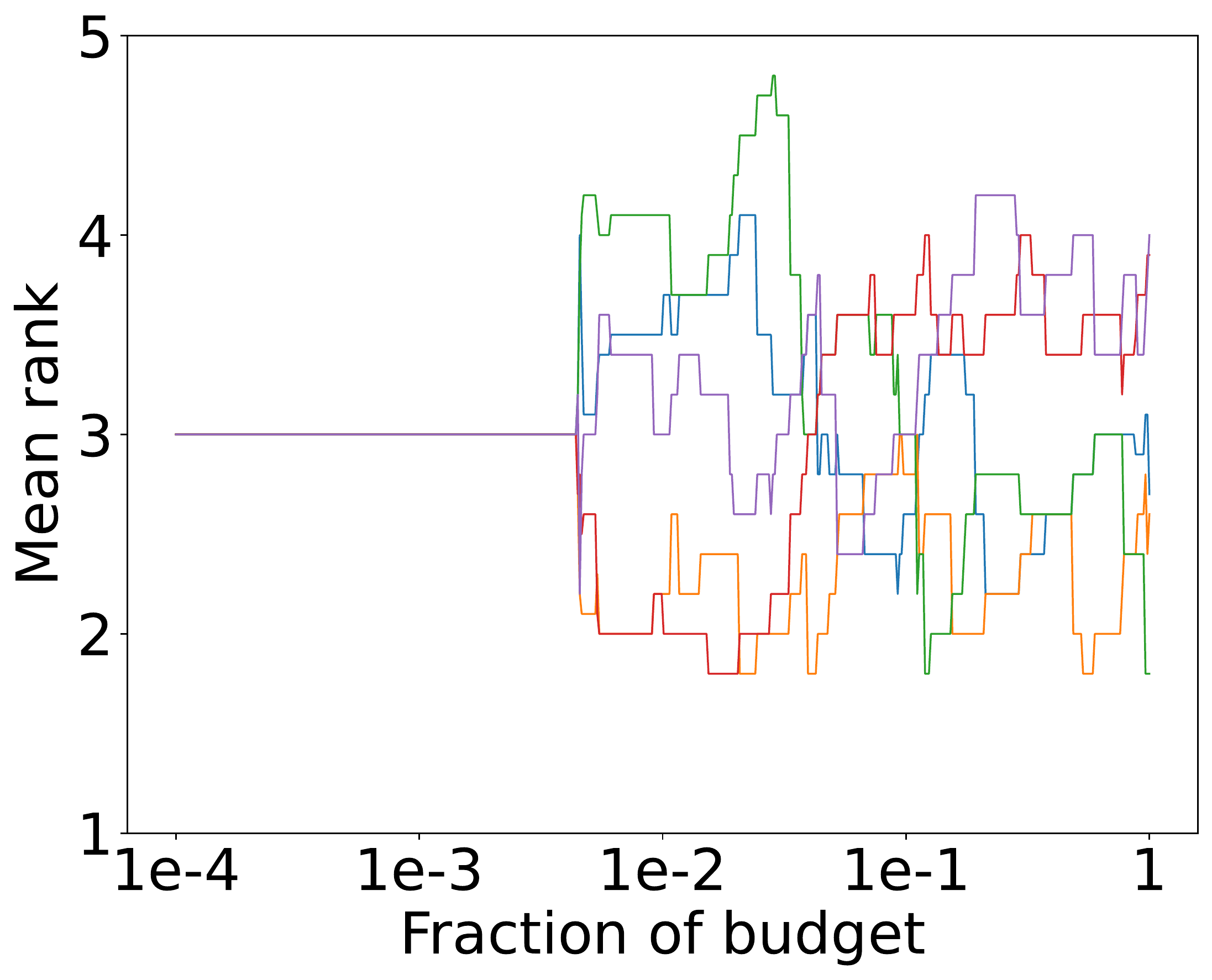}
		\caption{$\textit{BBO}_{31\text{OpenML}}$}
		\label{fig:BBO_31_openml_avg_MLP}
	\end{subfigure}
	\begin{subfigure}{0.25\linewidth}
		\centering
		\includegraphics[width=0.9\linewidth]{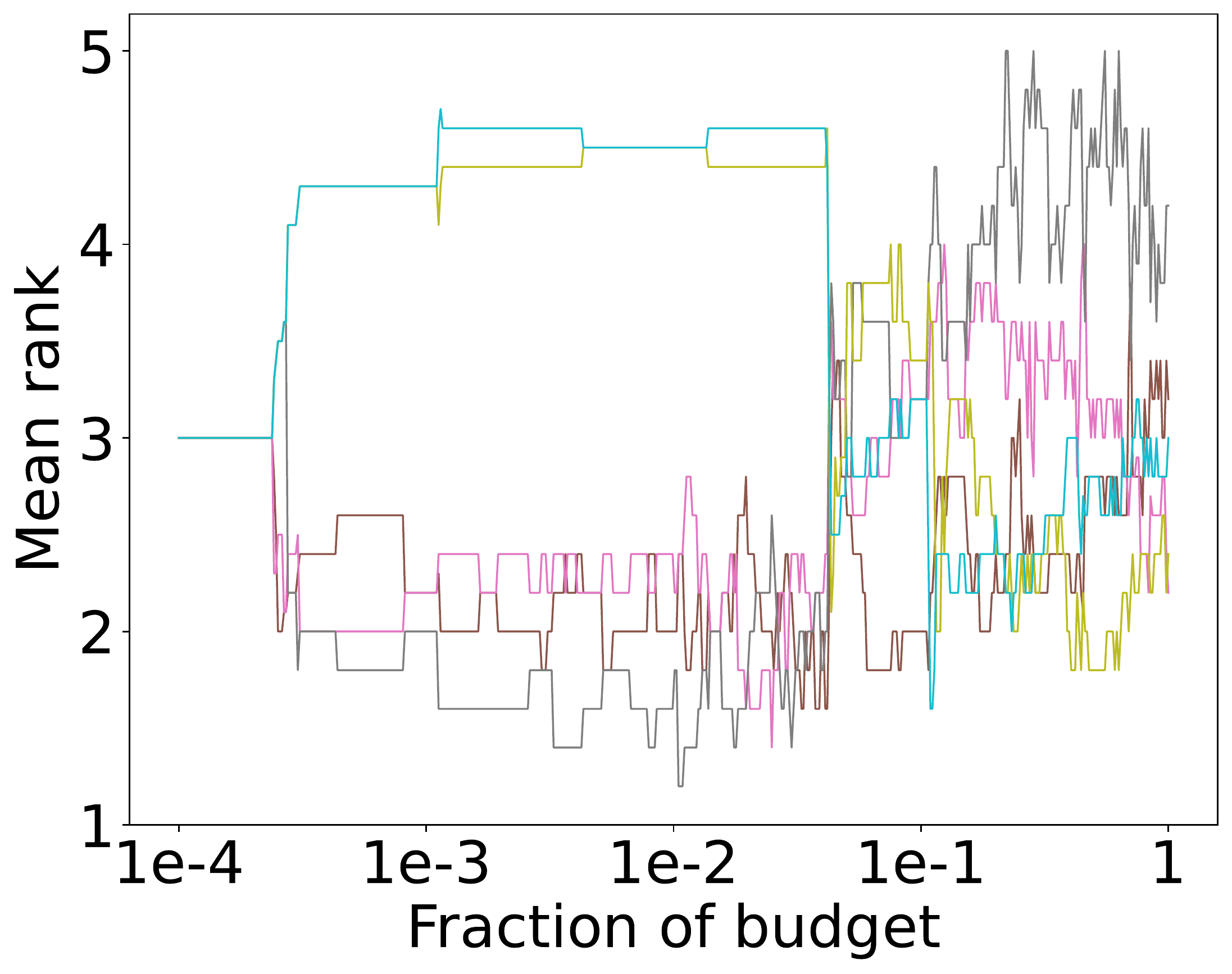}
		\caption{$\textit{MF}_{31\text{OpenML}}$}
		\label{fig:MF_31_openml_avg_MLP}
	\end{subfigure}
	
	\begin{subfigure}{0.25\linewidth}
		\centering
		\includegraphics[width=0.9\linewidth]{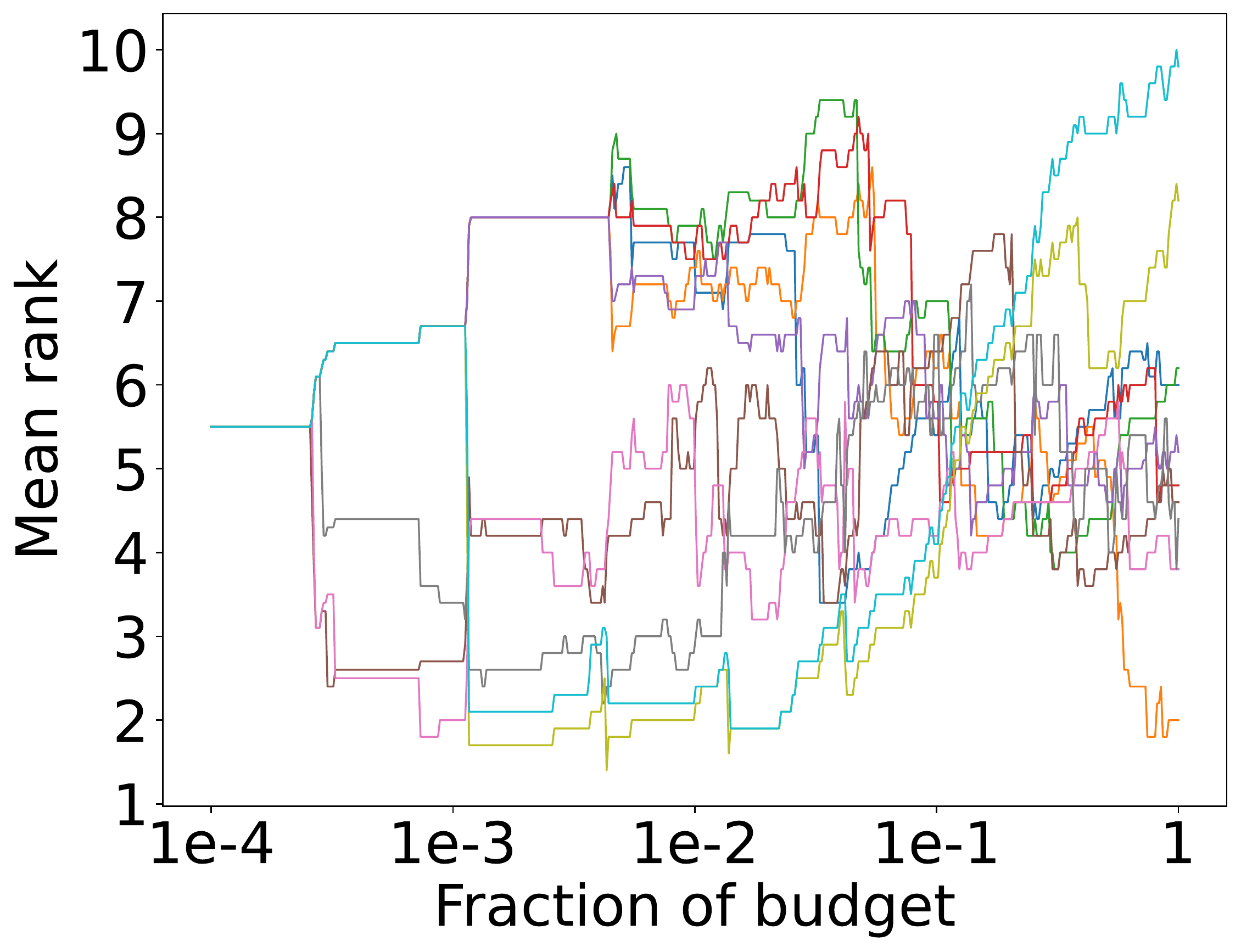}
		\caption{$\text{ALL}_{53\text{OpenML}}$}
		\label{fig:All_53_openml_avg_MLP}
	\end{subfigure}
	\begin{subfigure}{0.25\linewidth}
		\centering
		\includegraphics[width=0.9\linewidth]{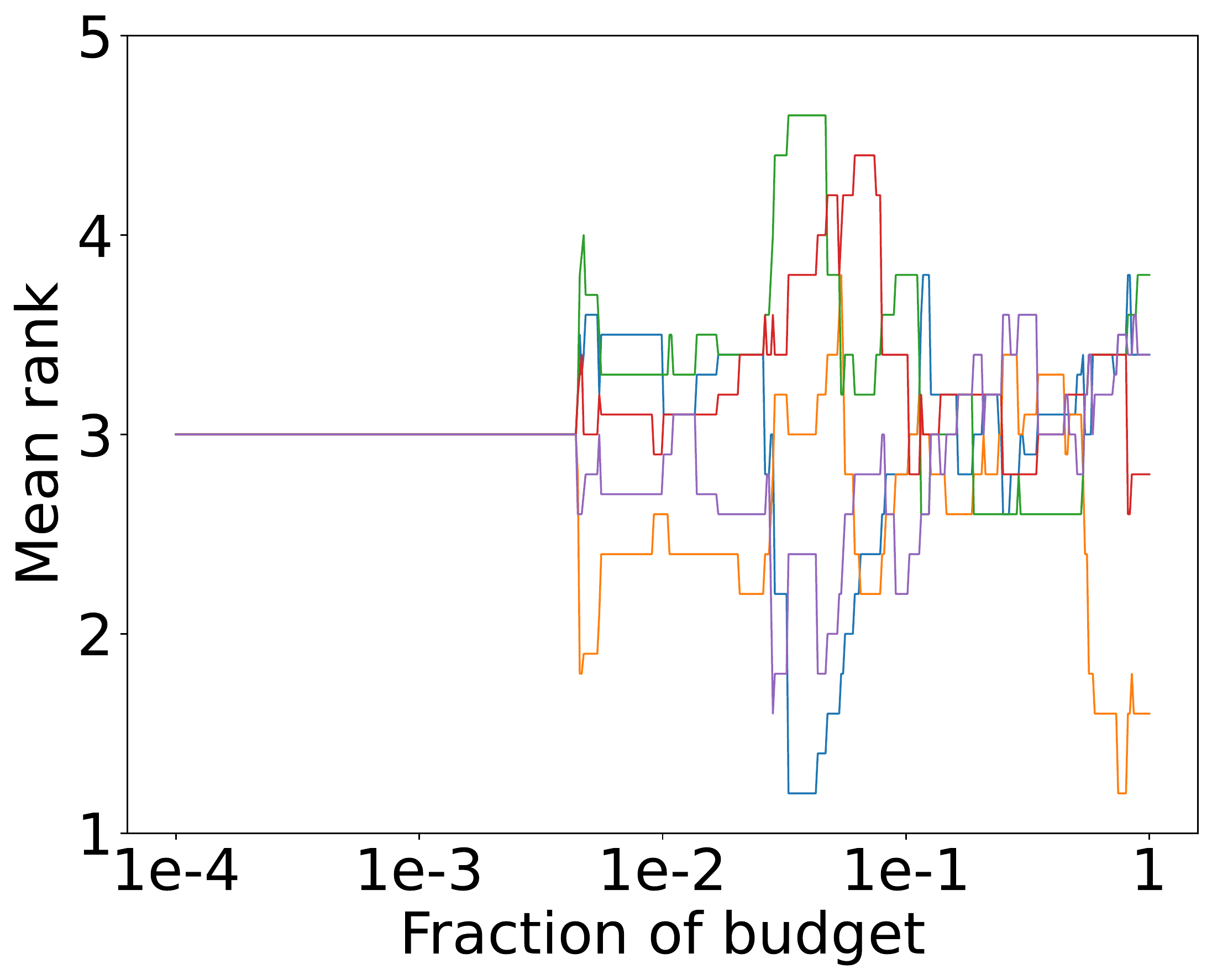}
		\caption{$\textit{BBO}_{53\text{OpenML}}$}
		\label{fig:BBO_53_openml_avg_MLP}
	\end{subfigure}
	\begin{subfigure}{0.25\linewidth}
		\centering
		\includegraphics[width=0.9\linewidth]{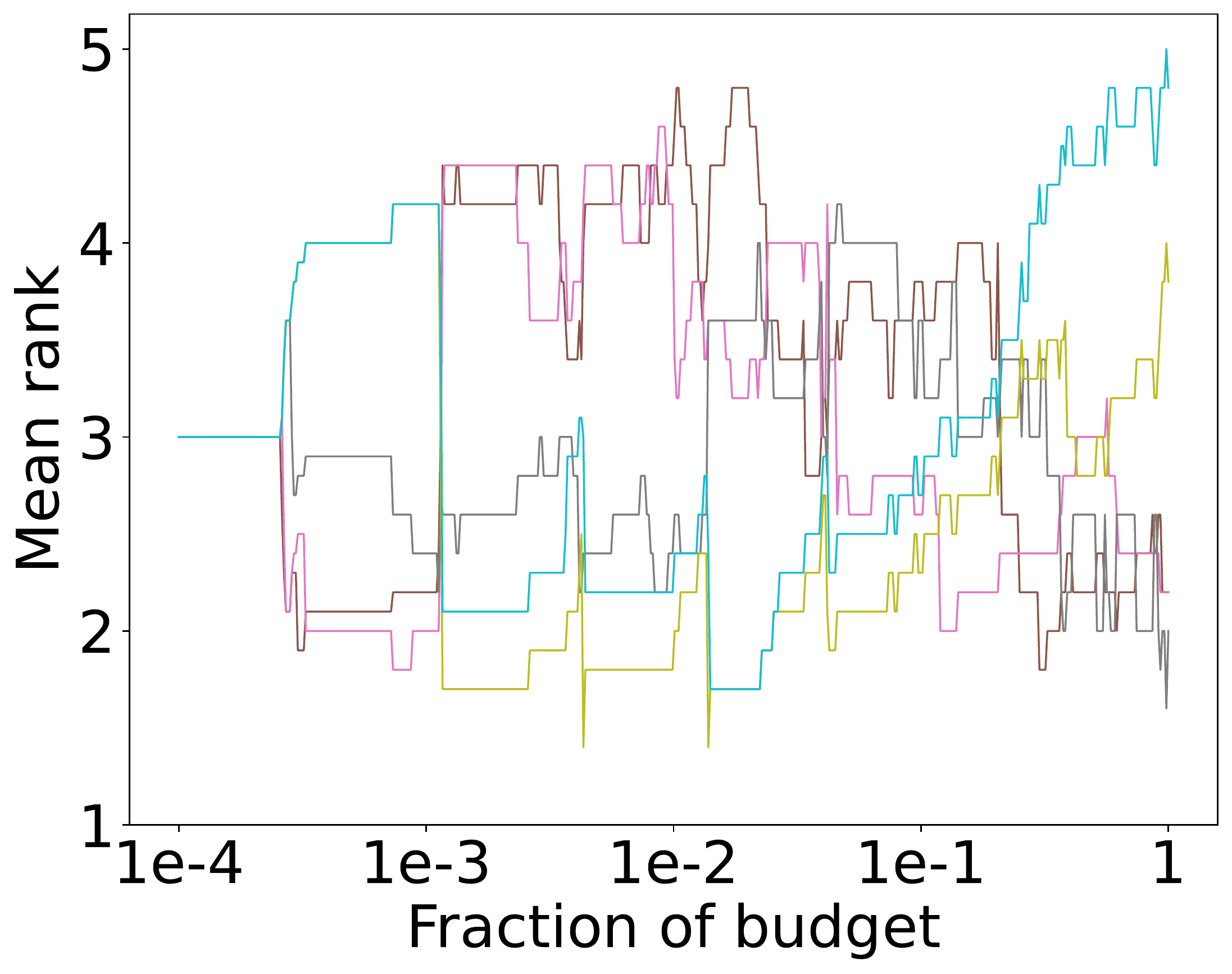}
		\caption{$\textit{MF}_{53\text{OpenML}}$}
		\label{fig:MF_53_openml_avg_MLP}
	\end{subfigure}
	
	\begin{subfigure}{0.25\linewidth}
		\centering
		\includegraphics[width=0.9\linewidth]{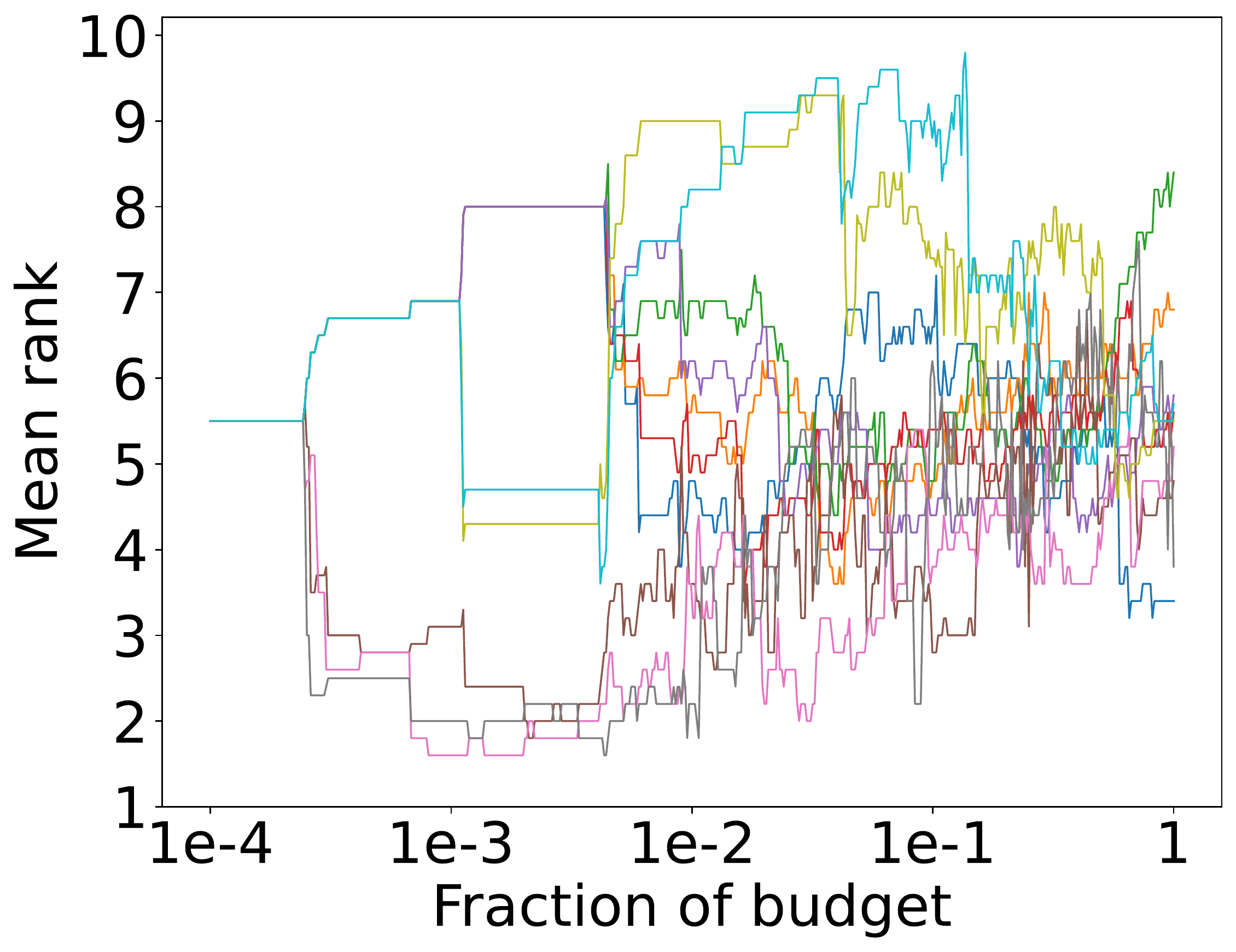}
		\caption{$\text{ALL}_{10101\text{OpenML}}$}
		\label{fig:All_10101_openml_avg_MLP}
	\end{subfigure}
	\begin{subfigure}{0.25\linewidth}
		\centering
		\includegraphics[width=0.9\linewidth]{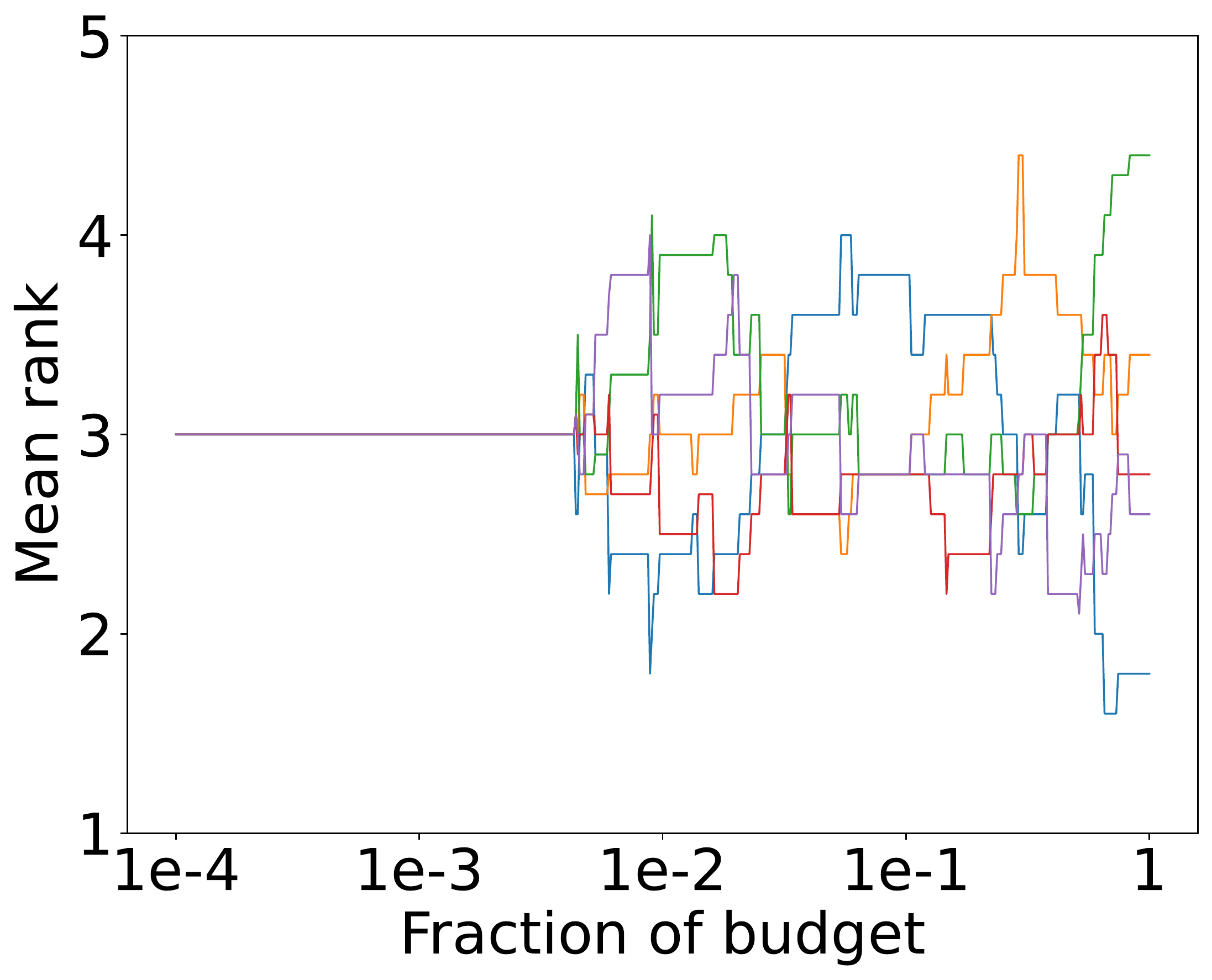}
		\caption{$\textit{BBO}_{10101\text{OpenML}}$}
		\label{fig:BBO_10101_openml_avg_MLP}
	\end{subfigure}
	\begin{subfigure}{0.25\linewidth}
		\centering
		\includegraphics[width=0.9\linewidth]{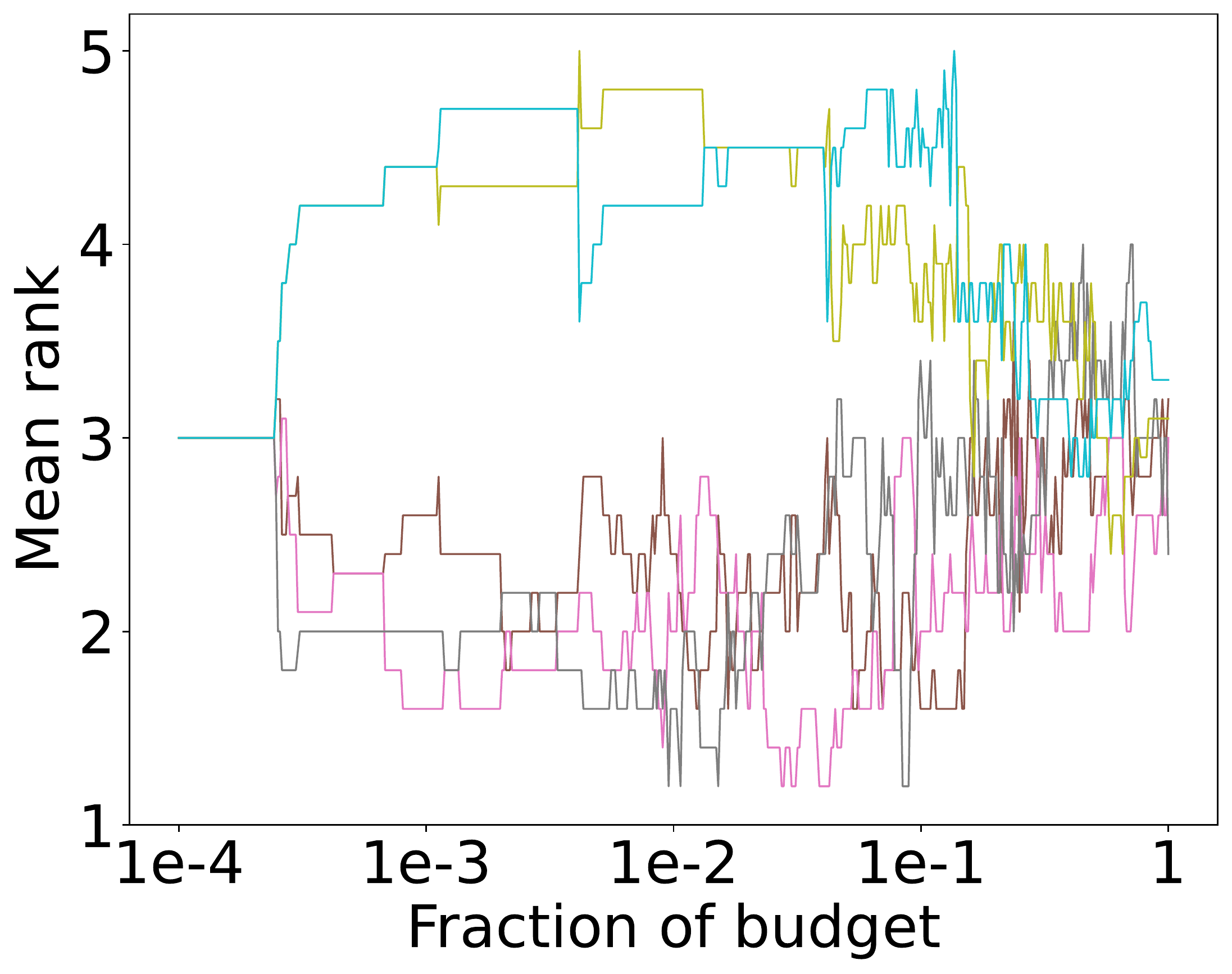}
		\caption{$\textit{MF}_{10101\text{OpenML}}$}
		\label{fig:MF_10101_openml_avg_MLP}
	\end{subfigure}
	
	\begin{subfigure}{0.25\linewidth}
		\centering
		\includegraphics[width=0.9\linewidth]{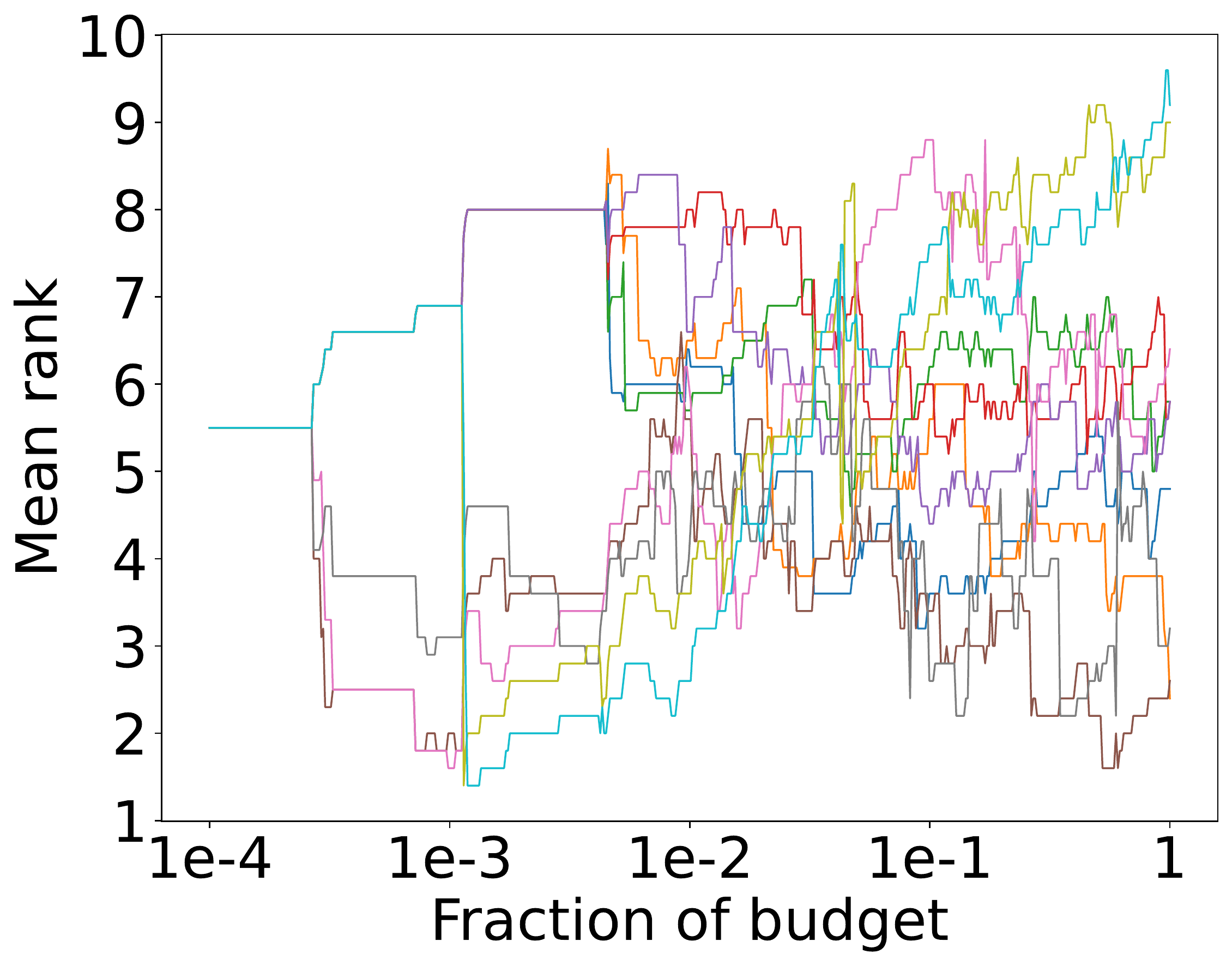}
		\caption{$\text{ALL}_{146818\text{OpenML}}$}
		\label{fig:All_146818_openml_avg_MLP}
	\end{subfigure}
	\begin{subfigure}{0.25\linewidth}
		\centering
		\includegraphics[width=0.9\linewidth]{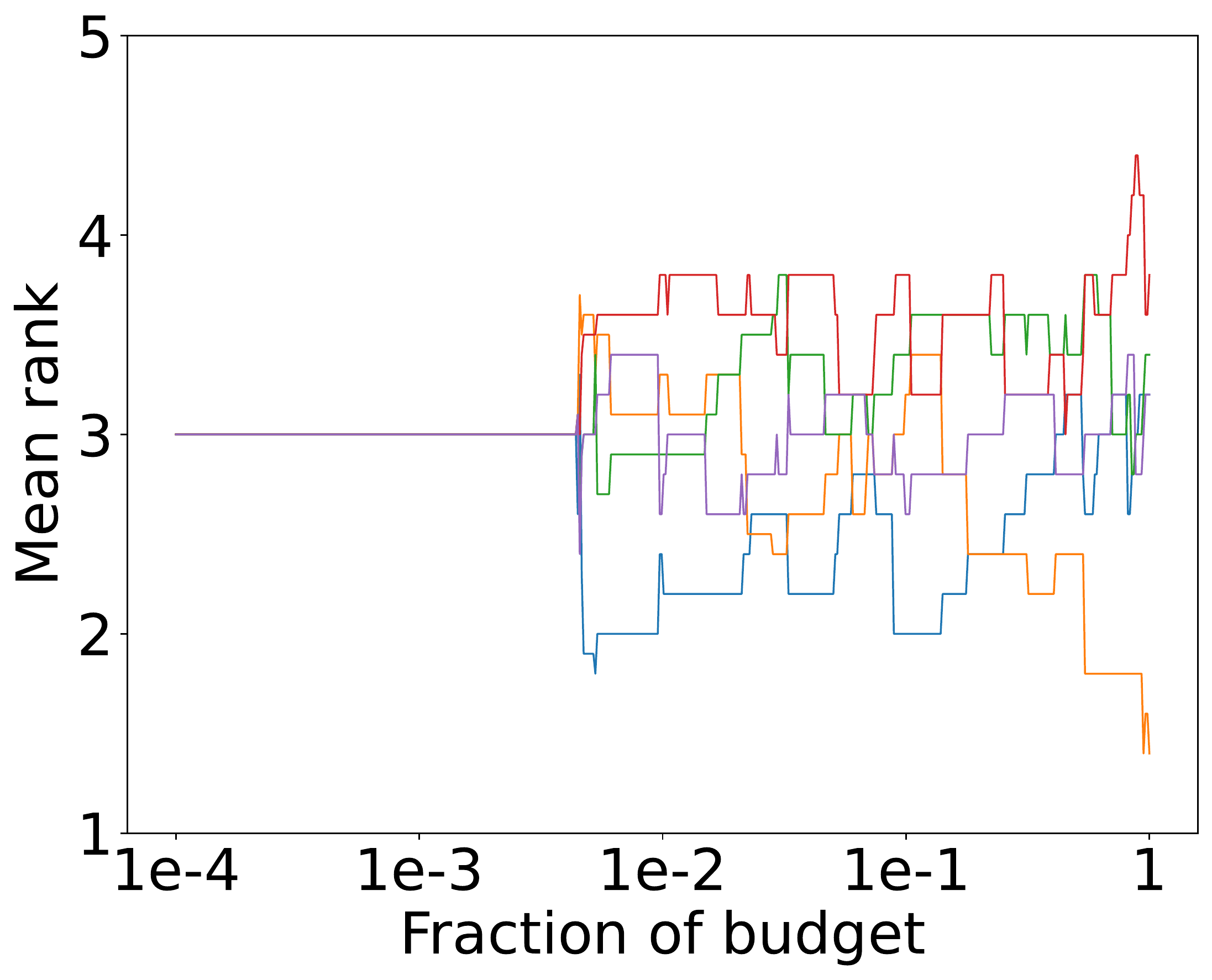}
		\caption{$\textit{BBO}_{146818\text{OpenML}}$}
		\label{fig:BBO_146818_openml_avg_MLP}
	\end{subfigure}
	\begin{subfigure}{0.25\linewidth}
		\centering
		\includegraphics[width=0.9\linewidth]{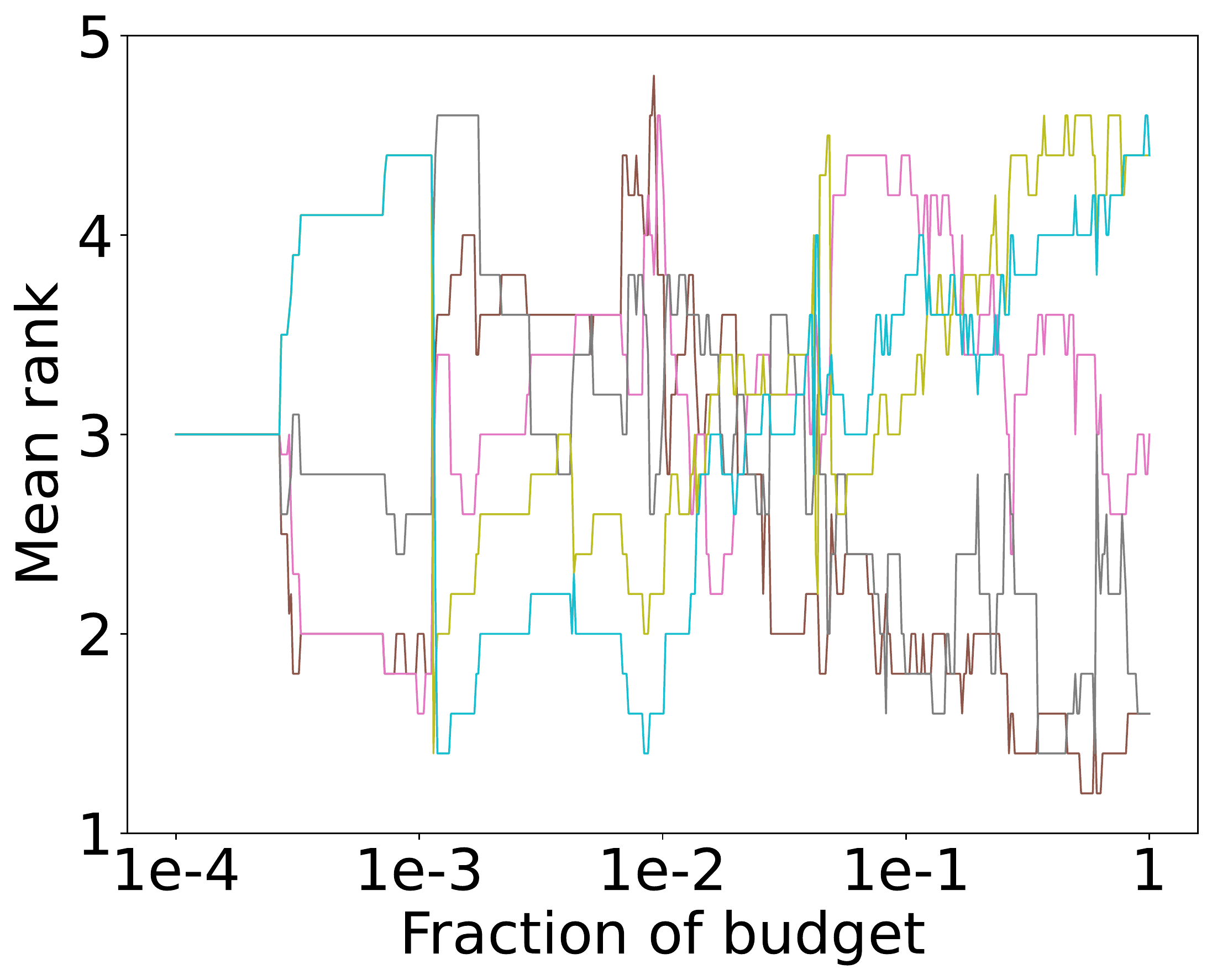}
		\caption{$\textit{MF}_{146818\text{OpenML}}$}
		\label{fig:MF_146818_openml_avg_MLP}
	\end{subfigure}
	
	\begin{subfigure}{0.25\linewidth}
		\centering
		\includegraphics[width=0.9\linewidth]{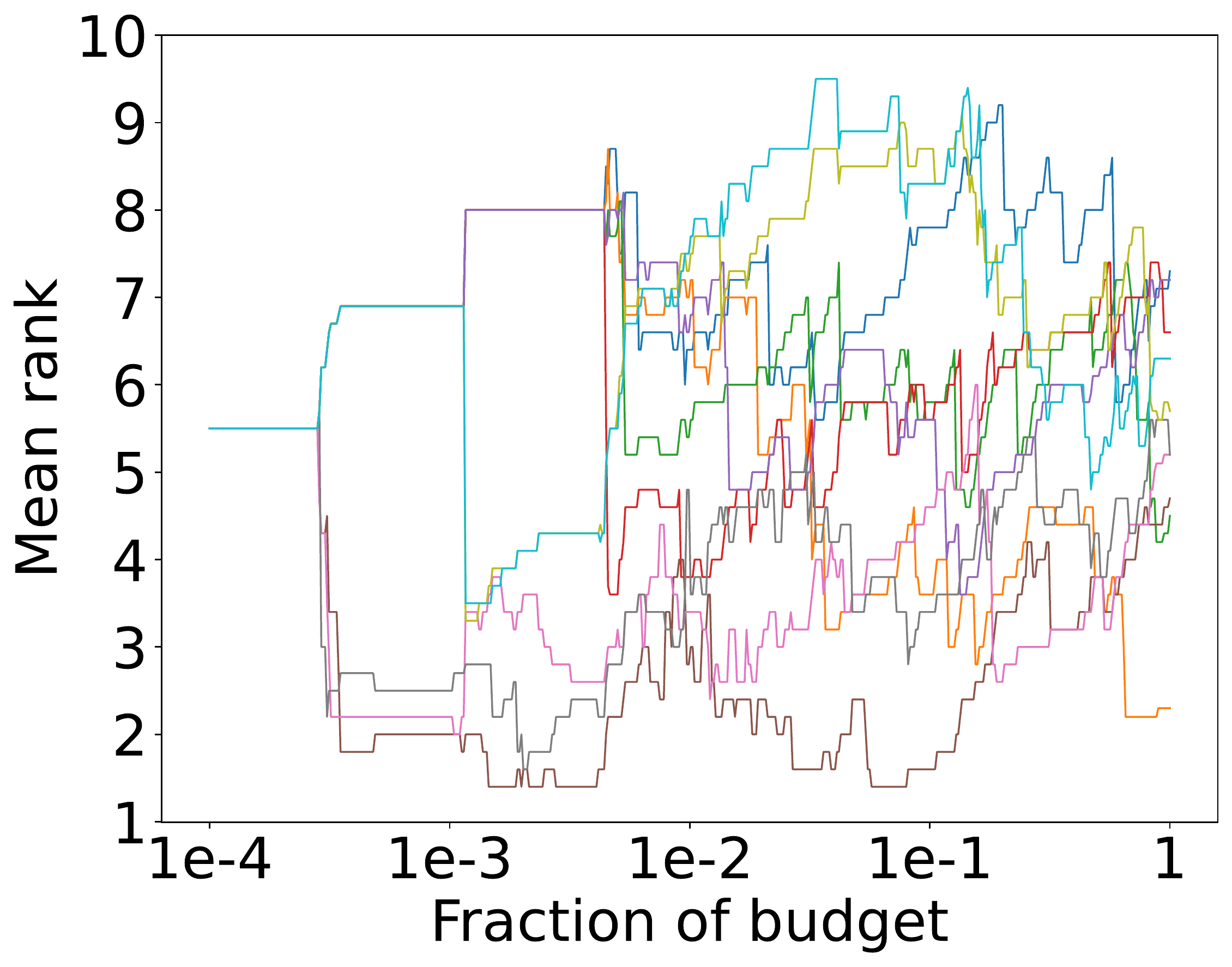}
		\caption{$\text{ALL}_{146821\text{OpenML}}$}
		\label{fig:All_146821_openml_avg_MLP}
	\end{subfigure}
	\begin{subfigure}{0.25\linewidth}
		\centering
		\includegraphics[width=0.9\linewidth]{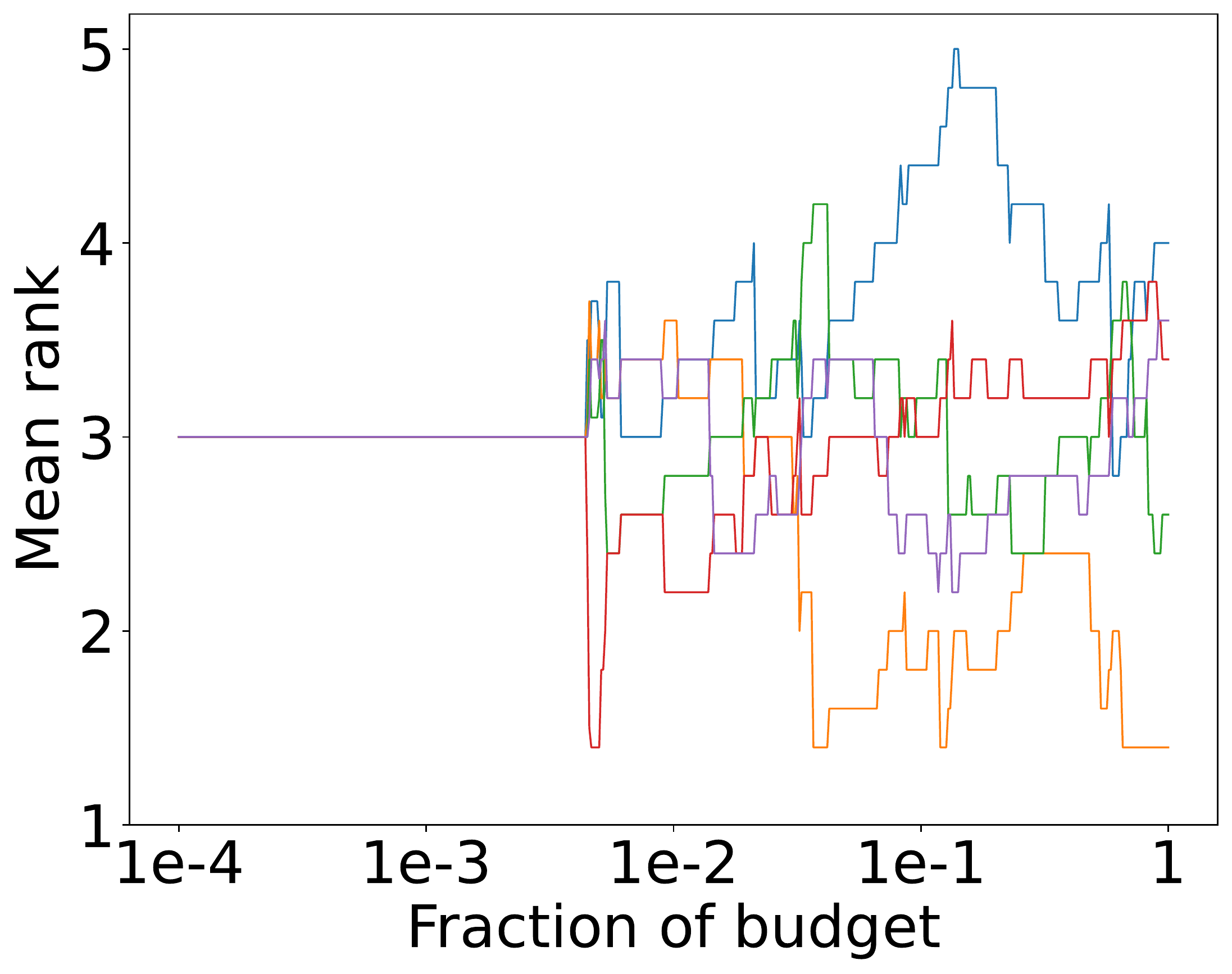}
		\caption{$\textit{BBO}_{146821\text{OpenML}}$}
		\label{fig:BBO_146821_openml_avg_MLP}
	\end{subfigure}
	\begin{subfigure}{0.25\linewidth}
		\centering
		\includegraphics[width=0.9\linewidth]{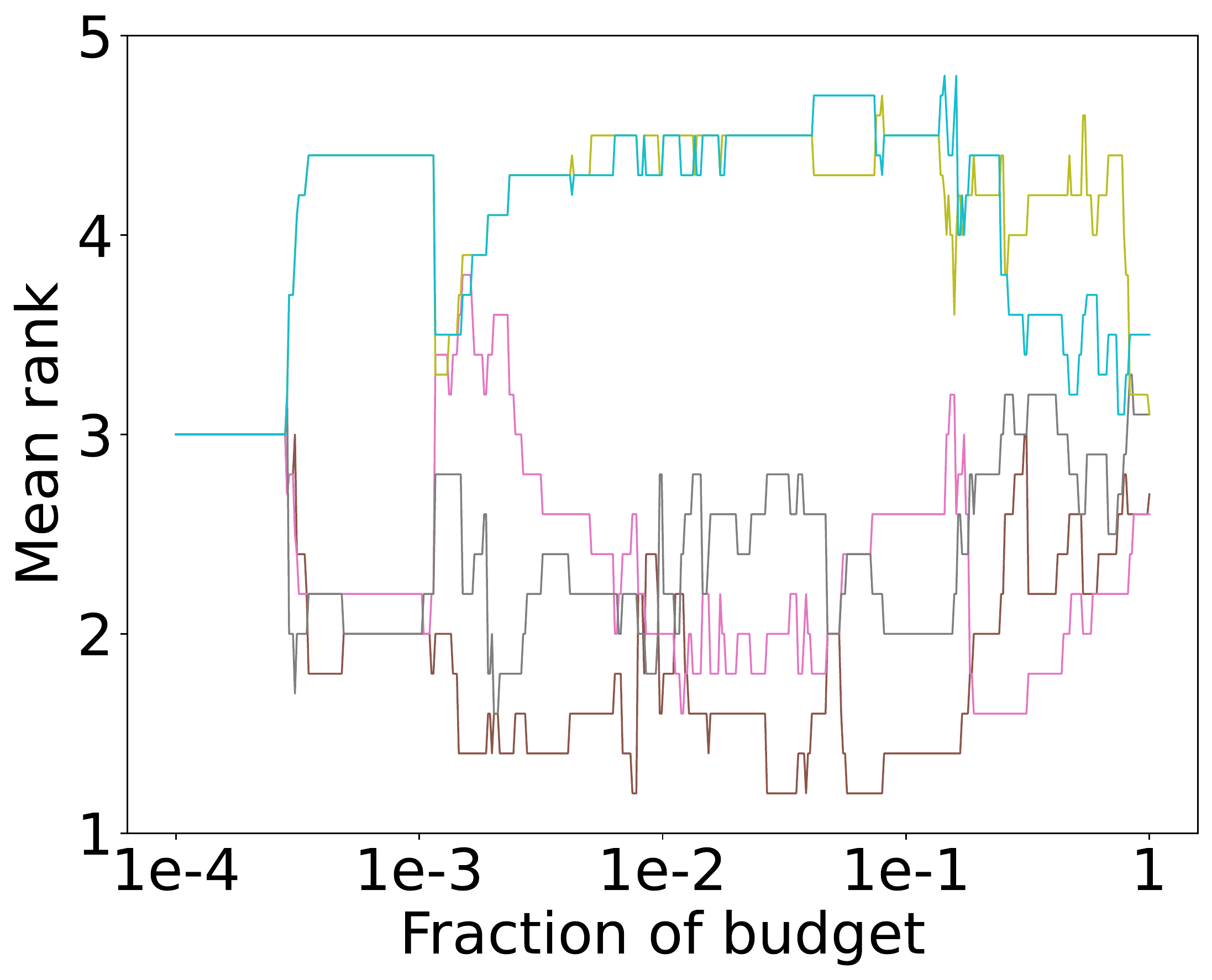}
		\caption{$\textit{MF}_{146821\text{OpenML}}$}
		\label{fig:MF_146821_openml_avg_MLP}
	\end{subfigure}
	
	\begin{subfigure}{0.25\linewidth}
		\centering
		\includegraphics[width=0.9\linewidth]{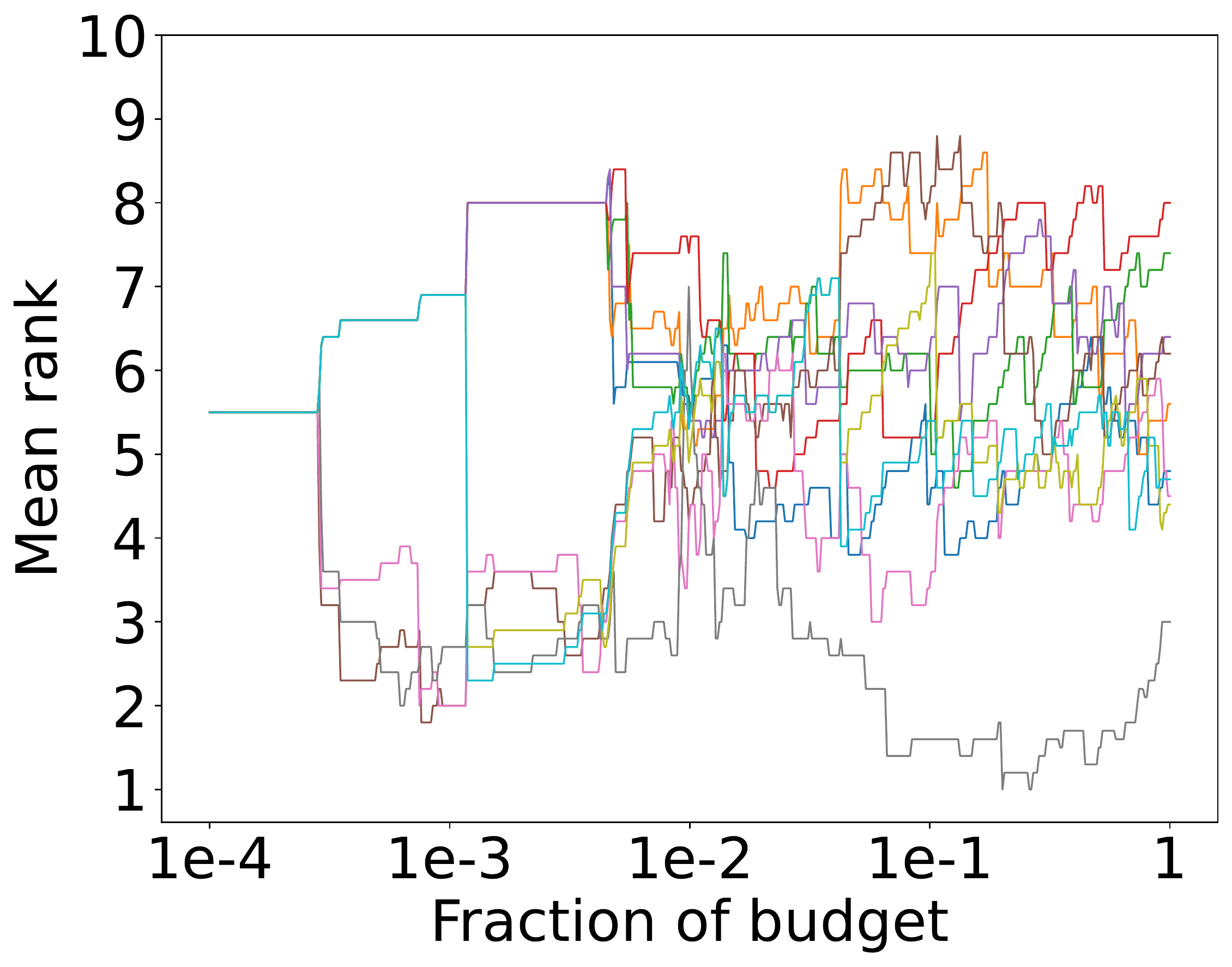}
		\caption{$\text{ALL}_{146822\text{OpenML}}$}
		\label{fig:All_146822_openml_avg_MLP}
	\end{subfigure}
	\begin{subfigure}{0.25\linewidth}
		\centering
		\includegraphics[width=0.9\linewidth]{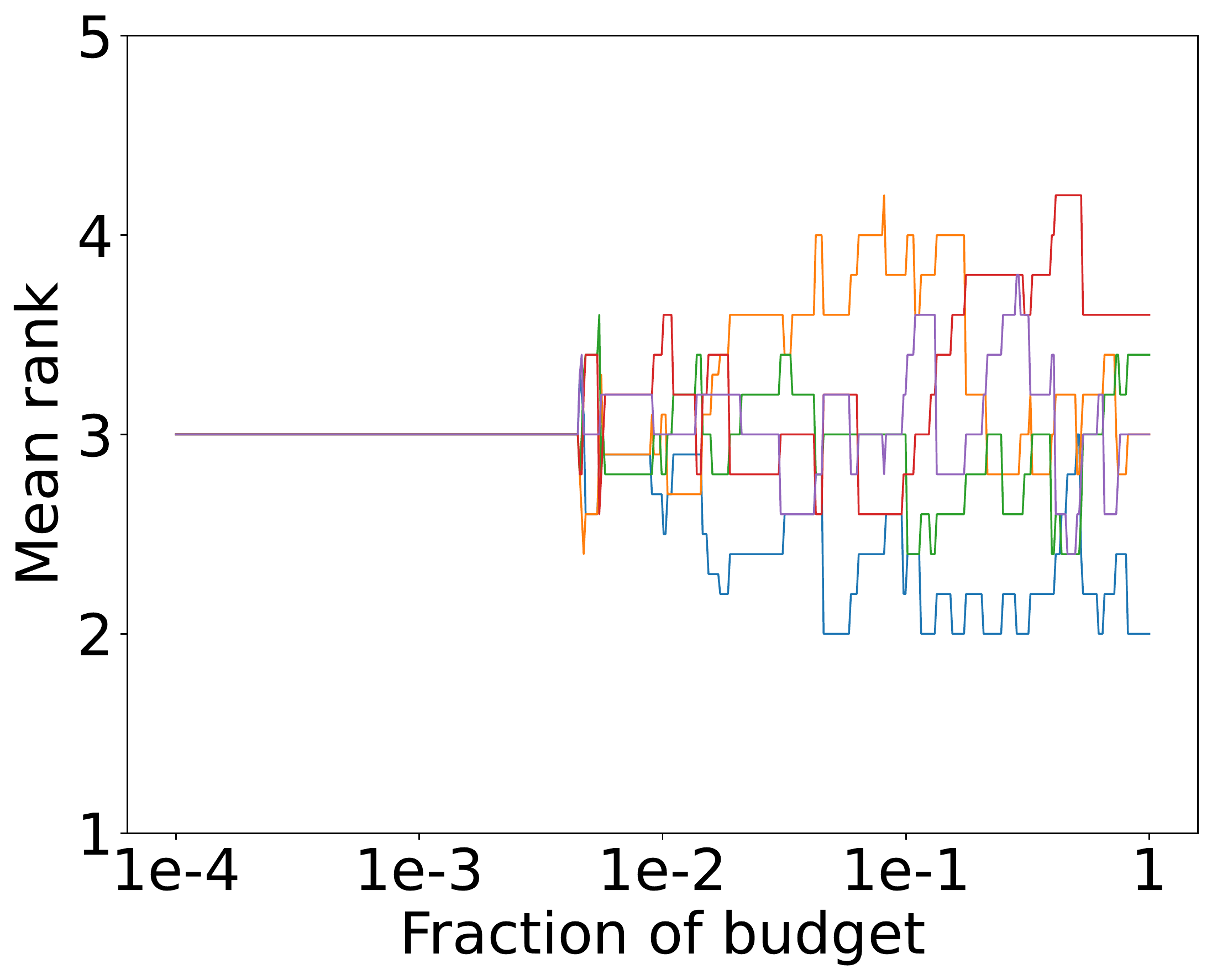}
		\caption{$\textit{BBO}_{146822\text{OpenML}}$}
		\label{fig:BBO_146822_openml_avg_MLP}
	\end{subfigure}
	\begin{subfigure}{0.25\linewidth}
		\centering
		\includegraphics[width=0.9\linewidth]{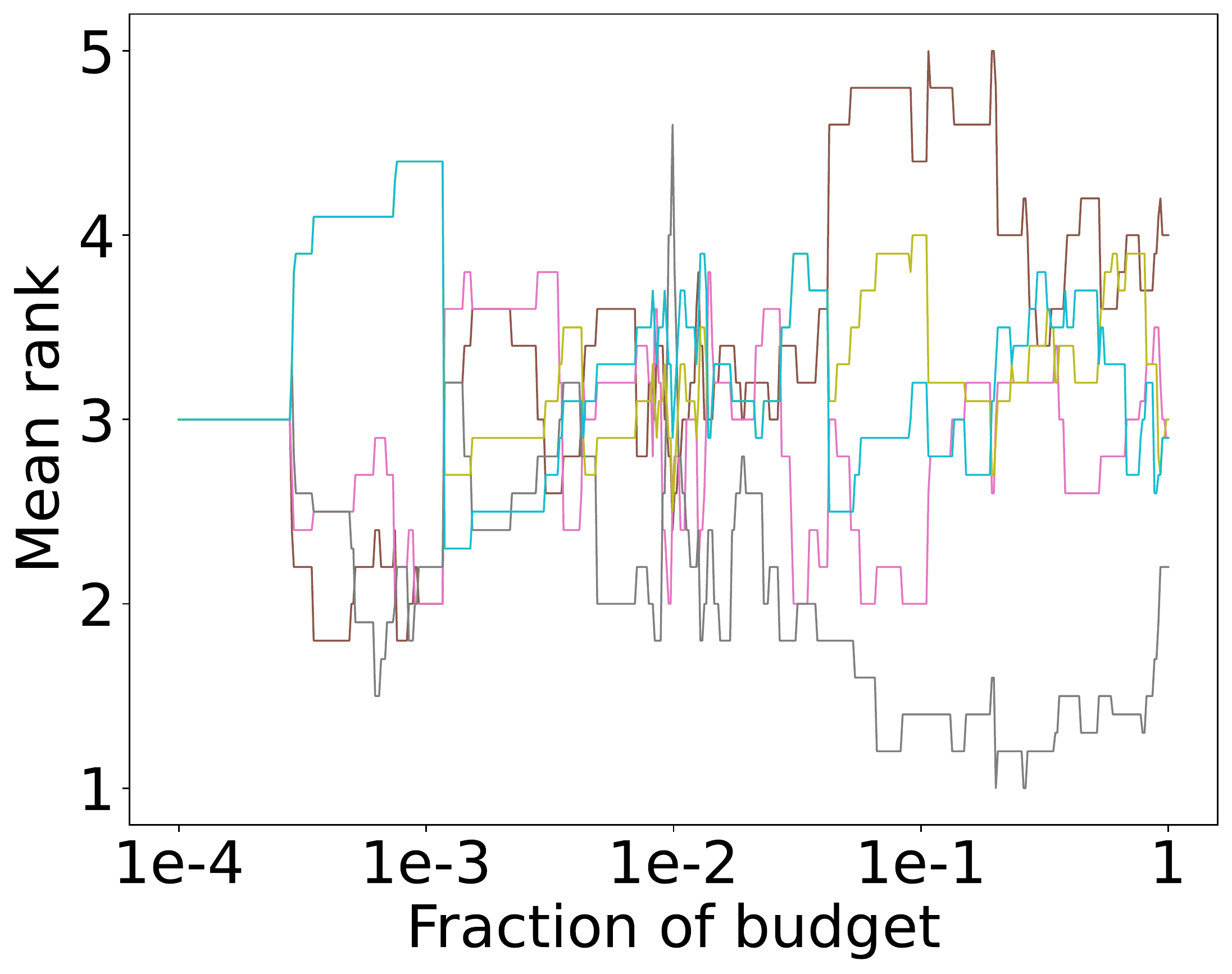}
		\caption{$\textit{MF}_{146822\text{OpenML}}$}
		\label{fig:MF_146822_openml_avg_MLP}
	\end{subfigure}
	
	\centering
	\hspace*{1.2cm}\begin{subfigure}{1.0\linewidth}
		\centering
		\includegraphics[width=0.95\linewidth]{materials/legend_rank_new.pdf}
	\end{subfigure}
	\vspace{-0.1in}
	
	\caption{Mean rank over time on MLP benchmark (FedAvg).}
	\label{fig:entire_MLP_tabular_avg_rank}
\end{figure}

\begin{figure}[htbp]
	\centering
	\begin{subfigure}{0.25\linewidth}
		\centering
		\includegraphics[width=0.9\linewidth]{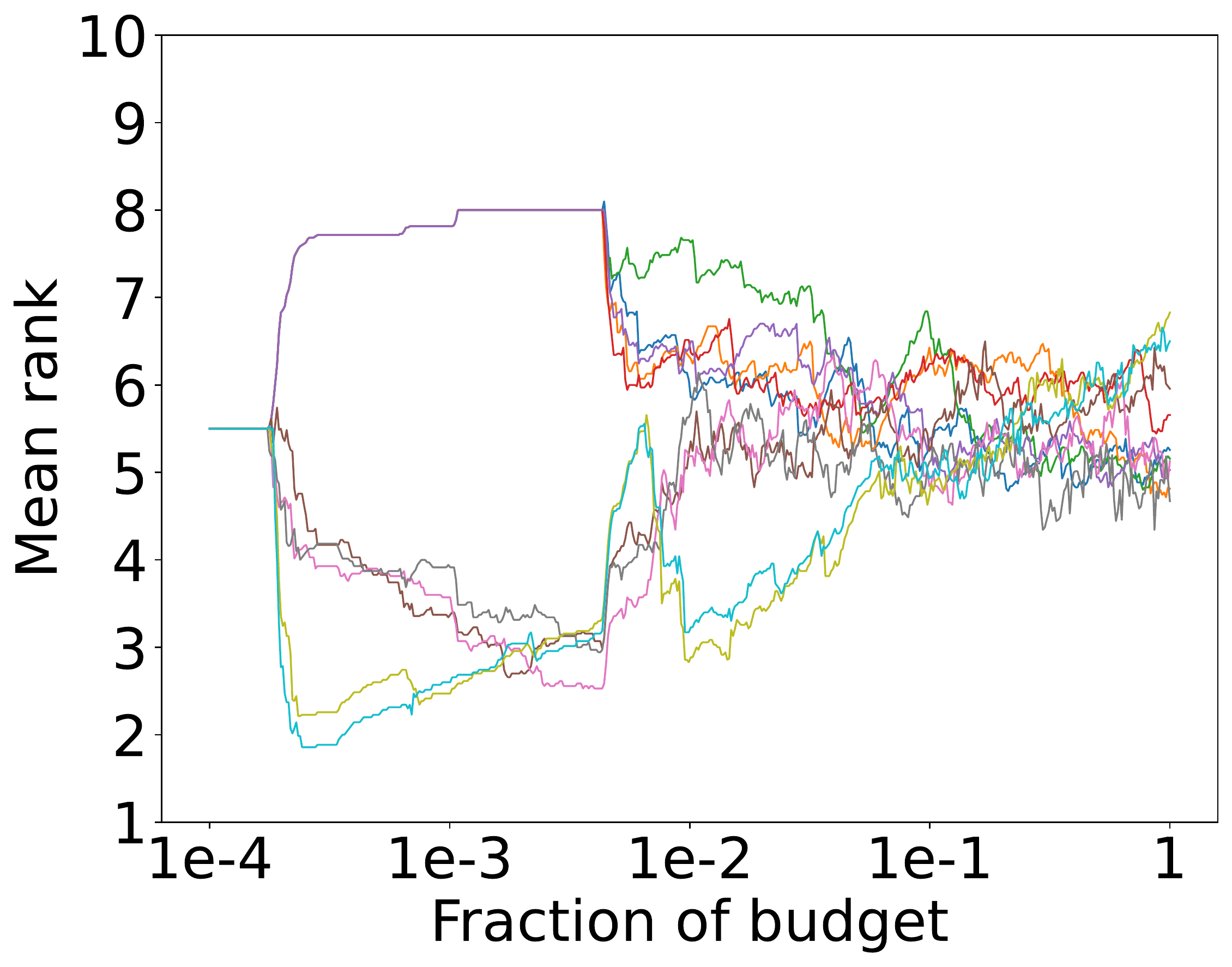}
		\caption{$\text{ALL}_{MLP}$}
		\label{fig:All_MLP_opt}
	\end{subfigure}
	\begin{subfigure}{0.25\linewidth}
		\centering
		\includegraphics[width=0.9\linewidth]{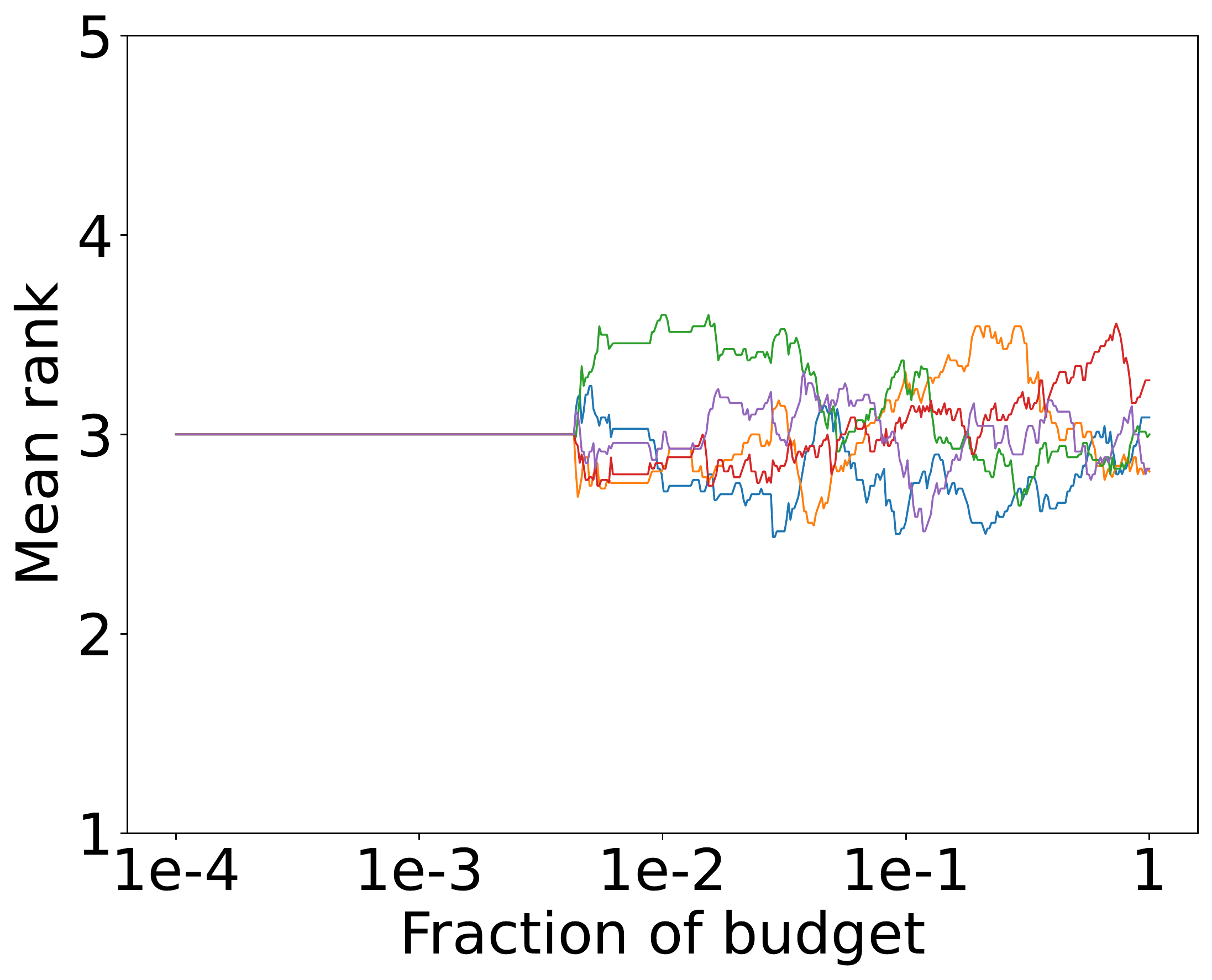}
		\caption{$\textit{BBO}_{MLP}$}
		\label{fig:BBO_MLP_opt}
	\end{subfigure}
	\begin{subfigure}{0.25\linewidth}
		\centering
		\includegraphics[width=0.9\linewidth]{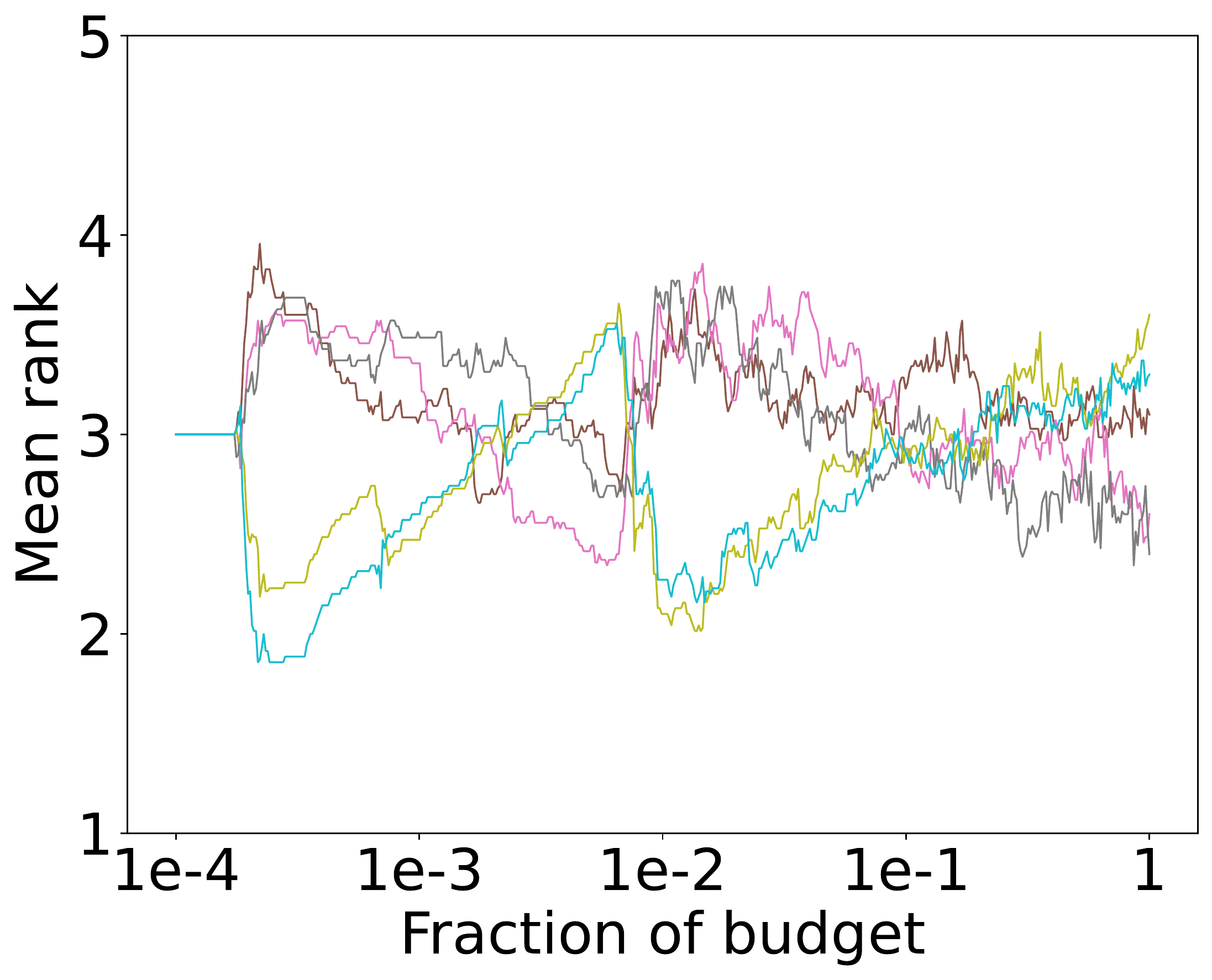}
		\caption{$\textit{MF}_{MLP}$}
		\label{fig:MF_MLP_opt}
	\end{subfigure}

	\begin{subfigure}{0.25\linewidth}
		\centering
		\includegraphics[width=0.9\linewidth]{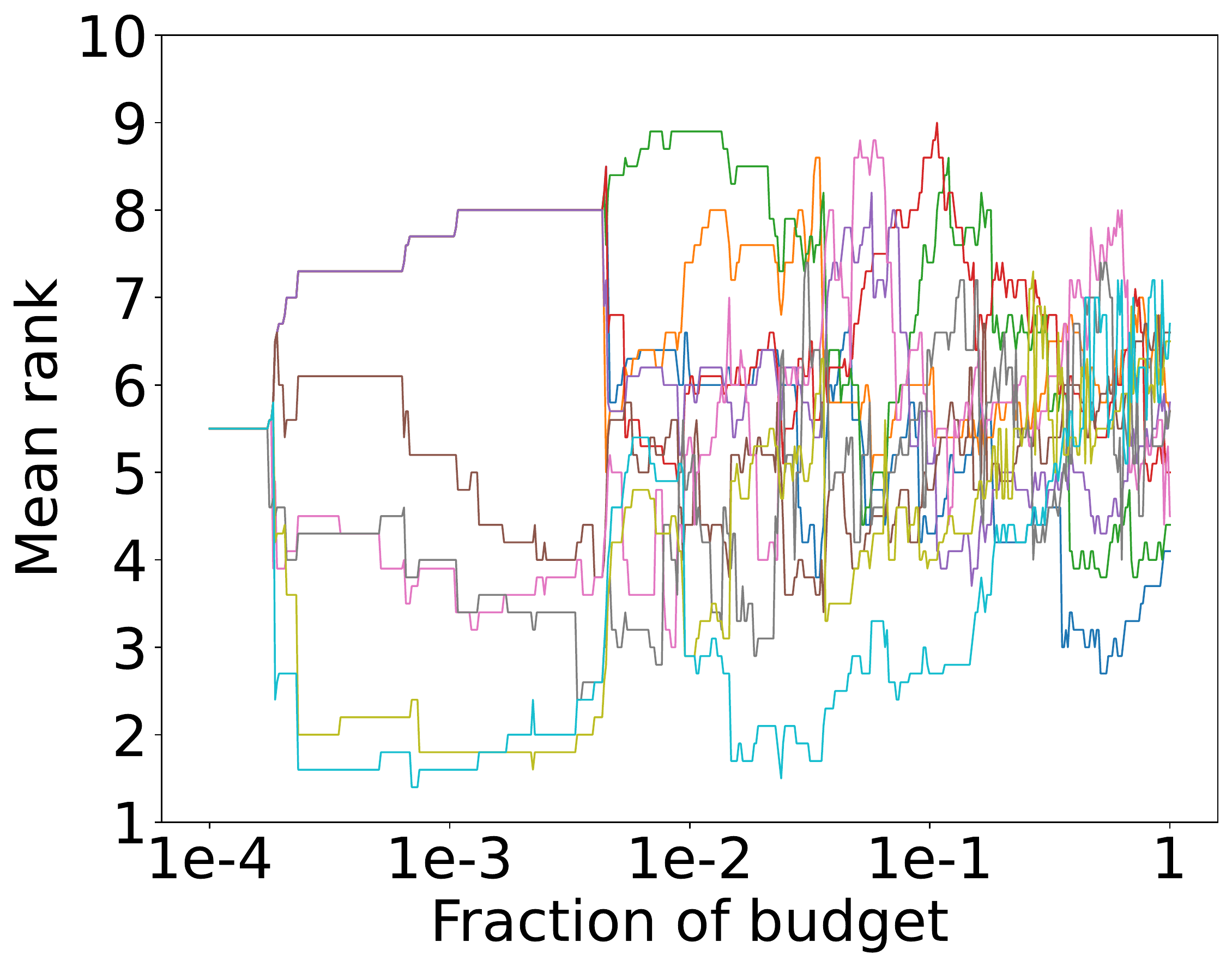}
		\caption{$\text{ALL}_{31\text{OpenML}}$}
		\label{fig:All_31_openml_opt_MLP}
	\end{subfigure}
	\begin{subfigure}{0.25\linewidth}
		\centering
		\includegraphics[width=0.9\linewidth]{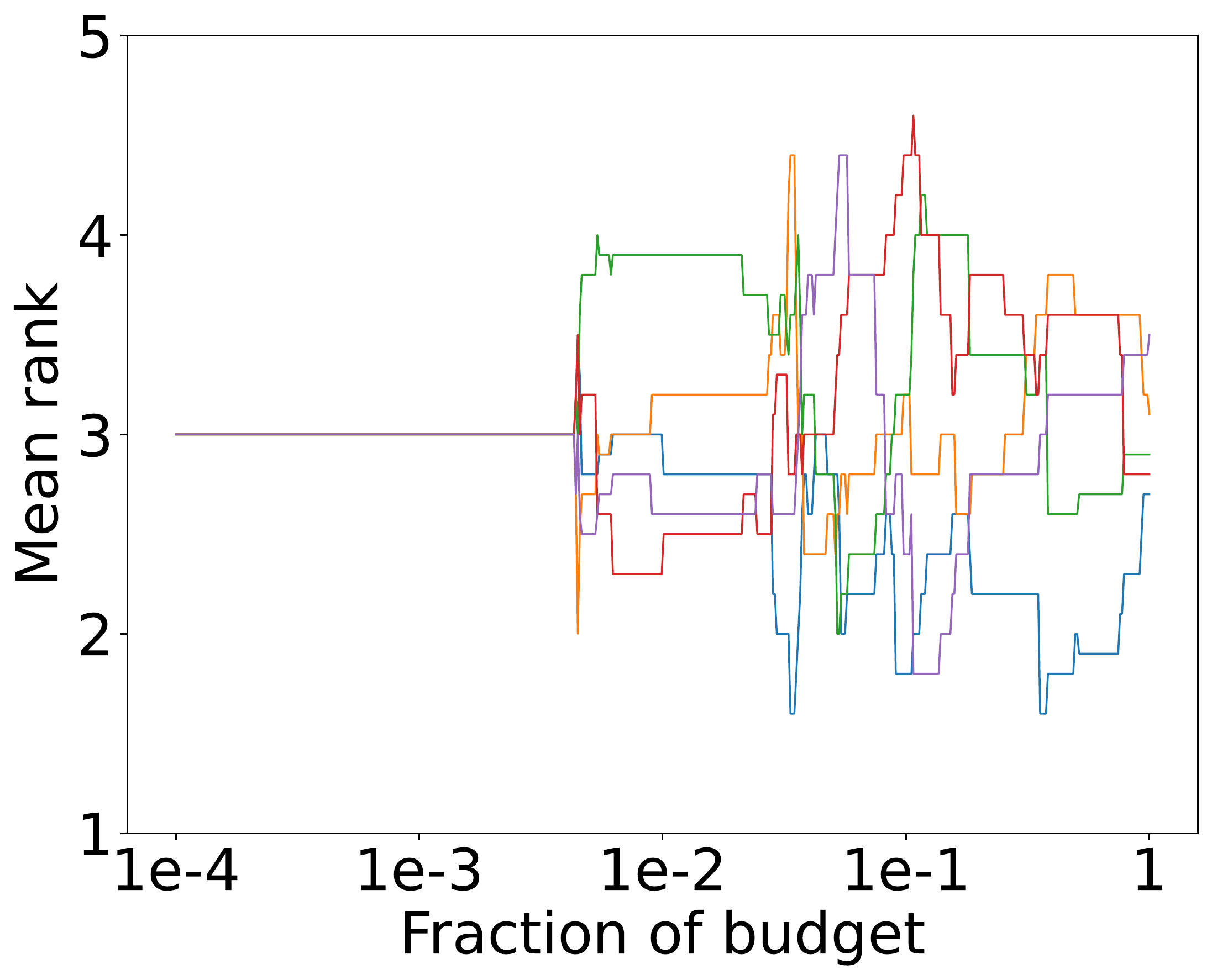}
		\caption{$\textit{BBO}_{31\text{OpenML}}$}
		\label{fig:BBO_31_openml_opt_MLP}
	\end{subfigure}
	\begin{subfigure}{0.25\linewidth}
		\centering
		\includegraphics[width=0.9\linewidth]{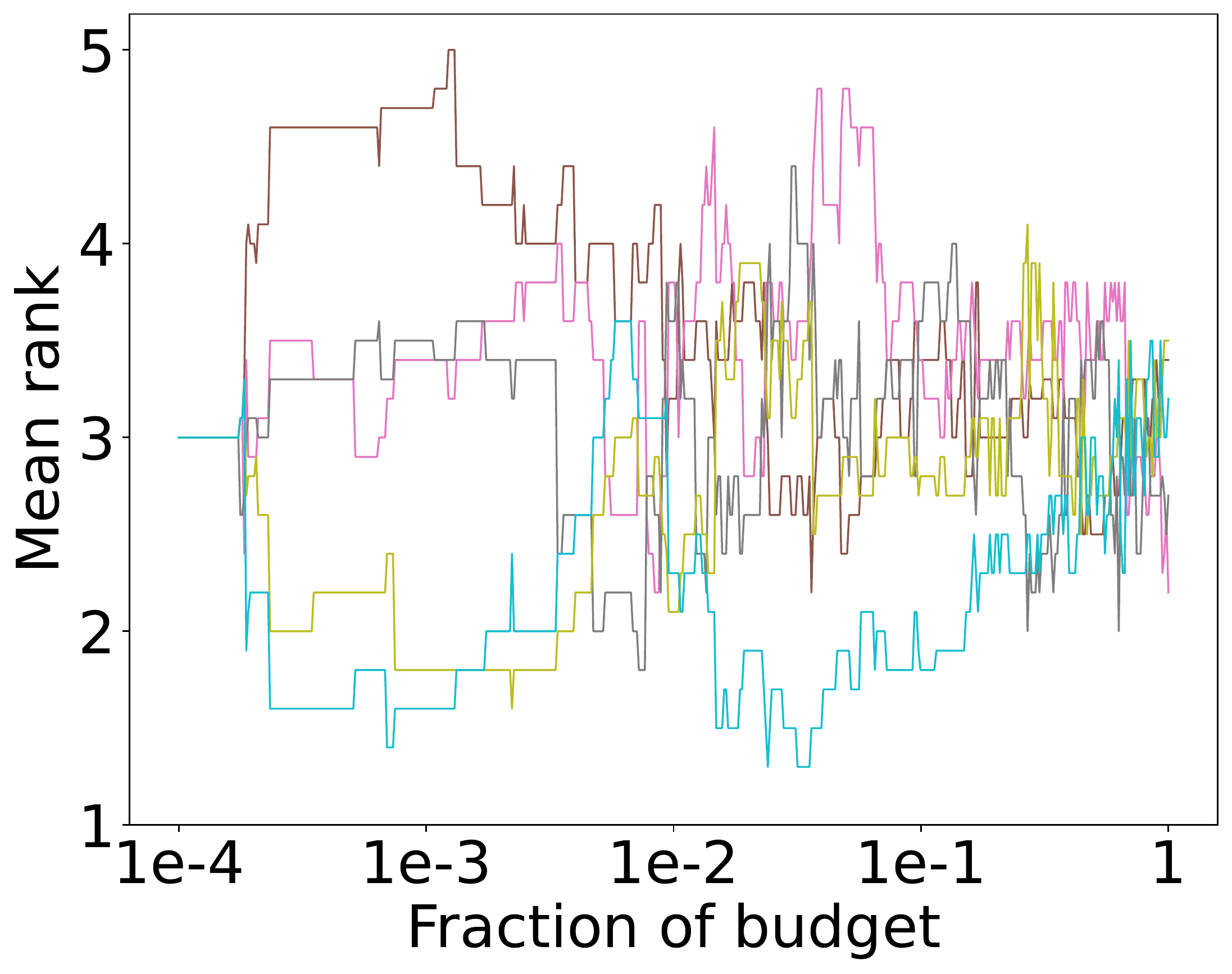}
		\caption{$\textit{MF}_{31\text{OpenML}}$}
		\label{fig:MF_31_openml_opt_MLP}
	\end{subfigure}
	
	\begin{subfigure}{0.25\linewidth}
		\centering
		\includegraphics[width=0.9\linewidth]{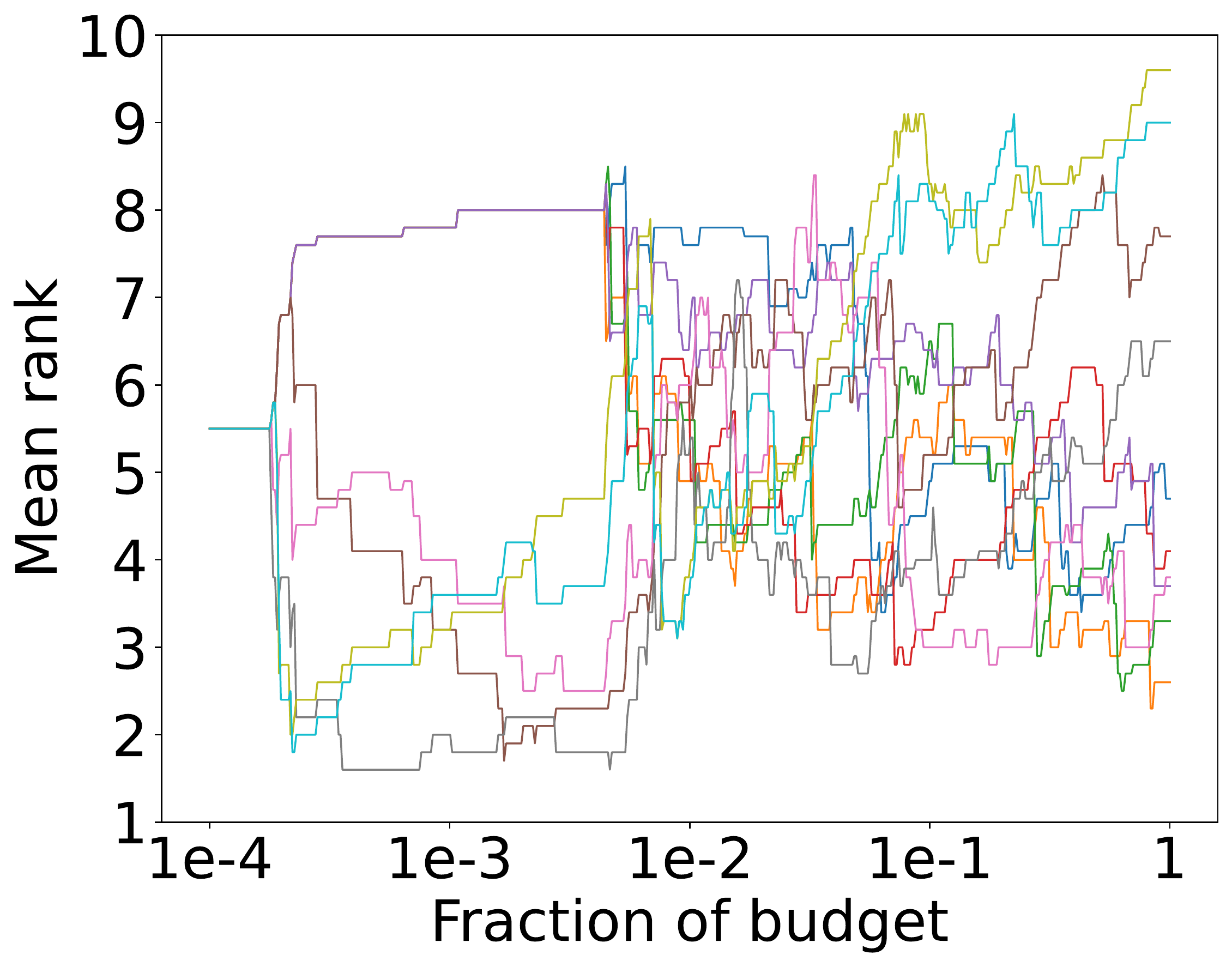}
		\caption{$\text{ALL}_{53\text{OpenML}}$}
		\label{fig:All_53_openml_opt_MLP}
	\end{subfigure}
	\begin{subfigure}{0.25\linewidth}
		\centering
		\includegraphics[width=0.9\linewidth]{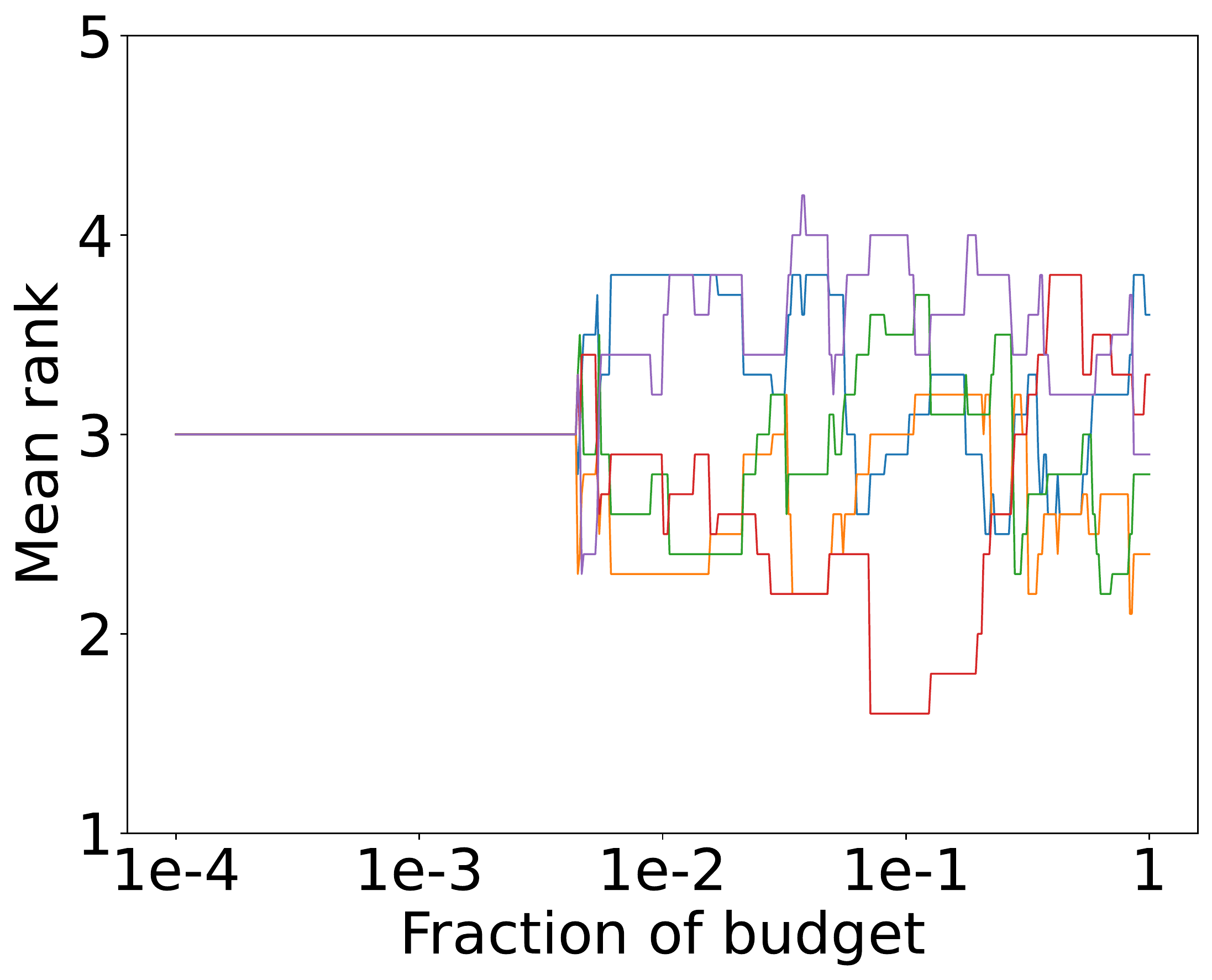}
		\caption{$\textit{BBO}_{53\text{OpenML}}$}
		\label{fig:BBO_53_openml_opt_MLP}
	\end{subfigure}
	\begin{subfigure}{0.25\linewidth}
		\centering
		\includegraphics[width=0.9\linewidth]{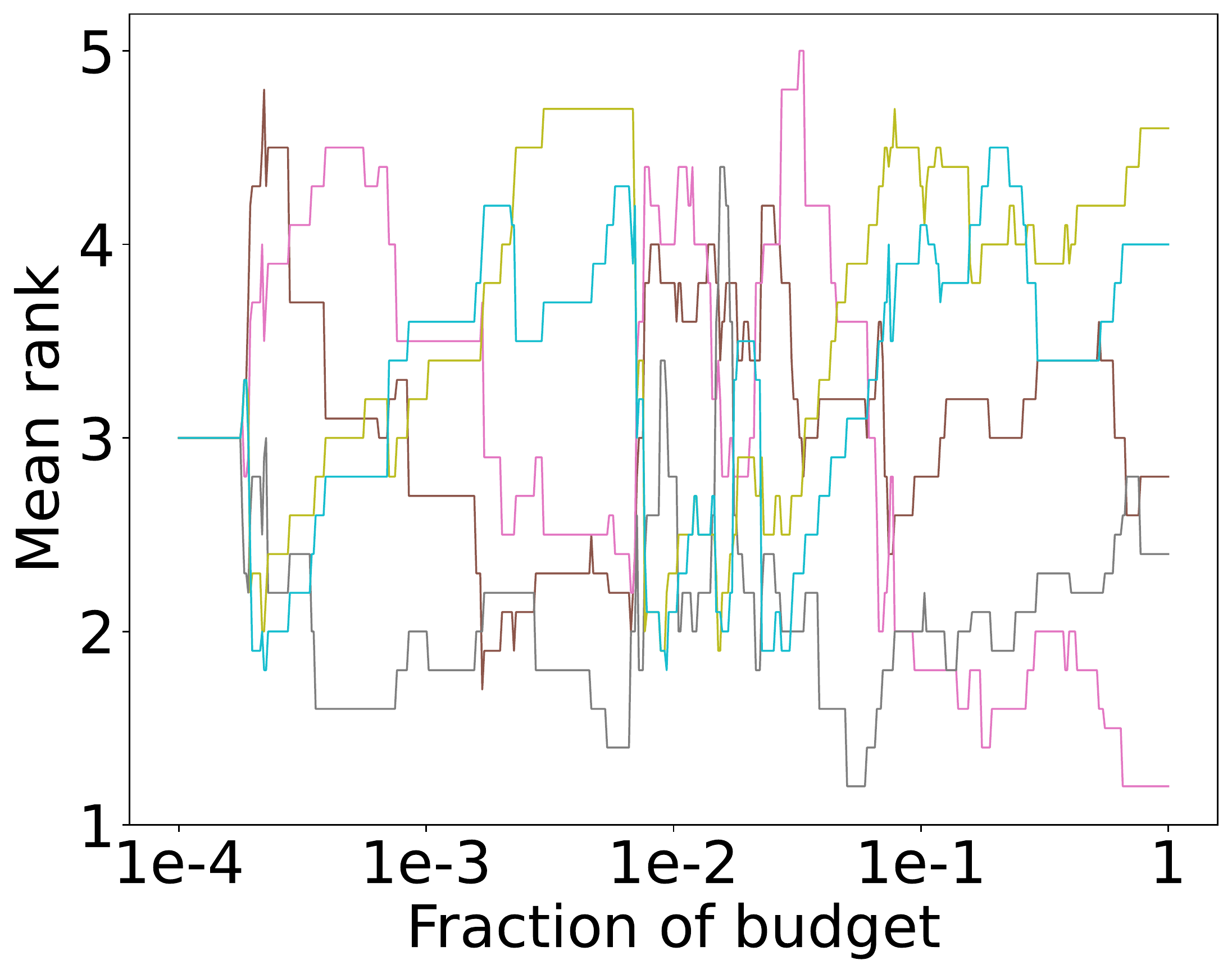}
		\caption{$\textit{MF}_{53\text{OpenML}}$}
		\label{fig:MF_53_openml_opt_MLP}
	\end{subfigure}
	
	\begin{subfigure}{0.25\linewidth}
		\centering
		\includegraphics[width=0.9\linewidth]{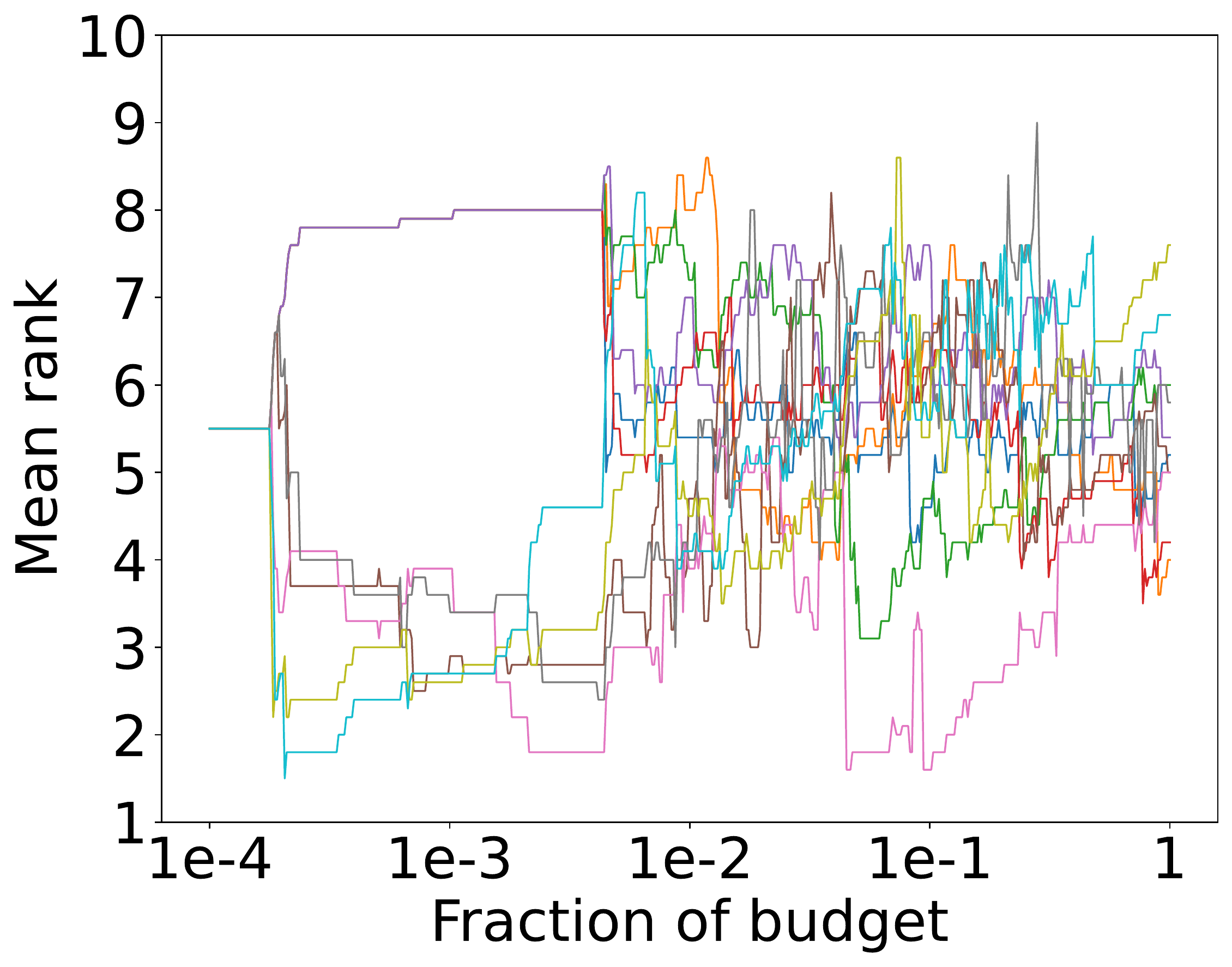}
		\caption{$\text{ALL}_{10101\text{OpenML}}$}
		\label{fig:All_10101_openml_opt_MLP}
	\end{subfigure}
	\begin{subfigure}{0.25\linewidth}
		\centering
		\includegraphics[width=0.9\linewidth]{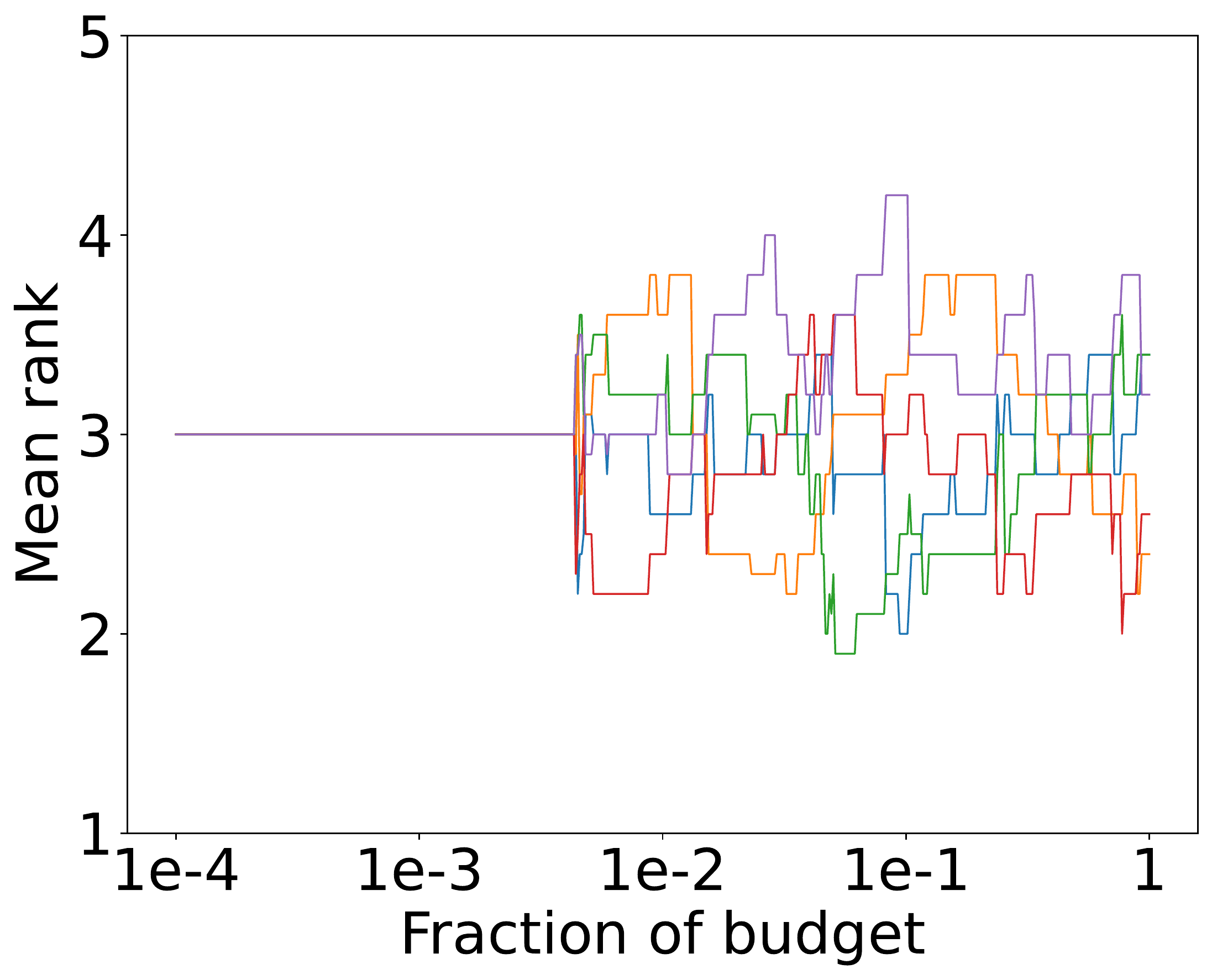}
		\caption{$\textit{BBO}_{10101\text{OpenML}}$}
		\label{fig:BBO_10101_openml_opt_MLP}
	\end{subfigure}
	\begin{subfigure}{0.25\linewidth}
		\centering
		\includegraphics[width=0.9\linewidth]{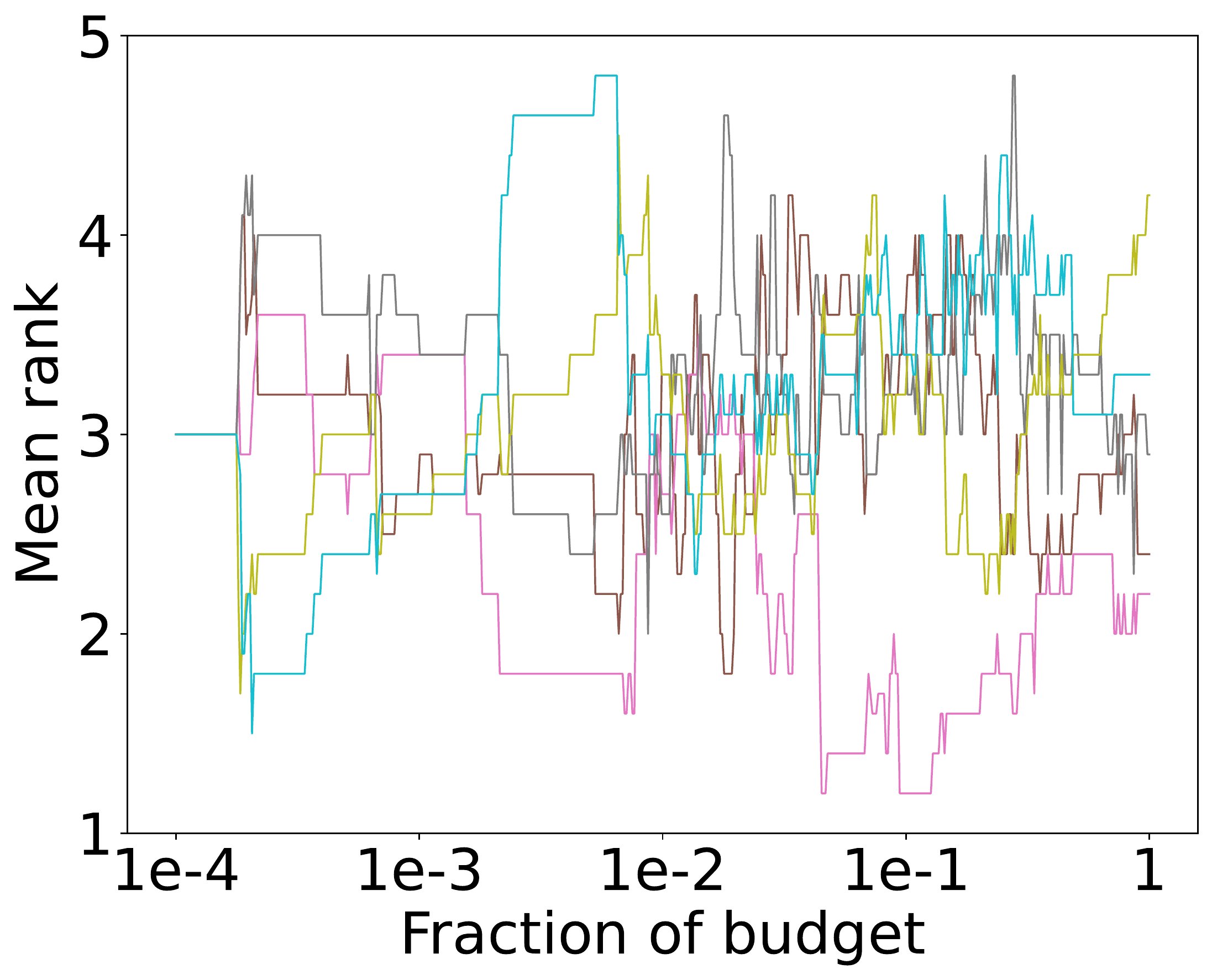}
		\caption{$\textit{MF}_{10101\text{OpenML}}$}
		\label{fig:MF_10101_openml_opt_MLP}
	\end{subfigure}
	
	\begin{subfigure}{0.25\linewidth}
		\centering
		\includegraphics[width=0.9\linewidth]{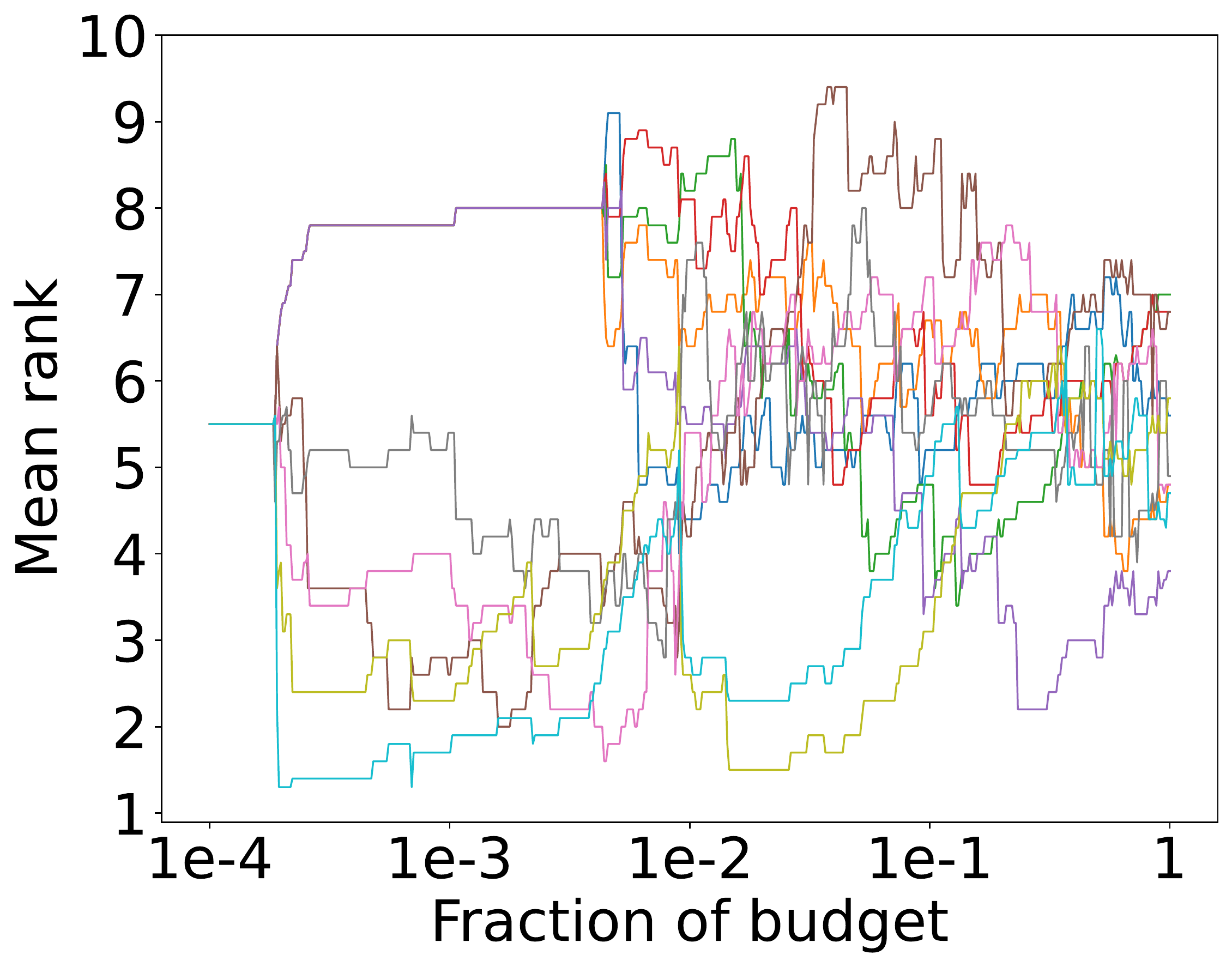}
		\caption{$\text{ALL}_{146818\text{OpenML}}$}
		\label{fig:All_146818_openml_opt_MLP}
	\end{subfigure}
	\begin{subfigure}{0.25\linewidth}
		\centering
		\includegraphics[width=0.9\linewidth]{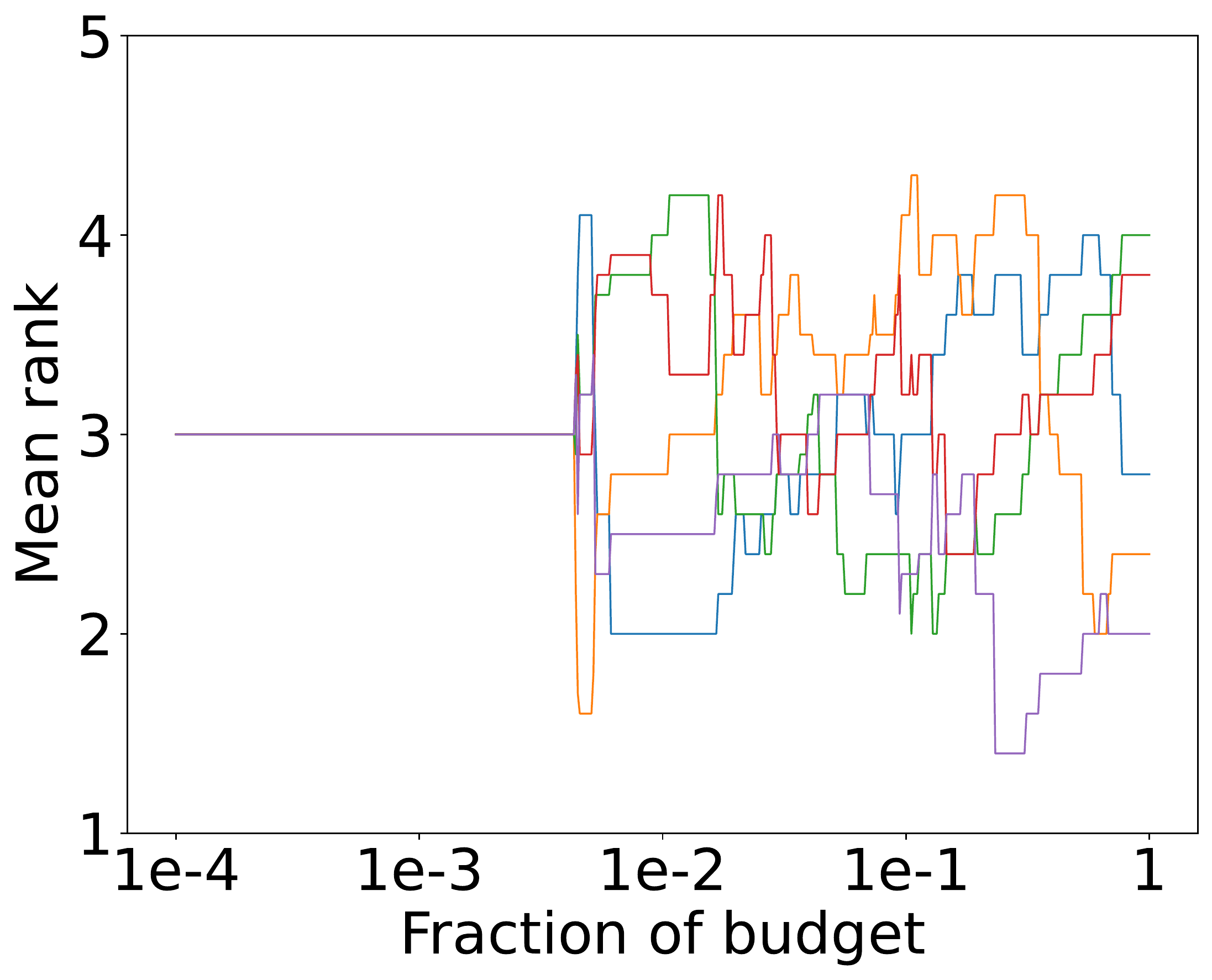}
		\caption{$\textit{BBO}_{146818\text{OpenML}}$}
		\label{fig:BBO_146818_openml_opt_MLP}
	\end{subfigure}
	\begin{subfigure}{0.25\linewidth}
		\centering
		\includegraphics[width=0.9\linewidth]{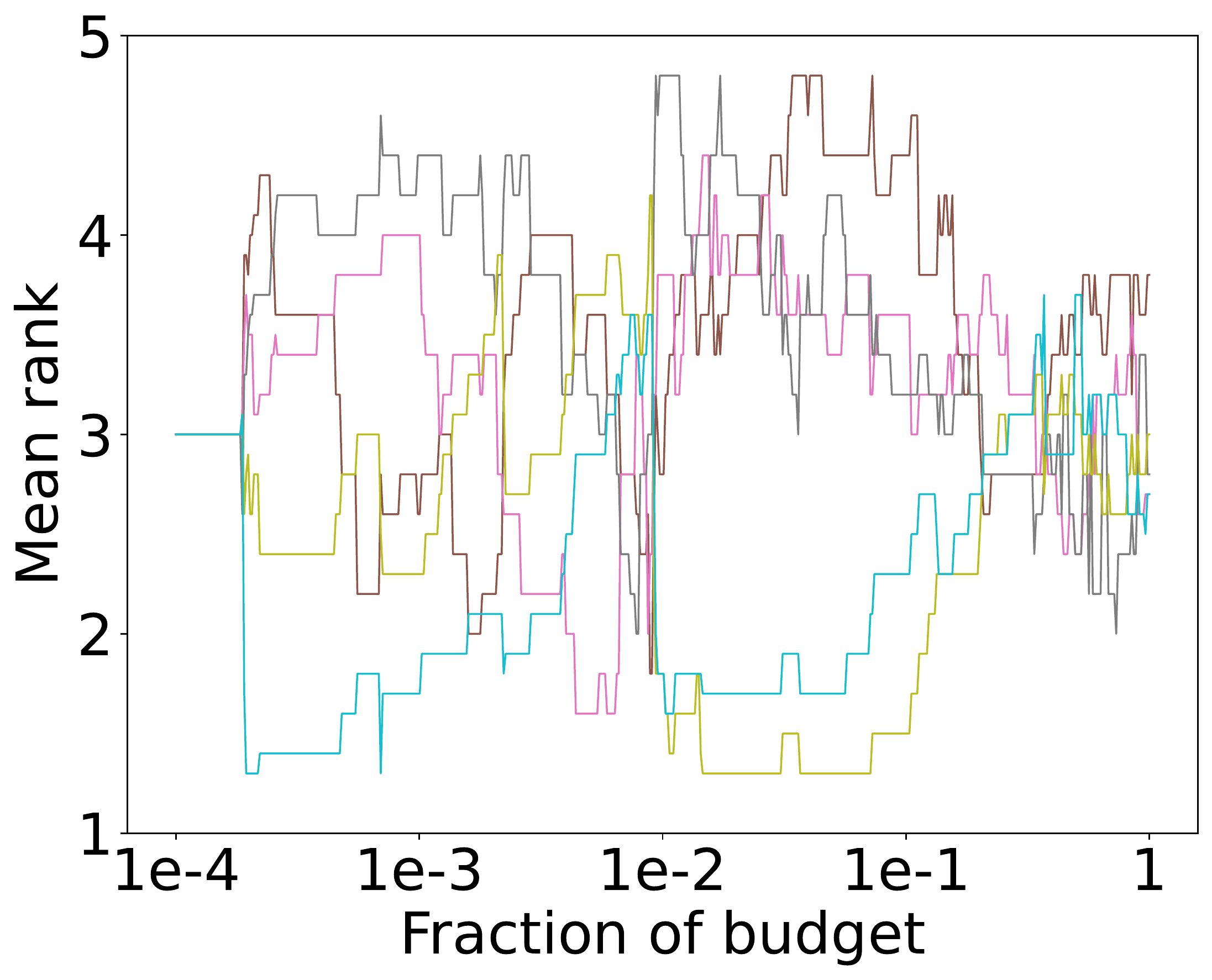}
		\caption{$\textit{MF}_{146818\text{OpenML}}$}
		\label{fig:MF_146818_openml_opt_MLP}
	\end{subfigure}
	
	\begin{subfigure}{0.25\linewidth}
		\centering
		\includegraphics[width=0.9\linewidth]{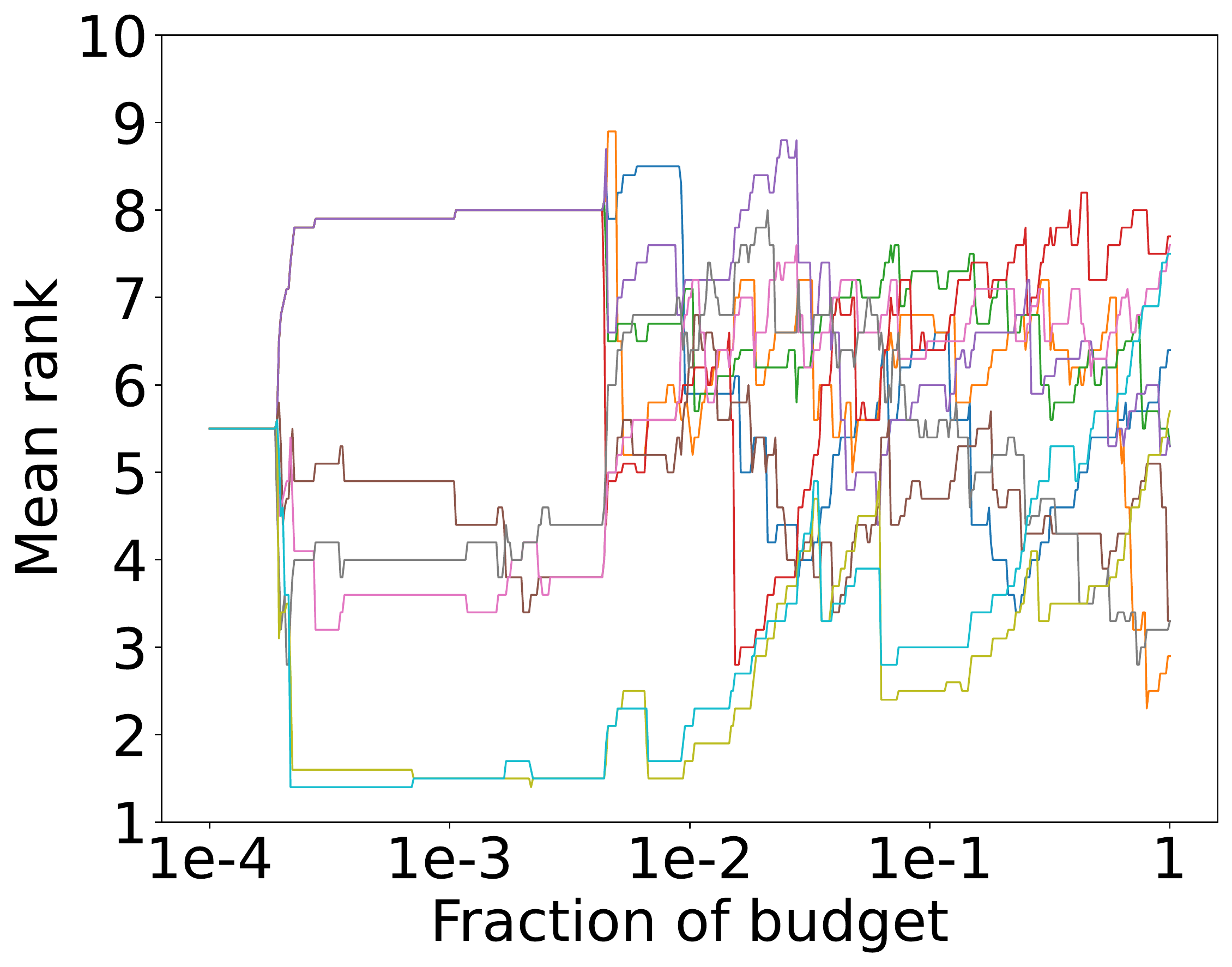}
		\caption{$\text{ALL}_{146821\text{OpenML}}$}
		\label{fig:All_146821_openml_opt_MLP}
	\end{subfigure}
	\begin{subfigure}{0.25\linewidth}
		\centering
		\includegraphics[width=0.9\linewidth]{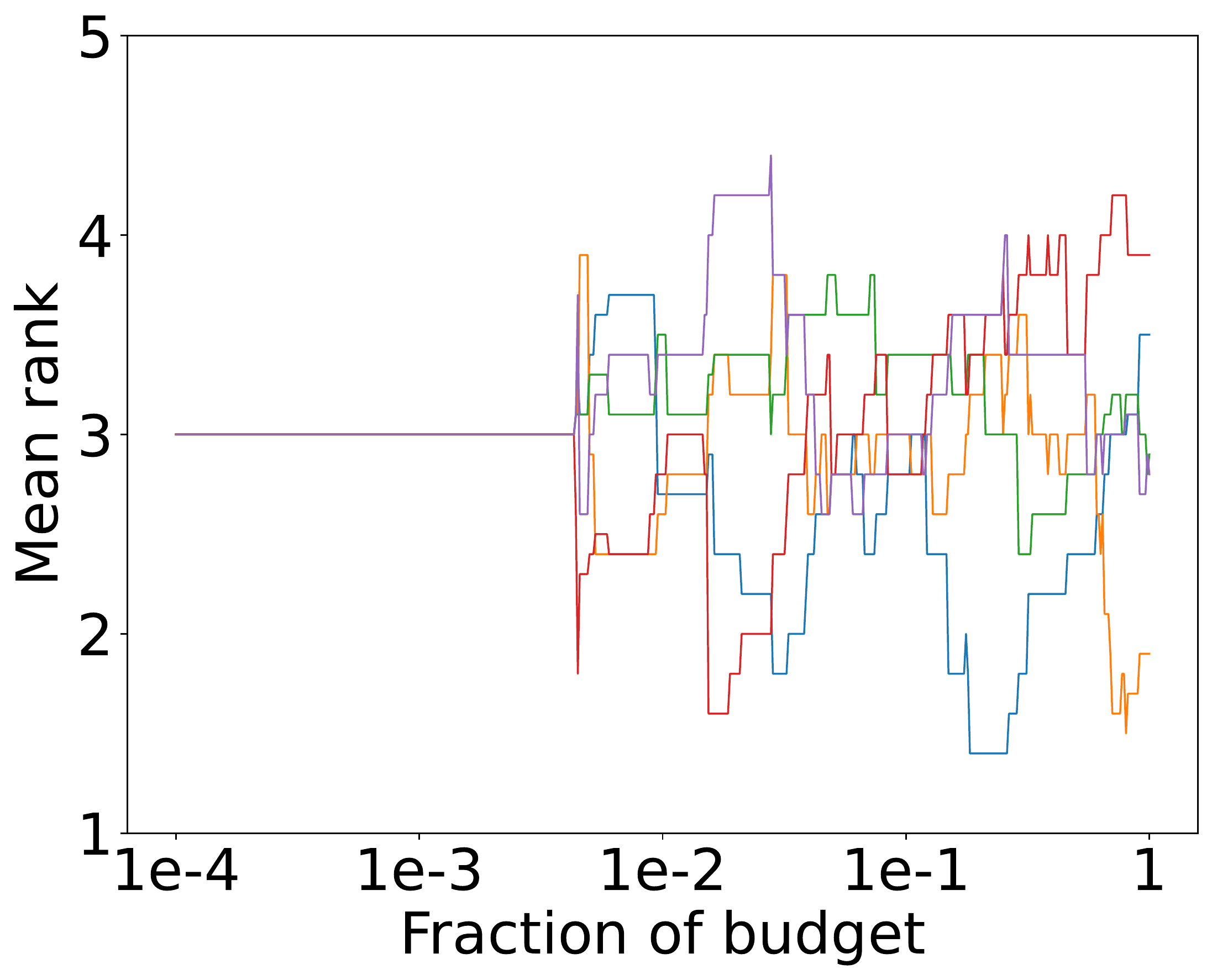}
		\caption{$\textit{BBO}_{146821\text{OpenML}}$}
		\label{fig:BBO_146821_openml_opt_MLP}
	\end{subfigure}
	\begin{subfigure}{0.25\linewidth}
		\centering
		\includegraphics[width=0.9\linewidth]{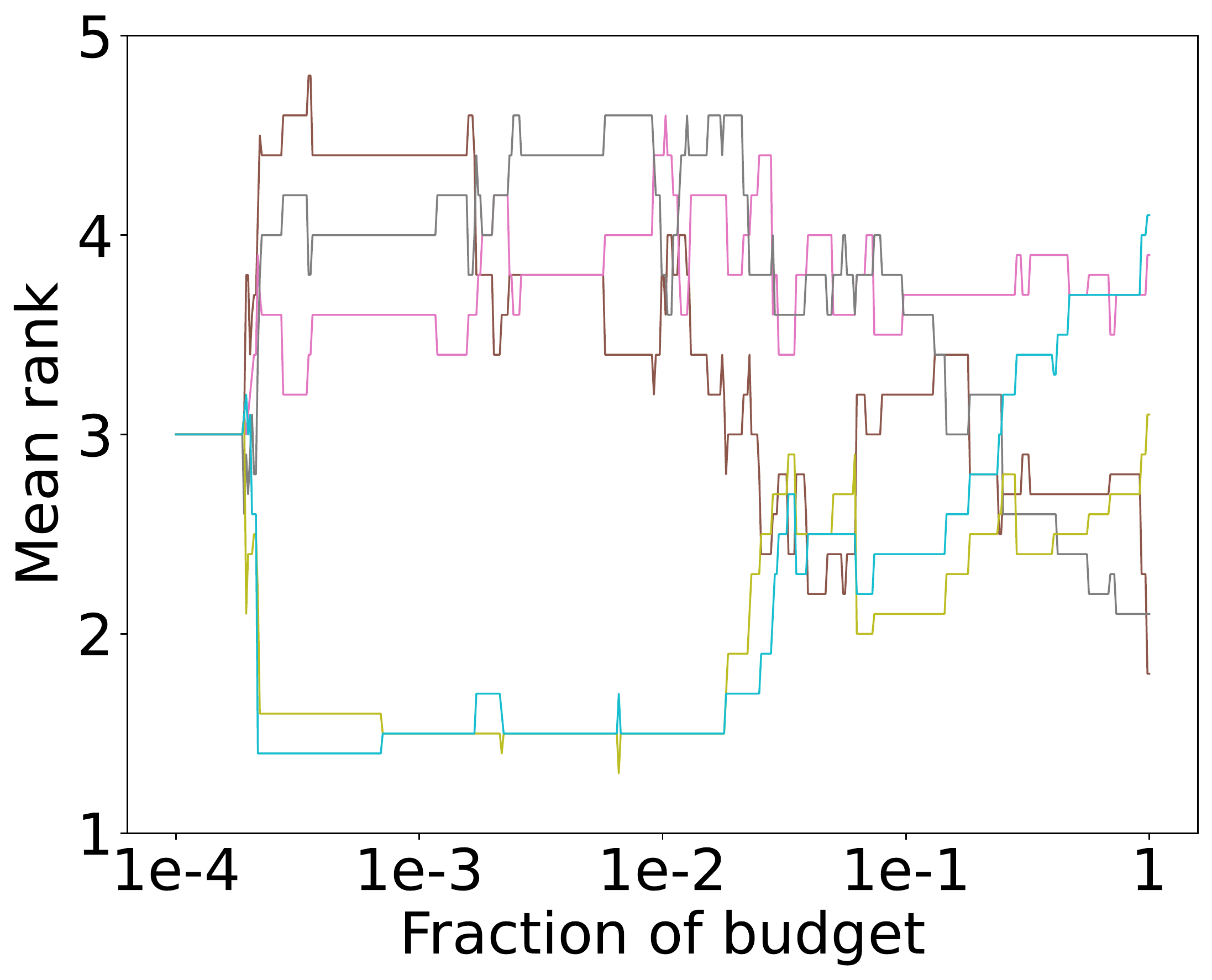}
		\caption{$\textit{MF}_{146821\text{OpenML}}$}
		\label{fig:MF_146821_openml_opt_MLP}
	\end{subfigure}
	
	\begin{subfigure}{0.25\linewidth}
		\centering
		\includegraphics[width=0.9\linewidth]{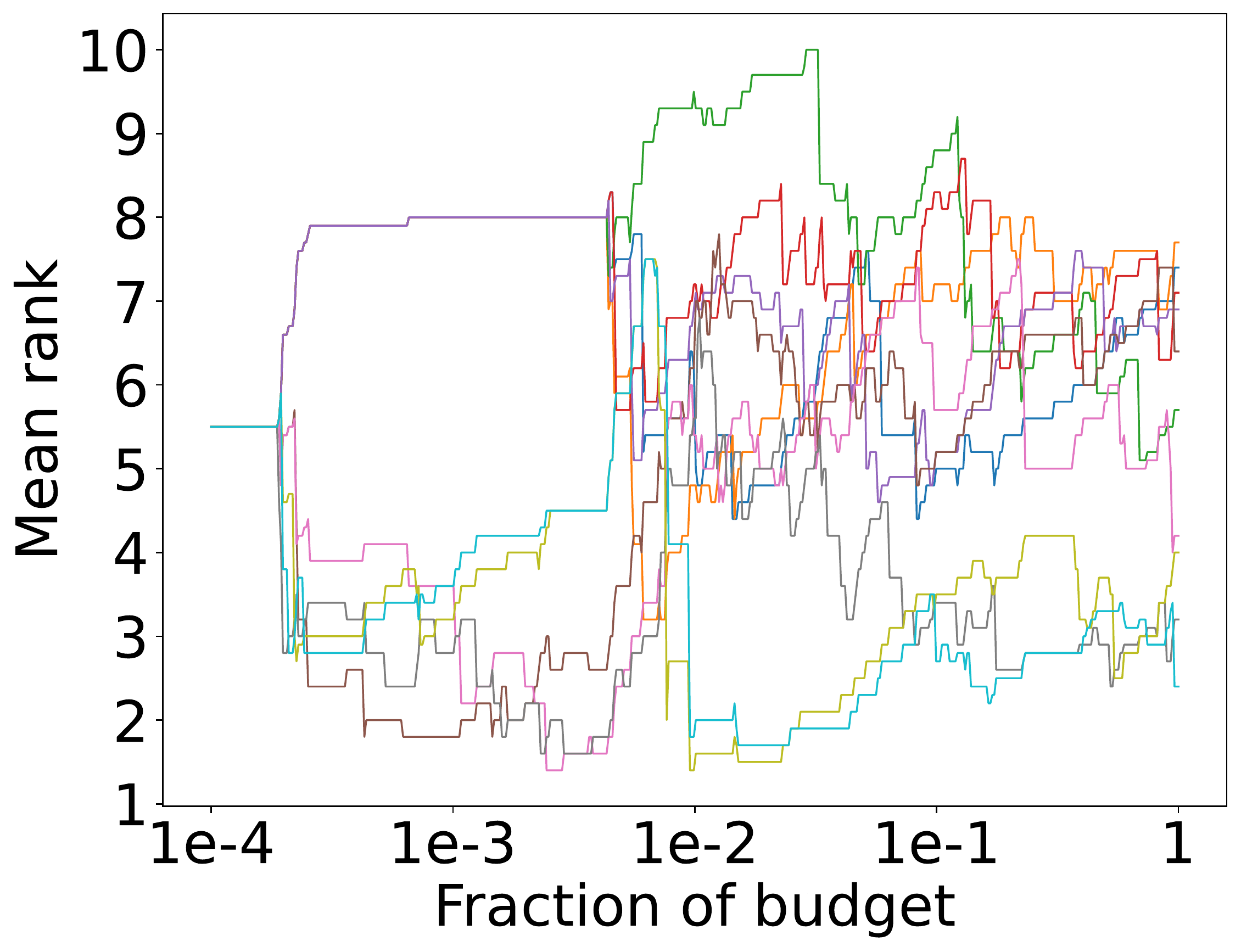}
		\caption{$\text{ALL}_{146822\text{OpenML}}$}
		\label{fig:All_146822_openml_opt_MLP}
	\end{subfigure}
	\begin{subfigure}{0.25\linewidth}
		\centering
		\includegraphics[width=0.9\linewidth]{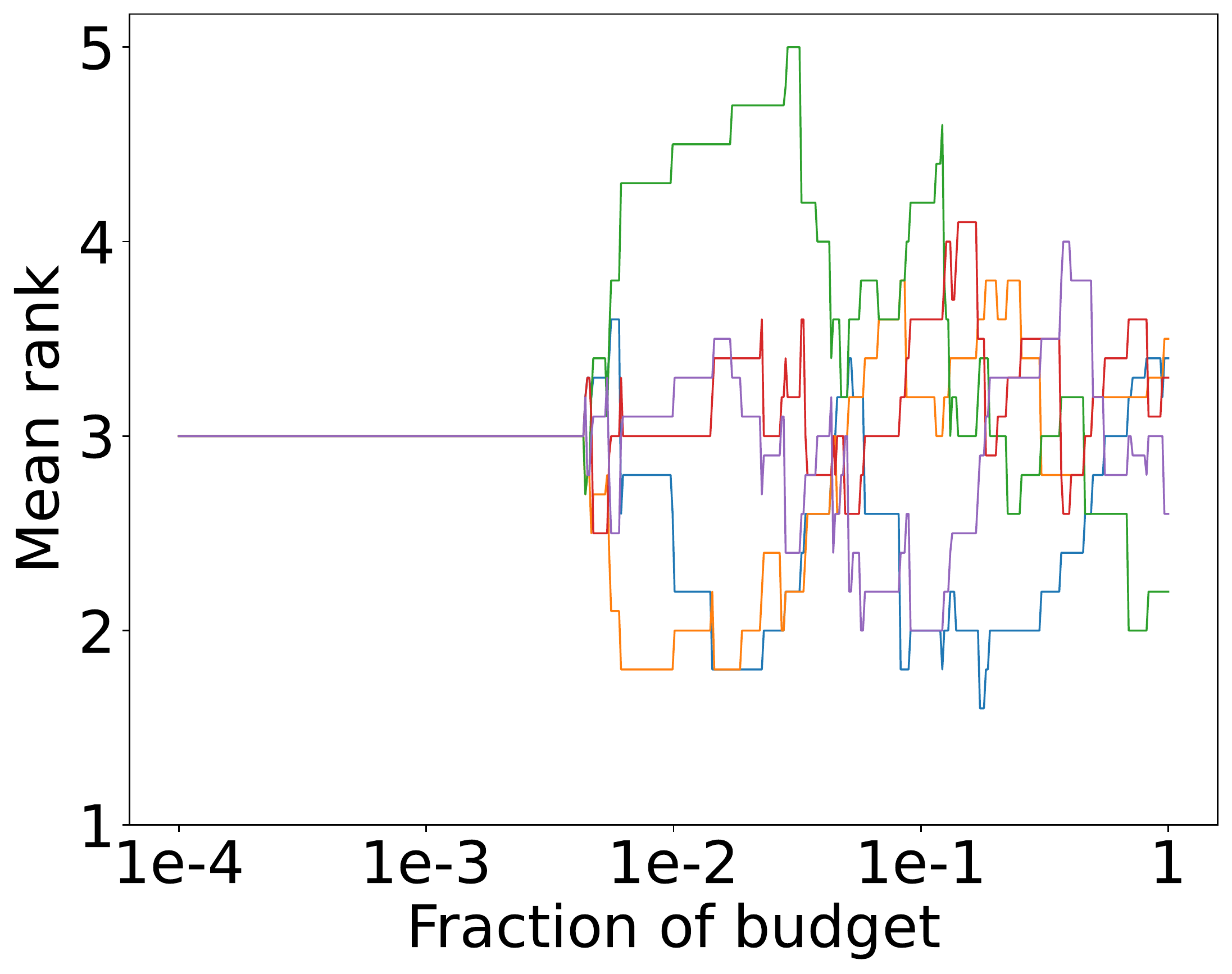}
		\caption{$\textit{BBO}_{146822\text{OpenML}}$}
		\label{fig:BBO_146822_openml_opt_MLP}
	\end{subfigure}
	\begin{subfigure}{0.25\linewidth}
		\centering
		\includegraphics[width=0.9\linewidth]{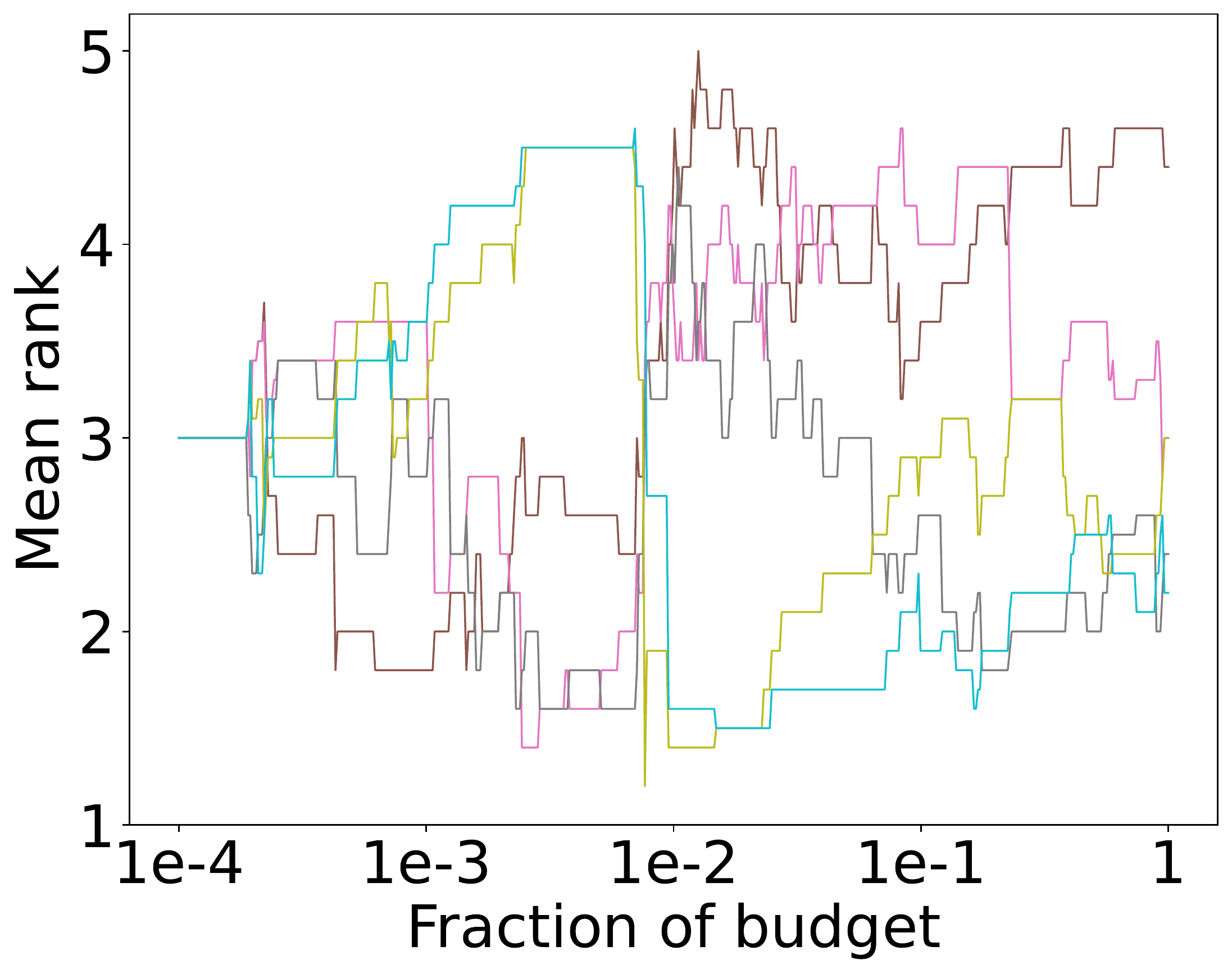}
		\caption{$\textit{MF}_{146822\text{OpenML}}$}
		\label{fig:MF_146822_openml_opt_MLP}
	\end{subfigure}
	
	\centering
	\hspace*{1.2cm}\begin{subfigure}{1.0\linewidth}
		\centering
		\includegraphics[width=0.95\linewidth]{materials/legend_rank_new.pdf}
	\end{subfigure}
	\vspace{-0.1in}
	
	\caption{Mean rank over time on MLP benchmark (FedOPT).}
	\label{fig:entire_MLP_tabular_opt_rank}
\end{figure}

Following Section~\ref{subsec:exp1}, we show the overall mean rank overtime on all \subj problems with FedOPT, whose pattern is similar to that of FedAvg in Figure~\ref{fig:entire_all_tabular_avg_rank}.
Then, we report the final results with FedAvg and FedOPT in Table~\ref{tab:final_results_tab_avg} and ~\ref{tab:final_results_tab_opt}, respectively.
Finally, we report the mean rank over time in Figure~\ref{fig:entire_cnn_tabular_avg_rank}-\ref{fig:entire_MLP_tabular_opt_rank}.
Due to time and computing resource constraints, the results on CNN benchmark are incomplete (lacking that with FedOPT), which we will supplement as soon as possible.

\begin{figure}[htbp]
	\centering
	\begin{subfigure}{0.32\linewidth}
		\centering
		\includegraphics[width=0.95\linewidth]{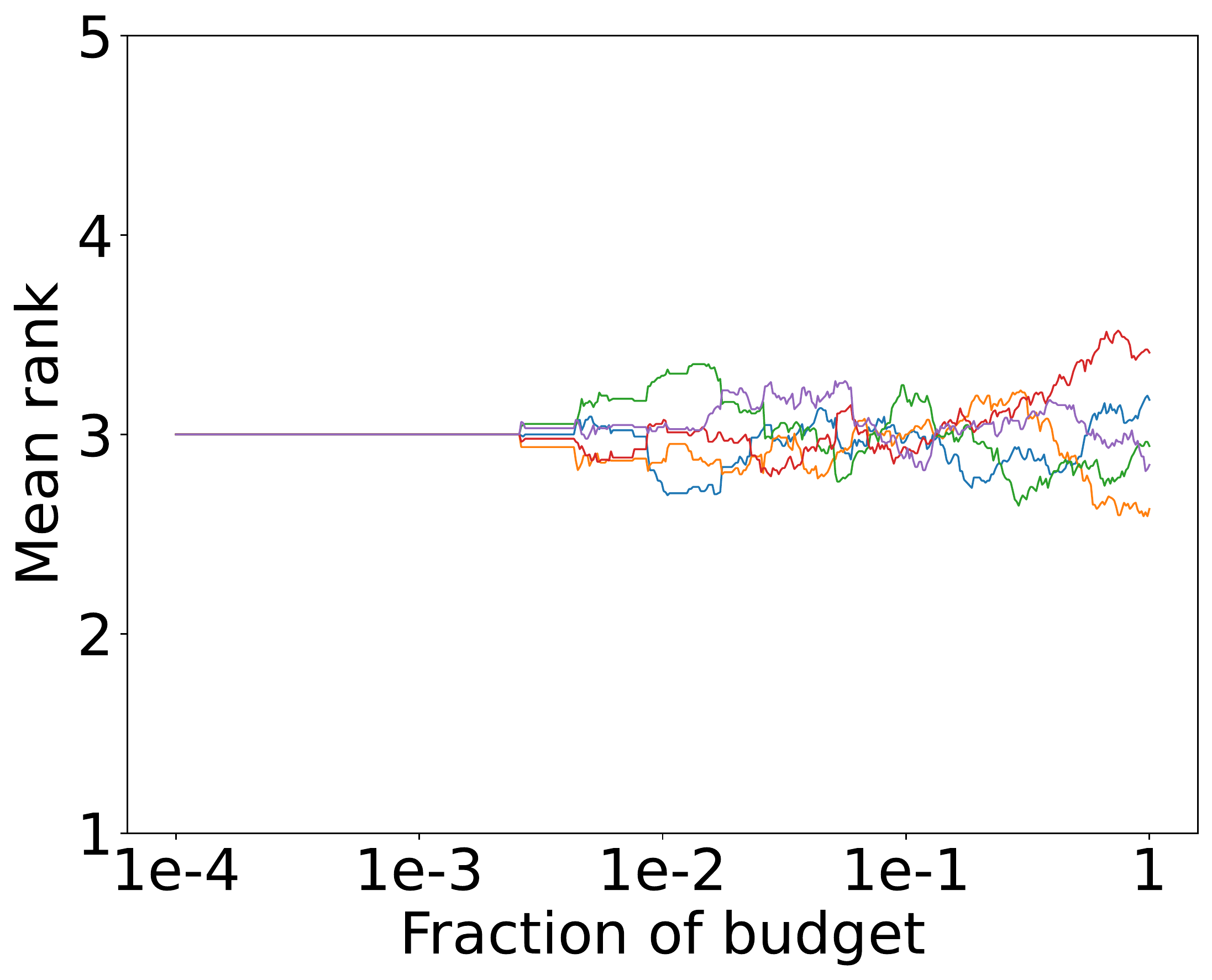}
		\caption{\textit{BBO}}
		\label{fig:BBO_All_opt}
	\end{subfigure}
	\begin{subfigure}{0.32\linewidth}
		\centering
		\includegraphics[width=0.95\linewidth]{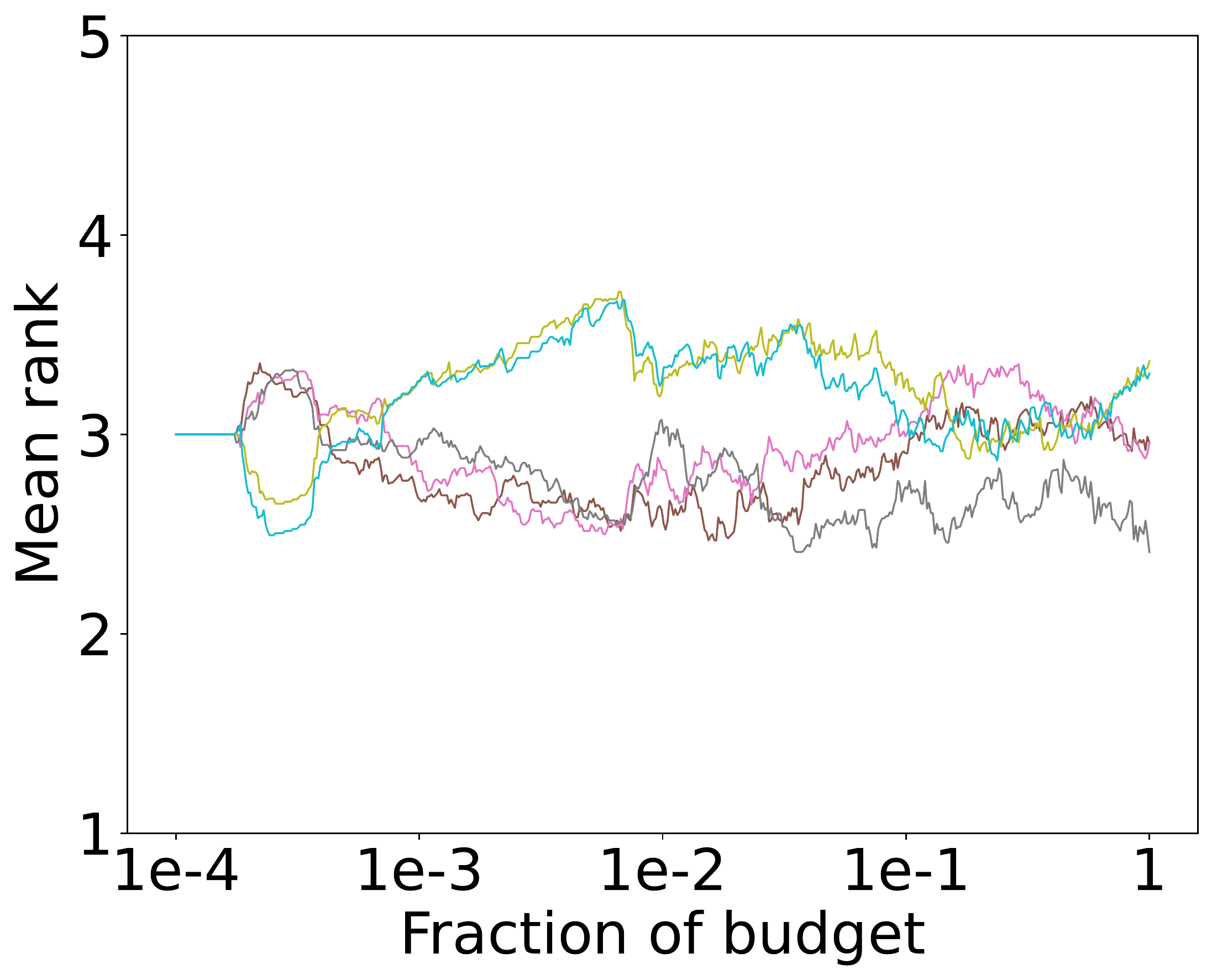}
		\caption{\textit{MF}}
		\label{fig:MF_All_opt}
	\end{subfigure}
	\begin{subfigure}{0.32\linewidth}
		\centering
		\includegraphics[width=0.95\linewidth]{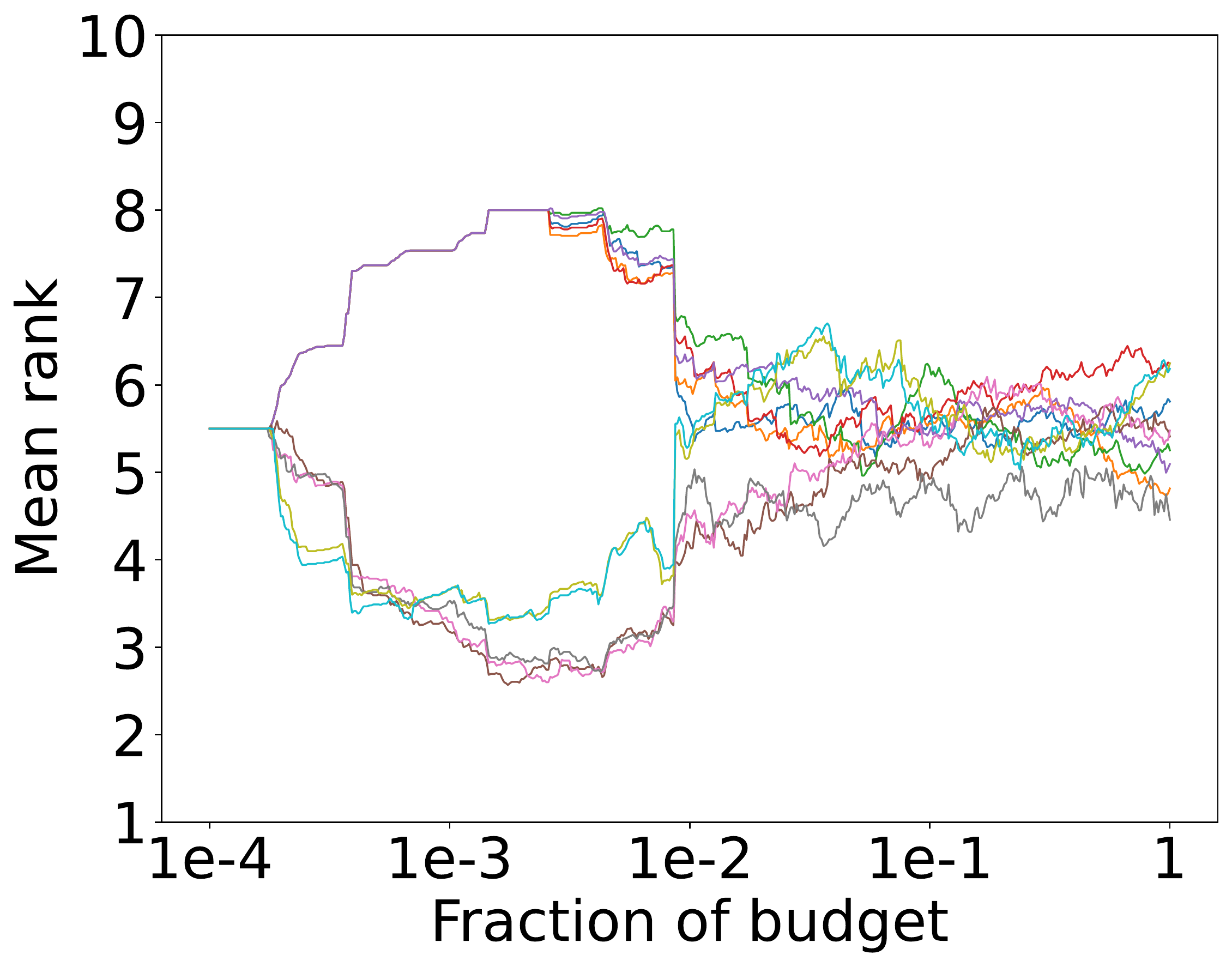}
		\caption{All}
		\label{fig:All_All_opt}
	\end{subfigure}
	\centering
	\hspace*{1.2cm}\begin{subfigure}{1.0\linewidth}
		\centering
		\includegraphics[width=0.95\linewidth]{materials/legend_rank_new.pdf}
	\end{subfigure}
	\vspace{-0.1in}
	\caption{Mean rank over time on all \subj problems (with FedOPT).}
	\label{fig:entire_all_tabular_opt_rank}
	\vspace{-0.2in}
\end{figure}

\begin{table}[htbp]
\centering
\caption{Final results of the optimizers on tabular mode with FedAvg (lower is better).}
\label{tab:final_results_tab_avg}
\resizebox{\textwidth}{!}{%
\begin{tabular}{lllllllllll}
\toprule
benchmark & \textit{RS}  & $\textit{BO}_{\textit{GP}}$ & $\textit{BO}_{\textit{RF}}$ & $\textit{BO}_{\textit{KDE}}$ & \textit{DE} & \textit{HB} & \textit{BOHB} & \textit{DEHB} & $\textit{TPE}_{\textit{MD}}$ & $\textit{TPE}_{\textit{HB}}$ \\ \midrule
$\text{CNN}_{\text{FEMNIST}}$      & 0.4969 & 0.4879 & 0.4885 & 0.5004 & 0.4928 & 0.4926 & 0.4945 & 0.498 & 0.5163 & 0.5148  \\ \hline
$\text{BERT}_{\text{SST-2}}$ &  0.435 & 0.4276 & 0.4294 & 0.4334 & 0.437 & 0.4311 & 0.4504 & 0.4319 & 0.4341 & 0.4251  \\
$\text{BERT}_{\text{CoLA}}$ &  0.6151 & 0.6148 & 0.6141 & 0.6133 & 0.6143 & 0.6143 & 0.6168 & 0.6178 & 0.6158 & 0.6146  \\ \hline
$\text{GNN}_{\text{Cora}}$         &  0.3265 & 0.3258 & 0.326 & 0.3347 & 0.3267 & 0.3324 & 0.3288 & 0.3225 & 0.3241 & 0.3249  \\
$\text{GNN}_{\text{CiteSeer}}$     &  0.6469 & 0.6442 & 0.6499 & 0.6442 & 0.6453 & 0.6387 & 0.6425 & 0.6452 & 0.6324 & 0.6371  \\
$\text{GNN}_{\text{PubMed}}$       &  0.5262 & 0.5146 & 0.5169 & 0.5311 & 0.5001 & 0.5006 & 0.5194 & 0.4934 & 0.506 & 0.5044  \\ \hline
$\text{LR}_{31}$              &  0.6821 & 0.6308 & 0.6382 & 0.6385 & 0.667 & 0.6492 & 0.6461 & 0.6145 & 0.7228 & 0.758  \\
$\text{LR}_{53}$              &  1.6297 & 1.7288 & 1.6116 & 1.7142 & 1.6062 & 1.5765 & 1.5634 & 1.4755 & 1.5506 & 1.5506  \\
$\text{LR}_{3917}$              &  1.8892 & 1.7561 & 1.7186 & 2.4271 & 1.7519 & 3.948 & 1.6384 & 3.1183 & 2.1344 & 2.6576  \\
$\text{LR}_{10101}$              &  0.548 & 0.5483 & 0.5482 & 0.5487 & 0.5481 & 0.5504 & 0.5505 & 0.5516 & 0.5483 & 0.5487  \\
$\text{LR}_{146818}$              &  0.5294 & 0.5291 & 0.5295 & 0.5289 & 0.5291 & 0.5292 & 0.529 & 0.5293 & 0.5328 & 0.5387  \\
$\text{LR}_{146821}$              &  0.4733 & 0.464 & 0.4722 & 0.4843 & 0.4971 & 0.4678 & 0.4747 & 0.4707 & 0.4792 & 0.4688  \\
$\text{LR}_{146822}$              &  0.4581 & 0.4481 & 0.4505 & 0.4731 & 0.4587 & 0.4478 & 0.4446 & 0.4304 & 0.4376 & 0.4419  \\ \hline
$\text{MLP}_{31}$             &  0.5899 & 0.5891 & 0.5808 & 0.5904 & 0.5925 & 0.5921 & 0.5929 & 0.593 & 0.593 & 0.593  \\
$\text{MLP}_{53}$             &  0.7795 & 0.7373 & 0.7849 & 0.8215 & 0.8068 & 0.769 & 0.7577 & 0.8173 & 0.9491 & 1.0567  \\
$\text{MLP}_{3917}$             &  0.3863 & 0.3937 & 0.3858 & 0.3958 & 0.383 & 0.3895 & 0.3911 & 0.4084 & 0.3979 & 0.3988  \\
$\text{MLP}_{10101}$             &  0.4054 & 0.4217 & 0.4361 & 0.4162 & 0.418 & 0.4137 & 0.4152 & 0.4102 & 0.4522 & 0.4352  \\
$\text{MLP}_{146818}$             &  0.5089 & 0.4997 & 0.5125 & 0.5112 & 0.5138 & 0.5009 & 0.5199 & 0.5039 & 0.5392 & 0.54  \\
$\text{MLP}_{146821}$             &  0.184 & 0.1251 & 0.155 & 0.1769 & 0.1851 & 0.1561 & 0.1683 & 0.1572 & 0.1654 & 0.1761  \\
$\text{MLP}_{146822}$             &  0.2839 & 0.2892 & 0.317 & 0.3586 & 0.2928 & 0.2927 & 0.2823 & 0.2549 & 0.2745 & 0.2755  \\
\bottomrule
\end{tabular}%
}
\end{table}

\begin{table}[htbp]
\centering
\caption{Final results of the optimizers on tabular mode with FedOPT (lower is better).}
\label{tab:final_results_tab_opt}
\resizebox{\textwidth}{!}{%
\begin{tabular}{lllllllllll}
\toprule
benchmark & \textit{RS}  & $\textit{BO}_{\textit{GP}}$ & $\textit{BO}_{\textit{RF}}$ & $\textit{BO}_{\textit{KDE}}$ & \textit{DE} & \textit{HB} & \textit{BOHB} & \textit{DEHB} & $\textit{TPE}_{\textit{MD}}$ & $\textit{TPE}_{\textit{HB}}$ \\ \midrule
$\text{BERT}_{\text{SST-2}}$ & 0.441 & 0.4325 & 0.4301 & 0.4463 & 0.4351 & 0.4403 & 0.4295 & 0.4285 & 0.4293 & 0.4332   \\
$\text{BERT}_{\text{CoLA}}$ & 0.616 & 0.616 & 0.6141 & 0.6137 & 0.6159 & 0.6154 & 0.6157 & 0.6176 & 0.6172 & 0.6168    \\ \hline
$\text{GNN}_{\text{Cora}}$         & 0.3264 & 0.3235 & 0.3268 & 0.3322 & 0.3256 & 0.3245 & 0.3347 & 0.3254 & 0.3405 & 0.3361    \\
$\text{GNN}_{\text{CiteSeer}}$     & 0.6483 & 0.6517 & 0.6497 & 0.6535 & 0.6458 & 0.6442 & 0.6543 & 0.6463 & 0.6488 & 0.6495    \\
$\text{GNN}_{\text{PubMed}}$       & 0.4777 & 0.4426 & 0.4718 & 0.4943 & 0.4318 & 0.4559 & 0.4699 & 0.4318 & 0.4368 & 0.4402   \\ \hline
$\text{LR}_{31}$              & 0.7358 & 0.6831 & 0.6849 & 0.8152 & 0.7085 & 0.6772 & 0.6877 & 0.6385 & 0.8652 & 0.7044   \\
$\text{LR}_{53}$              & 1.7838 & 1.5609 & 1.5241 & 1.5116 & 1.6208 & 1.6045 & 1.7236 & 1.3488 & 1.6654 & 1.7978    \\
$\text{LR}_{3917}$              & 2.254 & 2.0316 & 2.3952 & 1.9788 & 2.6261 & 2.3472 & 2.5452 & 2.3144 & 3.2131 & 2.0291   \\
$\text{LR}_{10101}$              & 0.5533 & 0.55 & 0.5505 & 0.5509 & 0.549 & 0.5504 & 0.5476 & 0.5522 & 0.5612 & 0.8567    \\
$\text{LR}_{146818}$              & 0.511 & 0.506 & 0.5034 & 0.5133 & 0.5007 & 0.5032 & 0.5086 & 0.4974 & 0.4983 & 0.5104    \\
$\text{LR}_{146821}$              & 0.4017 & 0.3599 & 0.4121 & 0.4134 & 0.4079 & 0.395 & 0.398 & 0.3902 & 0.4447 & 0.4625    \\
$\text{LR}_{146822}$              & 0.3972 & 0.4211 & 0.4037 & 0.4442 & 0.4075 & 0.4131 & 0.4008 & 0.3916 & 0.3878 & 0.3871    \\ \hline
$\text{MLP}_{31}$             & 0.5912 & 0.5914 & 0.5912 & 0.5912 & 0.5918 & 0.5923 & 0.5921 & 0.5911 & 0.5921 & 0.5921    \\
$\text{MLP}_{53}$             &  0.9096 & 0.8166 & 0.8111 & 0.8872 & 0.8546 & 1.0163 & 0.8565 & 0.9849 & 1.1276 & 1.0952   \\
$\text{MLP}_{3917}$             & 0.3798 & 0.3937 & 0.3862 & 0.3871 & 0.3867 & 0.4109 & 0.4262 & 0.3812 & 0.4003 & 0.4003    \\
$\text{MLP}_{10101}$             & 0.4219 & 0.4141 & 0.4197 & 0.4111 & 0.4303 & 0.4145 & 0.4256 & 0.4215 & 0.4502 & 0.4502    \\
$\text{MLP}_{146818}$             & 0.4943 & 0.4913 & 0.5022 & 0.5023 & 0.4884 & 0.4995 & 0.5046 & 0.4921 & 0.4978 & 0.4861    \\
$\text{MLP}_{146821}$             & 0.1169 & 0.0836 & 0.0915 & 0.1674 & 0.1079 & 0.0891 & 0.1389 & 0.0838 & 0.1051 & 0.1194    \\
$\text{MLP}_{146822}$             & 0.2963 & 0.2914 & 0.2705 & 0.3025 & 0.2779 & 0.2759 & 0.2621 & 0.2549 & 0.257 & 0.2518    \\
\bottomrule
\end{tabular}%
}
\end{table}

\subsection{Surrogate mode}
\begin{figure}[htbp]
	\centering

	\begin{subfigure}{0.25\linewidth}
		\centering
		\includegraphics[width=0.9\linewidth]{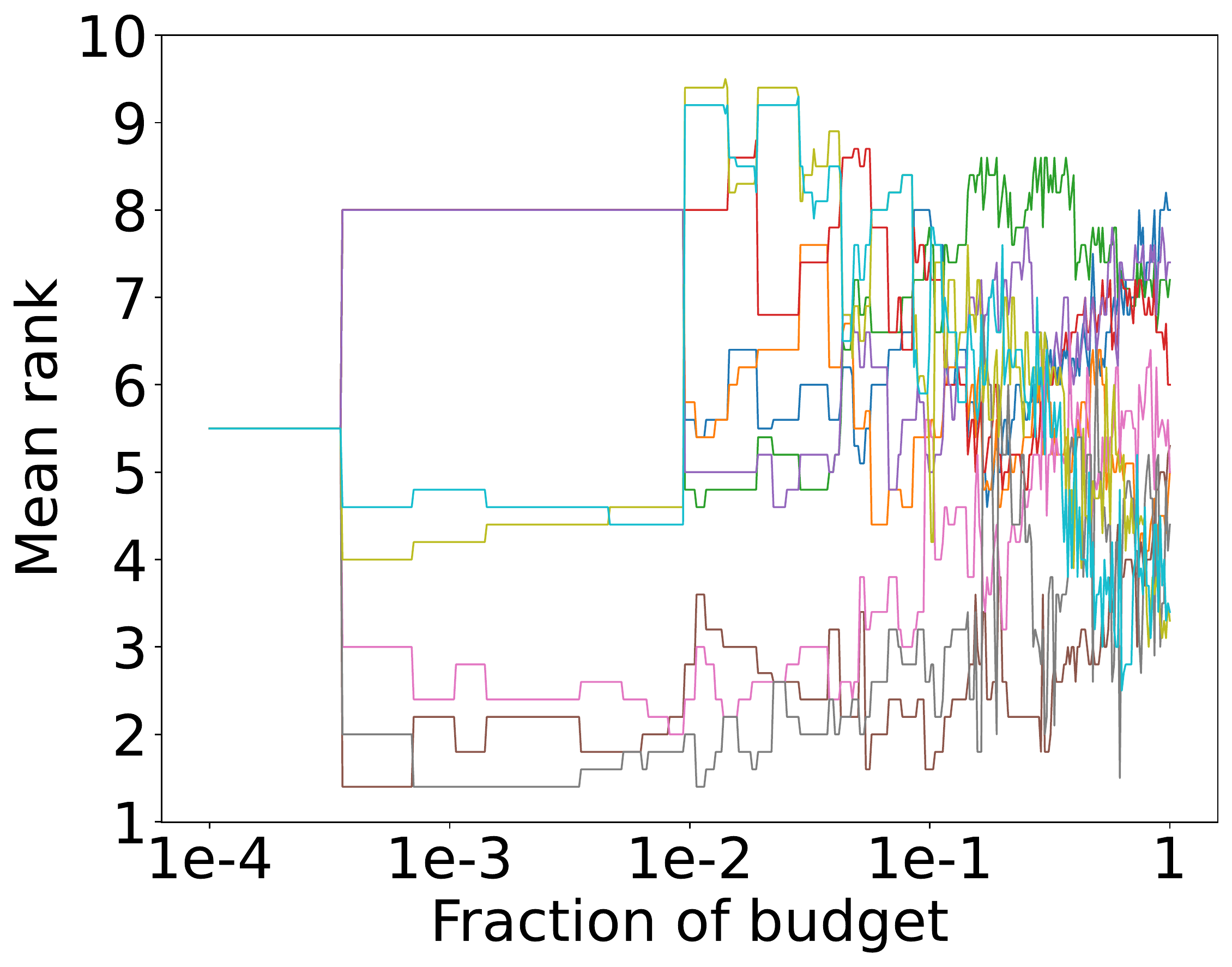}
		\caption{$\text{ALL}_{\text{FEMNIST}}$}
		\label{fig:All_FEMNIST_avg_surrogate}
	\end{subfigure}
	\begin{subfigure}{0.25\linewidth}
		\centering
		\includegraphics[width=0.9\linewidth]{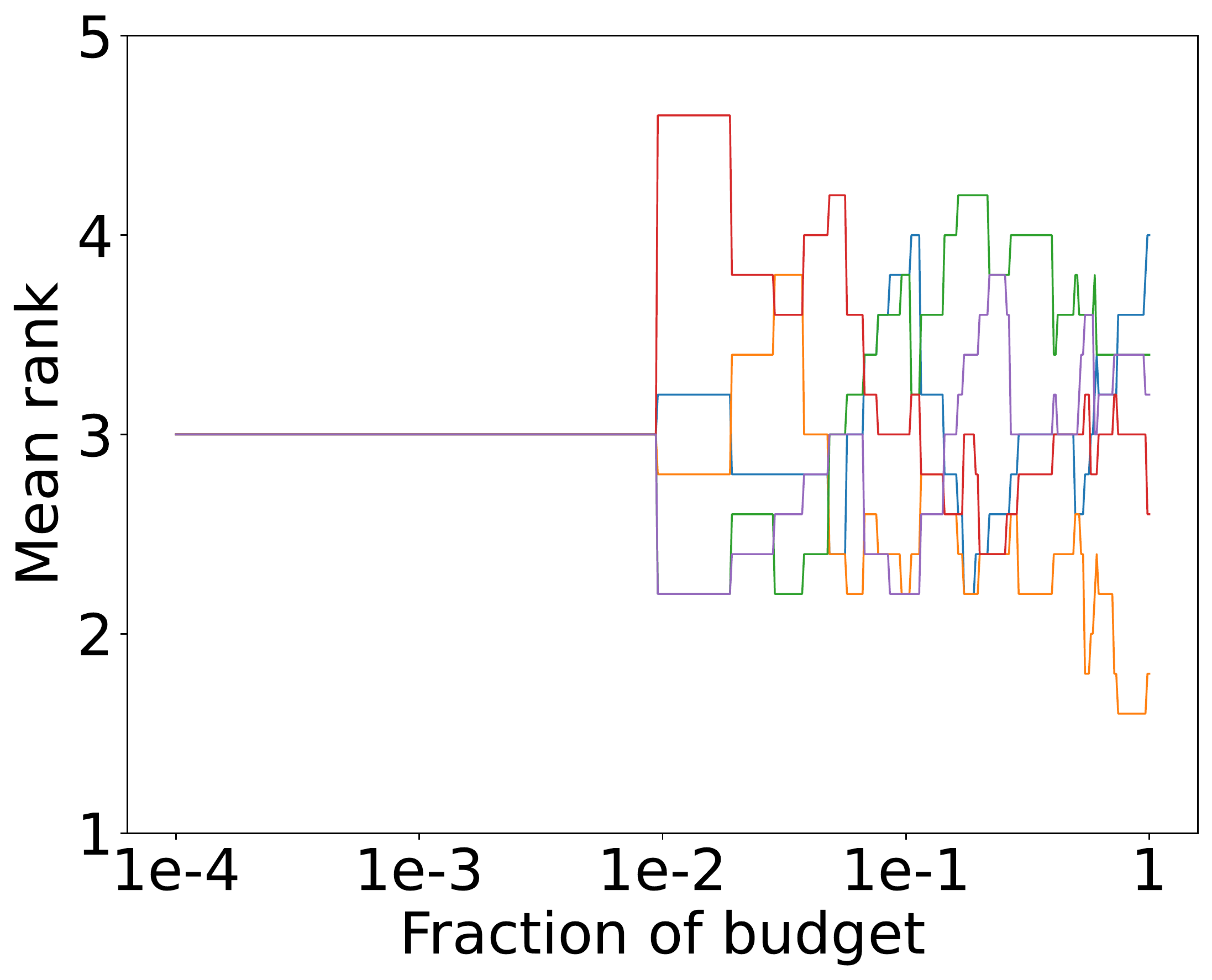}
		\caption{$\textit{BBO}_{\text{FEMNIST}}$}
		\label{fig:BBO_FEMNIST_avg_surrogate}
	\end{subfigure}
	\begin{subfigure}{0.25\linewidth}
		\centering
		\includegraphics[width=0.9\linewidth]{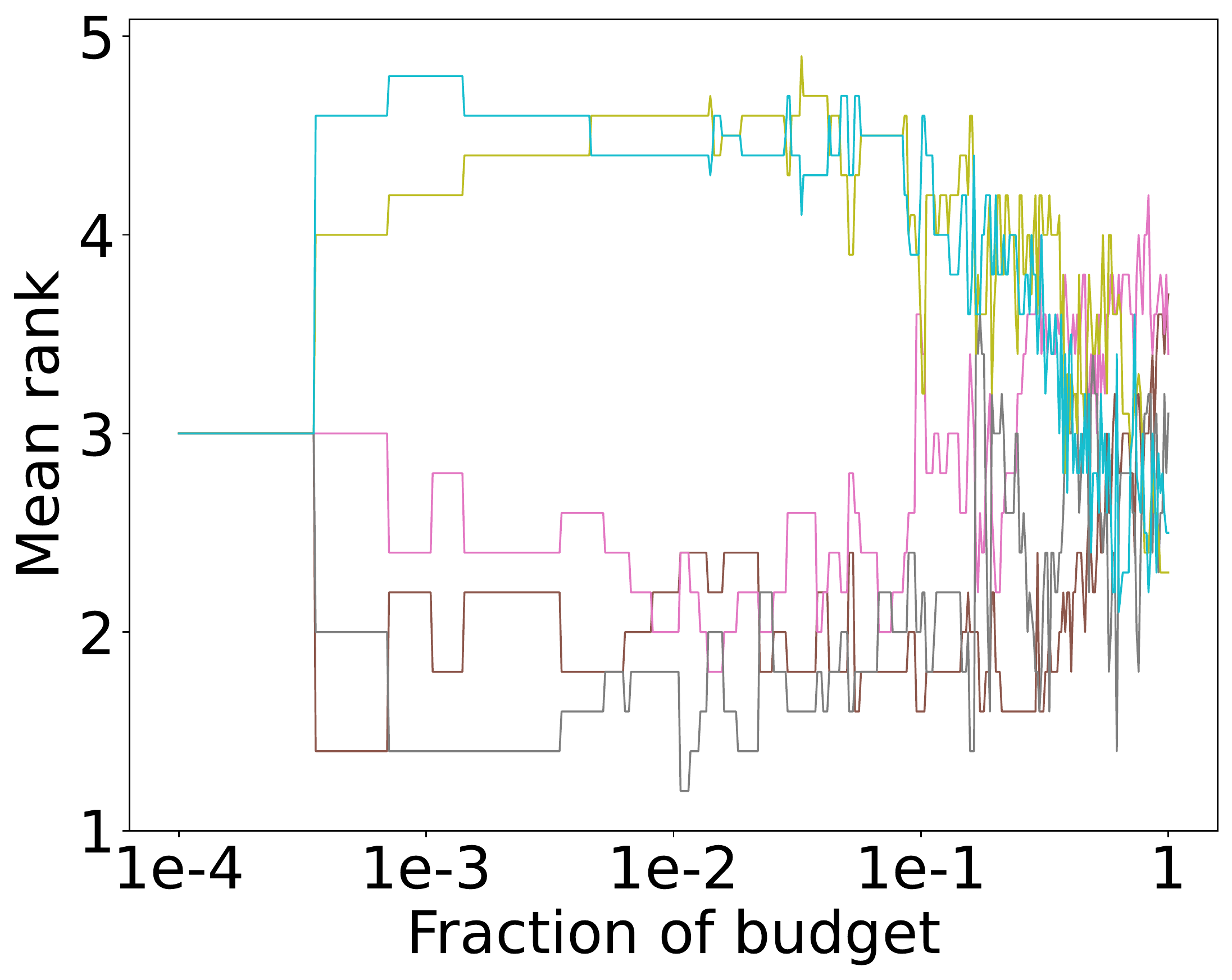}
		\caption{$\textit{MF}_{\text{FEMNIST}}$}
		\label{fig:MF_FEMNIST_avg_surrogate}
	\end{subfigure}
	\centering
	\hspace*{1.2cm}\begin{subfigure}{1.0\linewidth}
		\centering
		\includegraphics[width=0.95\linewidth]{materials/legend_rank_new.pdf}
	\end{subfigure}
	\vspace{-0.1in}

	\caption{Mean rank over time on CNN benchmark under surrogate mode (FedAvg).}
	\label{fig:entire_cnn_surrogate_avg_rank}
\end{figure}

\begin{figure}[htbp]
	\centering
	\begin{subfigure}{0.25\linewidth}
		\centering
		\includegraphics[width=0.9\linewidth]{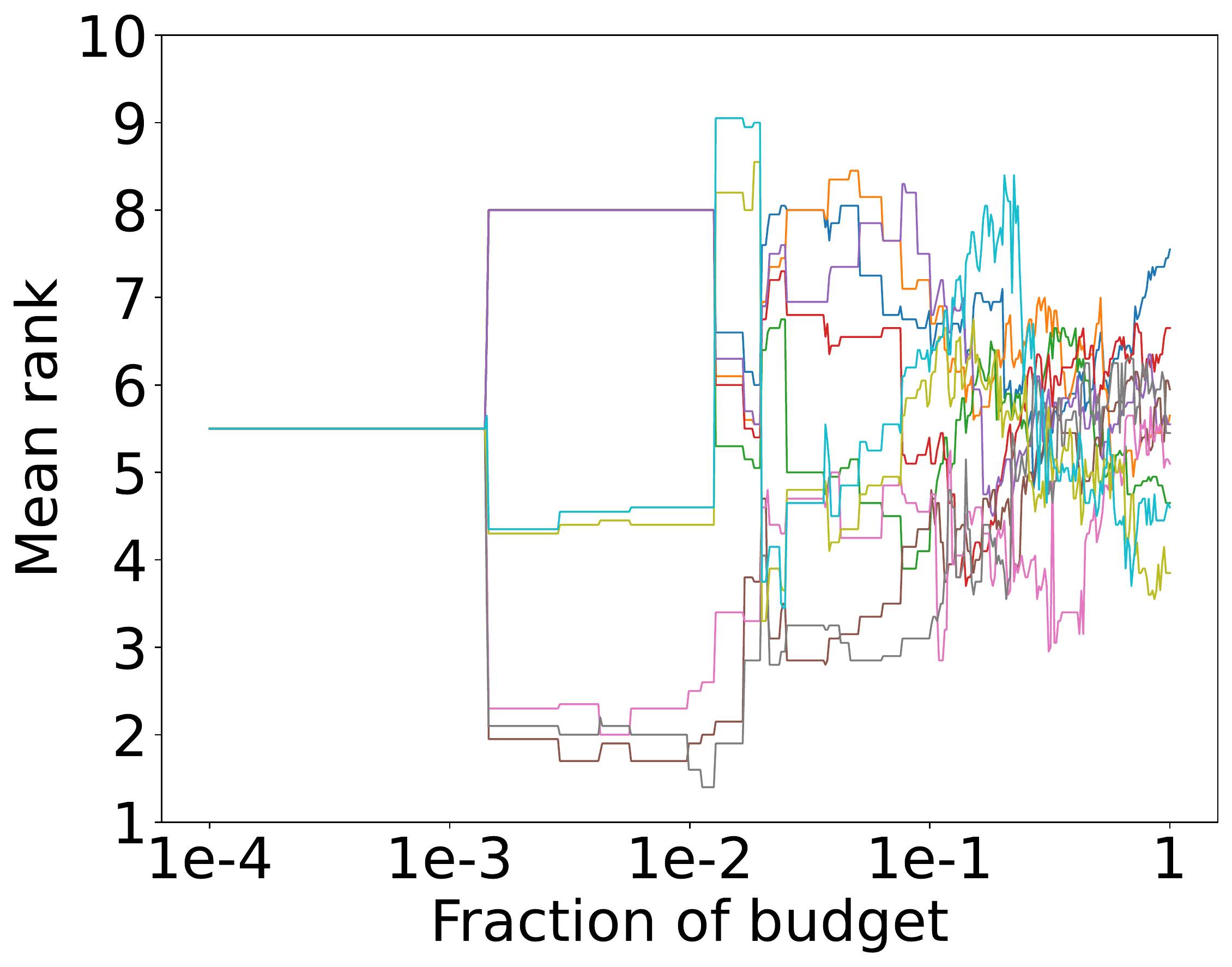}
		\caption{$\text{ALL}_\text{{BERT}}$}
		\label{fig:All_BERT_avg_surrogate}
	\end{subfigure}
	\begin{subfigure}{0.25\linewidth}
		\centering
		\includegraphics[width=0.9\linewidth]{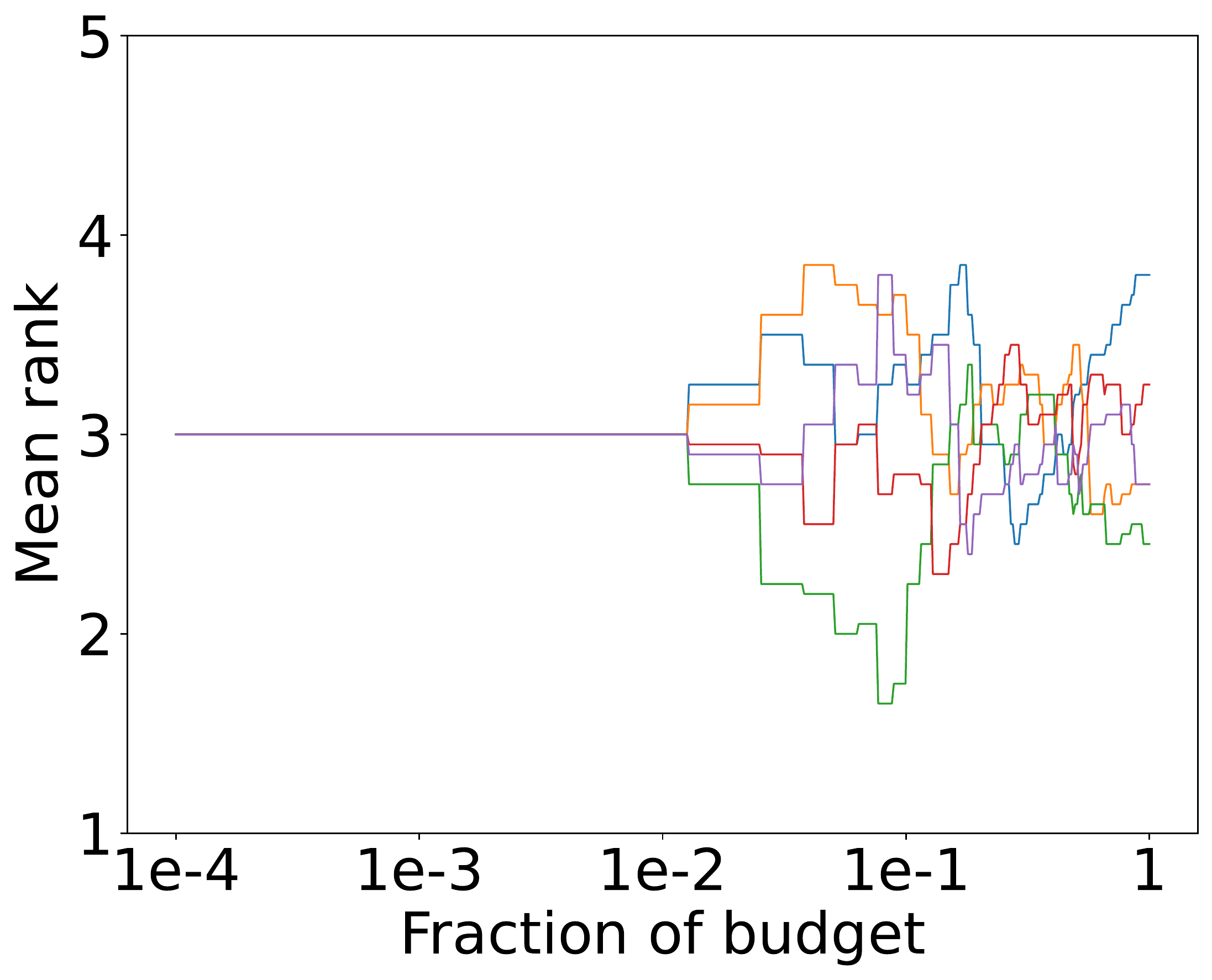}
		\caption{$\textit{BBO}_\text{{BERT}}$}
		\label{fig:BBO_BERT_avg_surrogate}
	\end{subfigure}
	\begin{subfigure}{0.25\linewidth}
		\centering
		\includegraphics[width=0.9\linewidth]{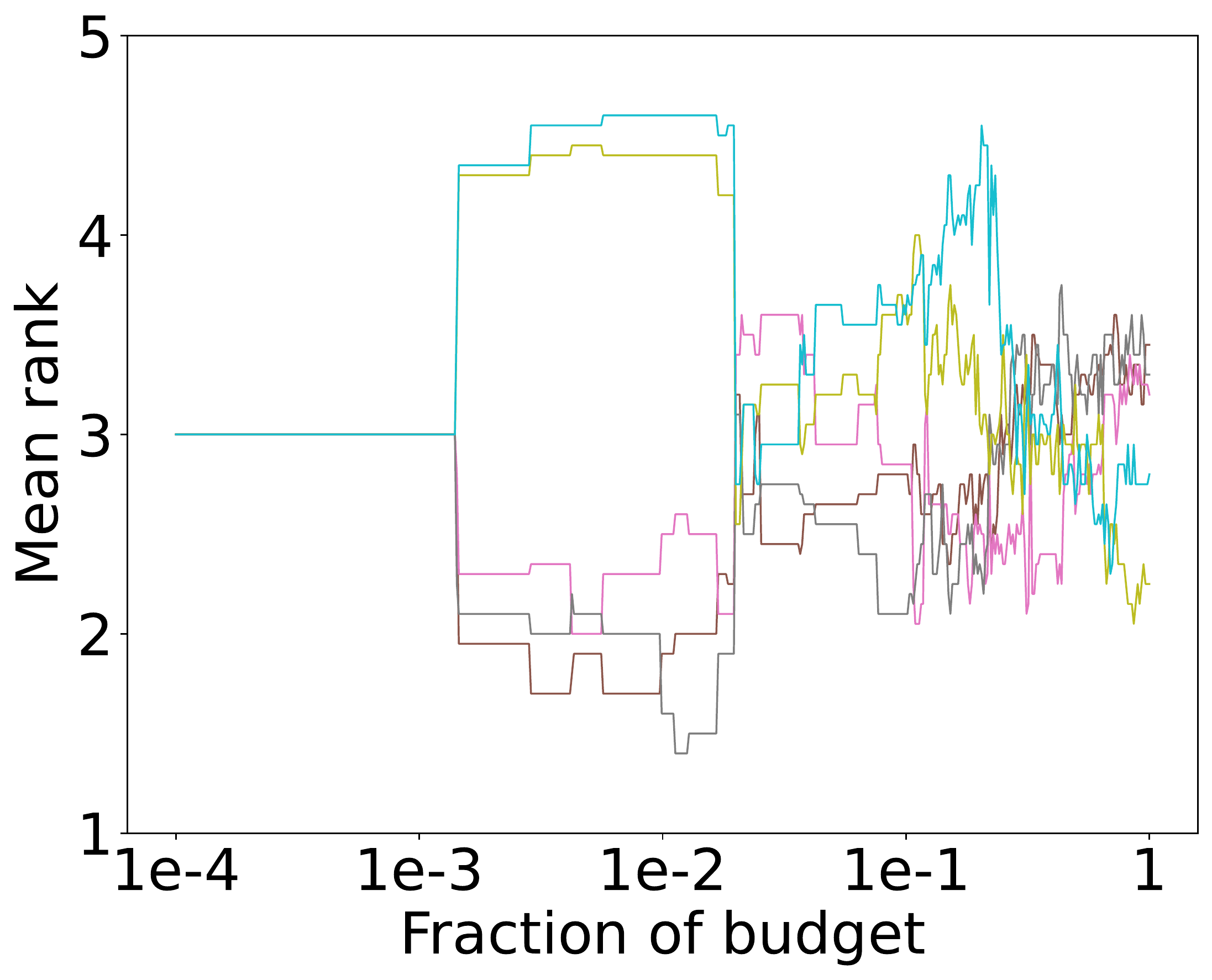}
		\caption{$\textit{MF}_\text{{BERT}}$}
		\label{fig:MF_BERT_avg_surrogate}
	\end{subfigure}

	\begin{subfigure}{0.25\linewidth}
		\centering
		\includegraphics[width=0.9\linewidth]{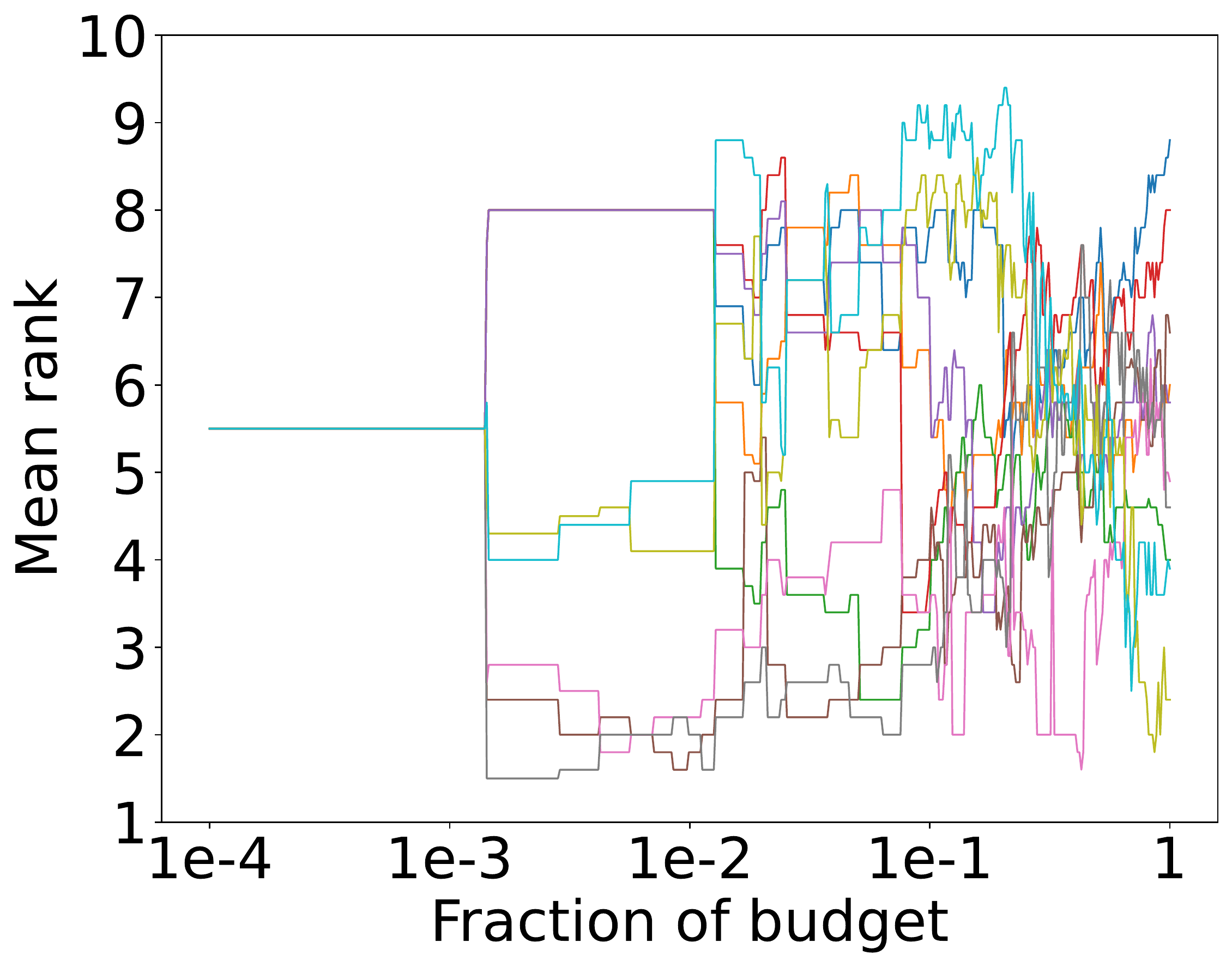}
		\caption{$\text{ALL}_{\text{CoLA}}$}
		\label{fig:All_CoLA_avg_surrogate}
	\end{subfigure}
	\begin{subfigure}{0.25\linewidth}
		\centering
		\includegraphics[width=0.9\linewidth]{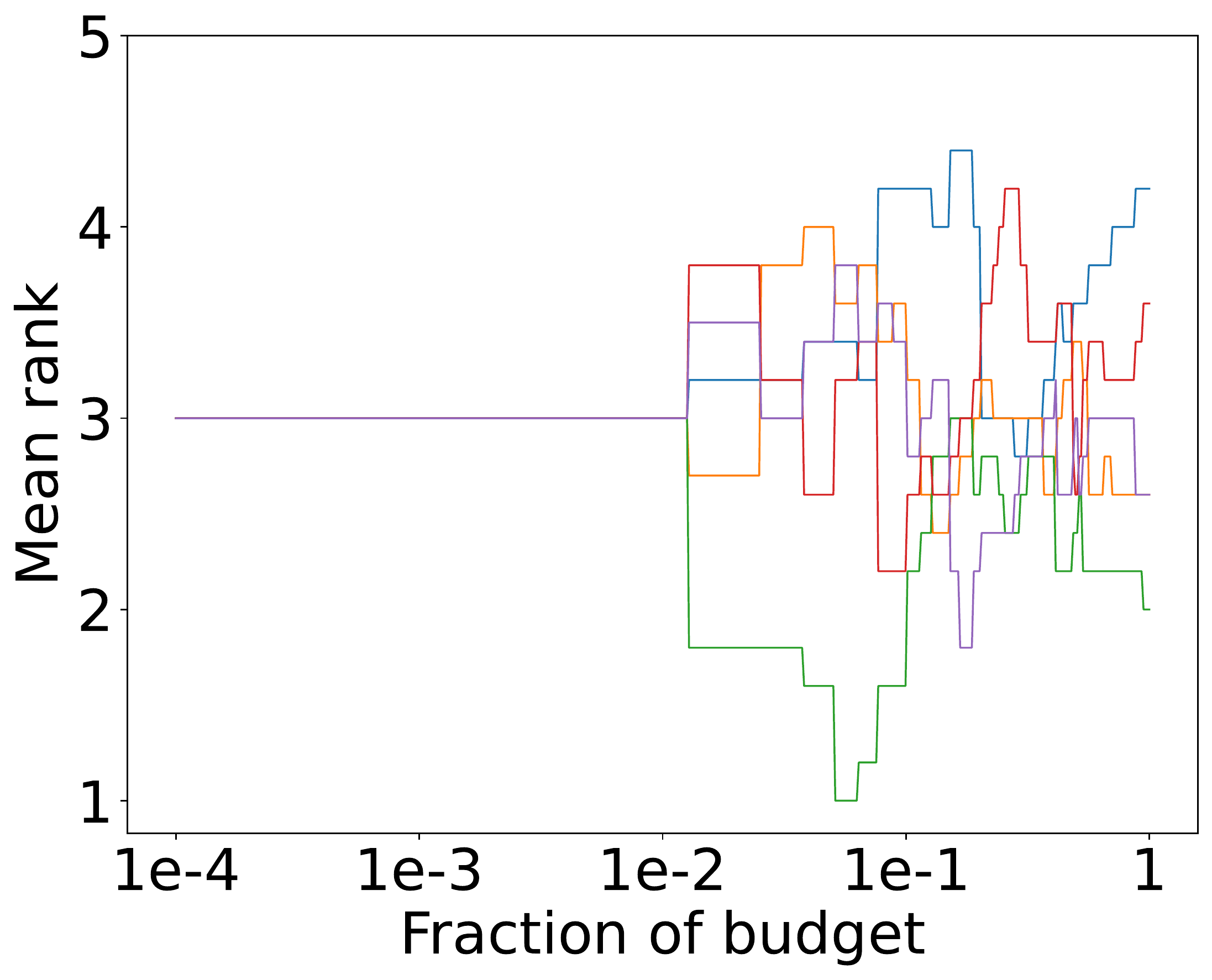}
		\caption{$\textit{BBO}_{\text{CoLA}}$}
		\label{fig:BBO_CoLA_avg_surrogate}
	\end{subfigure}
	\begin{subfigure}{0.25\linewidth}
		\centering
		\includegraphics[width=0.9\linewidth]{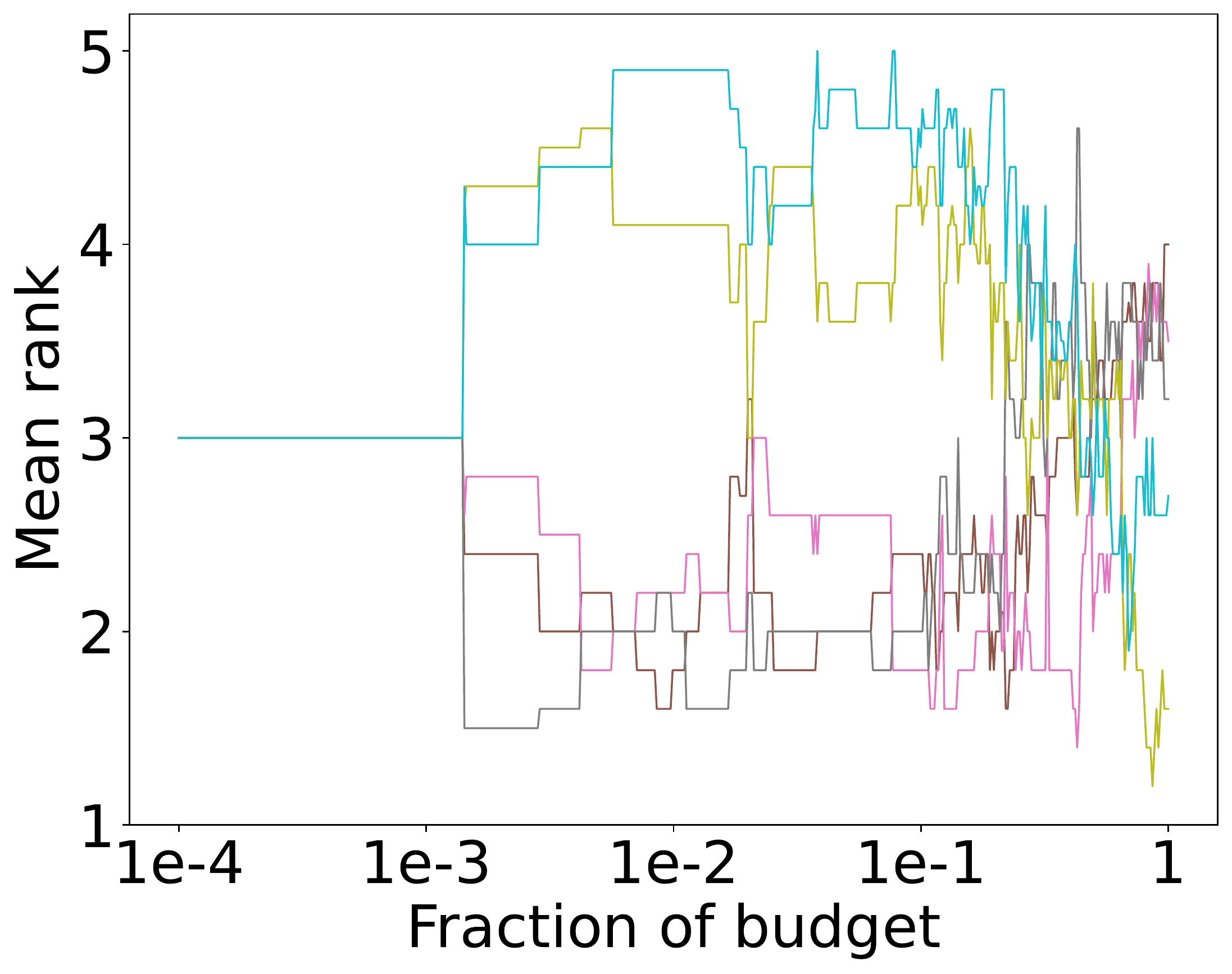}
		\caption{$\textit{MF}_{\text{CoLA}}$}
		\label{fig:MF_CoLA_avg_surrogate}
	\end{subfigure}
	
	\begin{subfigure}{0.25\linewidth}
		\centering
		\includegraphics[width=0.9\linewidth]{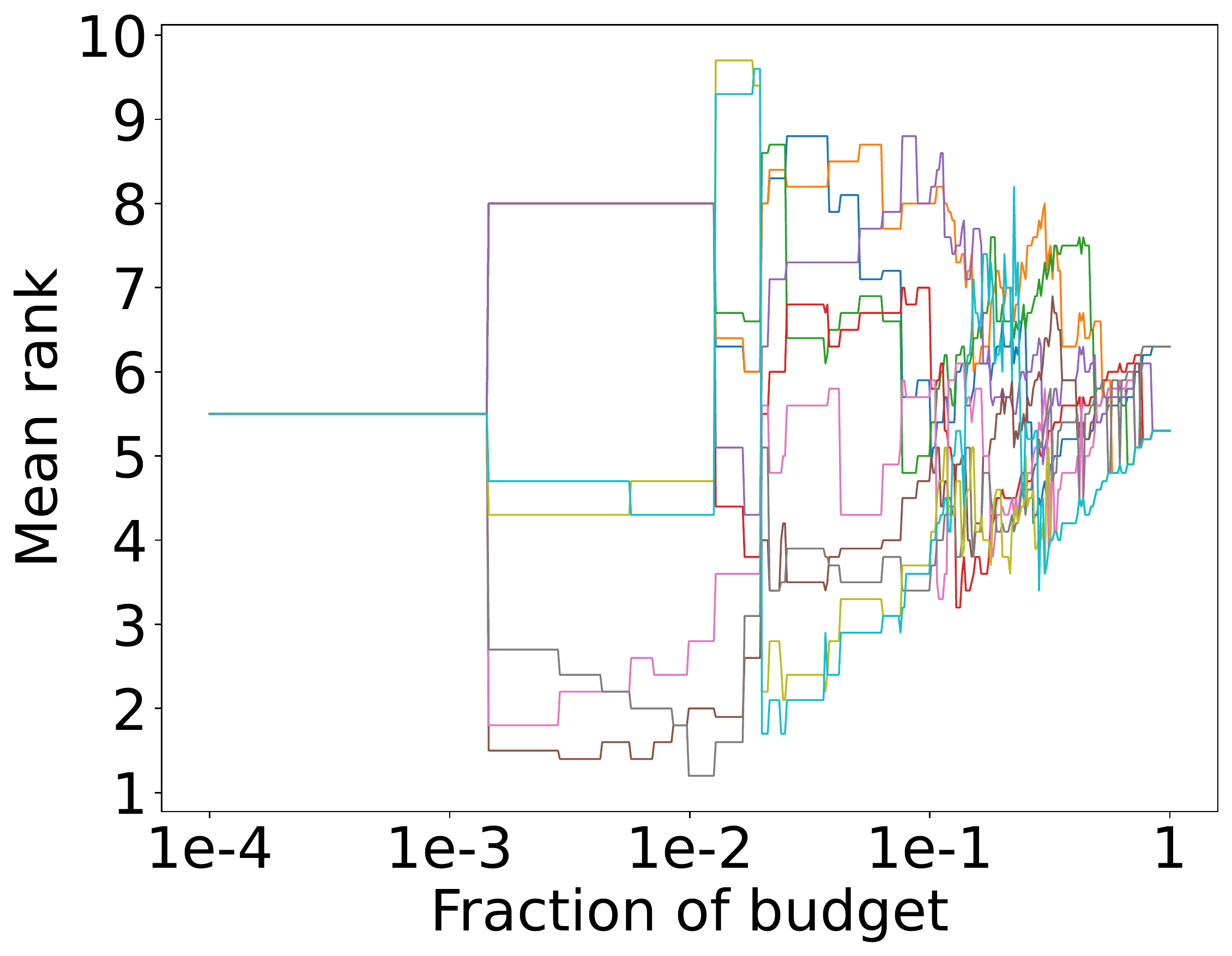}
		\caption{$\text{ALL}_{\text{SST-2}}$}
		\label{fig:All_SST-2_avg_surrogate}
	\end{subfigure}
	\begin{subfigure}{0.25\linewidth}
		\centering
		\includegraphics[width=0.9\linewidth]{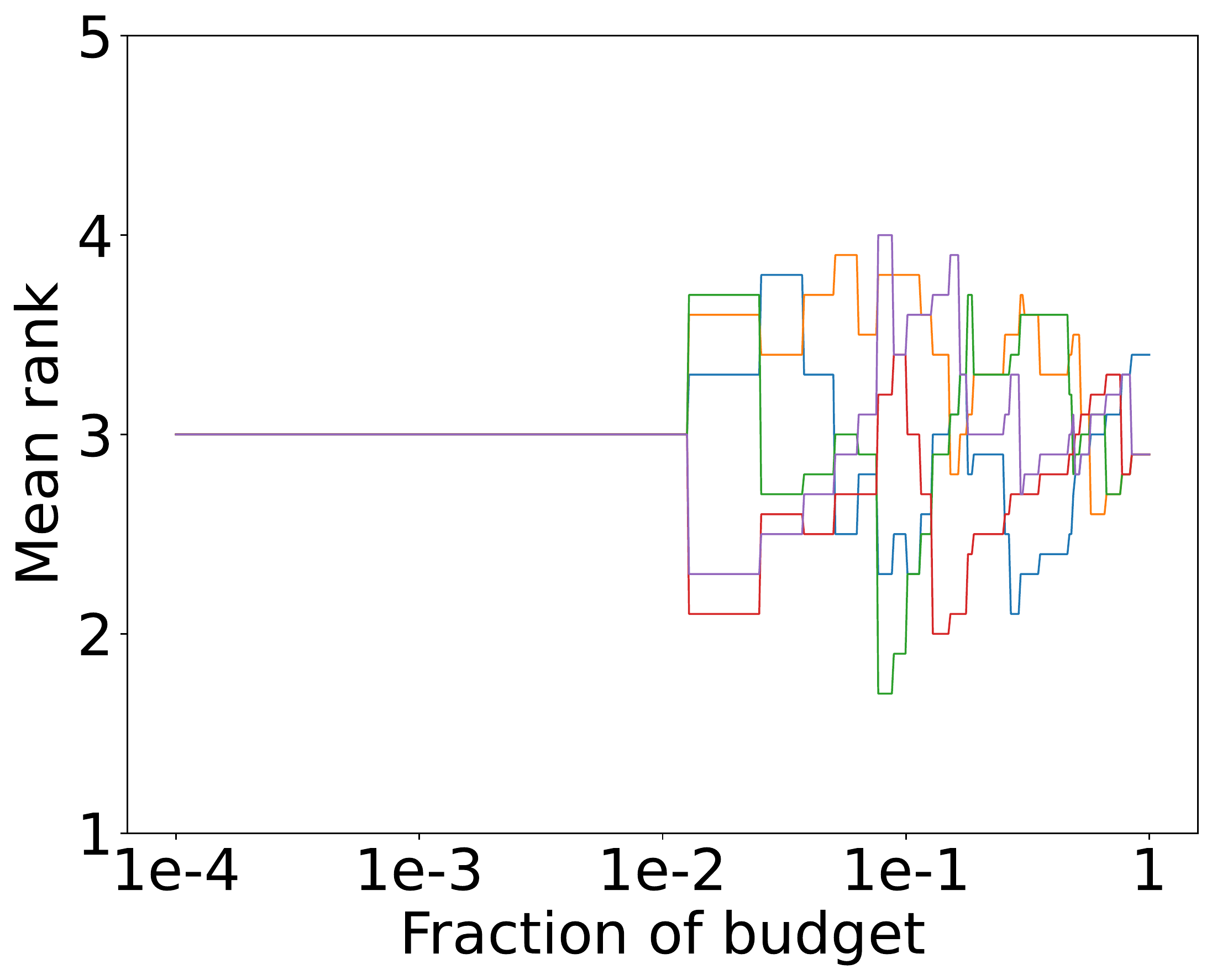}
		\caption{$\textit{BBO}_{\text{SST-2}}$}
		\label{fig:BBO_SST-2_avg_surrogate}
	\end{subfigure}
	\begin{subfigure}{0.25\linewidth}
		\centering
		\includegraphics[width=0.9\linewidth]{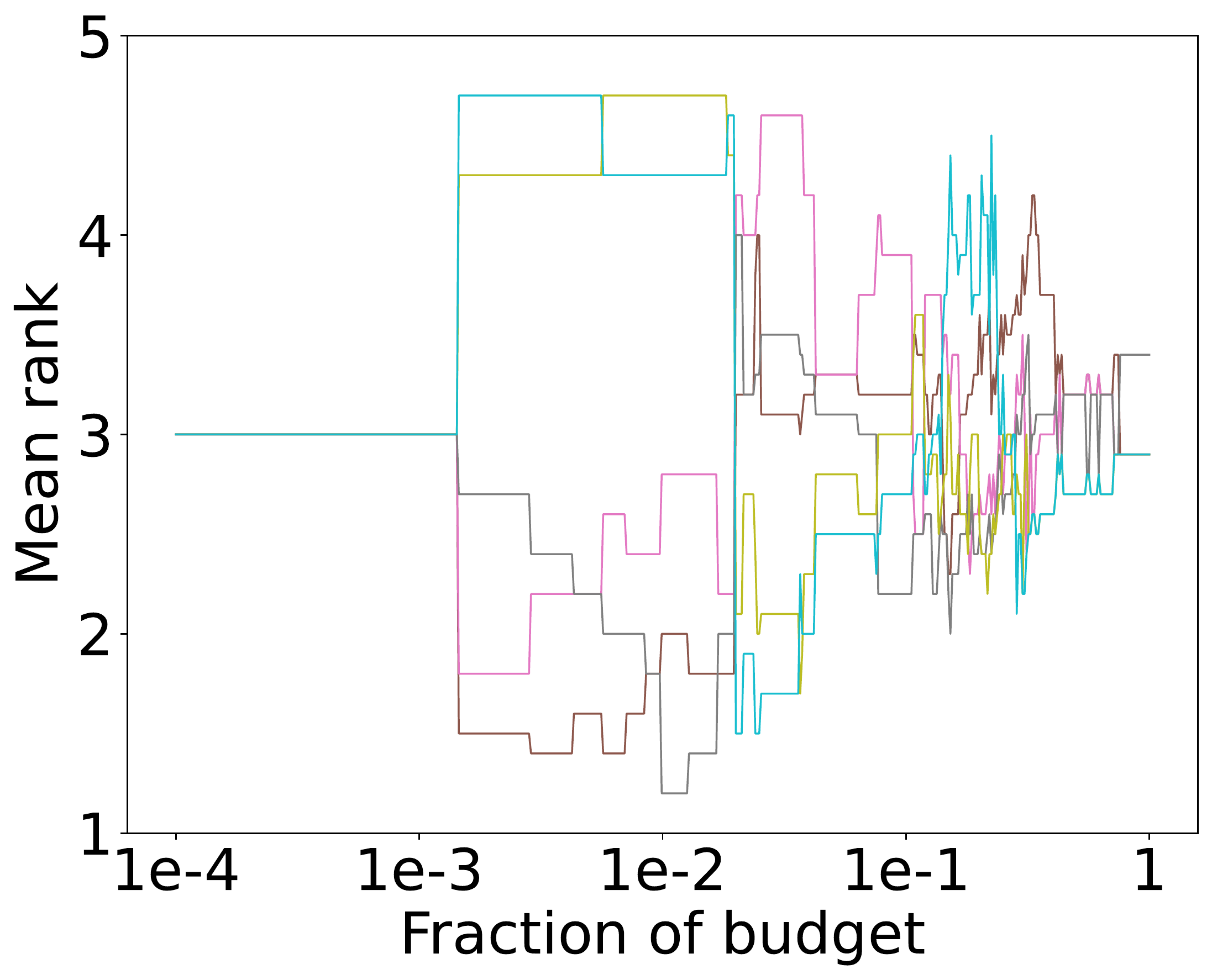}
		\caption{$\textit{MF}_{\text{SST-2}}$}
		\label{fig:MF_SST-2_avg_surrogate}
	\end{subfigure}
	
	\centering
	\hspace*{1.2cm}\begin{subfigure}{1.0\linewidth}
		\centering
		\includegraphics[width=0.95\linewidth]{materials/legend_rank_new.pdf}
	\end{subfigure}
	\vspace{-0.1in}
	
	\caption{Mean rank over time on BERT benchmark under surrogate mode (FedAvg).}
	\label{fig:entire_bert_surrogate_avg_rank}
\end{figure}

We report the the final results with FedAvg on FEMNIST and BERT benchmarks in Table~\ref{tab:final_results_sur_avg}. Then we present the mean rank over time of the optimizers in Figure~\ref{fig:entire_cnn_surrogate_avg_rank} and Figure~\ref{fig:entire_bert_surrogate_avg_rank}.

\begin{table}[htbp]
\centering
\caption{Final results of the optimizers in surrogate mode (lower is better).}
\label{tab:final_results_sur_avg}
\resizebox{\textwidth}{!}{%
\begin{tabular}{lllllllllll}
\toprule
benchmark & \textit{RS}  & $\textit{BO}_{\textit{GP}}$ & $\textit{BO}_{\textit{RF}}$ & $\textit{BO}_{\textit{KDE}}$ & \textit{DE} & \textit{HB} & \textit{BOHB} & \textit{DEHB} & $\textit{TPE}_{\textit{MD}}$ & $\textit{TPE}_{\textit{HB}}$ \\ \midrule
$\text{CNN}_{\text{FEMNIST}}$      &  0.0508 & 0.0478 & 0.0514 & 0.0492 & 0.0503 & 0.0478 & 0.048 & 0.0469 & 0.0471 & 0.0458  \\ \hline
$\text{BERT}_{\text{SST-2}}$ & 0.4909 & 0.4908 & 0.4908 & 0.4908 & 0.4908 & 0.4908 & 0.4908 & 0.4917 & 0.4908 & 0.4908  \\
$\text{BERT}_{\text{CoLA}}$ &  0.5013 & 0.4371 & 0.4113 & 0.487 & 0.444 & 0.4621 & 0.4232 & 0.4204 & 0.3687 & 0.3955  \\
\bottomrule
\end{tabular}%
}
\end{table}

\end{document}